%% file: main.tex
\documentclass[10pt,twocolumn,letterpaper]{article}

\usepackage[pagenumbers]{cvpr}              %

\usepackage{graphicx}
\usepackage{amsmath}
\usepackage{amssymb}
\usepackage{booktabs}
\usepackage{pifont}
\usepackage{caption}
\usepackage{subcaption}

\makeatletter
\@namedef{ver@everyshi.sty}{}
\makeatother
\usepackage{tikz}
\usepackage{pgfplots}
\usepackage{wrapfig}

\usepackage[accsupp]{axessibility}

\usepackage{algorithm}
\usepackage{algcompatible}
\usepackage{algorithmicx}
\usepackage{algpseudocode}

\usepackage{mathtools, cuted}
\usepackage{resizegather}

\usepackage{widetext}

\usepackage{rotating}
\usepackage{xr}

\usepackage[pagebackref,breaklinks,colorlinks]{hyperref}

\usepackage[capitalize]{cleveref}
\crefname{section}{Sec.}{Secs.}
\Crefname{section}{Section}{Sections}
\Crefname{table}{Table}{Tables}
\crefname{table}{Tab.}{Tabs.}

\usepackage{color} 
\definecolor{darkorange}{rgb}{1.0, 0.55, 0.0} 
\definecolor{darkgreen}{rgb}{0.0, 0.55, 0.0} 

\newcommand{\exy}[1]{\textcolor{darkorange}{#1}} 
\newcommand{\exz}[1]{\textcolor{darkgreen}{#1}} 
\newcommand{\eyz}[1]{\textcolor{blue}{#1}} 

\DeclareMathOperator*{\argmin}{arg\,min}

\begin{document}

\title{CCuantuMM: Cycle-Consistent Quantum-Hybrid Matching of Multiple Shapes} 

\author{Harshil Bhatia$^{1,2}$~~~Edith Tretschk$^2$~~~Zorah Lähner$^3$~~~Marcel Seelbach Benkner$^3$\\
Michael Moeller$^3$~~~Christian Theobalt$^2$~~~Vladislav Golyanik$^2$\\
$^1$Indian Institute of Technology, Jodhpur~~~~$^2$MPI for Informatics, SIC~~~~$^3$Universität Siegen
}
\maketitle

\begin{abstract}
   \input{sections/0_abstract}

\end{abstract}

\input{sections/1_introduction}
\input{sections/2_related_work}

\input{sections/3_background}

\input{sections/4_method}

\input{sections/5_experiments}

\input{sections/6_discussion_conclusion}

{\small
\bibliographystyle{ieee_fullname}
\bibliography{main}
}

~
\newpage
~
\newpage

\appendix

\begin{center}
\textbf{\Large Supplementary Material} 
\end{center} 

This supplementary material contains additional results.
We present them in the order they are mentioned in the main paper. 
First, Sec.~\ref{sec:notation} provides an overview of our notation. 
Sec.~\ref{sec:nshapealg} formally states the $N$-shape algorithm. 
Sec.~\ref{sec:100shapes} contains large-scale figures showing the $100$ matched FAUST shapes. 
Sec.~\ref{sec:higherorder} describes the elimination of higher-order terms in detail. 
Sec.~\ref{sec:Wtilde} states the weight matrix $\Tilde{W}$ of the final QUBO. 
Sec.~\ref{sec:evolution} shows that the total energy almost never decreases in practice. 
Sec.~\ref{sec:anchorchoice} compares our anchor scheme to using random triplets. 
Sec.~\ref{sec:timecomplexity} discusses the time complexity of our proposed method. 
Sec.~\ref{sec:implementationdetails} provides more implementation details on  initialisation and deciding trivial cycles. 
Sec.~\ref{sec:qpu} discusses the minor embeddings on real quantum hardware and contains further QPU experiments. 
Sec.~\ref{sec:ablations_and_analysis} presents more results, including more ablation experiments. 
Finally, Sec.~\ref{sec:related_work_appendix} compares existing quantum computer vision works for alignment tasks. 

In addition, the supplementary material contains a video showing the evolution of the matchings of a ten-shape FAUST instance over the course of the optimisation. 
We visualise the matchings by fixing the colouring of the top-left shape and transferring this colouring to all other shapes according to the estimated correspondences.

\section{Notations}\label{sec:notation}

Tab.~\ref{tab:notation} summarises the notation we use. 

\begin{table}[h]
    \centering
    \resizebox{\linewidth}{!}{
    \begin{tabular}{l|l}
    \hline
    Notation     &  Meaning \\ \hline \hline
    $\mathcal{I}$, $\mathcal{J}$, $\mathcal{X}$, $\mathcal{Y}$, $\cal Z$ & shapes (represented as meshes) \\ 
    $P_\mathcal{IJ}$ & permutation matrix from shape $\mathcal{I}$ to $\mathcal{J}$ \\ 
    $x$, $y$, $u$, $v$ & vertices of a shape\\
    $\mathcal{P}$ & set of permutations \\ 
    $\mathcal{S}$ & set of shapes  \\
    $W$ & energy matrix \\
    $N$ & number of shapes \\ 
    $I$ & identity matrix \\
    $E_{\cal I J}(P,Q)$ & $\textit{vec}(P)^\top W_{\cal I J} \textit{vec}(Q)$ \\
    $E_{\cal I J}(P)$ & $E_{\cal I J}(P,P)$ \\
    $F_{\cal I J}(A,B) $ & $E_{\cal I J}(A,B) + E_{\cal I J}(B,A)$ \\ 
    $\alpha, \beta$ & decision variables \\ 
    $k$ & size of decision variables $\alpha$,$\beta$ \\
    $m$ & number of worst vertices \\
    $I_{\cal X Y}(x)$ & relative inconsistency of vertex $x$ for $P_{\cal X Y}$ \\ 
    $V_{\cal X}$ & set of worst vertices for shape $\cal X$ \\
    $\mathcal{C_X}$ & set of cycles for shape $\cal X$\\ 
    $A$ & anchor shape \\ 
    $C_i $ &  $(c_i - I) P_{\cal X Y}$; single-cycle update of $P_{\cal X Y} $ \\
    $\tilde{C}_j $ & $(c_j - I)P_{\cal Y Z}$; single-cycle update of $P_{\cal Y Z} $\\
    $K_{ij}$ & $C_i \tilde{C}_j$ \\
    \hline
    \end{tabular}
    }
    \caption{Notation used in this work.}
    \label{tab:notation}
\end{table}

\section{$N$-Shape Algorithm}\label{sec:nshapealg}
We provide the formal $N$-shape algorithm as Alg.~\ref{Alg:nshape}. 
\begin{algorithm}
\def\NoNumber#1{{\def\alglinenumber##1{}\State #1}\addtocounter{ALG@line}{-1}}
\caption{CCuantuMM (Matching $N$ Shapes; Sec. 4.2)}
\label{Alg:nshape}
\textbf{Input:} $\mathcal{S},T$ \\ 
\textbf{Output:} $\mathcal{P}$
\begin{algorithmic}[1]
\State {initialise $\mathcal{P}^{\mathit{init}}$ from HKS descriptors}
\State {determine anchor $A\in\mathcal{S}$}  
\State {$\mathcal{P}^0\gets \{P^\mathit{init}_{\mathcal{I} A}\}_{\mathcal{I} \in \mathcal{S}, \mathcal{I} \neq A} $ } 
\State {randomly pick $\mathcal{X}^{-1}\in\mathcal{S}$}\hfill\Comment{technicality for $\mathcal{Z}^0$}
\For{$i=0$ to $2T(\lvert \mathcal{S} \rvert-1)$ } \hfill \Comment{iterations}
    \If{$i \mod (\lvert \mathcal{S} \rvert-1) == 0$}
        \State {$\mathcal{S}' \gets \mathcal{S}\setminus\{A\}$}\hfill\Comment{for stratified sampling}
    \EndIf
    \State {randomly pick $\mathcal{X}^i\in\mathcal{S}'$}
    \State {$\mathcal{S}' \gets \mathcal{S}' \setminus \{\mathcal{X}^i\}$}
    \State {$\mathcal{Y}^{i} \gets A$}
    \State {$\mathcal{Z}^{i} \gets \mathcal{X}^{i-1}$ } 
    \State {get $P^i_{\mathcal{X}^i A}, P^i_{\mathcal{Z}^i A}$ from $\mathcal{P}^i$}
    \State {$P^i_{\mathcal{X}^i \mathcal{Y}^i} \gets P^i_{\mathcal{X}^i A}$}
    \State {$P^i_{\mathcal{Y}^i \mathcal{Z}^i} \gets (P^i_{\mathcal{Z}^i A})^{-1}$}
    \State {$ P^i_{\mathcal{X}^i \mathcal{Z}^i} \gets P^i_{\mathcal{X}^i A}(P^i_{\mathcal{Z}^i A})^{-1}$}
    \State {$\operatorname{mode} \gets \operatorname{geodesic}$ \textbf{if} $i< T(\lvert \mathcal{S} \rvert-1) $ \textbf{else} $\operatorname{Gaussian}$ }
    \State {$\{P^{i+1}_{\mathcal{X}^i \mathcal{Y}^i},P^{i+1}_{\mathcal{Y}^i \mathcal{Z}^i},P^{i+1}_{\mathcal{X}^i \mathcal{Z}^i}\} \gets$ run Alg.~1 with $\operatorname{mode}$} 
    \NoNumber{on $(\{P^i_{\mathcal{X}^i \mathcal{Y}^i},P^i_{\mathcal{Y}^i \mathcal{Z}^i},P^i_{\mathcal{X}^i \mathcal{Z}^i}\}$, $\{\mathcal{X}^{i},\mathcal{Y}^{i},\mathcal{Z}^{i}\})$}
    \State {$P^{i+1}_{\mathcal{X}^i A} \gets P^{i+1}_{\mathcal{X}^i \mathcal{Y}^i}$}
    \State {$P^{i+1}_{\mathcal{Z}^i A} \gets (P^{i+1}_{\mathcal{Y}^i \mathcal{Z}^i})^{-1}$}
    \State {$\mathcal{P}^{i+1}\gets (\mathcal{P}^i \setminus \{P^i_{\mathcal{X}^i A}, P^i_{\mathcal{Z}^i A}\}) \cup \{P^{i+1}_{\mathcal{X}^i A},P^{i+1}_{\mathcal{Z}^i A}\} $}
\EndFor
\State\Return {$\mathcal{P}=\mathcal{P}^{2T(\lvert \mathcal{S} \rvert-1)}$}
\end{algorithmic}
\end{algorithm}

\section{Matching $\boldsymbol{100}$  Shapes}\label{sec:100shapes} 

\begin{figure*}
    \centering
    \includegraphics[width=\linewidth]{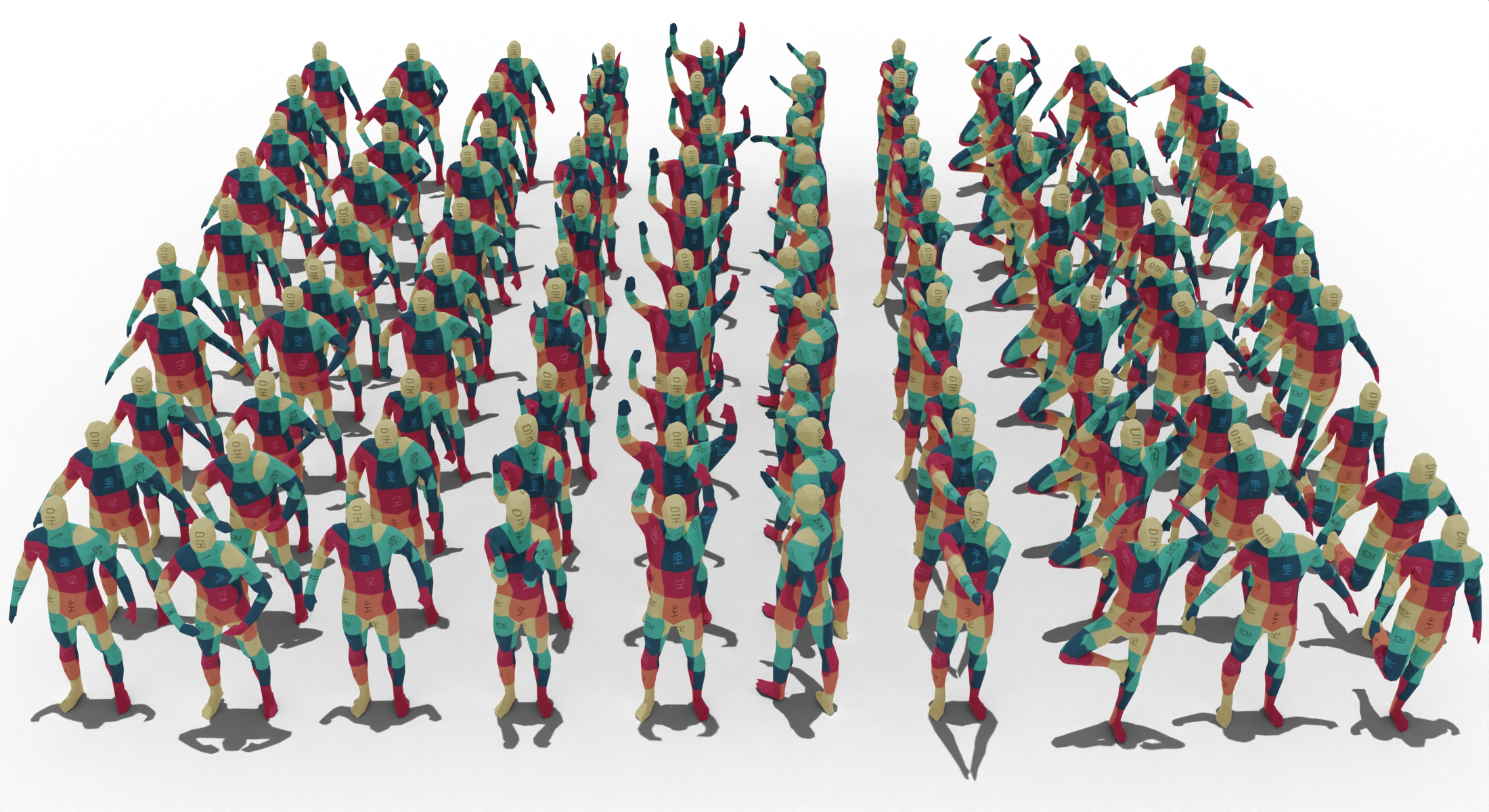}
    \caption{\textbf{Extended version of Fig.~1 in the main paper.} We visualise the matchings between all $100$ FAUST shapes via texture transfer. All correspondences are cycle-consistent by design.}
    \label{fig:teaser_full}
\end{figure*}

Fig.~\ref{fig:teaser_full} shows qualitative results of all $100$ matched FAUST shapes.

\section{Higher-Order Terms}\label{sec:higherorder}

In this section, we give the expansions for the third term of (7), namely $E_{\cal X Z} (P_{ \cal X Z}(\alpha,\beta))$, as it contains higher-order terms. 
Fig.~\ref{fig:higher_order_terms} shows the equations. 
The terms in red are cubic or bi-quadratic and we hence assume them to be 0, except for the summands that simplify to quadratic terms. 
For example, $\alpha_i \beta_j \alpha_i \beta_j  E_{\cal X Z}(C_i \tilde{C}_j, C_i \tilde{C}_j)$ simplifies to the linear summand $\alpha_i \beta_j E_{\cal X Z}(C_i \tilde{C}_j, C_i \tilde{C}_j)$. 
This then yields (8) from the main paper.

\section{Final QUBO}\label{sec:Wtilde}

Fig.~\ref{fig:WTilde} shows the weight matrix $\Tilde{W}\in\mathbb{R}^{2k \times 2k}$ that is used for the final QUBO.

\begin{figure*}
\begin{equation} 
\resizebox{\linewidth}{!}{
$
\tilde{W}_{ij} = \begin{cases}
E_{\cal YZ} (C_i,C_i) +E_{\cal XY}(P_{\cal XY},C_i) 
+ E_{\cal XY}(C_i, P_{\cal XY})
+ E_{ \cal XZ}(P_{\mathcal{XY}} P_{\mathcal{YZ}}, C_i P_{\mathcal{YZ}})
+ E_{ \cal XZ}(C_iP_{\mathcal{YZ}},P_{\mathcal{XY}} P_{\mathcal{YZ}}) 
+ E_{ \cal XZ}(C_i P_{\mathcal{YZ}}  , C_i P_{\mathcal{YZ}} )
 & \text{if } i = j  \leq  k \\
E_{\cal XY}(\tilde{C_j},\tilde{C_j})  +  E_{\cal YZ}(P_{\cal YZ}, \tilde{C}_j) 
+ E_{\cal YZ}(\tilde{C}_j,P_{\cal YZ})
+ E_{ \cal XZ}(P_{\mathcal{XY}} P_{\mathcal{YZ}}, P_{\mathcal{XY}} \tilde{C_j}) 
+ E_{ \cal XZ}(P_{\mathcal{XY}} \tilde{C_j},P_{\mathcal{XY}} P_{\mathcal{YZ}}) 
+ E_{ \cal XZ}(P_{\mathcal{XY}} \tilde{C_j}, P_{\mathcal{XY}} \tilde{C_j})
 & \text{if } i = j > k \\ 
E_{\cal YZ} (C_i,C_j) +  E_{ \cal XZ}(C_i P_{\mathcal{YZ}}, C_j P_{\mathcal{YZ}}) & \text{if } i \neq j ,i \leq k,j \leq k \\ 
E_{\cal XY}(\tilde{C_i},\tilde{C_j})  + E_{ \cal XZ}(P_{\mathcal{XY}} \tilde{ C_i}, P_{\mathcal{XY}} \tilde{ C_j} ) & \text{if } i \neq j , i >k , j > k \\
E_{\cal X Z}(P_{\cal X Y} P_{\cal Y Z},  C_i \tilde{C}_j )
+E_{ \cal XZ}(P_{\mathcal{XY}} \tilde{C_j},C_i P_{\mathcal{YZ}}) 
+E_{\cal X Z}( P_{\cal X Y}  \tilde{C}_j, C_i \tilde{C}_j)
+ E_{ \cal XZ}(C_i P_{\mathcal{YZ}}, P_{\mathcal{XY}} \tilde{ C_j}) \\
+ E_{\cal X Z}(C_i P_{\cal YZ}, C_i \tilde{C}_j)
+ E_{\cal X Z}(C_i \tilde{C}_j, P_{\cal X Y} P_{\cal Y Z})
+ E_{ \cal XZ}(C_i \tilde{C}_j, P_{ \cal XY} \tilde{C}_j) &\\
+ E_{ \cal XZ}(C_i \tilde{C}_j,C_i P_{ \cal YZ}) 
+ E_{ \cal XZ}(C_i \tilde{C}_j,C_i \tilde{C}_j) 
& \text{if } i \neq j , i \leq k ,  j>k  \\
0 & \text{otherwise}   
\end{cases}$
}
\end{equation} 
\caption{ The final QUBO weight matrix $\Tilde{W}$. For indices $i>k$, we define 
$\Tilde{C}_i=\Tilde{C}_{i-k}$. %
}
\label{fig:WTilde}
\end{figure*}

\section{Evolution During Optimisation}\label{sec:evolution}
Fig.~\ref{fig:Improvement} depicts how the total energy (11) and PCK evolve during optimisation. %
Importantly, the energy almost never increases in practice. 
Furthermore, our algorithm converges close to the ground truth.   
A sudden and significant improvement occurs as soon as the schedule switches from geodesics to Gaussians. 
We note that although the objective uses Gaussians in the second half of the schedule, both the total energy plotted here and the PCK are based on the (unfiltered) geodesic distances. 

\begin{figure}
    \centering
    \begin{subfigure}[b]{\linewidth}
            \resizebox{0.49\linewidth}{!}{
    \input{SupplimentFigures/PCK_Evolution}}
    \resizebox{0.49\linewidth}{!}{
    \input{SupplimentFigures/Energy_Change}}
        \caption{FAUST}
    \end{subfigure}

        \begin{subfigure}[b]{\linewidth}
    \resizebox{0.49\linewidth}{!}{
    \input{SupplimentFigures/PCK_Evolution_TOSCA.tex}}
    \resizebox{0.46\linewidth}{!}{
    \input{SupplimentFigures/EnergyTOSCA.tex}}
        \caption{TOSCA}
    \end{subfigure}

        \begin{subfigure}[b]{\linewidth}
    \resizebox{0.49\linewidth}{!}{
    \input{SupplimentFigures/PCK_Evolution_SMAL.tex}}
    \resizebox{0.463\linewidth}{!}{
    \input{SupplimentFigures/EnergySMAL.tex}}
        \caption{SMAL}
    \end{subfigure}

    \caption{Evolution during the optimisation (left) of the PCK, depicted with a colour bar gradient, and (right) of the total  energy. 
    The horizontal red-dashed line is the energy of the ground-truth solution. 
    The results are on (a)  a class of FAUST containing ten shapes, (b) the cat class of TOSCA containing eleven shapes, and (c) the cat class of SMAL containing nine shapes. 
    }
    \label{fig:Improvement}
\end{figure}

\section{Fixed Anchor vs. Random Triplets}\label{sec:anchorchoice}

Fig.~\ref{fig:AnchorAblation} compares our anchor-based scheme with using random triplets. 
We see a small improvement with our scheme. 

\begin{figure}%
    \centering
    \resizebox{0.59 \linewidth}{!}{
    \input{SupplimentFigures/PCK_vm}}
    \caption{PCK when using a fixed anchor (as our method does) and when using random triplets.  
    We average across all classes of FAUST. 
    }
    \label{fig:AnchorAblation}
\end{figure}

\section{Time Complexity}\label{sec:timecomplexity}
Our method mainly consists of constructing and solving the QUBO matrix $\tilde{W}$, which is based on $W$. 
However, standard meshes contain thousands of vertices, which makes na\"ively calculating the full $W \in \mathbb{R}^{n^2 \times n^2}$ not feasible due to the memory restrictions. 
Fortunately, we do not need to compute the full $W$ but only a small set of its entries. 
This is due to the extreme sparsity of the $c_i{-}I$ matrix (only four non-zero elements) since we only consider 2-cycles. 
Using $k$ 2-cycles leads to a worst-case time complexity of $\mathcal{O}(n k^2)$ for the three-shape Alg. 1 from the main paper (\textit{i.e.}, one sub-sub-iteration). %
For the sub-sub-iterations of a sub-iteration, $\tilde{W}$ is constant and we thus need to compute $\tilde{W}$ only once for each sub-iteration. 
In addition, each iteration has $k{-}1$ sub-iterations, resulting in a time complexity of $\mathcal{O}(n k^3)$ for each iteration. 
Furthermore, our implementation for computing $\tilde{W}$ is significantly faster in practice than the sub-sampling technique proposed in Q-Match.

\section{Implementation Details}\label{sec:implementationdetails}

\subsection{Initialisation}
The initial set of permutations $\mathcal{P}^\mathit{init}$ is computed using a descriptor-based similarity $\mathit{DS}_{\cal I J} \in \mathbb{R}^{n \times n }$ between all $n$ vertices of shape pairs $\cal I, J \in \mathcal{S}$. 
Specifically, $\mathit{DS}_{\cal I J}(u,v)$ contains the similarity (inner product) of the normalised heat-kernel-signatures (HKS)~\cite{bronstein2010scale} descriptors (which we extend by an additional dimension indicating whether a vertex lies on the left or right side of a shape) of vertex $u$ of $\cal I$ and vertex $v$ of $\cal J$. 
The left-right descriptors reduce left-right flips in the solution since the shape classes we consider are globally symmetric, which is not captured by the local HKS descriptors. 
This is standard practice in the shape-matching literature~\cite{Gao2021}. 
The solution of a linear assignment problem on $\mathit{DS}_{\cal I J}$ is then the initial $P^\mathit{init}_{\cal I J}$.

\subsection{Pre-Computing Trivial Cycles} \label{sec:kernelization} 
The matrix $\Tilde{W}$ consists of couplings (quadratic terms) and linear terms. 
Numerical experiments show that there often exist cycles with linear terms that dominate the corresponding coupling terms.  
This happens when a cycle is largely uncorrelated to the rest of the cycles in the current set of cycles. 
In this case, the decision for such a cycle can be made trivially. 

We now derive an inequality for how large the linear term has to be such that the couplings can be neglected in the optimisation problem. 
Consider the QUBO problem:
\begin{gather}
    \min_{\alpha \in\{0,1\}^k}  \alpha^T W \alpha+ b^T \alpha,
\end{gather}
where $\alpha$ are the decision variables, $b\in\mathbb{R}^k$ is a vector, and $W$ is a symmetric matrix with zeros on the diagonal representing the couplings. 
We can look at the terms that depend on $\alpha_q$ for fixed $q$ separately:
\begin{gather*}
    \alpha^T W \alpha = \alpha_q\sum_{i\neq q}W_{q,i} \alpha_i +  \alpha_q \sum_{i\neq q} \alpha_i W_{i,q}  \\ \nonumber + \sum_{j\neq q}(\sum_{i \neq q} \alpha_j  W_{j,i} \alpha_i + \alpha_j b_j) + \alpha_q b_q.
\end{gather*}
As $W$ is symmetric, $W_{i,q} = W_{q,i}$ holds and we can write
\begin{gather*}
    \alpha^T W \alpha = \alpha_q \left(b_q+ 2\sum_{i\neq q}W_{q,i} \alpha_i \right)  \\ \nonumber + \sum_{j\neq q}(\sum_{i \neq q} \alpha_j  W_{j,i} \alpha_i + \alpha_j b_j).
\end{gather*}
It follows that if:
\begin{equation}
    |b_q|\geq  \sum_{i\neq q} 2 |W_{q,i}|,
\end{equation}
then we can make the decision based on the sign of $b_q$: If $b_q$ is positive we do not choose the cycle as doing so would increase the energy.
This reduces the number of physical qubits required for the embedding.

\section{Minor Embeddings and Other QPU Experiments}\label{sec:qpu}

\subsection{QPU Processing and Annealing Time}

Optimising a single QUBO uses ${\sim}40 ms$ of total QPU processing time for 200 anneals. 
This is also called the \emph{QPU access time} \cite{Timing}. 
However, there are also several overheads that occur when solving QUBOs by accessing a D-Wave annealer via the cloud. 
For example, a latency when connecting to the D-Wave annealer and a post processing time. The QPU access time also includes a programming time.

These overheads can be orders of magnitude greater than the time taken by the actual annealing, which is very short since we use the default annealing schedule and the default annealing time of 20 $\mu s$. %

\subsection{Minor Embeddings}

As explained in the main paper, not all physical qubits on a real quantum processing unit (QPU) can be connected (coupled) with each other. 
Thus, a minor embedding of the logical-qubit graph (defined by non-zero entries of the QUBO matrix) into the physical-qubit graph (defined by the hardware) is required. 
This can lead to a chain of multiple physical qubits representing a single logical qubit. 

\begin{figure*}
    \centering
    \begin{subfigure}[b]{0.48\textwidth}
    \includegraphics[width = \textwidth]{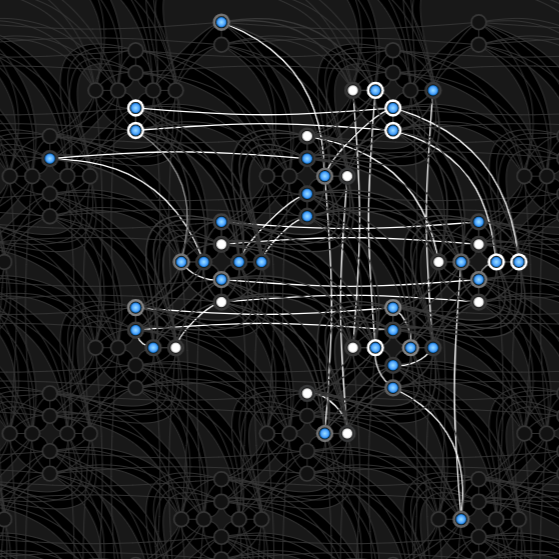}
    \end{subfigure}
    \begin{subfigure}[b]{0.48\textwidth}
    \includegraphics[width = \textwidth]{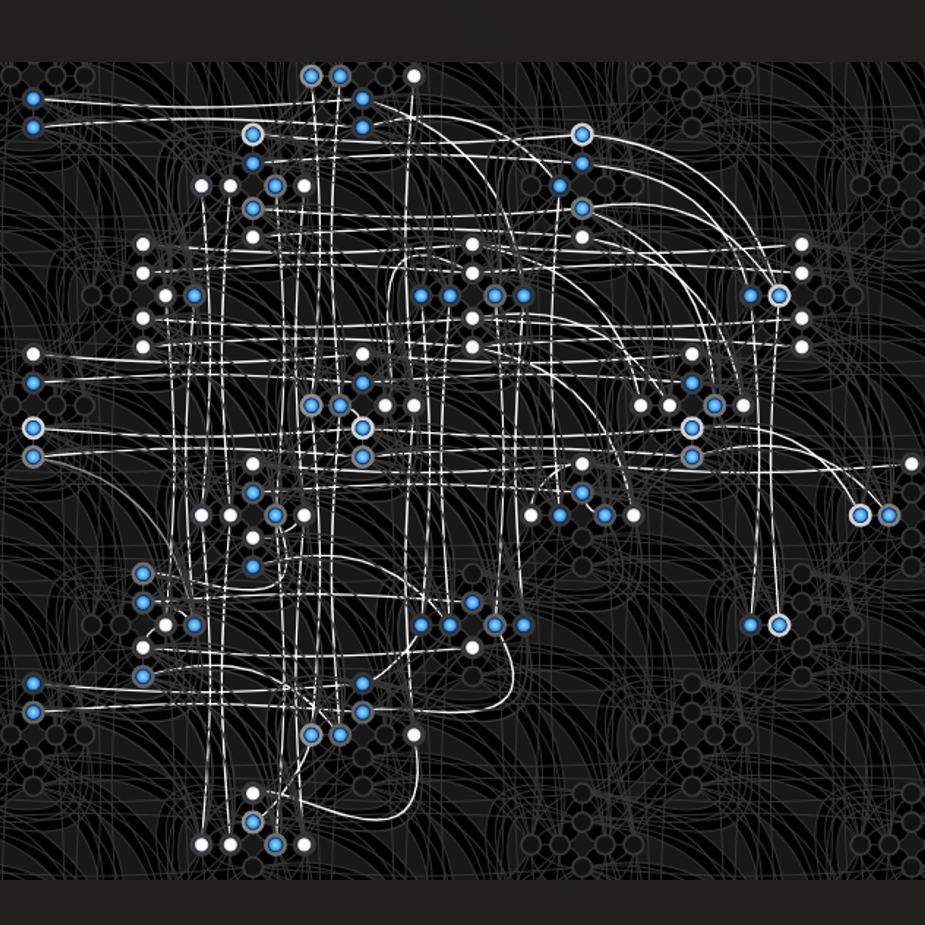}
    \end{subfigure}
    
    \begin{subfigure}[b]{0.48\textwidth}
    \includegraphics[width = \textwidth]{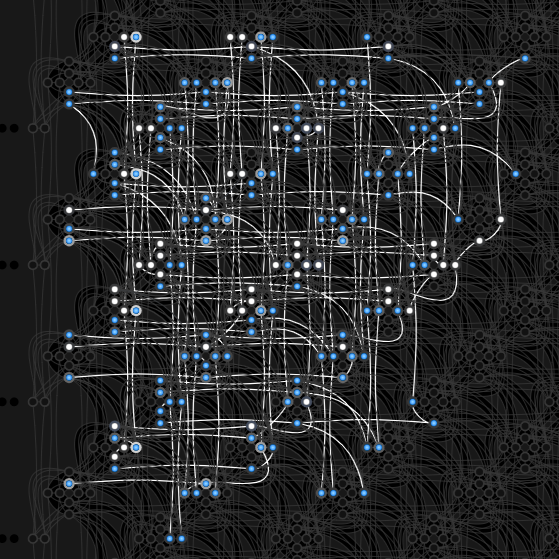}
    \end{subfigure}
    \begin{subfigure}[b]{0.48\textwidth}
    \includegraphics[width = \textwidth]{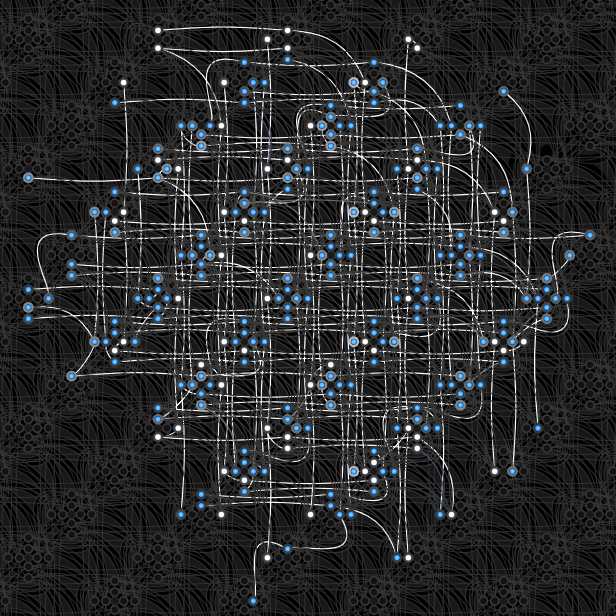}
    \end{subfigure}
    \caption{Visualisation of an example minor embedding on Pegasus (which we use in the main paper). %
    The visualisation is obtained via D-Wave Leap 2's problem inspector~\cite{DWave_Leap} for (upper left) 20, (upper right) 30, (lower left) 40, and (lower right) 50 worst vertices. 
    Each node depicts a physical qubit and the edges depict the chains of the minor embeddings.  
    }
    \label{fig:qpuembedding}
\end{figure*}

Our logical input graph is a clique. 
Due to the limited connectivity of current hardware, a clique cannot be directly embedded onto the physical annealer. 
We thus require a minor embedding, which is commonly computed using Cai \textit{et al.}'s method~\cite{cai2014practical}. 
Fig.~\ref{fig:qpuembedding} visualises an example minor embedding.
We note that, since our input graphs are cliques, a generalised embedding can be pre-computed and reused, not impacting the time complexity.

\begin{figure}
    \centering
    \resizebox{0.48\linewidth}{!}{
    \input{SupplimentFigures/PCK_Zephyr}}
    \resizebox{0.48\linewidth}{!}{
    \input{SupplimentFigures/PCK_10_Shapes_Zephyr}} 
    \caption{ 
    PCK curves for (left) a three-shape and (right) a ten-shape inter-class FAUST instance on both QPU architectures. %
    } 
    \label{fig:InterClassQPU} 
\end{figure} 

\begin{figure}
    \centering
    \resizebox{0.49\linewidth}{!}{
    \input{SupplimentFigures/qpuworstvertices}}
    \resizebox{0.49\linewidth}{!}{
    \input{SupplimentFigures/PCK_worst_vertices}}
    \caption{Quantitative results when using (solid) 20 and (dashed) 40 worst vertices on a three-shape FAUST instance. 
    We show (left) the energy evolution during optimisation and (right) the final PCK curves. 
    }
    \label{fig:Energy40}
\end{figure}

\begin{figure}[th]
    \centering
    \resizebox{0.48\linewidth}{!}{
    \input{SupplimentFigures/qubits}}
    \resizebox{0.48\linewidth}{!}{
    \input{SupplimentFigures/chains}}
    \caption{Structural changes of the minor embeddings when using more worst vertices. 
    We show (left) the number of physical qubits and (right) the average chain length, for both QPU topologies. %
    The number of \emph{logical} qubits equals the number of worst vertices. %
    }
    \label{fig:qubits_and_chains}
\end{figure}

\subsection{Minor Embeddings in Practice}

In this section, we investigate the empirical impact of minor embeddings on the solutions. 
The minor embeddings depend on the qubit topology, \textit{i.e.} the physical qubit connectivity pattern. 
Here, we show results on the D-Wave Advantage 4.1 with its Pegasus topology (used in the main paper) and also first results on a D-Wave Advantage2 prototype of the next-generation Zephyr topology, which has a higher connectivity than Pegasus. 

Fig.~\ref{fig:InterClassQPU} shows PCK curves on both topologies when using $m{=}2k{=}20$ worst vertices. 
Both architectures obtain similar results, although they are very slightly worse than SA. 
However, as discussed in the main paper, we find that the performance of CCuantuMM degrades significantly when using more than $20$ worst vertices with QA. 
Specifically, Fig.~\ref{fig:Energy40} shows that the quality of the matchings worsens when using $40$ worst vertices on both topologies, with Zephyr obtaining slightly better results. 
Still, in both cases, the quality is worse than the matching quality obtained by SA. 
Only SA shows the desired behaviour of improving when more worst vertices are used. 

The cause for these results lies with the structure of the minor embeddings and not the plain number of physical qubits. 
Fig.~\ref{fig:qubits_and_chains} shows how the structural properties of the minor embeddings evolve as the number of worst vertices increases. 
For $20$ worst vertices, they are similar. 
However, for $40$ worst vertices, Zephyr uses fewer physical qubits and smaller chains, which explains the very slight performance advantage in Fig.~\ref{fig:Energy40}. %
Physical qubits in a chain representing a single logical qubit are less likely to all anneal to the same value the longer the chain is. 
Longer chains become unstable and hence inconsistent, leading to inferior solution quality.

\section{Further Results and Ablations}\label{sec:ablations_and_analysis}

\subsection{Additional Qualitative Results} \label{sec:qualitative}

\begin{figure*}[!ht]
   \centering
    \resizebox{\linewidth}{!}{
    \begin{tabular}{c|cccccc}
    
    \raisebox{-0.5\height}{\includegraphics[width=.18\linewidth]{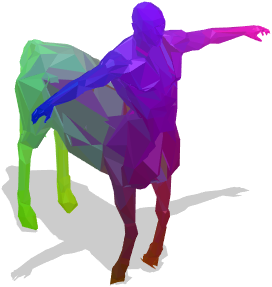}}
     &
    \raisebox{-0.5\height}{\includegraphics[width=.11\linewidth]{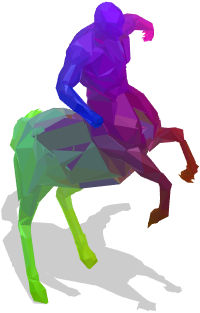}} &
    \raisebox{-0.5\height}{\includegraphics[width=.13\linewidth]{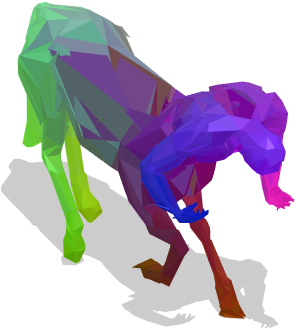}} & 
    \raisebox{-0.5\height}{\includegraphics[width=.14\linewidth]{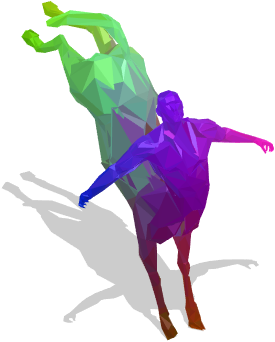}} &
    \raisebox{-0.5\height}{\includegraphics[width=.13\linewidth]{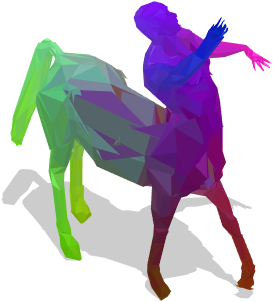}} & 
    \raisebox{-0.5\height}{\includegraphics[width=.13\linewidth]{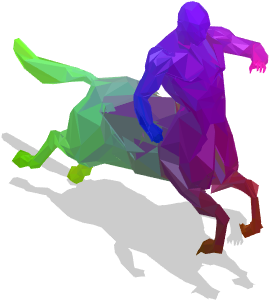}} &
    \rotatebox[origin=c]{270}{Ours} \\

    \raisebox{\height}{Source}
    &
    \raisebox{-0.5\height}{\includegraphics[width=.11\linewidth]{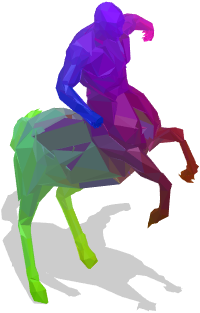}} &
    \raisebox{-0.5\height}{\includegraphics[width=.13\linewidth]{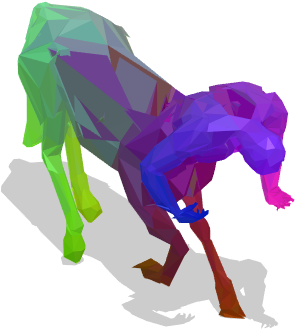}} &
    \raisebox{-0.5\height}{\includegraphics[width=.14\linewidth]{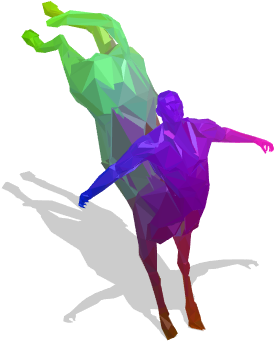}} &
    \raisebox{-0.5\height}{\includegraphics[width=.13\linewidth]{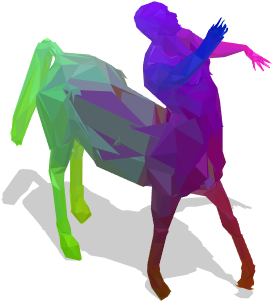}} &
    \raisebox{-0.5\height}{\includegraphics[width=.13\linewidth]{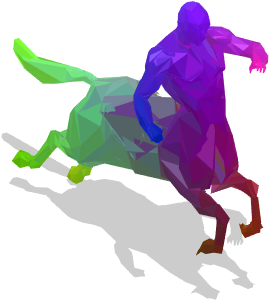}} &
    \rotatebox[origin=c]{270}{IsoMuSh} \\
    
    \ 
    &
    \raisebox{-0.5\height}{\includegraphics[width=.11\linewidth]{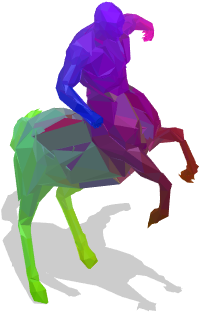}} &
    \raisebox{-0.5\height}{\includegraphics[width=.13\linewidth]{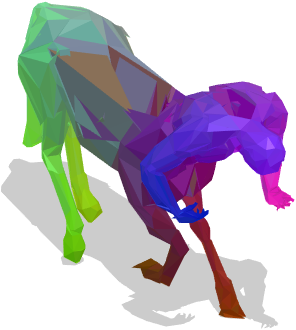}} &
    \raisebox{-0.5\height}{\includegraphics[width=.14\linewidth]{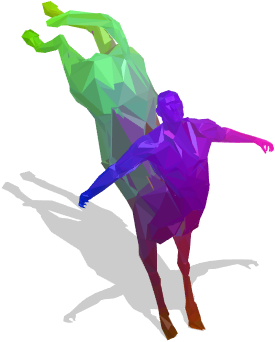}} &
    \raisebox{-0.5\height}{\includegraphics[width=.13\linewidth]{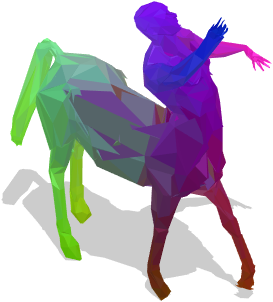}} &
    \raisebox{-0.5\height}{\includegraphics[width=.13\linewidth]{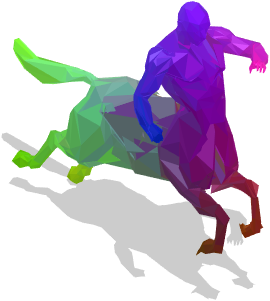}} &
    \rotatebox[origin=c]{270}{ZoomOut} \\
    
    \ 
    &
    \raisebox{-0.5\height}{\includegraphics[width=.11\linewidth]{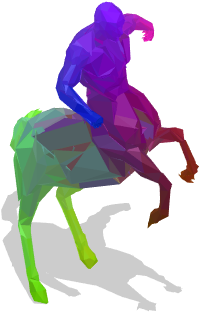}} &
    \raisebox{-0.5\height}{\includegraphics[width=.13\linewidth]{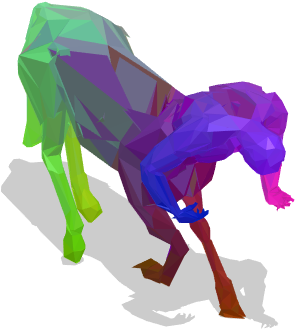}} &
    \raisebox{-0.5\height}{\includegraphics[width=.14\linewidth]{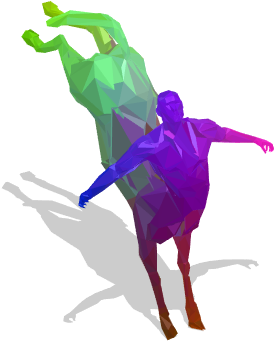}} &
    \raisebox{-0.5\height}{\includegraphics[width=.13\linewidth]{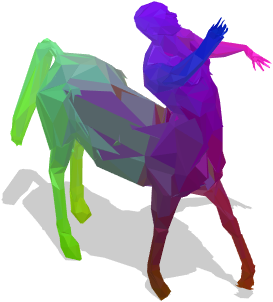}} &
    \raisebox{-0.5\height}{\includegraphics[width=.13\linewidth]{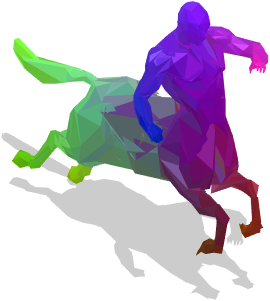}} &
    \rotatebox[origin=c]{270}{Q-MatchV2-nc} \\
    
    \ 
    &
    \raisebox{-0.5\height}{\includegraphics[width=.11\linewidth]{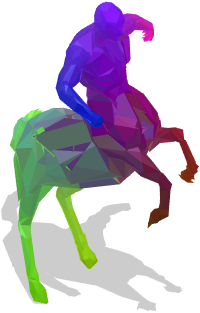}} &
    \raisebox{-0.5\height}{\includegraphics[width=.13\linewidth]{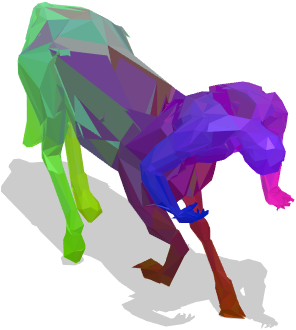}} &
    \raisebox{-0.5\height}{\includegraphics[width=.14\linewidth]{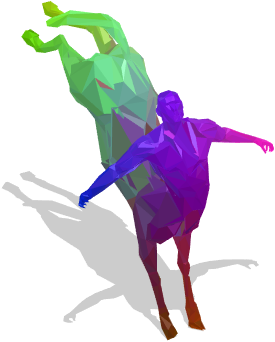}} &
    \raisebox{-0.5\height}{\includegraphics[width=.13\linewidth]{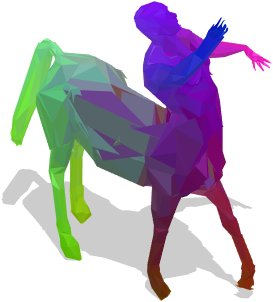}} &
    \raisebox{-0.5\height}{\includegraphics[width=.13\linewidth]{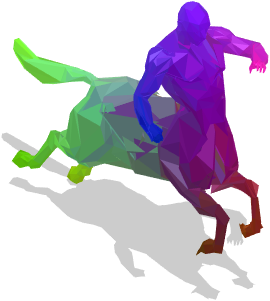}} &
    \rotatebox[origin=c]{270}{Q-MatchV2-cc} \\
    
    \end{tabular}
    }

    \caption{Qualitative results on the TOSCA  \cite{bronstein2008numerical} centaur class. 
    We colour a source shape %
    and transfer this colouring to target shapes via the matches estimated by our method and competitors.%
    } 
    \label{fig:tosca_centaur} 
\end{figure*}

\begin{figure*}[!ht]
   \centering
    \resizebox{\linewidth}{!}{
    \begin{tabular}{c|ccccccc}
    
    \raisebox{-0.5\height}{\includegraphics[width=.17\linewidth]{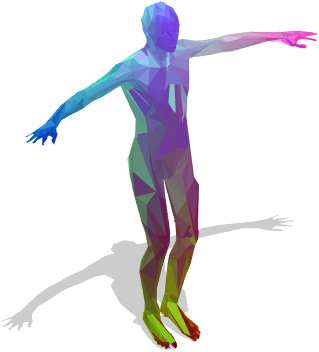}}
     &
    \raisebox{-0.5\height}{\includegraphics[width=.06\linewidth]{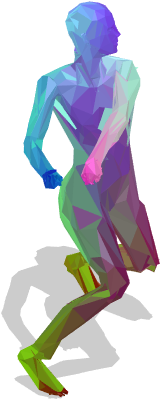}} &
    \raisebox{-0.5\height}{\includegraphics[width=.11\linewidth]{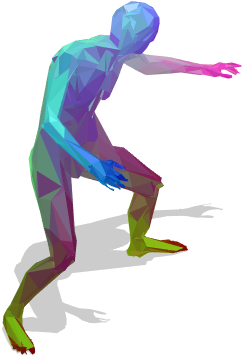}} & 
    \raisebox{-0.5\height}{\includegraphics[width=.09\linewidth]{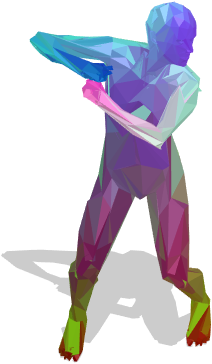}} &
    \raisebox{-0.5\height}{\includegraphics[width=.06\linewidth]{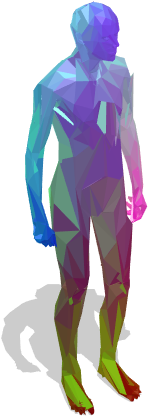}} & 
    \raisebox{-0.5\height}{\includegraphics[width=.15\linewidth]{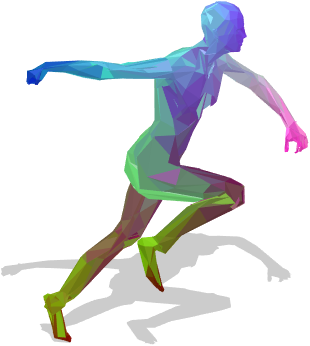}} &
    \raisebox{-0.5\height}{\includegraphics[width=.15\linewidth]{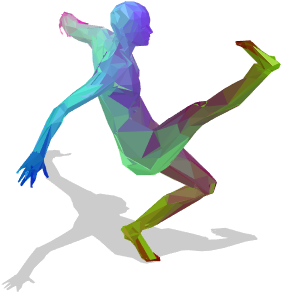}} &
    \rotatebox[origin=c]{270}{Ours} \\

    \raisebox{\height}{Source}
    &
    \raisebox{-0.5\height}{\includegraphics[width=.06\linewidth]{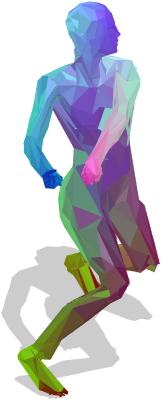}} &
    \raisebox{-0.5\height}{\includegraphics[width=.11\linewidth]{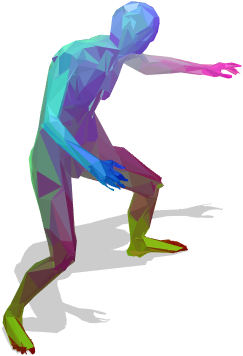}} &
    \raisebox{-0.5\height}{\includegraphics[width=.09\linewidth]{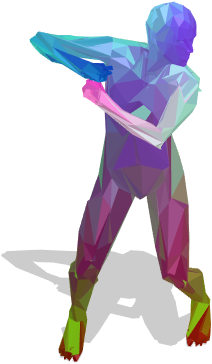}} &
    \raisebox{-0.5\height}{\includegraphics[width=.06\linewidth]{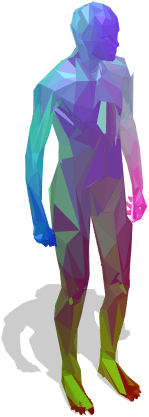}} &
    \raisebox{-0.5\height}{\includegraphics[width=.15\linewidth]{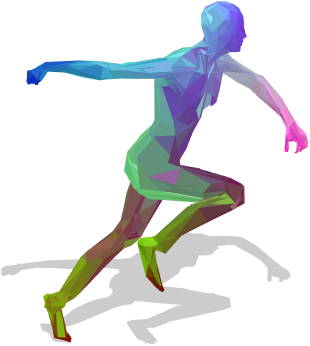}} &
    \raisebox{-0.5\height}{\includegraphics[width=.15\linewidth]{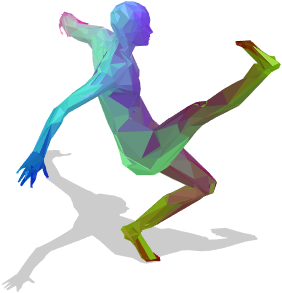}} &
    \rotatebox[origin=c]{270}{IsoMuSh} \\
    
    \ 
    &
    \raisebox{-0.5\height}{\includegraphics[width=.06\linewidth]{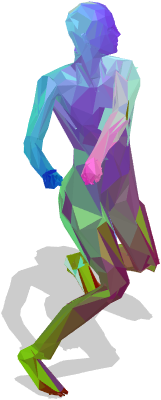}} &
    \raisebox{-0.5\height}{\includegraphics[width=.11\linewidth]{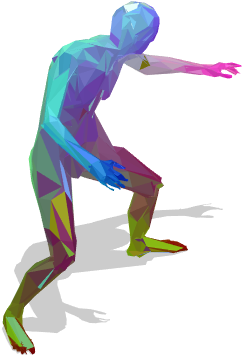}} &
    \raisebox{-0.5\height}{\includegraphics[width=.09\linewidth]{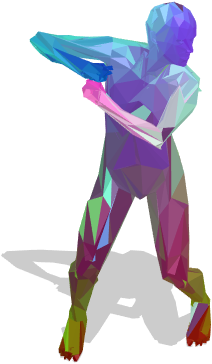}} &
    \raisebox{-0.5\height}{\includegraphics[width=.06\linewidth]{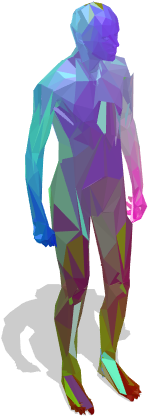}} &
    \raisebox{-0.5\height}{\includegraphics[width=.15\linewidth]{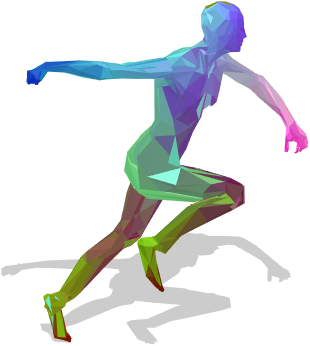}} &
    \raisebox{-0.5\height}{\includegraphics[width=.15\linewidth]{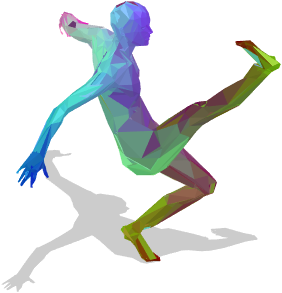}} &
    \rotatebox[origin=c]{270}{ZoomOut} \\
    
    \ 
    &
    \raisebox{-0.5\height}{\includegraphics[width=.06\linewidth]{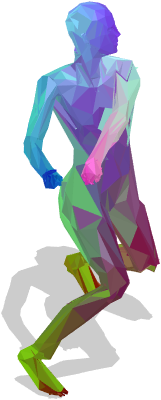}} &
    \raisebox{-0.5\height}{\includegraphics[width=.11\linewidth]{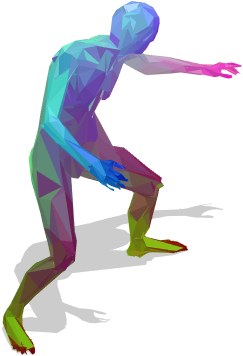}} &
    \raisebox{-0.5\height}{\includegraphics[width=.09\linewidth]{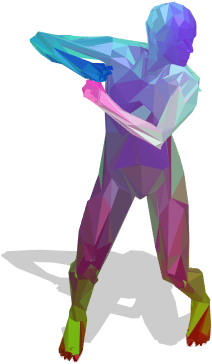}} &
    \raisebox{-0.5\height}{\includegraphics[width=.06\linewidth]{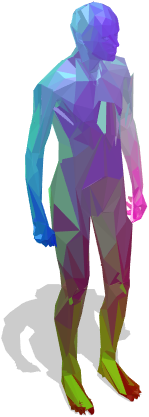}} &
    \raisebox{-0.5\height}{\includegraphics[width=.15\linewidth]{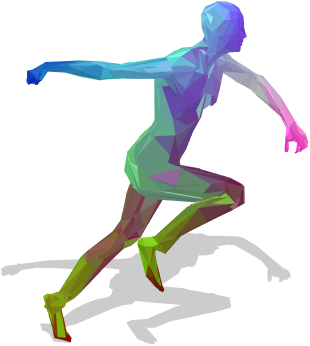}} &
    \raisebox{-0.5\height}{\includegraphics[width=.15\linewidth]{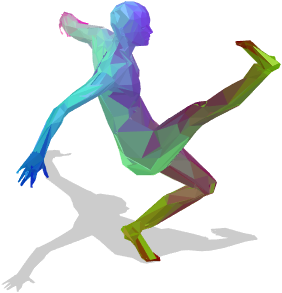}} &
    \rotatebox[origin=c]{270}{Q-MatchV2-nc} \\
    
    \ 
    &
    \raisebox{-0.5\height}{\includegraphics[width=.06\linewidth]{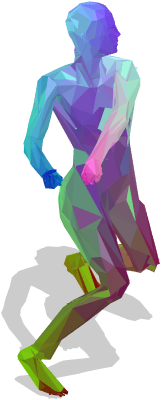}} &
    \raisebox{-0.5\height}{\includegraphics[width=.11\linewidth]{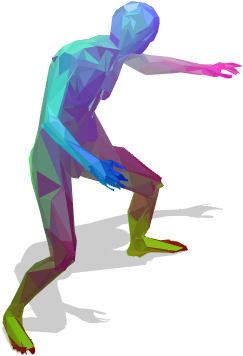}} &
    \raisebox{-0.5\height}{\includegraphics[width=.09\linewidth]{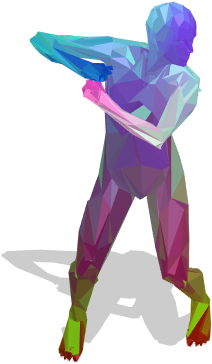}} &
    \raisebox{-0.5\height}{\includegraphics[width=.06\linewidth]{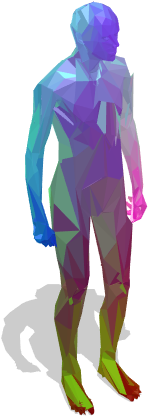}} &
    \raisebox{-0.5\height}{\includegraphics[width=.15\linewidth]{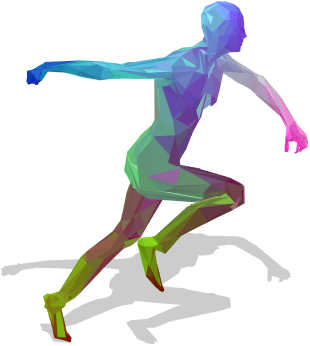}} &
    \raisebox{-0.5\height}{\includegraphics[width=.15\linewidth]{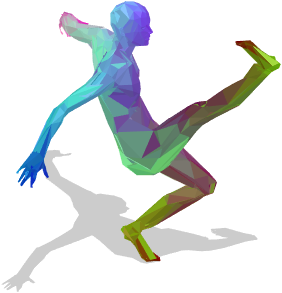}} &
    \rotatebox[origin=c]{270}{Q-MatchV2-cc} \\
    
    \end{tabular}
    }

    \caption{Qualitative results on the TOSCA  \cite{bronstein2008numerical} David class. 
    We colour a source shape %
    and transfer this colouring to target shapes via the matches estimated by our method and competitors. %
    } 
    \label{fig:tosca_david} 
\end{figure*}

\begin{figure*}[!ht]
   \centering
    \resizebox{\linewidth}{!}{
    \begin{tabular}{c|cccccc}
    
    \raisebox{-0.5\height}{\includegraphics[width=.1\linewidth]{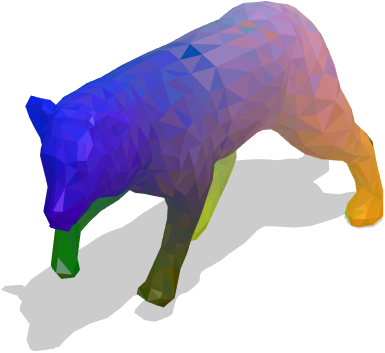}}
     &
    \raisebox{-0.5\height}{\includegraphics[width=.09\linewidth]{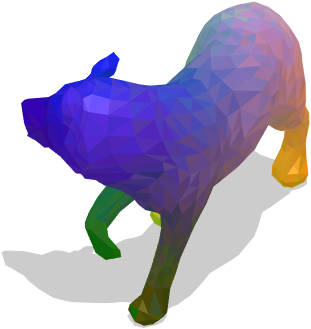}} &
    \raisebox{-0.5\height}{\includegraphics[width=.12\linewidth]{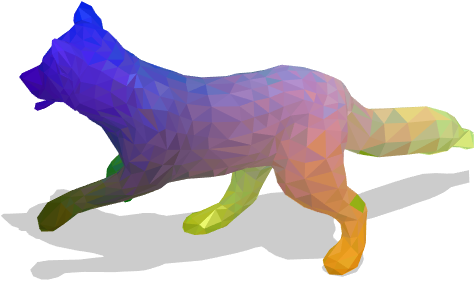}} & 
    \raisebox{-0.5\height}{\includegraphics[width=.14\linewidth]{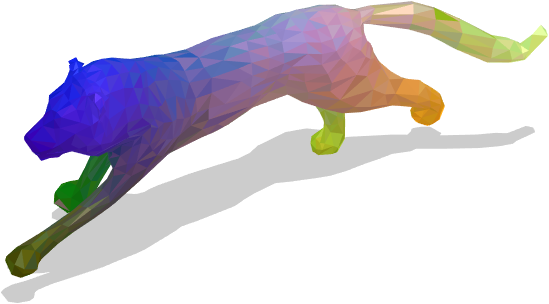}} &
    \raisebox{-0.5\height}{\includegraphics[width=.1\linewidth]{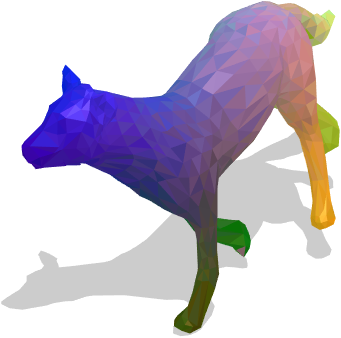}} & 
    \raisebox{-0.5\height}{\includegraphics[width=.1\linewidth]{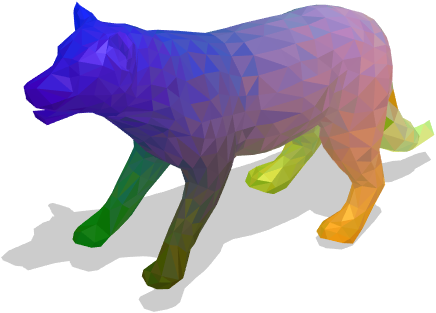}} &
    \rotatebox[origin=c]{270}{Ours} \\

    \raisebox{\height}{Source}
    &
    \raisebox{-0.5\height}{\includegraphics[width=.09\linewidth]{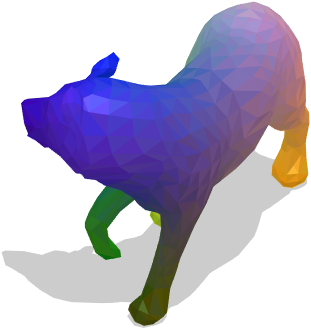}} &
    \raisebox{-0.5\height}{\includegraphics[width=.12\linewidth]{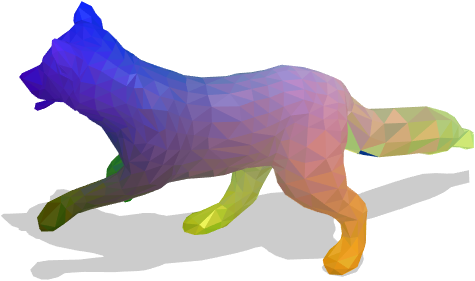}} &
    \raisebox{-0.5\height}{\includegraphics[width=.14\linewidth]{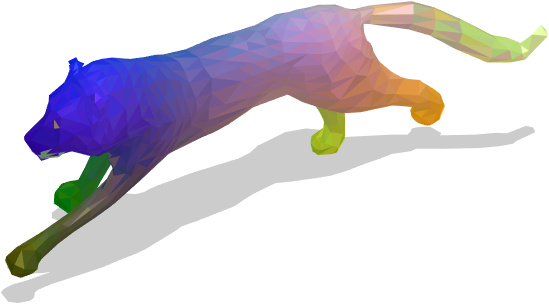}} &
    \raisebox{-0.5\height}{\includegraphics[width=.1\linewidth]{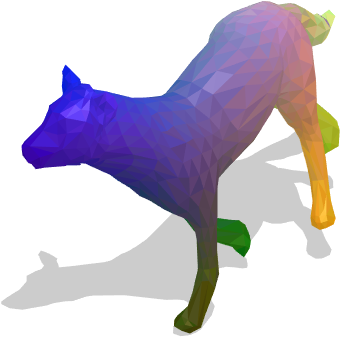}} &
    \raisebox{-0.5\height}{\includegraphics[width=.1\linewidth]{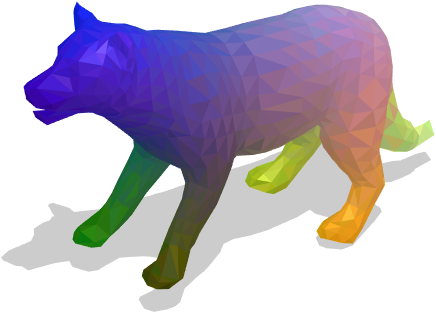}} &
    \rotatebox[origin=c]{270}{IsoMuSh} \\
    
    \ 
    &
    \raisebox{-0.5\height}{\includegraphics[width=.09\linewidth]{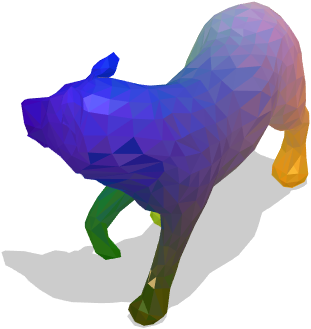}} &
    \raisebox{-0.5\height}{\includegraphics[width=.12\linewidth]{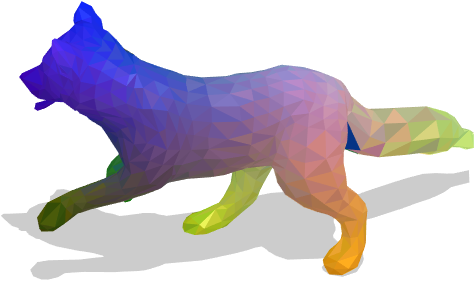}} &
    \raisebox{-0.5\height}{\includegraphics[width=.14\linewidth]{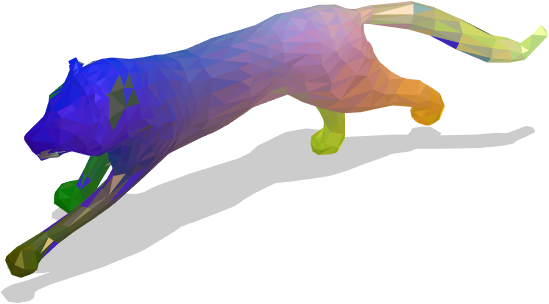}} &
    \raisebox{-0.5\height}{\includegraphics[width=.1\linewidth]{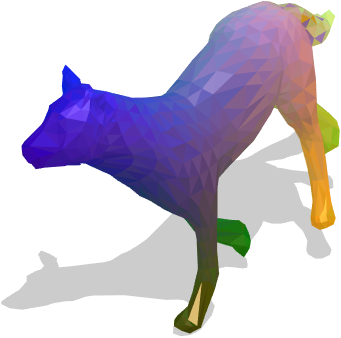}} &
    \raisebox{-0.5\height}{\includegraphics[width=.1\linewidth]{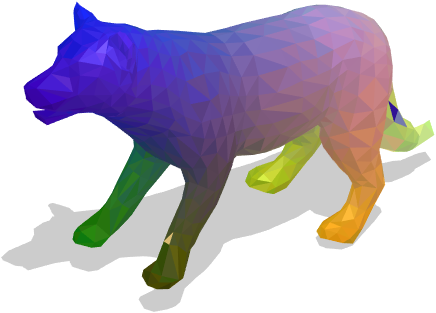}} &
    \rotatebox[origin=c]{270}{ZoomOut} \\
    
    \ 
    &
    \raisebox{-0.5\height}{\includegraphics[width=.09\linewidth]{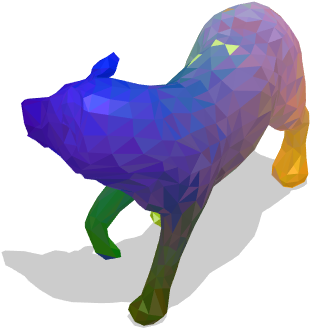}} &
    \raisebox{-0.5\height}{\includegraphics[width=.12\linewidth]{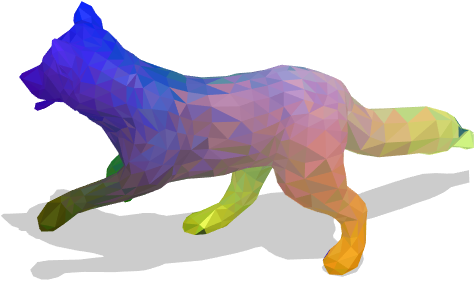}} &
    \raisebox{-0.5\height}{\includegraphics[width=.14\linewidth]{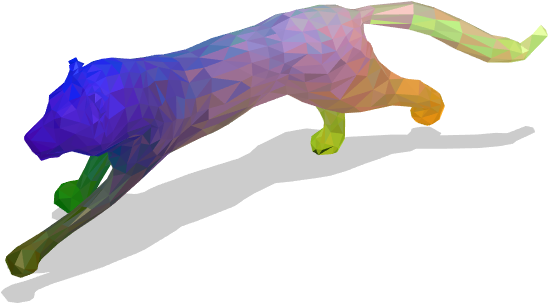}} &
    \raisebox{-0.5\height}{\includegraphics[width=.1\linewidth]{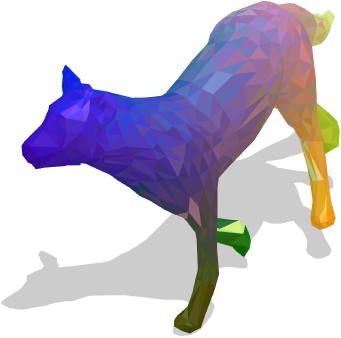}} &
    \raisebox{-0.5\height}{\includegraphics[width=.1\linewidth]{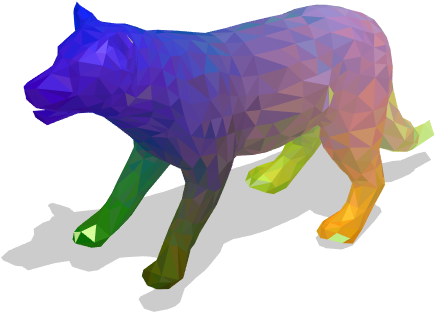}} &
    \rotatebox[origin=c]{270}{Q-MatchV2-nc} \\
    \ 
    &
    \raisebox{-0.5\height}{\includegraphics[width=.09\linewidth]{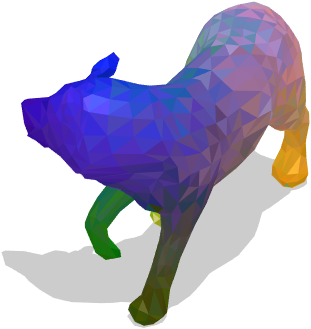}} &
    \raisebox{-0.5\height}{\includegraphics[width=.12\linewidth]{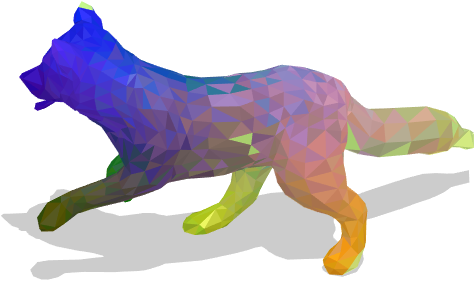}} &
    \raisebox{-0.5\height}{\includegraphics[width=.14\linewidth]{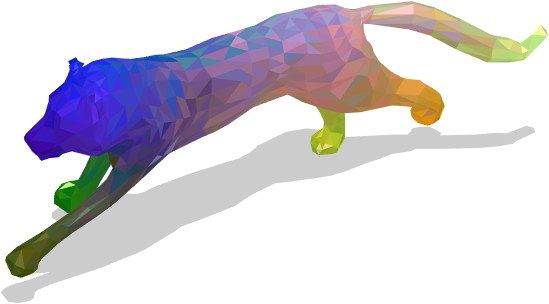}} &
    \raisebox{-0.5\height}{\includegraphics[width=.1\linewidth]{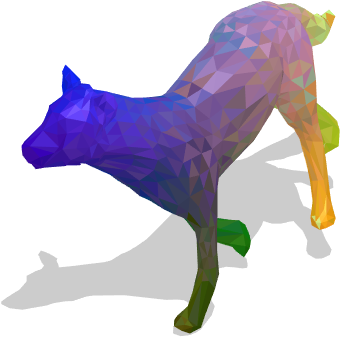}} &
    \raisebox{-0.5\height}{\includegraphics[width=.1\linewidth]{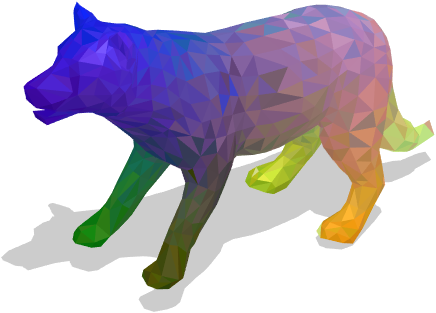}} &
    \rotatebox[origin=c]{270}{Q-MatchV2-cc} \\
    
    \end{tabular}
    }

    \caption{Qualitative results on a subset of the SMAL dog class.
    We colour a source shape %
    and transfer this colouring to target shapes via the matches estimated by our method and competitors. %
    } 
    \label{fig:smal_dog} 
\end{figure*}

Fig.~\ref{fig:tosca_centaur}, Fig.~\ref{fig:tosca_david}, and Fig.~\ref{fig:smal_dog} provide additional qualitative examples of matchings on TOSCA and SMAL calculated with our method and the competitors. %

\subsection{Variation Within a Dataset}

Within a dataset, some instances have more difficult deformations and are inherently harder to match than easier instances, independent of the method employed. 
We investigate the extent of this variation by taking a closer look at TOSCA. 
We observe a significant variation of PCK curves across different classes in Fig.~\ref{fig:SMAL_Variation} and of AUC in Tab.~\ref{Table:SMAL_variation}.

\begin{figure}[!h]%
    \centering
    \resizebox{\linewidth}{!}{
    \input{SupplimentFigures/PCK_TOSCA_class.tex}}
    \caption{PCK for seven different classes of TOSCA.
    We plot our method with solid lines and IsoMuSh~\cite{Gao2021} with dashed lines. 
    }
    \label{fig:SMAL_Variation}
\end{figure}

\begin{table}[h]
    \centering
    \resizebox{\linewidth}{!}{
    \begin{tabular}{|c|ccccccc|}
    \hline 
    & Dog & Cat & Horse & Michael & Victoria & David & Centaur \\ \hline
    Ours    & 0.957 & 0.917          & \textbf{0.990} & \textbf{0.992} & 0.912 & 0.989 & 0.983 \\ 
    IsoMush & \textbf{0.959} & \textbf{0.959} & 0.981         & 0.939 & \textbf{0.991} & \textbf{0.996} & \textbf{0.984} \\ \hline
\end{tabular}
}
    \caption{AUC across seven different classes of TOSCA. 
    }
    \label{Table:SMAL_variation}
\end{table}

\subsection{Influence of Descriptors}

We ablate the need for left-right indicators when using HKS descriptors for initialisation. %
Fig.~\ref{fig:noLR} contains results on the cat class of TOSCA.
Without left-right indicators, we observe flips in the matchings on TOSCA and partial flips on FAUST for inter-class instances. %
This is expected since both our method and IsoMuSh exploit intrinsic properties of the shapes, which are invariant to such symmetric flips. 

\begin{figure}[h]
    \centering
    \resizebox{0.5\linewidth}{!}{
    \input{SupplimentFigures/LRdesc.tex}
    }
    \caption{PCK with and without left-right descriptors.
    We plot our method with solid lines and IsoMuSh~\cite{Gao2021} with dashed lines. }
    \label{fig:noLR}
\end{figure}

\subsection{Noise Perturbation}

\begin{figure}[h]
    \centering
    \includegraphics[width=.3\linewidth]{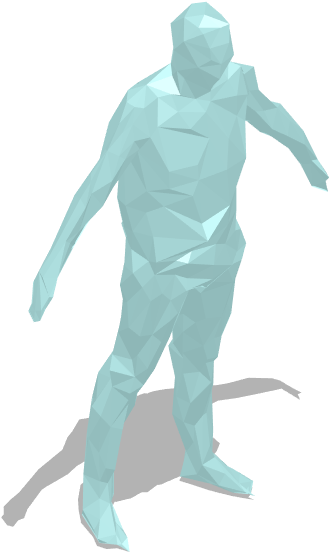}
    \quad
    \includegraphics[width=.3\linewidth]{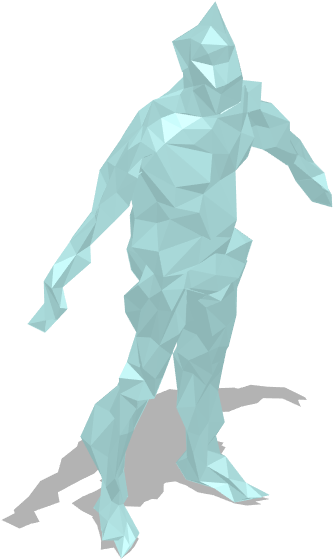}
    \caption{Example of perturbing the geometry on FAUST, with the noise variance set to $0.01$ (left) and $0.02$ (right).}
    \label{fig:noisy}
\end{figure}

We investigate the robustness of our approach to noise. 
To that end, we analyse the effect of adding synthetic perturbations to the shapes. 
Specifically, for each FAUST mesh, we add a Gaussian-distributed offset along the vertex normal to each vertex position.
This ensures the meshed structure of the shape does not change. 
Fig.~\ref{fig:noisy} visualises the amount of noise we experiment with.
Fig.~\ref{fig:NoiseVariation} shows that our method is significantly more robust to noise than IsoMuSh. 

\begin{figure}[h]%
    \centering
    \resizebox{0.5\linewidth}{!}{
    \input{SupplimentFigures/pckNoise}}
    \caption{PCK for different amounts of perturbation (Gaussian noise with variance $\sigma^2$).  
    We plot our method with solid lines and IsoMuSh~\cite{Gao2021} with dashed lines. 
    We report results on a class of FAUST. 
    }
    \label{fig:NoiseVariation}
\end{figure}

\section{Related Quantum Computer Vision  Works}\label{sec:related_work_appendix} 
Several quantum methods tackle alignment tasks, as  Tab.~\ref{tab:related_work} shows. 
However, only Q-Match is relevant for comparisons. 
Several works only consider two point clouds and cannot handle the multi-matching setting of our work, do not consider fully non-rigid transformations, or only operate on point clouds, not meshes. %
QGM~\cite{SeelbachBenkner2020} only considers two graphs with at most four vertices. 
Q-Sync~\cite{QuantumSync2021} similarly only works on at most five vertices, far fewer than what our method can handle.

\begin{sidewaysfigure*}[b]
\begin{equation} \label{eq:fullPXZ}
\resizebox{\linewidth}{!}{$
\begin{split} 
    E_{\cal X Z} (P_{ \cal X Z}(\alpha,\beta)) &= E_{\cal X Z}\left( %
    (P_{\cal XY} + \sum_{i=1}^k \alpha_i C_i) \cdot (P_{\cal YZ}+\sum_{j=1}^k \beta_j \tilde{C}_j),%
    (P_{\cal XY} + \sum_{q=1}^k \alpha_q C_q) \cdot (P_{\cal YZ} + \sum_{l=1}^k \beta_l\tilde{C}_l) %
    \right) \\ \\ 
    &= E_{\cal X Z} \left(P_{\cal X Y} P_{\cal Y Z}, (P_{\cal XY} + \sum_{q=1}^k \alpha_q C_q) \cdot (P_{\cal YZ} + \sum_{l=1}^k \beta_l \tilde{C}_l\right) %
    + E_{\cal X Z} \left(\sum_{j=1}^k \beta_j P_{\cal X Y}  \tilde{C} _j  ),(P_{\cal XY} + \sum_{q=1}^k \alpha_q C_q) \cdot (P_{\cal YZ} + \sum_{l=1}^k \beta_l \tilde{C}_l)\right) 
    \\ 
    &+ E_{\cal X Z} \left( \sum_{i=1}^k \alpha_i C_i P_{\cal YZ}, (P_{\cal XY} + \sum_{q=1}^k \alpha_q C_q) \cdot (P_{\cal YZ} + \sum_{l=1}^k \beta_l \tilde{C}_l)\right) %
    + E_{\cal X Z} \left(\sum_{i=1}^k \sum_{j=1}^k \alpha_i \beta_j C_i  \tilde{C}_j,(P_{\cal XY} + \sum_{q=1}^k \alpha_q C_q) \cdot (P_{\cal YZ} + \sum_{l=1}^k \beta_l \tilde{C}_l)\right)
    \\  \\ 
    &= E_{\cal X Z}(P_{\cal X Y} P_{\cal Y Z},P_{\cal X Y} P_{\cal Y Z}) %
    + \sum_{l=1}^k \beta_l E_{\cal X Z}(P_{\cal X Y} P_{\cal Y Z}, P_{\cal X Y} \tilde{C}_l )%
    + \sum_{q=1}^k \alpha_q E_{\cal X Z}(P_{\cal X Y} P_{\cal Y Z},  C_q P_{\cal YZ} )%
    \\
    & + \sum_{q=1}^k \sum_{l=1}^k \alpha_q  \beta_l E_{\cal X Z}(P_{\cal X Y} P_{\cal Y Z},  C_q \tilde{C}_l )%
    + \sum_{j=1}^k \beta_j E_{\cal X Z}( P_{\cal X Y} \tilde{C}_j, P_{\cal X Y} P_{\cal Y Z}) %
    + \sum_{j=1}^k \sum_{l=1}^k \beta_j \beta_l E_{\cal X Z}(P_{\cal X Y}   \tilde{C}_j, P_{\cal X Y} \tilde{C}_l) %
    \\
    &+ \sum_{j=1}^k \sum_{q=1}^k \beta_j \alpha_q E_{\cal X Z}(P_{\cal X Y} \tilde{C}_j,  C_q P_{\cal YZ}) %
    + \sum_{j=1}^k \sum_{q=1}^k \sum_{l=1}^k \beta_j \alpha_q \beta_l  E_{\cal X Z}( P_{\cal X Y}  \tilde{C}_j, C_q \tilde{C}_l)
    + \sum_{i=1}^k \alpha_i E_{\cal X Z}(C_i P_{\cal YZ}, P_{\cal X Y} P_{\cal Y Z}) %
    \\  %
    &+ \sum_{i=1}^k \sum_{l=1}^k \alpha_i \beta_l E_{\cal X Z}(C_i P_{\cal YZ}, P_{\cal X Y} \tilde{C}_l) %
    + \sum_{i=1}^k \sum_{q=1}^k \alpha_i \alpha_q E_{\cal X Z}(C_i P_{\cal YZ},  C_q P_{\cal YZ}) %
    + \sum_{i=1}^k \sum_{q=1}^k \sum_{l=1}^k \alpha_i \alpha_q \beta_l  E_{\cal X Z}(C_i P_{\cal YZ}, C_q \tilde{C}_l)
    \\  %
    &
    + \sum_{i=1}^k \sum_{j=1}^k \alpha_i \beta_j E_{\cal X Z}(C_i \tilde{C}_j, P_{\cal X Y} P_{\cal Y Z}) %
    + \sum_{i=1}^k \sum_{j=1}^k \sum_{l=1}^k \alpha_i \beta_j \beta_l E_{\cal X Z}(C_i \tilde{C}_j, P_{\cal X Y} \tilde{C}_l)%
    + \sum_{i=1}^k \sum_{j=1}^k \sum_{q=1}^k \alpha_i \beta_j \alpha_q E_{\cal X Z}(C_i\tilde{C}_j,  C_q P_{\cal YZ}) %
    \\
    & + \sum_{i=1}^k \sum_{j=1}^k \sum_{q=1}^k \sum_{l=1}^k \alpha_i \beta_j \alpha_q \beta_l  E_{\cal X Z}(C_i \tilde{C}_j, C_q \tilde{C}_l)%
    \end{split}
$}
\end{equation}
\caption{We expand the third term of (7) from the main paper, which yields higher-order terms (highlighted in red). 
When the summands constituting these terms are truly cubic or bi-quadratic, we assume them to be 0, which results in the QUBO (8). 
} 
\label{fig:higher_order_terms} 
\end{sidewaysfigure*}

\begin{sidewaystable*}[!htbp]
    \centering
    \resizebox{\textwidth}{!}{ 
    \begin{tabular}{ccccccccc}\hline  Method & Problem & Transformation & Input Type  & \# Inputs  & \# Points  & \begin{tabular}{@{}c@{}}\# Qubits\\(per sweep)\end{tabular} & QPU & Iterative \\\hline\hline
    QA, CVPR 2020 \cite{golyanik2020quantum}        &  TE, PSR  & A/R    & point clouds          & $2$               & ${\leq}5k^{\star}$      & ${\leq}140$     & 2000Q &                \\\hline
    IQT, CVPR 2022 \cite{Meli_2022_CVPR}            & TE  & R          & point clouds          & $2$               & ${\leq}1.5k^{\star}$      & ${\leq}10$     & 2000Q &   \checkmark  \\\hline
    qKC, ICASSP 2022 \cite{NoormandipourWang2022}  & PSR   & R         & point clouds          & $2$               & ${\leq}2k^{\star}$      & $4${-}$6$       & Rigetti \cite{Rigetti} &   \checkmark  \\\hline
    QGM, 3DV 2020 \cite{SeelbachBenkner2020}        & GM   & F         & graphs                & $2$               & ${\leq}4$               & ${\leq}50$               & 2000Q &               \\\hline
    QSync, CVPR 2021 \cite{QuantumSync2021}         & PS, GM & F       & perm.~matrices  & ${\leq}5$         & ${\leq}5$               & $\leq1.5k$      & Adv1.1&               \\\hline
    QuMoSeg, ECCV 2022 \cite{Arrigoni2022}         & MS & F       & segm.~matrices  & ${\leq}9$         & ${\leq}200$               & $\leq250$      & Adv\{1.1;4.1\}&               \\\hline
    Q-Match, ICCV 2021 \cite{SeelbachBenkner2021}   & MA  & R/NR          & meshes                & $2$               & ${\leq}500$             & ${\leq}200$                & Adv4.1 &   \checkmark  \\\hline
    CCuantuMM (Ours)                                   & MA & R/NR           & meshes                & ${\leq}100$        & ${\leq}1k$           &     ${\leq}50$        & \begin{tabular}{@{}c@{}}Adv4.1,\\Adv2 prototype\end{tabular}  &   \checkmark  \\\hline
    \end{tabular}
    }
    \caption{Overview of related quantum methods for alignment tasks on point sets, graphs and meshes. ``$^\star$'': according to the experiments reported in the paper; the methods can also process larger point clouds. Key: ``TE'': transformation estimation; ``PSR'': point set alignment; ``GM'': graph matching; ``PS'': permutation synchronisation; ``MS'': motion segmentation; ``MA'': mesh alignment; ``A'': affine transformation; ``R'': rigid transformation; ``NR'': non-rigid deformations; ``F'': the method operates on features extracted in a pre-processing step (and can support both rigid and non-rigid transformations). 
    Note that only Q-Match \cite{SeelbachBenkner2021} can be extended and applied to our data. 
    } 
    \label{tab:related_work} 
\end{sidewaystable*}

\end{document}

%% file: sections/0_abstract.tex
Jointly matching multiple, non-rigidly deformed 3D shapes is a challenging, $\mathcal{NP}$-hard problem. 
A perfect matching is necessarily cycle-consistent: Following the pairwise point correspondences along several shapes must end up at the starting vertex of the original shape. 
Unfortunately, existing quantum shape-matching methods do not support multiple shapes and even less cycle consistency. 
This paper addresses the open challenges and introduces the first  quantum-hybrid approach for 3D shape multi-matching; in addition, it is also  cycle-consistent. 
Its iterative formulation is admissible to modern adiabatic quantum hardware and scales linearly with the total number of input shapes. 
Both these characteristics are achieved by reducing the $N$-shape case to a sequence of three-shape matchings, the derivation of which is our main technical contribution. 
Thanks to quantum annealing, high-quality solutions with low energy are retrieved for the intermediate $\mathcal{NP}$-hard objectives. 
On benchmark datasets, the proposed approach significantly outperforms extensions to multi-shape matching of a previous quantum-hybrid two-shape matching method and is on-par with classical multi-matching methods. 
Our source code is available at \url{4dqv.mpi-inf.mpg.de/CCuantuMM/}.

%% file: sections/1_introduction.tex
\section{Introduction}\label{sec:intro}

Recently, there has been a growing interest in applying quantum computers in computer vision  \cite{QuantumSync2021,golyanik2020quantum,Meli_2022_CVPR}. 
Such quantum computer vision methods rely on quantum annealing (QA) that allows to solve $\mathcal{NP}$-hard quadratic unconstrained binary optimisation problems (QUBOs). 
While having to formulate a problem as a QUBO is rather 
inflexible, QA is, in the future, widely expected to solve QUBOs at speeds not achievable with classical hardware.  
Thus, casting a problem as a QUBO promises to outperform more unrestricted formulations in terms of tractable problem sizes and attainable accuracy through sheer speed. 

\begin{figure}
    \centering
    \includegraphics[width=1\linewidth]{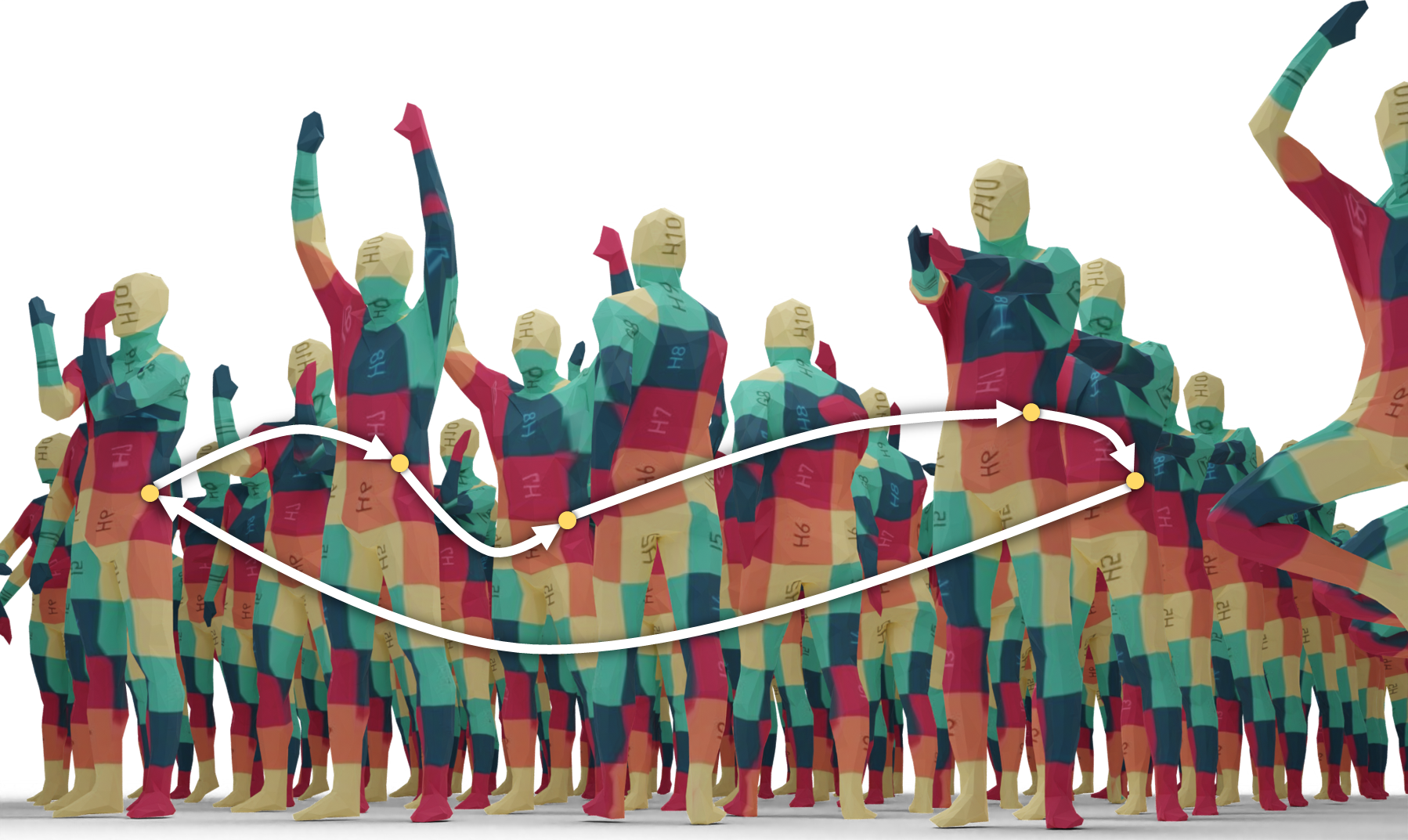}
    \caption{\textbf{Our quantum-hybrid method matches all $\boldsymbol{100}$ shapes of the FAUST collection \cite{Bogo:CVPR:2014} with guaranteed cycle consistency (white arrows).} 
    Here, we visualise the matchings via texture transfer between all shapes. 
    Our method scales linearly in the number of shapes. 
    See the full figure in the supplement. 
    }
    \label{fig:teaser}
\end{figure}

A recent example for such a problem is shape matching, where the goal is to estimate correspondences between two shapes. 
Accurate shape matching is a core element of many computer vision and graphics applications (\textit{i.e.,} texture transfer and statistical shape modelling). 
If non-rigid deformations are allowed, even pairwise matching is $\mathcal{NP}$-hard, leading to a wide area of research that approximates this problem, as a recent survey shows \cite{Deng2022}. 
Matching two shapes is one of the problems that was shown to benefit from quantum hardware: Q-Match~\cite{SeelbachBenkner2021} iteratively updates a subset of point correspondences using QA. 
Specifically, its cyclic $\alpha$-expansion allows to parametrise changes to permutation matrices without relaxations.

The question we ask in this work is: How can we design a \emph{multi-shape} matching algorithm in the style of Q-Match that has the same benefits? 
As we show in the experiments, where we introduce several na\"ive multi-shape extensions of Q-Match, this is a highly non-trivial task. 
Despite tweaking them, our proposed method significantly outperforms them.

If $N{>}2$ shapes have to be matched, the computational complexity of na\"ive exhaustive pairwise matching increases quadratically with $N$, which %
does not scale to large $N$. 
Furthermore, these pairwise matchings can easily turn out to be inconsistent with each other, thereby violating cycle consistency. 
For example, chaining the matchings $P_{\mathcal{X}\mathcal{Y}}$ from shape $\mathcal{X}$ to $\mathcal{Y}$ and $P_{\mathcal{Y}\mathcal{Z}}$ from $\mathcal{Y}$ to $\mathcal{Z}$ can give very different correspondences between $\mathcal{X}$ and $\mathcal{Z}$ than the direct, pairwise matching $P_{\mathcal{X}\mathcal{Z}}$ of $\mathcal{X}$ and $\mathcal{Z}$: $P_{\mathcal{X}\mathcal{Z}} \neq P_{\mathcal{X}\mathcal{Y}}P_{\mathcal{Y}\mathcal{Z}}$. 
(We apply the permutation matrix $P_{\mathcal{X}\mathcal{Y}}$ to the one-hot vertex index vector $x\in\mathcal{X}$ as $x^\top P_{\mathcal{X}\mathcal{Y}}=y\in\mathcal{Y}$.) 
Thus, how can we achieve cycle consistency by design? 
A simple solution %
would be to match a few pairs in the collection to create a spanning tree covering all shapes and infer the remaining correspondences by chaining along the tree. 
Despite a high accuracy of methods for matching two shapes, this correspondence aggregation policy 
is prone to error accumulation \cite{sahillioglu2014multiple}. 
A special case of this policy is pairwise matching against a single anchor shape, which also guarantees cycle-consistent solutions by construction \cite{Gao2021}. 
We build on this last option in our method as it avoids error accumulation.

\begin{figure}
    \centering
    \includegraphics[width=\linewidth]{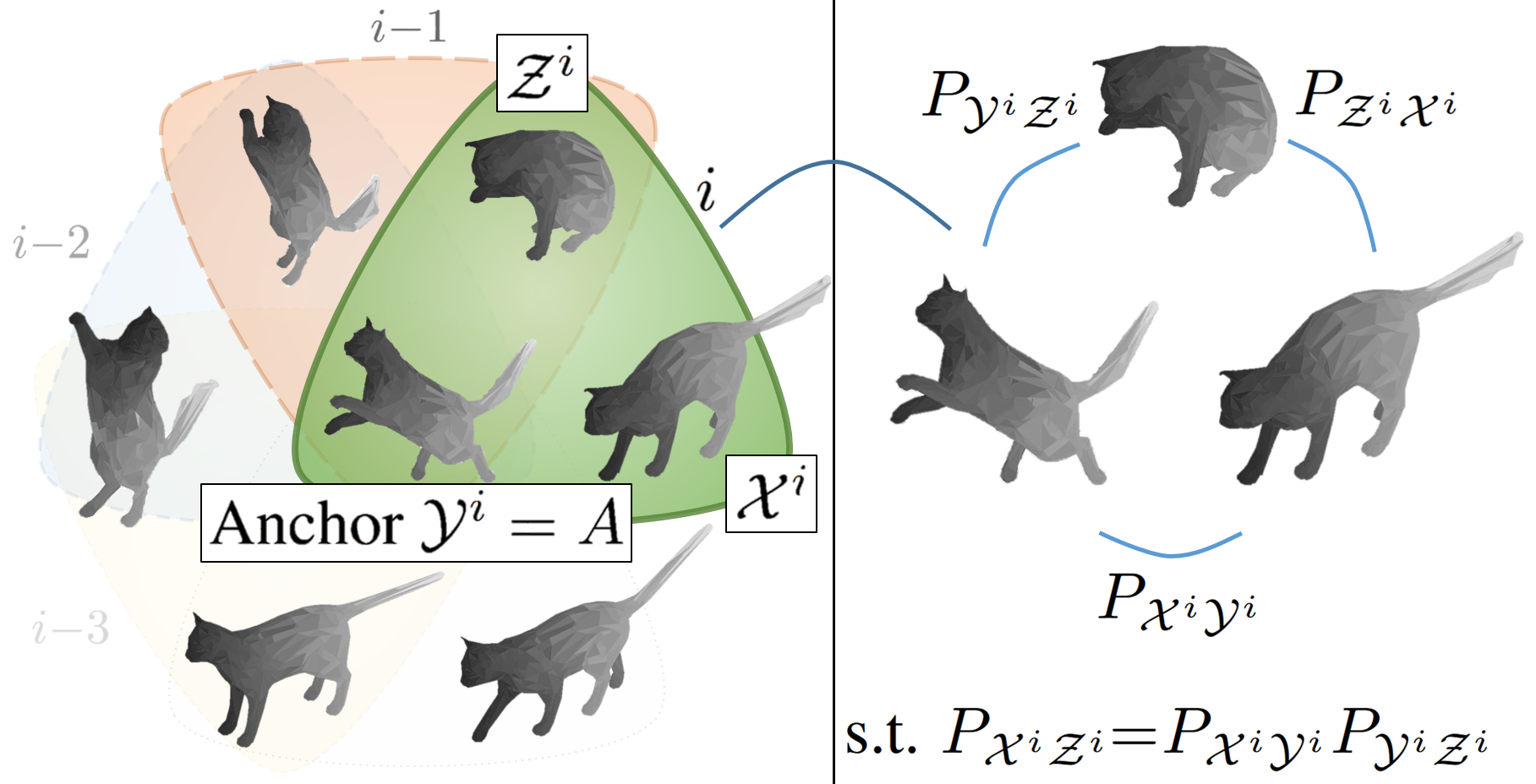}  
    \caption{We match $N$ shapes by iteratively matching triplets. %
    }
    \label{fig:main_figure}
\end{figure}

This paper, in contrast to purely classical methods, leverages the advantages of quantum computing for multi-shape matching 
and introduces a new method for simultaneous alignment of multiple meshes with guaranteed cycle consistency; see Fig.~\ref{fig:teaser}. 
It makes a significant step forward compared to Q-Match and other methods utilising adiabatic quantum computing (AQC), the basis for QA. 
Our \underline{c}ycle-\underline{c}onsistent q\underline{uantu}m-hybrid \underline{m}ulti-shape \underline{m}atching (CCuantuMM; pronounced ``quantum'') 
approach relies on the computational power of modern quantum hardware. 
Thus, our main challenge lies in casting our problem in QUBO form, which is necessary for compatibility with AQC. 
To that end, two design choices are crucial: 
(1) Our method reduces the $N$-shapes case to a series of three-shape matchings; see Fig.~\ref{fig:main_figure}. 
Thus, CCuantuMM is iterative and hybrid, \textit{i.e.,} it alternates in every iteration between preparing a QUBO problem on the CPU and sampling a QUBO solution on the AQC. 
(2) It discards negligible higher-order terms, which makes mapping the three-shape objective to quantum hardware possible. 
In summary, the core technical contributions of this paper are as follows: 
\vspace{-3pt}
\begin{itemize}
    \setlength{\itemsep}{1pt}
    \setlength{\parskip}{0pt}
    \setlength{\parsep}{0pt}
    \item CCuantuMM, \textit{i.e.,} a new quantum-hybrid method for shape  multi-matching relying on cyclic $\alpha$-expansion. 
    CCuantuMM produces cycle-consistent matchings and scales linearly with the number of shapes $N$. 
    \item A new formulation of the optimisation objective for the three-shapes case that is mappable to modern QA. 
    \item A new policy in shape multi-matching to address the $N$-shape case relying on a three-shapes formulation and adaptive choice of an anchor shape. 
\end{itemize} 

Our experiments show that CCuantuMM significantly outperforms several variants of the previous quantum-hybrid method Q-Match \cite{SeelbachBenkner2021}. 
It is even competitive with several non-learning-based classical state-of-the-art shape methods \cite{melzi2019zoomout, Gao2021} and can match more shapes than them. 
In a broader sense, this paper demonstrates the very high potential of applying (currently available and future) quantum hardware in computer vision. 

%% file: sections/2_related_work.tex
\section{Related Work}\label{sec:related} 

\noindent\textbf{Quantum Computer Vision (QCV).} Several algorithms for computer vision relying on quantum hardware were proposed over the last three years for such problems as shape matching \cite{golyanik2020quantum, SeelbachBenkner2021, Meli_2022_CVPR}, object tracking \cite{LiGhosh2020, Zaech_2022_CVPR}, fundamental matrix estimation, point triangulation \cite{Doan_2022_CVPR} and motion segmentation \cite{Arrigoni2022}, among others. 
The majority of 
them 
address various types of alignment problems, \textit{i.e.,} transformation estimation \cite{golyanik2020quantum, Meli_2022_CVPR}, 
point set \cite{golyanik2020quantum, NoormandipourWang2022} and mesh alignment  \cite{SeelbachBenkner2021}, 
graph matching \cite{SeelbachBenkner2020, QuantumSync2021} and permutation synchronisation \cite{QuantumSync2021}.  

Only one of them, QSync \cite{QuantumSync2021}, can operate on more than two inputs and ensure cycle consistency for the underlying matchings. 
In contrast to QSync, we can align inputs with substantially larger (by two orders of magnitude) shapes in the number of vertices. 
Furthermore, we address a different problem, \textit{i.e.,} mesh alignment, for which an algorithm for two-mesh alignment with the help of AQC exists, namely Q-Match \cite{SeelbachBenkner2021}, as we discuss in the introduction. 

To maintain the valid structure of permutation matrices, Quantum Graph Matching, QGM\cite{SeelbachBenkner2020} and Q-Sync \cite{QuantumSync2021} impose linear constraints. %
However, this requires that 
the corresponding penalty parameter is carefully chosen. 
If the parameter is chosen too big and the linear constraints are enforced too strongly, this
severely limits QGM and Q-Sync's ability to handle large sets of vertices. 
On the other hand, if the linear constraints are enforced too weakly, there is no guarantee to obtain valid permutations as solutions. 
As discussed in the introduction, our approach follows Q-Match to ensure valid permutation matrices by construction.

\noindent\textbf{Multi-Shape Matching.} 
We focus this section on multi-shape and non-learning methods as CCuantuMM falls in this category. 
As our approach is not learning-based, it trivially generalises to unknown object categories without a need for training data. 
For a general survey of recent advances in shape matching, see Sahillioglu  \cite{sahillioglu2019survey}. 

Matching shape pairs is a classical problem in geometry processing \cite{melzi2019zoomout}. 
When more than two shapes of the same class exist, stronger geometric cues can be leveraged to improve results by matching all of them simultaneously. 
Unfortunately, the already very high problem complexity increases even further the more shapes are used. 
Hence, existing multi-shape matching methods 
limit the total number of shapes and their resolution \cite{cosmo2017consistent,Gao2021}, work in spectral space \cite{huang2020consistent}, or relax the permutation constraints  \cite{kezurer2015tight}. 
Early multi-matching methods computed pair-wise matchings and subsequently used permutation synchronisation to establish cycle consistency~\cite{pachauri2013solving, maset2017practical,shen2016normalized}.
Still, permutation synchronisation requires the eigendecomposition of a matrix with quadratically increasing dimensions \cite{pachauri2013solving}. 

HiPPI \cite{bernard2019hippi} is a  computationally efficient method that takes geometric relations into account while generalising permutation  synchronization but is still limited in resolution. 
Instead of looking at permutations directly, ZoomOut~\cite{melzi2019zoomout} reduces the  dimensionality of the problem by projecting it onto the spectral decomposition. 
This idea has been extended to take cycle consistency within the spectral space into account  \cite{huang2013consistent}, which does not guarantee a point-wise consistent matching. 
To circumvent this issue, IsoMuSh~\cite{Gao2021} jointly optimises point and functional correspondences. 
The method detangles the optimisation into smaller subproblems by using a so-called \textit{universe shape} that all shapes are mapped to instead of each other, as Cao and Bernard do \cite{cao2022unsupervised}. 
Using a \textit{universe} is similar to requiring a template shape, as many learning-based approaches  do \cite{groueix2018b,deprelle2019learning,sundararaman2022implicit}: Both synchronise all correspondences by matching them through a unified space. 
This is similar to the concept of anchor shape we use but inherently less flexible because the universe size or template have to be given \emph{a priori}. 
Our anchor is chosen from the given collection as part of the method. 
Although using an anchor slightly improves our results, we note that our method does not necessarily require one for operation.  
Hence, a random shape could be picked instead in each iteration without an increase in complexity if using an anchor is not feasible or does not represent the shape collection well.

%% file: sections/3_background.tex
\section{Background}\label{sec:background}

\subsection{Adiabatic Quantum Computing (AQC)}\label{sec:aqc}
AQC is a model of computation that leverages quantum effects to obtain high-quality solutions to the $\mathcal{NP}$-hard problem class of Quadratic Unconstrained Binary Optimisation (QUBO) problems:
    $\min_{x\in \{0,1 \}^k }  x^T Qx$,  
for $k \in \mathbb{N}$ and a QUBO weight matrix $Q\in \mathbb{R}^{k \times k}$. 
Each entry of $x$ corresponds to its own logical qubit, the quantum equivalent of a classical bit. 
The diagonal of $Q$ consists of linear terms, while the off-diagonals are inter-qubit \emph{coupling weights}. 
A QUBO can be classically tackled with simulated annealing (SA)~\cite{van1987simulated} or a variety of other discrete optimisation techniques \cite{kochenberger2014unconstrained,dunning2018works}, which, for large $k$, typically yield only approximate solutions as QUBOs are in general $\mathcal{NP}$-hard. 
AQC holds the potential to systematically outperform classical approaches such as SA, see \cite{denchev2016computational,king2021scaling} for an example. 
AQC exploits the adiabatic theorem of quantum mechanics \cite{BornFock1928}: If, when starting from an equal superposition state of the qubits (where all solutions $\{0,1 \}^k$ have the same probability of being measured) and imposing external influences corresponding to the QUBO matrix on the qubits sufficiently slowly (called \emph{annealing}), they will end up in a quantum state that, when measured, yields a minimizer $x$ of the QUBO. 
Not all physical qubits on a real quantum processing unit (QPU) can be %
connected (coupled) with each other. 
Thus, a \emph{minor embedding} of the logical-qubit graph (defined by non-zero entries of the QUBO matrix) into the physical-qubit graph (defined by the hardware) is required \cite{cai2014practical}. 
This can lead to a chain of multiple physical qubits representing a single logical qubit. 
For details of quantum annealing on D-Wave machines, we recommend \cite{mcgeoch2014adiabatic}.

\subsection{Shape Matching}\label{sec:q_match}
The problem of finding a matching for non-rigidly deformed shapes having $n$ vertices can be formulated as an $\cal N P$-hard
Quadratic Assignment Problem (QAP) \cite{kezurer2015tight,burghard17lifted}:
\begin{align} 
    \min_{P\in \mathbb{P}_{n}}  p^T W p,
\end{align}
where $p = \text{vec}(P)\in \{0,1\}^{n^2}$ is a flattened permutation matrix, %
and $W%
\in \mathbb{R}^{n^2 \times n^2}$ is an energy matrix describing how well certain pairwise properties are conserved between two pairs of matches. %
If two shapes $\cal X, \cal Y$ are discretised with $n$ vertices each, $W$ is often chosen as \cite{kezurer2015tight}: 
\begin{equation}
    W_{x_1\cdot n + y_1, x_2\cdot n +y_2} = \| d_{\cal X}^g(x_1,x_2) - d_{\cal Y}^g(y_1,y_2) \|,
    \label{eq:W}
\end{equation}
where  $x_1,x_2$ are vertices on $\cal X$; $y_1,y_2$ are vertices on $\cal Y$; and %
$d_{\cal I}^g(\cdot, \cdot)$ 
is the geodesic distance on the shape $\cal I$. 
Therefore, $W_{x_1\cdot n + y_1, x_2\cdot n +y_2}$ represents how well the geodesic distance is preserved between corresponding pairs of vertices on the two shapes. %
Instead of pure geodesics, Gaussian-filtered geodesics are also a popular choice for $W$ \cite{kernel17}: 
\begin{equation}
    g_{\cal X}(x_1,x_2) = \frac{1}{\rho \sqrt{2 \pi}} \exp {\left( -\frac{1}{2} \left(\frac{d^g_{\cal X}(x_1, x_2)}{\rho} \right)^2\right)}.
\end{equation}
$g_{\mathcal{I}}$ can be used to directly replace $d^g_{\mathcal{I}}$ in \eqref{eq:W}.
A small value of $\rho$ focuses the energy on a local neighbourhood around the vertex, while a large value increases the receptive field. 
Using Gaussian kernels in $W$ places more emphasis on local geometry whereas geodesics have higher values far away from the source vertex. 
Thus, geodesics work well for global alignment and  Gaussians for local fine-tuning. 

\subsection{Cyclic {\large $\boldsymbol{\alpha}$}-Expansion (CAE) } \label{sec:CAE}
CCuantuMM represents matchings as permutation matrices. 
In order to
update 
them, we build on Seelbach~\textit{et al.}'s CAE algorithm \cite{SeelbachBenkner2021} (similar to a fusion move \cite{hutschenreiter2021fusion}), which we describe here. 
A permutation matrix $P$ is called an \emph{$r$-cycle}, if there exist $r$ disjoint indices $i_1, \ldots , i_r$ such that $P_{i_j i_{(j+1) \% r}} = 1$ for all $j \in \{1,\ldots, r\}$ and $P_{l,l} = 1$ for all $l \notin \{i_1, \ldots, i_r\}$, in which case $P = (i_1 i_2 \ldots i_r)$ 
is a common notation.
Two cycles, \emph{i.e.}\ two permutation matrices, are \emph{disjoint} if these indices are pairwise disjoint.
We know that disjoint cycles %
commute, which allows us to represent any permutation $P$ as $P = \left(\prod_{i=1}^{k} c_i\right) \left(\prod_{i=1}^{l} \Tilde{c}_i \right) $, where $\{c_i\}_i$ and $\{\Tilde{c}_i\}_i$ each are sets of disjoint $2$-cycles.

Given a set $\{c_i\}_{i=1}^k$ of $k$ disjoint $2$-cycles, an \emph{update}, or modification, of $P$ can therefore be parameterised as: $P(\alpha) = \prod_{i=1}^{k} c_i^{\alpha_i} P,$ where $\alpha \in \{0,1\}^{k}$ is a binary decision vector determining the update. 
(Note that $\alpha_i$ in $c_i^{\alpha_i}$ is an  exponent, not an index.) 
Crucially, to make this parameterisation compatible with QUBOs, we need to make it linear in $\alpha$. 
To this end, CAE uses the following equality: 
\begin{equation} \label{eq:CAE}
    P(\alpha) = P + \sum_{i=1}^{k} \alpha_i (c_i - I ) P. 
\end{equation}

%% file: sections/4_method.tex
\section{Our CCuantuMM Method} \label{sec:method}

Previous adiabatic quantum computing methods \cite{SeelbachBenkner2020, SeelbachBenkner2021} can only match two shapes. %
We present a method for matching $N$ shapes. 
To ensure cycle consistency on $N$ shapes, %
it is sufficient that all triplets of shapes are matched cycle consistently~\cite{huang2013consistent}. 
CCuantuMM iteratively solves three-shape problems, which preserve cycle consistency by construction and fit on existing quantum annealers with limited  resources. 
We introduce our formulation for matching three shapes in Sec.~\ref{sec:threeshapes} and then extend it to $N$ shapes in Sec.~\ref{sec:n_shapes}. 

\subsection{Matching Three Shapes}\label{sec:threeshapes}
Consider the problem of matching three non-rigidly deformed shapes $\mathcal{S} = \{\mathcal{X,Y,Z}\}$ of $n$ vertices each, while preserving cycle consistency.
We formulate this as an energy minimisation with respect to $ \mathcal{P} = \{ P_{\cal I J}\in\mathbb{P}_n | \cal I,J \in \cal S \} $, the set of permutations between all pairs in $\mathcal{S}$: %
\begin{equation} 
\begin{aligned} \label{eq:ThreeMatch} 
    &\min_\mathcal{P}  \sum_{\cal I, J \in \mathcal{S}; \cal I \neq \cal J} \operatorname{vec}(P_{\cal I J})^\top W_{\cal I J} \operatorname{vec}(P_{\cal I J}),  \\
    &\textrm{s.t. } \quad P_{\cal X Z} = P_{\cal X Y}P_{\cal Y Z}, 
\end{aligned} 
\end{equation} 
where $W_{\cal I J} \in \mathbb{R}^{n^2 \times n^2}$ is the energy matrix describing how well certain pairwise properties are conserved between shapes $\cal I$ and $\cal J$ (see Sec.~\ref{sec:q_match}), and $P_{\cal X Z} = P_{\cal X Y} P_{\cal Y Z} $ enforces cyclic consistency. %
An overview of the algorithm for three shapes is shown in Alg.~\ref{Alg:3QMatch}.

\subsubsection{QUBO Derivation} 

To perform optimisation on the quantum annealer, we need to transform \eqref{eq:ThreeMatch} into a QUBO problem. 
We adapt the CAE formulation from \cite{SeelbachBenkner2021} (see Sec.~\ref{sec:CAE}) and iteratively update the permutations to decrease the value of \eqref{eq:ThreeMatch}.
Given a set $C = \{c_i\}_{i=1}^k $ of $k$ disjoint 2-cycles and binary decision variables $\alpha$, we can parameterise our permutation matrices as $P_{\cal I J} (\alpha) = P_{\cal I J} + \sum_{i=1}^k \alpha_i (c_i - I) P_{\cal I J}$.
However, CAE alone is not sufficient to transform \eqref{eq:ThreeMatch} into a QUBO as cyclic consistency is still missing. 
A simple solution would be to encourage cyclic consistency as a quadratic soft penalty, but then there are no guarantees on the solution.
Instead, we enforce cyclic consistency by construction:
{\small
\begin{align}
    &P_{\cal XZ}(\alpha, \beta) = \\ \nonumber &(P_{\cal XY} + \sum_{i=1}^k \alpha_i (c_i -I) P_{\cal X Y}) \cdot (P_{\cal YZ} + \sum_{j=1}^k \beta_j( \tilde{c}_j - I) P_{\cal YZ}), 
\end{align}
}
\hspace{-3pt}where $\{c_i\}_i$ $\left(\{\tilde{c}_j\}_j\right)$ are cycles and $\alpha$ ($\beta$) are decision variables for the updates to $P_{\cal X Y}$ ($P_{\cal Y Z}$).
For brevity, we write $C_i = (c_i - I) P_{\cal X Y}$ and $\tilde{C_j} = (\tilde{c}_j - I) P_{\cal Y Z}$. 
We explain how we construct the cycles in  Secs.~\ref{sec:vertices}-\ref{sec:cycles}. 
Thus, we iteratively solve \eqref{eq:ThreeMatch} via a sequence of problems of the form: 
\begin{equation} \label{eq:intermediate_three}
    \min_{\alpha,\beta \in \{0,1\}^k} \exy{E_{\cal X Y}(P_{\cal X Y}(\alpha))} + \eyz{E_{\cal Y Z}(P_{\cal Y Z}(\beta))} + \exz{E_{\cal X Z}({P_{\cal X Z}(\alpha,\beta))}},
\end{equation} 
where $E_{\cal I J}(P,Q) =$ $ \operatorname{vec}(P)^\top W_{\cal I J} \operatorname{vec}(Q)$ and $E_{\cal I J}(P) = $  $ E_{\cal I J}(P,P)$. %
While the first two terms are in QUBO form, the third term contains cubic and bi-quadratic terms (see the supplement for details) which are not compatible with current quantum annealer architectures.
\paragraph{Higher-Order Terms.} 
All of these higher-order terms come from $P_{\cal X Z}(\alpha, \beta)$, specifically from the term $H = \sum_i \sum_j \alpha_i  \beta_j C_i \tilde{C}_j$.
As we only consider 2-cycles, $C_i$ and $\tilde{C}_j$ each have only four non-zero elements. 
Due to this extreme sparsity, %
most summands of $H$ become $0$.

We could tackle these undesirable terms by decomposing them into quadratic terms by using ancilla variables and adding penalty terms \cite{dattani2019quadratization}. 
This gives exact solutions for sufficiently high weights of the penalty terms.  
However, multiple reasons speak against this: (1) the QUBO matrix is already dense (a clique) under the current formulation (as we will see in \eqref{eq:WTildeForm}) and adding ancilla qubits scales quadratically in $k$, 
(2) adding penalties makes the problem harder to solve%
, and (3) $H$ is sparsely non-zero and in practise we observe no drastic influence on the quality of the solution. %

Alternatively, we could assume $H{=}0$. 
However, this is unnecessarily strong since (1) $H$ also contributes to quadratic terms ($E(H,\cdot)$), and (2) higher-order terms operating on the same decision variable trivially reduce to quadratic terms: $\alpha_i \cdot \alpha_i = \alpha_i$ for binary $\alpha_i$. 
We thus keep those two types of terms and merely assume all truly cubic and bi-quadratic terms to be zero. 

\paragraph{Cycle-Consistent CAE.} After eliminating the higher-order terms and ignoring constants from \eqref{eq:intermediate_three}, we obtain (with the same colour coding): 
\begin{equation} 
{\small
\label{eq:FinalOptimisation}
\footnotesize   
\begin{aligned} 
&\min_{\alpha,\beta} \quad %
\sum_{i=1}^{k} \alpha_i \bigg(%
\exy{F_{\cal X Y}(P_{\cal X Y}, C_i)} + 
\exz{F_{\cal X Z}(P_{\cal XZ}, C_i P_{\cal YZ} )}\bigg) \\
&+ 
\sum_{j=1}^{k} \beta_j \bigg(%
\eyz{F_{\cal Y Z}(P_{\cal Y Z}, \tilde{C}_j)} + 
\exz{F_{\cal X Z}(P_{\cal XZ}, P_{\cal X Y}  \tilde{C}_j )} 
\bigg)  \\  
&+\sum_{i=1}^{k} \sum_{l=1}^{k} \alpha_i \alpha_l \bigg(
\exy{E_{\cal X Y}(C_i,C_l)} + %
\exz{E_{\cal X Z}(C_i P_{\cal YZ},  C_l P_{\cal YZ})}
\bigg)\\
&+ \sum_{j=1}^{k} \sum_{l=1}^{k} \beta_j \beta_l \bigg(
\eyz{E_{\cal Y Z}(\tilde{C}_j,\tilde{C}_l)} + %
\exz{E_{\cal X Z}(P_{\cal X Y} \tilde{C}_j  , P_{\cal X Y} \tilde{C}_l)} 
\bigg) 
\\ %
&+\sum_{i=1}^{k} \sum_{j=1}^{k} \alpha_i \beta_j \bigg( %
\exz{F_{\cal X Z}(P_{\cal X Y} \tilde{C}_j, C_i P_{\cal YZ})} + %
\exz{F_{\cal X Z}(K_{ij},P_{\cal XZ})} \\
&+
\exz{F_{\cal X Z}(K_{ij}, P_{\cal X Y} \tilde{C}_j)} +
\exz{F_{\cal X Z}(K_{ij}, C_i P_{\cal YZ})} %
+\exz{E_{\cal X Z}(K_{ij},K_{ij})} %
\bigg), \\
\end{aligned}
}
\end{equation}
where %
$P_{\cal XZ}=$ $P_{\cal XZ}(\mathbf{0}, \mathbf{0})= $ $ P_{\cal XY}P_{\cal YZ}$, and we use the shorthands $F_{\cal I J}(A,B) =$ $ E_{\cal I J}(A,B) + E_{\cal I J}(B,A)$ and $K_{ij} = $ $C_i \tilde{C}_j$. 
Denoting $\alpha_{k+j} = \beta_j$ for an expanded $\alpha\in \{0,1\}^{2k}$, 
\eqref{eq:FinalOptimisation} can be written in the form:
\begin{equation} \label{eq:WTildeForm}
    \min_{\alpha\in \{0,1\}^{2k}} \quad \alpha^\top \tilde{W} \alpha.
\end{equation}
The full formula for $\tilde{W}$ is provided in the supplement.  
\eqref{eq:WTildeForm}~is finally in QUBO form and we can optimise it classically or on real quantum hardware (see Sec.~\ref{sec:aqc}). 

\subsubsection{Choosing Vertices} \label{sec:vertices}
The question of how to choose the sets of cycles $\{c_i\}_i, \{\tilde{c}_j\}_j$ is still open. 
We first choose a subset of vertices using the ``worst vertices'' criterion introduced in~\cite{SeelbachBenkner2021} based on the relative inconsistency $I_{\cal XY}$ of a vertex $x\in\mathcal{X}$ under the current permutation: 
\begin{equation} \label{eq:Inconsistency}
    I_{\cal XY}(x) = \sum_{w \in \cal X} W_{x \cdot n + x^\top P_{\cal XY}, w\cdot n + w^\top P_{\cal XY} },
\end{equation}
where we treat the one-hot vector $x^\top P_{\cal X Y}$ as a vertex index on $\cal Y$.
A high value indicates that $x$ is inconsistent with many other matches under $P_{\cal X Y}$ and swapping it will likely improve the matching. %
We denote the set of the $m{=}2k$ vertices with the highest $I_{\cal X Y} (\cdot)$ as $V_{\cal X}$. 
Finally, we follow the permutations to get $V_{\cal Y} = \{ x^\top P_{\cal X Y} | x \in V_{\cal X} \} \subset {\cal Y}$. %

In practice, we observe a systematic improvement in the matchings when considering all three possibilities (using $I_{\cal X Y}$, $I_{\cal Y Z}$, or $I_{\cal X Z}$ %
as the starting point). 
We thus use three ``sub''-iterations per iteration, one for each possibility.

\subsubsection{Choosing Cycles} \label{sec:cycles}

\begin{figure}
    \centering
    \resizebox{\columnwidth}{!}{\input{figures/subiteration.tex}}
    \caption{
    We depict the sub-iteration that starts from $I_{\cal X Y}$, from which we construct $V_\mathcal{X}$, then $\mathcal{C}^\mathit{all}_{\cal X}$, and finally $\mathcal{C}_{\cal X } = $ $\{\mathcal{C}^0_{\cal X},$ $ \mathcal{C}^1_{\cal X}, $ $ \mathcal{C}^2_{\cal X} \}$. 
    We also build $V_\mathcal{Y}$ from $V_\mathcal{X}$  %
    and construct $\mathcal{C}_{\cal Y}$ analogously. 
    Matching each element of $\mathcal{C}_{\cal X}$ with one from $\mathcal{C}_{\cal Y}$ (visualised via matching colours) leads to three sub-sub-iterations. 
    }
    \label{fig:cycleconstruction}
\end{figure}

\begin{figure}
    \centering
    \resizebox{0.8\columnwidth}{!}{\input{figures/subsubiteration.tex}}    
    \caption{
    We depict the sub-sub-iteration for $\mathcal{C}_{\cal X}^{2} = $ $ \{ (u_x,v_x), $ $(w_x,t_x) \} $ %
    and $\mathcal{C}_{\cal Y}^{1} =\{ (u_y,w_y),(v_y,t_y) \} $ %
    from Fig.~\ref{fig:cycleconstruction}. %
    } 
    \label{fig:choosing cycle}
\end{figure}
Given the worst vertices $V_{\cal X}$ and $V_{\cal Y}$ of any sub-iteration, %
we construct the cycles $\{c_i\}_i, \{\tilde{c}_j\}_j$ from them. 
Fig.~\ref{fig:cycleconstruction} visualises this process. 
Focusing on $V_{\cal X}$ for the moment, we want to use all possible 2-cycles $\mathcal{C}_{\mathcal{X}}^\mathit{all} = \{(uv) | u,v\in V_{\cal X}, u\neq v \}$ in each sub-iteration. 
We cannot use all of these cycles at once since they are not disjoint, as CAE requires. 
Instead, we next construct a set $\mathcal{C}_{\cal X}$ by partitioning $\mathcal{C}_{\mathcal{X}}^\mathit{all}$ into $m{-}1$ sets of cycles with each containing $m/2{=}k$ disjoint cycles. 
An analogous methodology is used for $\mathcal{C_Y}$. 

We now have $\mathcal{C_X}$ and $\mathcal{C_Y}$. 
Since we want to consider each cycle of $\mathcal{C}_{\mathcal{X}}^\mathit{all}$ and $\mathcal{C}_{\mathcal{Y}}^\mathit{all}$ once, we need several ``sub-sub'' iterations. 
Thus, we next need to pick one set of cycles from each $\mathcal{C_X}$ and $\mathcal{C_Y}$ for each sub-sub-iteration. 
There are $(m{-}1)^2$ possible pairs between elements of $\mathcal{C_X}$ and $\mathcal{C_Y}$.
Considering all possible pairs is redundant, does not provide significant performance advantage, and increases the computational complexity quadratically. 
Hence, we randomly pair each element of $\mathcal{C}_{\cal X}$ with one element of $\mathcal{C}_{\cal Y}$ (without replacement). 
This leads to $m{-}1$ sub-sub-iterations, with each one solving \eqref{eq:WTildeForm} with its respective cycles; see
Fig.~\ref{fig:choosing cycle}. %

\begin{algorithm}
\caption{Hybrid Three-Shape Matching} 
\label{Alg:3QMatch}
\textbf{Input:} $\mathcal{P}^{i},\mathcal{S}$ \\
\textbf{Output:} $\mathcal{P}^{i+1}$
\begin{algorithmic}[1]
\For {$I \in \{I_{\cal X Y}, I_{\cal Y Z}, I_{\cal X Z}\}$} \hfill \Comment{sub-iterations}
    \State {construct $V_{\cal X}, V_{\cal Y}, V_{\cal Z}$ (see Sec.~\ref{sec:vertices})}
    \State {construct $\cal C_{X}, C_{Y}, C_{Z}$ (see Sec.~\ref{sec:cycles})} 
    \For{$l{=}1$ to $m{-}1$ } \hfill \Comment{sub-sub-iterations}
        \State {compute $\tilde{W}$}
        \State {optimise QUBO \eqref{eq:WTildeForm}} \hfill \Comment{quantum}
    \EndFor
    \State {$P_{\cal XY}=\prod_{i=1}^{k} c_i^{\alpha_i} P_{\cal XY}$ }
    \State {$P_{\cal YZ} = \prod_{j=1}^{k} \tilde{c}_j^{\alpha_{m + j}}P_{\cal YZ}$}
    \State {$ P_{\cal XZ} = P_{\cal XY} \cdot P_{\cal YZ}$ }
\EndFor
\State \Return  {$\mathcal{P}^{i+1} = \{P_{\cal XY}, P_{\cal YZ}, P_{\cal XZ}\}$}
\end{algorithmic}
\end{algorithm}

\subsection{Matching {\large $N$} Shapes }\label{sec:n_shapes}

In this section, we extend our model to matching a shape collection $\mathcal{S}$ with $N$ elements by iteratively matching three shapes while still guaranteeing cycle consistency. 
Similar to the three-shape case \eqref{eq:ThreeMatch}, this can be formulated as an energy minimisation problem with respect to the set of permutations $\mathcal{P}$, except $\mathcal{S}$ now has cardinality $N$:
\begin{equation} 
\begin{aligned} \label{eq:multimatching}
    \min_\mathcal{P} &\sum_{\mathcal{I},\mathcal{J} \in \mathcal{S}; \mathcal{I}\neq \mathcal{J}} E_{\cal I J}(P_{\cal I J}), %
    \\ 
    \textrm{s.t. } \quad & P_{\cal I K} = P_{\cal I J } P_{\cal J K } \quad \forall \mathcal{I,J,K} \in \mathcal{S}. 
\end{aligned}
\end{equation} 
The energy contains summands for each possible pair of shapes.
Solving all of them jointly would be computationally expensive and even more complicated than \eqref{eq:FinalOptimisation}.
This is the reason most multi-shape matching methods apply relaxations at this point or cannot scale to a large $N$.
However, the cycle-consistency constraints still only span over three shapes; triplets are sufficient for global consistency \cite{huang2013consistent}.

We thus iteratively focus on a triplet $\mathcal{X}, \mathcal{Y}, \mathcal{Z} \in \mathcal{S}$ and its set of permutations $\mathcal{P}' = \{P_{\mathcal{X} \mathcal{Y}}, P_{\mathcal{X} \mathcal{Z}}, P_{\mathcal{Y} \mathcal{Z}} \}$. 
We could then minimise \eqref{eq:multimatching} over $\mathcal{P}'$, leading to a block-coordinate descent optimisation of \eqref{eq:multimatching} over $\mathcal{P}$. 
This would make the problem tractable on current quantum hardware since it keeps the number of decision variables limited. %
It would also formally guarantee that our iterative optimisation would never increase the total energy. 
However, each iteration would be linear in $N$ due to the construction of the QUBO matrix, preventing scaling to large $N$ in practice. 
We therefore instead restrict \eqref{eq:multimatching} to those terms that depend \emph{only} on permutations from $\mathcal{P}'$. 
This leads to the same energy as for the three-shape case \eqref{eq:ThreeMatch}, where the minimisation is now over $\mathcal{P}'$. 
Importantly, the computational complexity per triplet becomes independent of $N$, allowing to scale to large $N$. 
While this foregoes the formal guarantee that the \emph{total} energy never increases, we crucially find that it still only rarely increases in practice; see the supplement.

By iterating over different triples $\mathcal{X}^i, \mathcal{Y}^i, \mathcal{Z}^i$, we cover the entire energy term and reduce it iteratively. 
Specifically, one iteration $i$ of the $N$-shape algorithm runs Alg.~\ref{Alg:3QMatch} on $\mathcal{P}'=\{P^i_{\mathcal{X}^i \mathcal{Y}^i}, P^i_{\mathcal{Y}^i \mathcal{Z}^i}, P^i_{\mathcal{X}^i \mathcal{Z}^i}\}$. 
Here, $\mathcal{X}^i$ $\in\mathcal{S}$ is chosen randomly (we use stratified sampling to pick all shapes equally often), the \emph{anchor} $\mathcal{Y}^i =$ $ A\in\mathcal{S}$ is fixed, and $\mathcal{Z}^i = $ $\mathcal{X}^{i-1}$. 
In practice, we saw slightly better results with this scheme instead of choosing the triplet randomly; see the supplement. 
We note that we only need to explicitly keep track of permutations into the anchor: $\mathcal{P}^{i}=\{P^i_{\mathcal{I} A}\}_{\mathcal{I}\in\mathcal{S},\mathcal{I}\neq A}$. %
We then get $\mathcal{P}^{i+1}$ from $\mathcal{P}^{i}$ by replacing $P^i_{\mathcal{X}^i A}$ and $P^i_{\mathcal{Z}^i A}$ with their updated versions from Alg.~\ref{Alg:3QMatch}. %

\noindent\textbf{Initialisation.} 
We compute an initial set of pairwise permutations $\mathcal{P}^\mathit{init}$ using a descriptor-based similarity of the normalised heat-kernel-signatures (HKS)~\cite{bronstein2010scale} extended by a dimension indicating whether a vertex lies on the left or right side of a shape (a standard practice in the shape-matching literature \cite{Gao2021}). 
Instead of using a random shape as anchor, the results improve when using the following shape: %
\begin{equation}
    A = \argmin_{A \in \mathcal{S}} \sum_{\mathcal{I} \in \mathcal{S}; \mathcal{I} \neq A} E_{\mathcal{I} A}(P^\mathit{init}_{\mathcal{I} A}), %
\end{equation}
where $P^\mathit{init}_{\mathcal{I} A}\in \mathcal{P}^\mathit{init}$. %
We thus have $\mathcal{P}^0 = \{P^\mathit{init}_{\mathcal{I} A}\}_{\mathcal{I} \in \mathcal{S}, \mathcal{I} \neq A}$.

\noindent\textbf{Time Complexity.} 
Our algorithm scales linearly with the number of shapes. 
Each iteration of Alg.~\ref{Alg:3QMatch} has worst-case time complexity $\mathcal{O}(nk^3)$, as we discuss in the supplement.

\noindent\textbf{Energy Matrix Schedule.} %
In practise, we first use pure geodesics for a coarse matching and then Gaussian-filtered geodesics to fine-tune. 
Specifically, for a shape collection of three shapes, we use a schedule with $2T$ geodesics iterations followed by $2T$ Gaussian iterations. 
For each additional shape in the shape collection, we add $T$ iterations to both schedules. 
We exponentially decrease the variance of the Gaussians every $N{-}1$ iterations to $\rho(i) = c_2 \exp(\frac{c_1}{i-T})$ where $c_1$ and $c_2$ are chosen such that the variance decreases from $25 \%$ to $ 5 \%$ of the shape diameter over the iterations. %
Thus, \emph{all} shapes undergo one iteration with the same specific variance.
We refer to the supplement for more details.

%% file: figures/subiteration.tex
\begin{tikzpicture}
\LARGE
\node (u1x) at (-3,0.5) {$u_x$};
\node (v1x) at (-1,0.5) {$v_x$};

\node (u2x) at (-3,1.5) {$u_x$};
\node (w2x) at (-1,1.5) {$w_x$};

\node (u3x) at (-3,2.5) {$u_x$};
\node (t3x) at (-1,2.5) {$t_x$};

\node (w1x) at (3,0.5) {$w_x$};
\node (t1x) at (1,0.5) {$t_x$};

\node (v2x) at (3,1.5) {$v_x$};
\node (t2x) at (1,1.5) {$t_x$};

\node (v3x) at (3,2.5) {$v_x$};
\node (w3x) at (1,2.5) {$w_x$};

\draw[->, to path={-| (\tikztotarget)}]
  (w1x) edge[bend left] (t1x) (t1x) edge[bend left] (w1x) (v2x) edge[bend left] (t2x) (t2x) edge[bend left] (v2x) (v3x) edge[bend left] (w3x) (w3x) edge[bend left] (v3x);

\draw[->, to path={-| (\tikztotarget)}]
  (u1x) edge[bend left] (v1x) (v1x) edge[bend left] (u1x) (u2x) edge[bend left] (w2x) (w2x) edge[bend left] (u2x) (u3x) edge[bend left] (t3x) (t3x) edge[bend left] (u3x);

\node (u1y) at (-3,-0.5) {$u_y$};
\node (v1y) at (-1,-0.5) {$v_y$};

\node (u2y) at (-3,-1.5) {$u_y$};
\node (w2y) at (-1,-1.5) {$w_y$};

\node (u3y) at (-3,-2.5) {$u_y$};
\node (t3y) at (-1,-2.5) {$t_y$};

\node (w1y) at (3,-0.5) {$w_y$};
\node (t1y) at (1,-0.5) {$t_y$};

\node (v2y) at (3,-1.5) {$v_y$};
\node (t2y) at (1,-1.5) {$t_y$};

\node (v3y) at (3,-2.5) {$v_y$};
\node (w3y) at (1,-2.5) {$w_y$};

\node (CXAll) at (0,1.5) {\huge $\mathcal{C}^{\textit{all}}_{\cal X}$};
\node (CYAll) at (0,-1.5) {\huge $\mathcal{C}^{\textit{all}}_{\cal Y}$};

\draw[->, to path={-| (\tikztotarget)}]
  (w1y) edge[bend left] (t1y) (t1y) edge[bend left] (w1y) (v2y) edge[bend left] (t2y) (t2y) edge[bend left] (v2y) (v3y) edge[bend left] (w3y) (w3y) edge[bend left] (v3y);

\draw[->, to path={-| (\tikztotarget)}]
  (u1y) edge[bend left] (v1y) (v1y) edge[bend left] (u1y) (u2y) edge[bend left] (w2y) (w2y) edge[bend left] (u2y) (u3y) edge[bend left] (t3y) (t3y) edge[bend left] (u3y);
  
 \node (u1y2) at (4.5,-0.5) {$u_y$};
\node (v1y2) at (6.5,-0.5) {$v_y$};

\node (u2y2) at (4.5,-1.5) {$u_y$};
\node (w2y2) at (6.5,-1.5) {$w_y$};

\node (u3y2) at (4.5,-2.5) {$u_y$};
\node (t3y2) at (6.5,-2.5) {$t_y$};

\node (w1y2) at (10.5,-0.5) {$w_y$};
\node (t1y2) at (8.5,-0.5) {$t_y$};

\node (v2y2) at (10.5,-1.5) {$v_y$};
\node (t2y2) at (8.5,-1.5) {$t_y$};

\node (v3y2) at (10.5,-2.5) {$v_y$};
\node (w3y2) at (8.5,-2.5) {$w_y$};

\draw[->, to path={-| (\tikztotarget)}]
  (w1y2) edge[bend left] (t1y2) (t1y2) edge[bend left] (w1y2) (v2y2) edge[bend left] (t2y2) (t2y2) edge[bend left] (v2y2) (v3y2) edge[bend left] (w3y2) (w3y2) edge[bend left] (v3y2);

\draw[->, to path={-| (\tikztotarget)}]
  (u1y2) edge[bend left] (v1y2) (v1y2) edge[bend left] (u1y2) (u2y2) edge[bend left] (w2y2) (w2y2) edge[bend left] (u2y2) (u3y2) edge[bend left] (t3y2) (t3y2) edge[bend left] (u3y2);

\node (u1x2) at (4.5,0.5) {$u_x$};
\node (v1x2) at (6.5,0.5) {$v_x$};

\node (u2x2) at (4.5,1.5) {$u_x$};
\node (w2x2) at (6.5,1.5) {$w_x$};

\node (u3x2) at (4.5,2.5) {$u_x$};
\node (t3x2) at (6.5,2.5) {$t_x$};

\node (w1x2) at (10.5,0.5) {$w_x$};
\node (t1x2) at (8.5,0.5) {$t_x$};

\node (v2x2) at (10.5,1.5) {$v_x$};
\node (t2x2) at (8.5,1.5) {$t_x$};

\node (v3x2) at (10.5,2.5) {$v_x$};
\node (w3x2) at (8.5,2.5) {$w_x$};

\draw[->, to path={-| (\tikztotarget)}]
  (w1x2) edge[bend left] (t1x2) (t1x2) edge[bend left] (w1x2) (v2x2) edge[bend left] (t2x2) (t2x2) edge[bend left] (v2x2) (v3x2) edge[bend left] (w3x2) (w3x2) edge[bend left] (v3x2);

\draw[->, to path={-| (\tikztotarget)}]
  (u1x2) edge[bend left] (v1x2) (v1x2) edge[bend left] (u1x2) (u2x2) edge[bend left] (w2x2) (w2x2) edge[bend left] (u2x2) (u3x2) edge[bend left] (t3x2) (t3x2) edge[bend left] (u3x2);

\draw[thick] (-7.5,3) rectangle (11.25,-3);
\draw[thick] (-7.5,0) -- (11.25,0);
\draw[thick] (3.75,3) -- (3.75,-3);
\draw[thick] (-3.75,3) -- (-3.75,-3);

\draw[dashed,red,thick] (4,2.95) rectangle (11,2.05);
\draw[dashed,brown,thick] (4,1.95) rectangle (11,1.05);
\draw[dashed,darkgreen,thick] (4,0.95) rectangle (11,0.05);
\node (CX0) at (7.5,2.475) {$\mathcal{C}^{0}_{\cal X}$};
\node (CX1) at (7.5,1.475) {$\mathcal{C}^{1}_{\cal X}$};
\node (CX2) at (7.5,0.475) {$\mathcal{C}^{2}_{\cal X}$};

\draw[dashed,brown,thick] (4,-2.95) rectangle (11,-2.05);
\draw[dashed,darkgreen,thick] (4,-1.95) rectangle (11,-1.05);
\draw[dashed,red,thick] (4,-0.95) rectangle (11,-0.05);

\node (CY0) at (7.5,-2.475) {$\mathcal{C}^{2}_{\cal Y}$};
\node (CY1) at (7.5,-1.475) {$\mathcal{C}^{1}_{\cal Y}$};
\node (CY2) at (7.5,-0.475) {$\mathcal{C}^{0}_{\cal Y}$};

\node (vx) at (-7,1.5) {\huge $V_{\mathcal{X}}$};
\node (shapeux) at (-4.1,0.6) {$u_x$};
\node (shapevx) at (-4.2,1.8) {$v_x$};
\node (shapewx) at (-5.2,2.4) {$w_x$};
\node (shapetx) at (-6,1) {$t_x$};

\draw (-4.5,1.8) circle (0.05cm);
\draw (-4.4,1) circle (0.05cm);
\draw (-5.2,2) circle (0.05cm);
\draw (-5.8,1.3) circle (0.05cm);

\node (vx) at (-7,-1.5) {\huge $V_{\mathcal{Y}}$};
\node (shapeuy) at (-4.5,-1) {$u_y$};
\node (shapevy) at (-4.2,-1.8) {$v_y$};
\node (shapewy) at (-5.2,-2.4) {$w_y$};
\node (shapety) at (-6,-1.3) {$t_y$};

\draw (-4.5,-1.8) circle (0.05cm);
\draw (-4.9,-1.2) circle (0.05cm);
\draw (-5.2,-2) circle (0.05cm);
\draw (-5.5,-1.3) circle (0.05cm);

\draw[->, to path={-| (\tikztotarget)}]
  (-5.7,1.2) edge[bend left] (-5.4,-1.2);
\node (PXY) at (-5.82,0.3) {$P_\mathcal{{XY}}$};

\end{tikzpicture}

%% file: figures/subsubiteration.tex
\definecolor{turquoise}{rgb}{0.45,0.55,0.65}
\definecolor{red}{rgb}{0.83,0.53,0.42}
\definecolor{blue}{rgb}{0.83,0.75,0.42}

\begin{tikzpicture}
\LARGE
\fill[blue] (0,0) circle (1.9);
\node[minimum size= 4pt] (X) at (0,0) {\huge $\mathcal{X}$};

\node (a0) at (-1, 0) {$ \alpha_0$};

\node (a1) at (1, 0) {$ \alpha_1$};
\node (ux) at (-1, 1) {$u_x$};
\node (wx) at (1, 1) {$w_x$};
\node (vx) at (-1, -1) {$v_x$};
\node (tx) at (1, -1) {$t_x$};
\draw[->, to path={-| (\tikztotarget)}]
  (ux) edge[bend right] (vx) (vx) edge[bend right] (ux) (wx) edge[bend right] (tx) (tx) edge[bend right] (wx);
  
\node[rotate = 45] (PXY) at (0.8,2.8) {\huge$P_\mathcal{{XY}(\alpha)}$};

\fill[turquoise] (4,3) circle (1.9);
\node (Y) at (4,3) {\huge $\mathcal{Y}$};

\node (b0) at (3, 3) {$ \beta_0$};
\node (b1) at (5, 3) {$ \beta_1$};
\node (uy) at (3, 4) {$u_y$};
\node (vy) at (5, 4) {$v_y$};
\node (wy) at (3, 2) {$w_y$};
\node (ty) at (5, 2) {$t_y$};
\draw[->, to path={-| (\tikztotarget)}]
  (uy) edge[bend right] (wy) (wy) edge[bend right] (uy) (vy) edge[bend right] (ty) (ty) edge[bend right] (vy);

\fill[red] (8,0) circle (1.9);

\node[rotate = -40] (PYZ) at (7.2,3) {\huge $P_\mathcal{{YZ}}(\beta)$};

\node (Z) at (8,0) {\huge $\mathcal{Z}$};
\node (uz) at (7, -1) {$u_z$};
\node (wz) at (9, -1) {$w_z$};
\node (vz) at (7, 1) {$v_z$};
\node (tz) at (9, 1) {$t_z$};

\node (PXZ) at (4,-2) {\huge $P_\mathcal{{XZ}} = P_\mathcal{XY}(\alpha) P_\mathcal{YZ}(\beta)$};

\draw[->, dotted, to path={-| (\tikztotarget)}]
  (ux) edge[bend left] (uy) (vx) edge[bend left] (vy) (tx) edge[bend left] (ty) (wx) edge[bend left] (wy) ;

\draw[<-,dotted, to path={-| (\tikztotarget)}]
  (uz) edge[bend right] (uy) (vz) edge[bend right] (vy) (tz) edge[bend right] (ty) (wz) edge[bend right] (wy) ;

\draw[<-,dotted, to path={-| (\tikztotarget)}]
  (uz) edge[bend left] (ux) (vz) edge[bend left] (vx) (tz) edge[bend left] (tx) (wz) edge[bend left] (wx) ;
    
\end{tikzpicture}

%% file: sections/5_experiments.tex
\section{Experimental Evaluation}\label{sec:experiments}

We compare against state-of-the-art multi-matching methods %
with a focus on quantum methods. 
We consider classical works for reference. 
All experiments use Python 3.9 on an Intel Core i7-8565U CPU with 8GB RAM and the D-Wave Advantage System 4.1 (accessed via Leap 2). 
We will release our code, which is accelerated using Numba.

\noindent\textbf{Hyperparameters.} We set $T{=}11$. %
We set the number of worst vertices $m$ to  $16 \%$ of the number of vertices $n$.

\noindent\textbf{Quantum Comparisons.} 
The closest quantum work, Q-Match \cite{SeelbachBenkner2021}, matches only two shapes. 
We consider two adaptations to multi-matching: 1) Q-MatchV2-cc, similar to our CCuantuMM, chooses an anchor and matches the other shapes pairwise to it, implicitly enforcing cycle consistency; and 2) Q-MatchV2-nc matches all pairs of shapes directly, without guaranteed cycle consistency. 
In both cases, we use our faster implementation and adapt our energy matrix schedule, which gives significantly better results.

\noindent\textbf{Classical Comparisons.} For reference, we also compare against the classical, non-learning-based  multi-matching state of the art: IsoMuSh~\cite{Gao2021}  and the \emph{synchronised} version of ZoomOut \cite{melzi2019zoomout}, which both guarantee vertex-wise cycle consistency across multiple shapes.

\noindent\textbf{Evaluation Metric.} 
We evaluate the correspondences using the Princeton benchmark protocol \cite{VladimirGPCK}. 
Given the ground-truth correspondences $P_{\cal I J}^{*}$ for matching the shape $\mathcal{I}$ to $\mathcal{J}$, the error of vertex $v \in \mathcal{I}$ under our estimated matching $P_{\cal I J}$ is given by the normalised geodesic distance: 
\begin{equation}
    e_v(P_{\cal I J}) = \frac{d_\mathcal{J}^{g}(v^\top P_{\cal I J},v^\top P_{\cal I J}^{*})}{\operatorname{diam}(\mathcal{J})}, \quad
\end{equation}
where $\operatorname{diam}(\cdot)$ is the shape diameter. %
We plot the fraction of errors that is below a threshold in a percentage-of-correct-keypoints (PCK) curve, where the threshold varies along the x-axis. %
As a summary metric, we also report the area-under-the-curve (AUC) of these PCK curves. 

\noindent\textbf{Datasets.} 
The \emph{FAUST} dataset~\cite{Bogo:CVPR:2014} contains real scans of ten humans in different poses. 
We use the registration subset with ten poses for each class and downsample to $500$ vertices. 
\emph{TOSCA}~\cite{bronstein2008numerical} has $76$ shapes from eight classes of humans and animals. 
We downsample to ${\sim}1000$ vertices. 
\emph{SMAL}~\cite{Zuffi:CVPR:2017} has scans of toy animals in arbitrary poses, namely %
$41$ non-isometric shapes from five classes %
registered to the same template. 
(\textit{E.g.,} the felidae (cats) class contains scans of lions, cats, and tigers.)
We downsample to $1000$ vertices. 
We use the same number of vertices as IsoMuSh~\cite{Gao2021}, except that they use $1000$ vertices for FAUST.

\subsection{Experiments on Real Quantum Annealer} 
\begin{figure}
    \centering
    \resizebox{0.49\linewidth}{!}{
    \input{figures/PCK_QPU_3_Shapes}}
    \resizebox{0.49\linewidth}{!}{
    \input{figures/PCK_QPU_10_shapes}}
    \caption{
    PCK curves for (left) two three-shape and (right) two ten-shape instances using QA and SA. %
    In each plot, we denote one instance by normal lines and the other one by dotted lines. 
    }
    \label{fig:QPU_three_ten}
\end{figure}
We run two three-shapes and two ten-shapes experiments with FAUST on a real QPU. 
However, since our QUBO matrices are dense, we effectively need to embed a clique on the QPU. 
(The supplement contains a detailed analysis of the minor embeddings and the solution quality.) %
Hence, we test a reduced version of our method with 20 worst vertices per shape (40 virtual qubits in total), as more would worsen results significantly on current hardware. 
To compensate for this change, we use more  iterations for the ten-shape experiments. 
We use 200 anneals per QUBO, the default annealing path, and the default annealing time of $20\mu\mathit{s}$. 
As standard chain strength, we choose $1.0001$ times the largest absolute value of entries in $Q$. 
Each ten-shape experiment takes about 10 minutes of QPU time. 
In total, our results took about 30 minutes of QPU time for a total of $5.5 \cdot 10^4$ QUBOs. 
QA under these settings achieves a similar performance as SA under the same settings (Fig.~\ref{fig:QPU_three_ten}). %
As QPU time is expensive and since we have just shown that SA performs comparably to a QPU in terms of result quality, we perform the remaining experiments with SA under our default settings, on classical hardware. 
This is common practice~\cite{SeelbachBenkner2021, Arrigoni2022, Zaech_2022_CVPR } since SA is conceptually close to QA.
For additional results, including results on the new
Zephyr hardware \cite{Zephyr2021}, we refer to the supplement.

\subsection{Comparison to Quantum and Classical SoTA} 

\begin{table}[h]
\centering

\resizebox{\linewidth}{!}{
\begin{tabular}{|c|ccc|ccc|}\hline
      & Ours            & Q-MatchV2-cc & Q-MatchV2-nc & IsoMuSh        & ZoomOut  & HKS \\ \hline\hline
FAUST & \textbf{0.989}  & 0.886        & 0.879        & \textit{0.974}          & 0.886    & 0.746\\ 
TOSCA & \textbf{0.967}  & 0.932        & 0.940        & \textit{0.952}          & 0.864    & 0.742 \\ 
SMAL  & \textit{0.866}  & 0.771        & 0.813        & \textbf{0.926}          & 0.851    & 0.544 \\ \hline 
\end{tabular}
}
\caption{AUC averaged over all classes of each dataset. For reference, we also include classical methods on the right. 
}
\label{Table:FAUST_TOSCA_SMAL}
\end{table}

\begin{figure}
   \centering
    \resizebox{\linewidth}{!}{
    \begin{tabular}{c|cccccc}
    
    \raisebox{-0.5\height}{\includegraphics[width=.18\linewidth]{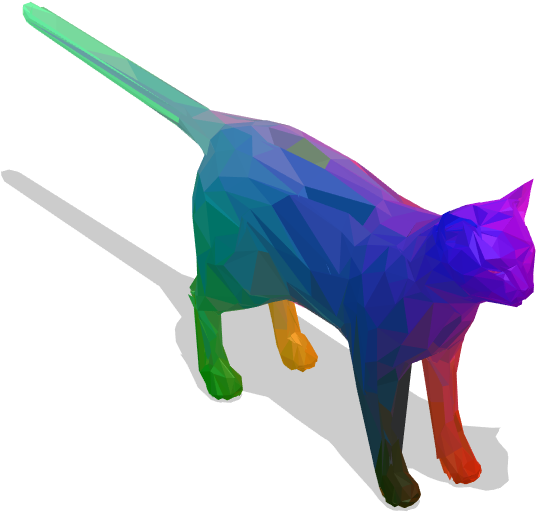}}
     &
    \raisebox{-0.5\height}{\includegraphics[width=.11\linewidth]{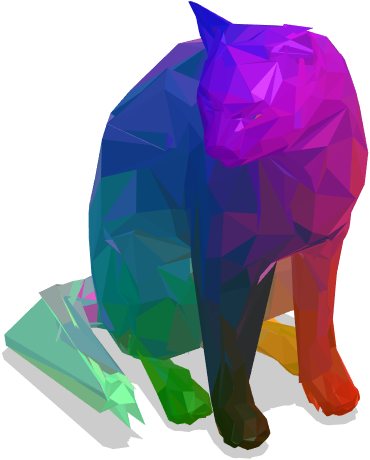}} &
    \raisebox{-0.5\height}{\includegraphics[width=.17\linewidth]{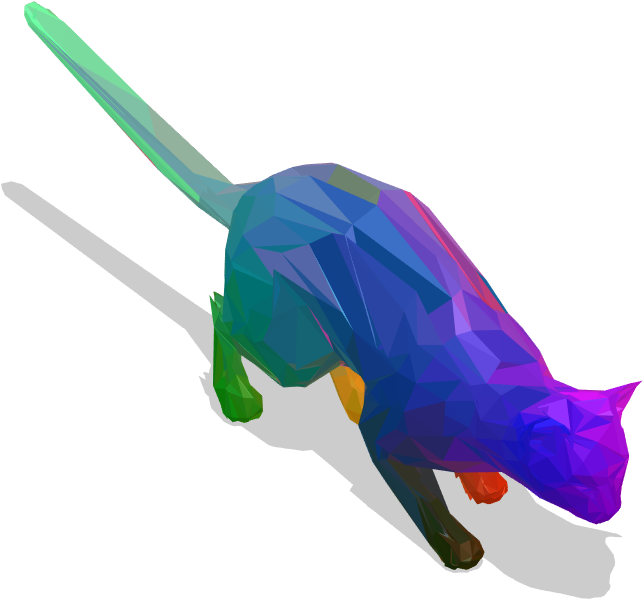}} & 
    \raisebox{-0.5\height}{\includegraphics[width=.13\linewidth]{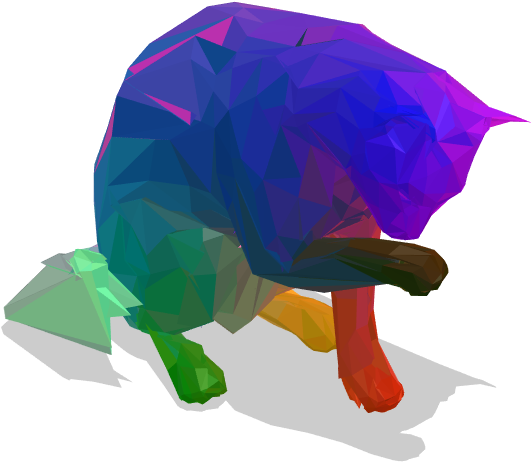}} &
    \raisebox{-0.5\height}{\includegraphics[width=.16\linewidth]{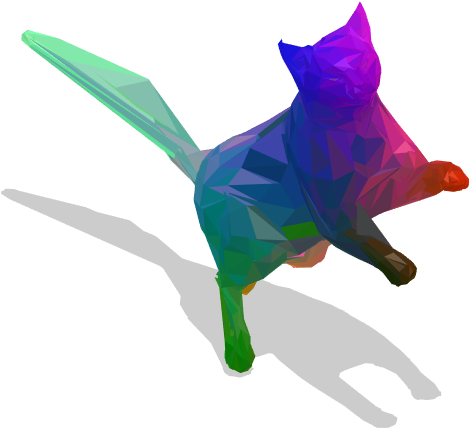}} & 
    \raisebox{-0.5\height}{\includegraphics[width=.11\linewidth]{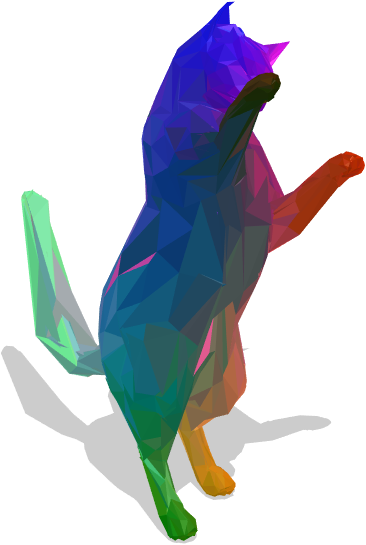}} &
    \rotatebox[origin=c]{270}{Ours} \\

    \raisebox{\height}{Source}
    &
    \raisebox{-0.5\height}{\includegraphics[width=.11\linewidth]{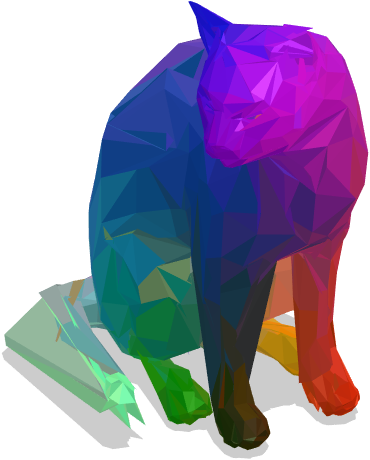}} &
    \raisebox{-0.5\height}{\includegraphics[width=.17\linewidth]{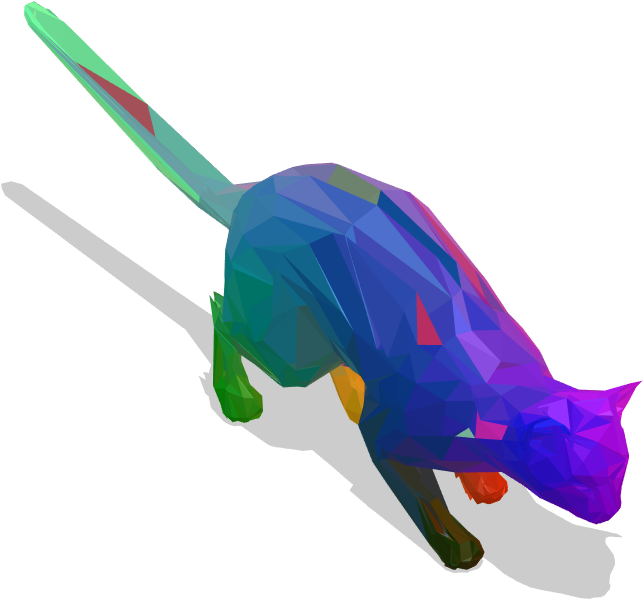}} &
    \raisebox{-0.5\height}{\includegraphics[width=.13\linewidth]{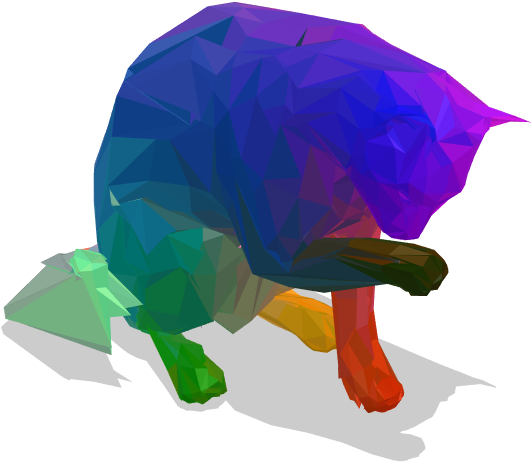}} &
    \raisebox{-0.5\height}{\includegraphics[width=.16\linewidth]{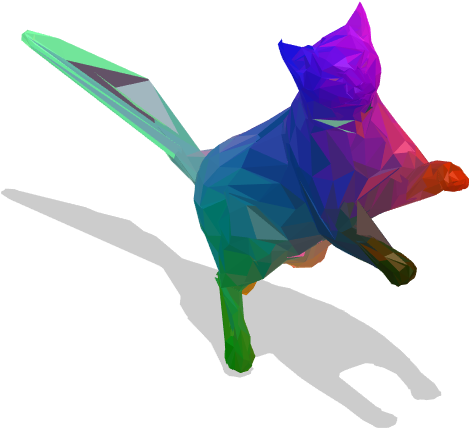}} &
    \raisebox{-0.5\height}{\includegraphics[width=.11\linewidth]{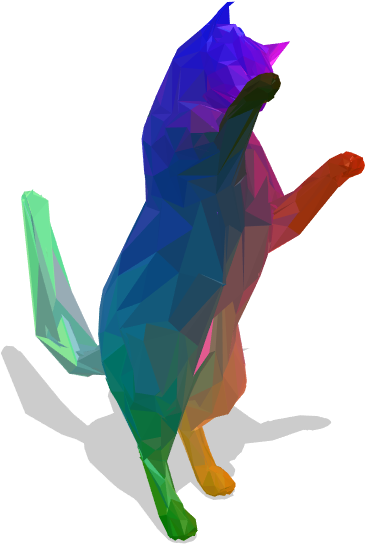}} &
    \rotatebox[origin=c]{270}{IsoMuSh}
    
    \end{tabular}
    }

    \caption{Qualitative results on the TOSCA  \cite{bronstein2008numerical} cat class. 
    We colour a source shape %
    and transfer this colouring to target shapes via the matches estimated by our method or IsoMuSh \cite{Gao2021}. %
    } 
    \label{fig:tosca_qualitative} 
\end{figure}

\begin{figure*}
    \centering
    
    \begin{subfigure}{0.32 \textwidth}
    \resizebox{\textwidth}{!}{
    \input{figures/PCK_FAUST}
     }
     \caption{FAUST \cite{Bogo:CVPR:2014}}
     \label{fig:FAUSTquant}
    \end{subfigure}
    \begin{subfigure}{0.32 \textwidth}
    \resizebox{\textwidth}{!}{
    \input{figures/PCK_TOSCA}
    }
     \caption{TOSCA \cite{bronstein2008numerical}}
     \label{fig:TOSCAquant}
    \end{subfigure}
    \begin{subfigure}{0.32 \textwidth}
    \resizebox{\textwidth}{!}{
     \input{figures/PCK_SMAL}}
      \caption{SMAL \cite{Zuffi:CVPR:2017}}
      \label{fig:SMALquant}
    
    \end{subfigure}
    \vspace{-1em}
    \caption{
    Quantitative results on all three datasets. %
    For each dataset, we match all shapes within a class and then plot the average PCK curve across classes. 
    We plot classical methods with dashed lines as they are only for reference. 
    HKS is our initialisation (see Sec.~\ref{sec:n_shapes}). 
    }
    \label{fig:FAUST_TOSCA_SMAL}
    \vspace{-1em}
\end{figure*}

\noindent\textbf{FAUST.} 
We outperform both quantum and classical prior work, as Fig.~\ref{fig:FAUSTquant} and Tab.~\ref{Table:FAUST_TOSCA_SMAL} show. 
Because we downsample FAUST more, IsoMuSh's results are better in our experiments than what Gao \textit{et al.}~\cite{Gao2021} report.

\noindent\textbf{Matching $\mathbf{100}$ Shapes.} 
Next, we demonstrate that, unlike IsoMuSh and ZoomOut, our approach can scale to matching all 100 shapes of FAUST. %
Fig.~\ref{fig:teaser} contains qualitative results.
Tab.~\ref{tab:runtime} compares the runtime of our method (using SA) to others. 
Only ours and Q-MatchV2-cc scale well to $100$ shapes while ZoomOut and IsoMuSh cannot. 

\begin{table}[h]
    \centering
    \resizebox{\linewidth}{!}{
    \begin{tabular}{|c|ccccc|}
    \hline
    \# Shapes       & Ours & Q-MatchV2-cc & Q-MatchV2-nc         & IsoMuSh & ZoomOut    \\
    \hline\hline
    10              & 97   & 16           & 81                   & $(4+)0.3$ & \textbf{4} \\
    100             & 1137 & \textbf{175} & ${\sim}8000^\dagger$ & OOM     & OOM        \\
    \hline
    \end{tabular}
    }
    \caption{Runtime (in \textit{min}) for FAUST. IsoMuSh uses ZoomOut for initialisation. 
    ``OOM''  (out of memory): memory requirements are infeasible. ``$^\dagger$'' denotes an estimate. %
    }
    \label{tab:runtime}
\end{table}

\noindent\textbf{TOSCA.} 
Fig.~\ref{fig:TOSCAquant} and Tab.~\ref{Table:FAUST_TOSCA_SMAL} show that our method achieves state-of-the-art results.
While IsoMuSh's PCK curve starts higher (better), the AUC in Tab.~\ref{Table:FAUST_TOSCA_SMAL} suggests that our method performs better overall. 
Fig.~\ref{fig:tosca_qualitative} has qualitative examples.

\noindent\textbf{SMAL.} 
Our CCuantuMM outperforms the quantum baselines, both in terms of PCK (Fig.~\ref{fig:SMALquant}) and AUC (Tab.~\ref{Table:FAUST_TOSCA_SMAL}). 
At the same time, it achieves performance on par with ZoomOut and below IsoMuSh. 
SMAL is considered the most difficult of the three datasets due to the challenging non-isometric deformations of its shapes. 
All methods thus show worse performance compared to FAUST and TOSCA.

\subsection{Ablation Studies} 
We perform an ablation study on FAUST to analyse how different components of our method affect the quality of the matchings. 
We refer to the supplement for more ablations. 

\begin{figure}%
    \resizebox{0.49\linewidth}{!}{
    \input{figures/PCK_Geo_Ablation}
    }
    \resizebox{0.49\linewidth}{!}{
    \input{figures/PCK_Shapes_Ablation}
    }
    \caption{We ablate (left) the usage of Gaussian kernels, and (right) the large-scale multi-shape setting. 
    Gaussian kernels improve the results greatly. 
    Matching more shapes improves results. 
    }
    \label{fig:Ablation}
\end{figure}

\noindent\textbf{Gaussian Energy Schedule.} 
Our schedule, which starts with geodesics and afterwards uses Gaussians, 
provides a significant performance gain over using only geodesics, under the same number of iterations, see Fig~\ref{fig:Ablation}.
That is because Gaussians better correct local errors in our approach. 

\noindent\textbf{Does Using More Shapes Improve Results?} 
We analyse what effect increasing the number of shapes $N$ has on the  matchings' quality. %
We first randomly select three shapes and run our method on them, to obtain the baseline. 
Next, we run our method again and again from scratch, each time adding one more shape to the previously used shapes. 
This isolates the effect of using more shapes from all other factors. 
In Fig.~\ref{fig:Ablation}, we plot the PCK curves \emph{for the three selected shapes}. %
We repeat this experiment for several randomly sampled instances. %
Our results show that including more shapes improves the matchings noticeably overall.  %

\subsection{Discussion and Limitations} 
Our method and all considered methods are based on intrinsic properties like geodesic distances. 
Thus, without left-right labels for initialisation, they would produce partial flips for inter-class instances in FAUST and intra-class instances in TOSCA and SMAL. 
For a large worst-vertices set, contemporary quantum hardware leads to embeddings (see Sec.~\ref{sec:aqc}) with long chains, which are unstable, degrading the result quality. 
Finally, while our method is currently slower in practice than SA, it would immediately benefit from the widely expected quantum advantage in the future. %

%% file: figures/PCK_QPU_3_Shapes.tex
\begin{tikzpicture}

\definecolor{darkslategray38}{RGB}{38,38,38}
\definecolor{lightgray204}{RGB}{204,204,204}
\definecolor{orange}{RGB}{255,165,0}

\definecolor{darkorange25512714}{RGB}{255,127,14}
\definecolor{steelblue31119180}{RGB}{31,119,180}

\large
\begin{axis}[
axis line style={lightgray204},
legend cell align={left},
legend style={
  fill opacity=0.8,
  draw opacity=1,
  text opacity=1,
  at={(0.97,0.03)},
  anchor=south east,
  draw=lightgray204
},
xtick = {0,0.05,0.10,0.15},
tick align=outside,
tick pos=left,
x grid style={lightgray204},
xlabel=\textcolor{darkslategray38}{\Large Geodesic Error},
xticklabel style={yshift= 5pt},
xmajorgrids,
xmin=0, xmax=0.15,
xtick style={color=darkslategray38},
y grid style={lightgray204},
ylabel=\textcolor{darkslategray38}{\Large \% Correspondences},
ymajorgrids,
ymin=40, ymax=100,
ytick style={color=darkslategray38},
yticklabel style={xshift= 5pt},
]

\addplot [ultra thick, darkorange25512714]
table {%
0 41.0358565737052
0.0015 41.0358565737052
0.003 41.0358565737052
0.0045 41.0358565737052
0.006 41.0358565737052
0.0075 41.1686586985392
0.009 41.8990703851262
0.0105 42.6958831341301
0.012 43.6918990703851
0.0135 44.8871181938911
0.015 46.0491367861886
0.0165 47.0451527224436
0.018 47.9747675962815
0.0195 49.3691899070385
0.021 51.2948207171315
0.0225 52.8884462151394
0.024 54.4488711819389
0.0255 55.7436918990704
0.027 57.6029216467464
0.0285 59.3957503320053
0.03 61.6201859229748
0.0315 63.7450199203187
0.033 65.5046480743692
0.0345 67.0318725099602
0.036 68.7583001328021
0.0375 69.9535192563081
0.039 71.2151394422311
0.0405 72.675962815405
0.042 74.1035856573705
0.0435 75.398406374502
0.045 76.4276228419655
0.0465 77.456839309429
0.048 78.6188579017264
0.0495 79.3824701195219
0.051 80.3120849933599
0.0525 81.3745019920319
0.054 82.2377158034529
0.0555 83.3333333333333
0.057 84.0969455511288
0.0585 84.8605577689243
0.06 85.6905710491368
0.0615 86.2549800796813
0.063 87.1513944223108
0.0645 87.5498007968127
0.066 88.0478087649402
0.0675 88.7450199203187
0.069 89.4754316069057
0.0705 90.2390438247012
0.072 90.9030544488712
0.0735 91.3346613545817
0.075 91.9654714475432
0.0765 92.5298804780876
0.078 93.2270916334661
0.0795 93.7250996015936
0.081 94.1567065073041
0.0825 94.3559096945551
0.084 94.6215139442231
0.0855 95.0531208499336
0.087 95.5843293492696
0.0885 95.8831341301461
0.09 96.0823373173971
0.0915 96.3479415670651
0.093 96.5139442231076
0.0945 96.6799468791501
0.096 96.8127490039841
0.0975 97.2111553784861
0.099 97.4103585657371
0.1005 97.5099601593625
0.102 97.5763612217795
0.1035 97.808764940239
0.105 97.875166002656
0.1065 98.00796812749
0.108 98.140770252324
0.1095 98.273572377158
0.111 98.472775564409
0.1125 98.67197875166
0.114 98.7051792828685
0.1155 98.738379814077
0.117 98.804780876494
0.1185 98.8379814077025
0.12 98.9043824701195
0.1215 99.0371845949535
0.123 99.0371845949535
0.1245 99.136786188579
0.126 99.203187250996
0.1275 99.203187250996
0.129 99.2363877822045
0.1305 99.269588313413
0.132 99.33598937583
0.1335 99.33598937583
0.135 99.402390438247
0.1365 99.4355909694555
0.138 99.468791500664
0.1395 99.468791500664
0.141 99.468791500664
0.1425 99.468791500664
0.144 99.5019920318725
0.1455 99.5019920318725
0.147 99.5019920318725
0.1485 99.5019920318725
0.15 99.5019920318725
};
\addplot [ultra thick, steelblue31119180]
table {%
0 53.5192563081009
0.0015 53.5192563081009
0.003 53.5192563081009
0.0045 53.5192563081009
0.006 53.5192563081009
0.0075 53.5856573705179
0.009 54.0504648074369
0.0105 54.6480743691899
0.012 55.5776892430279
0.0135 57.0385126162019
0.015 58.2005312084993
0.0165 58.9641434262948
0.018 59.6945551128818
0.0195 60.6573705179283
0.021 62.0185922974768
0.0225 63.9110225763612
0.024 65.3386454183267
0.0255 66.7330677290837
0.027 68.3598937583001
0.0285 70.1859229747676
0.03 71.9123505976095
0.0315 73.738379814077
0.033 75.199203187251
0.0345 76.460823373174
0.036 77.5564409030545
0.0375 78.9508632138114
0.039 80.2456839309429
0.0405 81.1752988047809
0.042 82.2045152722444
0.0435 83.3001328021248
0.045 84.0969455511288
0.0465 85.2921646746348
0.048 86.1553784860558
0.0495 87.1181938911023
0.051 87.7822045152722
0.0525 88.6454183266932
0.054 89.2098273572377
0.0555 90.3386454183267
0.057 90.7038512616202
0.0585 91.3014608233732
0.06 91.9654714475432
0.0615 92.2310756972112
0.063 92.5298804780877
0.0645 92.8286852589641
0.066 93.3266932270916
0.0675 93.9243027888446
0.069 94.4555112881806
0.0705 94.7543160690571
0.072 95.1859229747676
0.0735 95.6175298804781
0.075 95.8831341301461
0.0765 96.1487383798141
0.078 96.4807436918991
0.0795 96.7463479415671
0.081 96.9787516600266
0.0825 97.0451527224436
0.084 97.0783532536521
0.0855 97.2111553784861
0.087 97.4103585657371
0.0885 97.5763612217795
0.09 97.7091633466136
0.0915 97.9415670650731
0.093 98.0079681274901
0.0945 98.074369189907
0.096 98.1739707835325
0.0975 98.3067729083665
0.099 98.339973439575
0.1005 98.339973439575
0.102 98.3731739707835
0.1035 98.406374501992
0.105 98.472775564409
0.1065 98.5059760956175
0.108 98.5723771580345
0.1095 98.67197875166
0.111 98.7051792828685
0.1125 98.738379814077
0.114 98.804780876494
0.1155 98.8379814077025
0.117 98.9707835325365
0.1185 98.9707835325365
0.12 99.0371845949535
0.1215 99.1035856573705
0.123 99.1035856573705
0.1245 99.136786188579
0.126 99.203187250996
0.1275 99.3027888446215
0.129 99.402390438247
0.1305 99.4355909694555
0.132 99.4355909694555
0.1335 99.4355909694555
0.135 99.4355909694555
0.1365 99.4355909694555
0.138 99.4355909694555
0.1395 99.4355909694555
0.141 99.468791500664
0.1425 99.5019920318725
0.144 99.5019920318725
0.1455 99.535192563081
0.147 99.535192563081
0.1485 99.535192563081
0.15 99.535192563081
};
\addplot [ultra thick, darkorange25512714, dashed]
table {%
0 81.8061088977424
0.0015 81.9389110225764
0.003 81.9389110225764
0.0045 81.9389110225764
0.006 81.9389110225764
0.0075 82.1049136786189
0.009 82.6029216467464
0.0105 83.3997343957503
0.012 84.5617529880478
0.0135 85.2589641434263
0.015 86.0889774236388
0.0165 86.7197875166003
0.018 87.1181938911023
0.0195 87.7822045152723
0.021 88.9774236387782
0.0225 89.5750332005312
0.024 90.8366533864542
0.0255 91.5338645418327
0.027 92.1646746347942
0.0285 93.0942895086321
0.03 93.8579017264276
0.0315 94.4223107569721
0.033 94.8207171314741
0.0345 95.1527224435591
0.036 95.5179282868526
0.0375 95.6839309428951
0.039 96.0823373173971
0.0405 96.6135458167331
0.042 96.9787516600266
0.0435 97.2111553784861
0.045 97.4103585657371
0.0465 97.7423638778221
0.048 97.875166002656
0.0495 98.207171314741
0.051 98.539176626826
0.0525 98.738379814077
0.054 98.871181938911
0.0555 98.9043824701195
0.057 99.003984063745
0.0585 99.070385126162
0.06 99.136786188579
0.0615 99.2363877822045
0.063 99.33598937583
0.0645 99.33598937583
0.066 99.468791500664
0.0675 99.535192563081
0.069 99.535192563081
0.0705 99.6347941567065
0.072 99.7011952191235
0.0735 99.7011952191235
0.075 99.7011952191235
0.0765 99.734395750332
0.078 99.734395750332
0.0795 99.734395750332
0.081 99.734395750332
0.0825 99.734395750332
0.084 99.734395750332
0.0855 99.734395750332
0.087 99.734395750332
0.0885 99.734395750332
0.09 99.734395750332
0.0915 99.734395750332
0.093 99.734395750332
0.0945 99.734395750332
0.096 99.734395750332
0.0975 99.734395750332
0.099 99.734395750332
0.1005 99.734395750332
0.102 99.734395750332
0.1035 99.734395750332
0.105 99.734395750332
0.1065 99.734395750332
0.108 99.734395750332
0.1095 99.734395750332
0.111 99.734395750332
0.1125 99.734395750332
0.114 99.734395750332
0.1155 99.734395750332
0.117 99.734395750332
0.1185 99.734395750332
0.12 99.734395750332
0.1215 99.734395750332
0.123 99.800796812749
0.1245 99.800796812749
0.126 99.800796812749
0.1275 99.800796812749
0.129 99.800796812749
0.1305 99.800796812749
0.132 99.800796812749
0.1335 99.800796812749
0.135 99.8339973439575
0.1365 99.867197875166
0.138 99.867197875166
0.1395 99.867197875166
0.141 99.867197875166
0.1425 99.867197875166
0.144 99.867197875166
0.1455 99.867197875166
0.147 99.867197875166
0.1485 99.867197875166
0.15 99.867197875166
};

\addplot [ultra thick, steelblue31119180, dashed]
table {%
0 83.1341301460823
0.0015 83.2005312084993
0.003 83.2005312084993
0.0045 83.2669322709163
0.006 83.2669322709163
0.0075 83.7649402390438
0.009 84.2629482071713
0.0105 85.0597609561753
0.012 86.0889774236388
0.0135 87.0517928286853
0.015 87.4169986719788
0.0165 87.8154050464807
0.018 88.1806108897742
0.0195 89.3094289508632
0.021 90.4050464807437
0.0225 90.8698539176627
0.024 91.6334661354582
0.0255 92.1978751660026
0.027 92.4634794156706
0.0285 92.9946879150066
0.03 93.3266932270916
0.0315 93.9243027888446
0.033 94.5883134130146
0.0345 94.9535192563081
0.036 95.4847277556441
0.0375 95.7835325365206
0.039 96.0159362549801
0.0405 96.2815405046481
0.042 96.6135458167331
0.0435 97.1779548472775
0.045 97.4103585657371
0.0465 98.074369189907
0.048 98.5059760956175
0.0495 98.6387782204515
0.051 98.7715803452855
0.0525 98.7715803452855
0.054 98.871181938911
0.0555 98.871181938911
0.057 98.871181938911
0.0585 98.871181938911
0.06 98.9707835325365
0.0615 98.9707835325365
0.063 99.136786188579
0.0645 99.203187250996
0.066 99.203187250996
0.0675 99.203187250996
0.069 99.269588313413
0.0705 99.4355909694555
0.072 99.4355909694555
0.0735 99.4355909694555
0.075 99.468791500664
0.0765 99.5019920318725
0.078 99.5019920318725
0.0795 99.5019920318725
0.081 99.5019920318725
0.0825 99.535192563081
0.084 99.535192563081
0.0855 99.5683930942895
0.087 99.5683930942895
0.0885 99.5683930942895
0.09 99.5683930942895
0.0915 99.5683930942895
0.093 99.5683930942895
0.0945 99.5683930942895
0.096 99.601593625498
0.0975 99.6347941567065
0.099 99.7011952191235
0.1005 99.7011952191235
0.102 99.7011952191235
0.1035 99.7011952191235
0.105 99.7011952191235
0.1065 99.734395750332
0.108 99.734395750332
0.1095 99.734395750332
0.111 99.734395750332
0.1125 99.734395750332
0.114 99.734395750332
0.1155 99.734395750332
0.117 99.734395750332
0.1185 99.734395750332
0.12 99.734395750332
0.1215 99.734395750332
0.123 99.734395750332
0.1245 99.734395750332
0.126 99.734395750332
0.1275 99.734395750332
0.129 99.734395750332
0.1305 99.734395750332
0.132 99.734395750332
0.1335 99.734395750332
0.135 99.734395750332
0.1365 99.734395750332
0.138 99.734395750332
0.1395 99.734395750332
0.141 99.734395750332
0.1425 99.734395750332
0.144 99.734395750332
0.1455 99.734395750332
0.147 99.734395750332
0.1485 99.734395750332
0.15 99.734395750332
};
\end{axis}

\end{tikzpicture}

%% file: figures/PCK_QPU_10_shapes.tex
\begin{tikzpicture}

\definecolor{darkslategray38}{RGB}{38,38,38}
\definecolor{lightgray204}{RGB}{204,204,204}
\definecolor{orange}{RGB}{255,165,0}
\definecolor{darkorange25512714}{RGB}{255,127,14}
\definecolor{steelblue31119180}{RGB}{31,119,180}

\Large 
\begin{axis}[
axis line style={lightgray204},
legend cell align={left},
legend style={
  fill opacity=0.8,
  draw opacity=1,
  text opacity=1,
  at={(0.97,0.03)},
  anchor=south east,
  draw=lightgray204
},
xtick = {0,0.05,0.1,0.15},
tick align=outside,
tick pos=left,
x grid style={lightgray204},
xticklabel style={yshift= 5pt},
xlabel=\textcolor{darkslategray38}{Geodesic Error},
xmajorgrids,
xmin=0, xmax=0.15,
xtick style={color=darkslategray38},
y grid style={lightgray204},
ylabel=\textcolor{darkslategray38}{\% Correspondences},
ymajorgrids,
ymin=46.2039619300576, ymax=100,
ytick style={color=darkslategray38},
yticklabel style={xshift= 5pt},
]
\addplot [ultra thick, darkorange25512714]
table {%
0 54.763169544046
0.0015 54.7941567065073
0.003 54.7941567065073
0.0045 54.8317839752102
0.006 54.8605577689243
0.0075 55.3497122620629
0.009 56.3789287295263
0.0105 57.153607791058
0.012 58.0787959274015
0.0135 59.4510845506861
0.015 60.522355024347
0.0165 61.5316511730855
0.018 62.9681274900398
0.0195 64.6790615316512
0.021 66.4763169544046
0.0225 68.1474103585657
0.024 70.0177069499778
0.0255 71.7020805666224
0.027 73.481629039398
0.0285 75.1947764497565
0.03 76.8260292164674
0.0315 78.444001770695
0.033 79.8162903939796
0.0345 81.3036741921204
0.036 82.6715360779106
0.0375 83.8490482514387
0.039 85.0265604249668
0.0405 86.0358565737052
0.042 86.9853917662683
0.0435 87.8331119964586
0.045 88.5613103142984
0.0465 89.3426294820717
0.048 89.9513058875609
0.0495 90.5953961930057
0.051 91.3457281983178
0.0525 91.9344842850819
0.054 92.6228419654714
0.0555 93.0035413899956
0.057 93.4373616644533
0.0585 93.8291279327136
0.06 94.1722000885348
0.0615 94.572819831784
0.063 94.9513058875609
0.0645 95.2102700309872
0.066 95.4448871181939
0.0675 95.6883576803895
0.069 96.0292164674634
0.0705 96.281540504648
0.072 96.5073041168658
0.0735 96.6888003541389
0.075 96.8348826914563
0.0765 96.9654714475431
0.078 97.1536077910579
0.0795 97.3660911907924
0.081 97.5276671093404
0.0825 97.6029216467463
0.084 97.7467906153165
0.0855 97.846392208942
0.087 97.9570606463037
0.0885 98.0168216024789
0.09 98.074369189907
0.0915 98.1363435148295
0.093 98.1584772023019
0.0945 98.235945108455
0.096 98.3045595396193
0.0975 98.3554670208056
0.099 98.3820274457724
0.1005 98.3997343957503
0.102 98.435148295706
0.1035 98.4550686144311
0.105 98.4772023019035
0.1065 98.4904825143869
0.108 98.5413899955732
0.1095 98.5723771580345
0.111 98.5945108455068
0.1125 98.6033643204958
0.114 98.6188579017264
0.1155 98.6321381142098
0.117 98.6852589641434
0.1185 98.7118193891101
0.12 98.729526339088
0.1215 98.7472332890659
0.123 98.7649402390438
0.1245 98.7782204515272
0.126 98.8069942452412
0.1275 98.8424081451969
0.129 98.8733953076582
0.1305 98.8888888888889
0.132 98.9243027888446
0.1335 98.9397963700752
0.135 98.9508632138114
0.1365 98.9840637450199
0.138 99.0061974324922
0.1395 99.0239043824701
0.141 99.0393979637007
0.1425 99.0593182824258
0.144 99.0681717574147
0.1455 99.0836653386454
0.147 99.1146525011067
0.1485 99.1212926073484
0.15 99.1212926073484
};
\addlegendentry{QA one instance}
\addplot [ultra thick, steelblue31119180]
table {%
0 64.2762284196547
0.0015 64.3094289508632
0.003 64.3094289508632
0.0045 64.3315626383356
0.006 64.3470562195662
0.0075 64.601593625498
0.009 65.3386454183267
0.0105 66.1310314298362
0.012 67.211155378486
0.0135 68.738379814077
0.015 70.037627268703
0.0165 70.8277999114652
0.018 72.0274457724657
0.0195 73.5834440017707
0.021 74.9092518813634
0.0225 76.0756972111554
0.024 77.6361221779548
0.0255 78.9685701637893
0.027 80.5511288180611
0.0285 81.7795484727755
0.03 83.0101814962373
0.0315 84.4355909694555
0.033 85.6595838866755
0.0345 86.6356795042054
0.036 87.6073483842408
0.0375 88.5590969455511
0.039 89.2850818946437
0.0405 90.1195219123506
0.042 90.7613988490483
0.0435 91.3899955732625
0.045 91.8459495351926
0.0465 92.3284639220894
0.048 92.8773793714033
0.0495 93.3311199645861
0.051 93.6830455953962
0.0525 94.1057990261178
0.054 94.6746347941567
0.0555 95.0553342186808
0.057 95.4072598494909
0.0585 95.7702523240372
0.06 96.0823373173971
0.0615 96.385568835768
0.063 96.6091190792386
0.0645 96.7684816290394
0.066 96.9610447100487
0.0675 97.1336874723329
0.069 97.3882248782647
0.0705 97.5409473218237
0.072 97.6648959716688
0.0735 97.7689243027888
0.075 97.8707392651615
0.0765 97.9969012837538
0.078 98.1496237273129
0.0795 98.2293050022133
0.081 98.3355467020805
0.0825 98.4041611332447
0.084 98.5126162018592
0.0855 98.5635236830455
0.087 98.6299247454625
0.0885 98.6675520141655
0.09 98.7073926516157
0.0915 98.7250996015935
0.093 98.795927401505
0.0945 98.8689685701637
0.096 98.9065958388667
0.0975 98.9243027888445
0.099 98.966356795042
0.1005 98.9884904825143
0.102 99.0305444887117
0.1035 99.0349712262062
0.105 99.0504648074368
0.1065 99.0571049136785
0.108 99.0792386011508
0.1095 99.1146525011066
0.111 99.1456396635678
0.1125 99.1522797698095
0.114 99.1589198760512
0.1155 99.1633466135457
0.117 99.1699867197874
0.1185 99.1788401947763
0.12 99.194333776007
0.1215 99.1987605135014
0.123 99.2054006197431
0.1245 99.2120407259848
0.126 99.2275343072154
0.1275 99.2363877822044
0.129 99.2408145196988
0.1305 99.2496679946878
0.132 99.260734838424
0.1335 99.2651615759184
0.135 99.2695883134129
0.1365 99.2762284196546
0.138 99.2806551571491
0.1395 99.2939353696325
0.141 99.2939353696325
0.1425 99.2939353696325
0.144 99.3005754758742
0.1455 99.3050022133686
0.147 99.3116423196103
0.1485 99.3337760070827
0.15 99.3337760070827
};
\addlegendentry{SA one instance}
\addplot [ultra thick, darkorange25512714, dashed]
table {%
0 48.7339530765826
0.0015 48.7339530765826
0.003 48.7339530765826
0.0045 48.7516600265604
0.006 48.8468348826915
0.0075 49.2452412571934
0.009 49.904825143869
0.0105 50.7326250553342
0.012 51.6998671978752
0.0135 53.0079681274901
0.015 54.3691899070385
0.0165 55.7348384240815
0.018 57.1425409473218
0.0195 58.6077910579902
0.021 60.028773793714
0.0225 61.4320495794599
0.024 63.0787959274015
0.0255 64.8517042939353
0.027 66.3988490482514
0.0285 68.2558654271802
0.03 69.9070385126162
0.0315 71.6777335104028
0.033 73.2669322709163
0.0345 74.8583444001771
0.036 76.1620185922975
0.0375 77.7777777777778
0.039 79.3824701195219
0.0405 80.7547587428065
0.042 82.2797698096503
0.0435 83.6210712704736
0.045 84.9136786188579
0.0465 86.157591854803
0.048 87.09606020363
0.0495 87.8773793714033
0.051 88.6542718016822
0.0525 89.3758300132802
0.054 90.0553342186808
0.0555 90.7724656927844
0.057 91.4032757857459
0.0585 91.9853917662683
0.06 92.4015050907481
0.0615 92.8508189464365
0.063 93.302346170872
0.0645 93.6564851704294
0.066 93.9597166888004
0.0675 94.2585214696769
0.069 94.5573262505533
0.0705 94.8251438689686
0.072 95.1106684373617
0.0735 95.3740593182824
0.075 95.5710491367862
0.0765 95.8410801239487
0.078 96.0292164674635
0.0795 96.1974324922532
0.081 96.319167773351
0.0825 96.4873837981408
0.084 96.6533864541833
0.0855 96.8016821602479
0.087 96.9300575475874
0.0885 97.0473660911908
0.09 97.1226206285967
0.0915 97.1735281097831
0.093 97.3173970783532
0.0945 97.4037184594953
0.096 97.4789729969013
0.0975 97.5542275343072
0.099 97.6029216467463
0.1005 97.6803895528995
0.102 97.7357237715803
0.1035 97.8308986277113
0.105 97.866312527667
0.1065 97.9371403275785
0.108 98.00796812749
0.1095 98.0677290836653
0.111 98.0898627711376
0.1125 98.1274900398406
0.114 98.1562638335546
0.1155 98.1872509960158
0.117 98.2536520584329
0.1185 98.2824258521469
0.12 98.3156263833555
0.1215 98.3444001770695
0.123 98.3820274457724
0.1245 98.4152279769809
0.126 98.4395750332005
0.1275 98.444001770695
0.129 98.4528552456839
0.1305 98.4772023019035
0.132 98.5059760956175
0.1335 98.5214696768481
0.135 98.5502434705622
0.1365 98.5568835768039
0.138 98.5635236830456
0.1395 98.5834440017706
0.141 98.5989375830013
0.1425 98.6033643204957
0.144 98.6166445329791
0.1455 98.625498007968
0.147 98.6542718016821
0.1485 98.6896857016378
0.15 98.6896857016378
};
\addlegendentry{QA another instance}

\addplot [ultra thick, steelblue31119180, dashed]
table {%
0 51.5670650730412
0.0015 51.5670650730412
0.003 51.5670650730412
0.0045 51.5825586542718
0.006 51.6865869853918
0.0075 52.1359008410801
0.009 52.7379371403276
0.0105 53.643204957946
0.012 54.5374059318282
0.0135 55.8277999114653
0.015 57.2244355909694
0.0165 58.5480301018149
0.018 59.8140770252324
0.0195 61.4718902169101
0.021 62.8950863213812
0.0225 64.0925188136344
0.024 65.7835325365206
0.0255 67.6272687029659
0.027 69.0814519698982
0.0285 70.8742806551572
0.03 72.286409915892
0.0315 73.9729969012838
0.033 75.5821159805223
0.0345 77.0075254537406
0.036 78.1230633023461
0.0375 79.5683930942895
0.039 80.8676405489155
0.0405 82.0562195661797
0.042 83.3266932270916
0.0435 84.4245241257193
0.045 85.4625940681717
0.0465 86.498450641877
0.048 87.4767596281541
0.0495 88.273572377158
0.051 89.1235059760957
0.0525 89.7742363877822
0.054 90.4603806994245
0.0555 91.2439132359451
0.057 91.8680832226649
0.0585 92.4302788844621
0.06 92.8995130588756
0.0615 93.2514386896857
0.063 93.6387782204515
0.0645 93.9729969012838
0.066 94.269588313413
0.0675 94.6038069942452
0.069 94.8893315626383
0.0705 95.1571491810536
0.072 95.5444887118194
0.0735 95.7613988490483
0.075 95.9650287737937
0.0765 96.2416998671979
0.078 96.3811420982736
0.0795 96.5227976980965
0.081 96.6555998229305
0.0825 96.8238158477202
0.084 97.0340858787074
0.0855 97.0894200973882
0.087 97.2355024347056
0.0885 97.3638778220451
0.09 97.4745462594068
0.0915 97.5830013280212
0.093 97.6582558654271
0.0945 97.7733510402833
0.096 97.8906595838867
0.0975 97.9482071713147
0.099 98.0013280212483
0.1005 98.0389552899512
0.102 98.065515714918
0.1035 98.0832226648959
0.105 98.1606905710491
0.1065 98.2005312084993
0.108 98.2470119521912
0.1095 98.2868525896414
0.111 98.3134130146082
0.1125 98.3488269145639
0.114 98.3953076582558
0.1155 98.4285081894643
0.117 98.4661354581673
0.1185 98.4949092518813
0.12 98.5170429393537
0.1215 98.5546702080566
0.123 98.5812306330234
0.1245 98.5945108455068
0.126 98.6277113767153
0.1275 98.6586985391766
0.129 98.6808322266489
0.1305 98.700752545374
0.132 98.7162461266047
0.1335 98.7273129703408
0.135 98.7472332890659
0.1365 98.7627268702965
0.138 98.767153607791
0.1395 98.7915006640106
0.141 98.8379814077025
0.1425 98.8689685701637
0.144 98.8955289951305
0.1455 98.9021691013722
0.147 98.9243027888446
0.1485 98.9619300575475
0.15 98.9619300575475
};
\addlegendentry{SA another instance}
\end{axis}

\end{tikzpicture}

%% file: figures/PCK_FAUST.tex
\begin{tikzpicture}

\definecolor{crimson2143940}{RGB}{214,39,40}
\definecolor{darkorange25512714}{RGB}{255,127,14}
\definecolor{darkslategray38}{RGB}{38,38,38}
\definecolor{forestgreen4416044}{RGB}{44,160,44}
\definecolor{lightgray204}{RGB}{204,204,204}
\definecolor{mediumpurple148103189}{RGB}{148,103,189}
\definecolor{sienna1408675}{RGB}{140,86,75}
\definecolor{steelblue31119180}{RGB}{31,119,180}

\Large 

\begin{axis}[
axis line style={lightgray204},
legend cell align={left},
legend style={
  fill opacity=0.8,
  draw opacity=1,
  text opacity=1,
  at={(0.97,0.03)},
  anchor=south east,
  draw=lightgray204,
},
xtick = {0,0.05,0.10,0.15},
xticklabel style={yshift= 5pt},
yticklabel style={xshift= 5pt},
ytick = {60,70,80,90,100},
tick align=outside,
tick pos=left,
x grid style={lightgray204},
xlabel=\textcolor{darkslategray38}{Geodesic Error},
xmajorgrids,
xmin=0, xmax=0.15,
xtick style={color=darkslategray38},
y grid style={lightgray204},
ylabel=\textcolor{darkslategray38}{\% Correspondences},
ymajorgrids,
ymin=47.0120185922975, ymax=100,
ytick style={color=darkslategray38},
]
\addplot [ultra thick, steelblue31119180]
table {%
0 93.4019477644976
0.0015 93.4021691013723
0.003 93.4074811863656
0.0045 93.4229747675963
0.006 93.4645861000443
0.0075 93.6702080566622
0.009 93.9803010181496
0.0105 94.4156706507304
0.012 94.7726870296591
0.0135 95.1188579017264
0.015 95.4712262062859
0.0165 95.8612217795485
0.018 96.199203187251
0.0195 96.4645861000443
0.021 96.8025675077468
0.0225 97.1790615316512
0.024 97.4922532093847
0.0255 97.7833111996458
0.027 98.0336432049579
0.0285 98.2386011509517
0.03 98.4420097388225
0.0315 98.6250553342187
0.033 98.7990261177512
0.0345 98.9389110225763
0.036 99.0894200973882
0.0375 99.1963258078796
0.039 99.2682602921646
0.0405 99.3530323151836
0.042 99.4156706507303
0.0435 99.4873837981407
0.045 99.5559982293049
0.0465 99.6177512173528
0.048 99.6757414785303
0.0495 99.7153607791058
0.051 99.7563081009296
0.0525 99.7886232846392
0.054 99.8065515714918
0.0555 99.8247011952191
0.057 99.839973439575
0.0585 99.847941567065
0.06 99.8601150951748
0.0615 99.8634351482957
0.063 99.8691899070385
0.0645 99.8747233289066
0.066 99.881584772023
0.0675 99.8851261620185
0.069 99.8860115095174
0.0705 99.8906595838866
0.072 99.893536963258
0.0735 99.8968570163789
0.075 99.9026117751217
0.0765 99.9052678176184
0.078 99.9061531651173
0.0795 99.9065958388667
0.081 99.9096945551128
0.0825 99.9163346613546
0.084 99.9196547144754
0.0855 99.9225320938468
0.087 99.9260734838424
0.0885 99.9267374944666
0.09 99.9280655157149
0.0915 99.9285081894643
0.093 99.9285081894643
0.0945 99.9293935369632
0.096 99.9318282425852
0.0975 99.9322709163346
0.099 99.9324922532093
0.1005 99.9329349269588
0.102 99.935148295706
0.1035 99.9393536963258
0.105 99.9393536963258
0.1065 99.9424524125719
0.108 99.9424524125719
0.1095 99.9431164231961
0.111 99.9446657813191
0.1125 99.945551128818
0.114 99.9459938025675
0.1155 99.9464364763169
0.117 99.9475431606906
0.1185 99.9510845506861
0.12 99.951969898185
0.1215 99.9528552456839
0.123 99.9532979194334
0.1245 99.954404603807
0.126 99.9552899513059
0.1275 99.95595396193
0.129 99.9561752988048
0.1305 99.9572819831784
0.132 99.9581673306773
0.1335 99.9586100044267
0.135 99.9586100044267
0.1365 99.9586100044267
0.138 99.9586100044267
0.1395 99.9588313413014
0.141 99.9590526781762
0.1425 99.9610447100487
0.144 99.9612660469234
0.1455 99.9612660469234
0.147 99.9612660469234
0.1485 99.9614873837981
0.15 99.9614873837981
};
\addplot [ultra thick, mediumpurple148103189]
table {%
0 53.2649402390438
0.0015 53.2689243027888
0.003 53.2749003984064
0.0045 53.3067729083665
0.006 53.4820717131474
0.0075 53.99203187251
0.009 54.804780876494
0.0105 55.898406374502
0.012 57.1294820717132
0.0135 58.7529880478088
0.015 60.2490039840637
0.0165 61.8705179282868
0.018 63.6713147410359
0.0195 65.304780876494
0.021 67.0019920318725
0.0225 68.7211155378486
0.024 70.6035856573705
0.0255 72.3824701195219
0.027 74.1314741035857
0.0285 75.8027888446215
0.03 77.5298804780876
0.0315 79.292828685259
0.033 80.8466135458167
0.0345 82.2729083665339
0.036 83.5577689243028
0.0375 84.7370517928287
0.039 85.996015936255
0.0405 87.0776892430279
0.042 88.0239043824701
0.0435 88.8545816733068
0.045 89.6812749003984
0.0465 90.4282868525897
0.048 91.1394422310757
0.0495 91.7430278884462
0.051 92.3346613545817
0.0525 92.8784860557768
0.054 93.4143426294821
0.0555 93.8346613545817
0.057 94.2908366533864
0.0585 94.6713147410359
0.06 94.9880478087649
0.0615 95.2490039840637
0.063 95.5278884462151
0.0645 95.7669322709163
0.066 95.9601593625498
0.0675 96.1454183266932
0.069 96.3446215139442
0.0705 96.5358565737051
0.072 96.6733067729083
0.0735 96.8187250996015
0.075 96.9521912350597
0.0765 97.0617529880478
0.078 97.1533864541832
0.0795 97.2470119521912
0.081 97.3147410358565
0.0825 97.3685258964142
0.084 97.4581673306772
0.0855 97.5318725099601
0.087 97.6055776892429
0.0885 97.6713147410358
0.09 97.7609561752987
0.0915 97.8147410358565
0.093 97.8705179282868
0.0945 97.9302788844621
0.096 97.9800796812748
0.0975 98.0338645418326
0.099 98.0776892430278
0.1005 98.121513944223
0.102 98.1693227091633
0.1035 98.2051792828685
0.105 98.2410358565736
0.1065 98.2689243027888
0.108 98.2948207171314
0.1095 98.3187250996015
0.111 98.3525896414342
0.1125 98.3824701195218
0.114 98.4083665338645
0.1155 98.418326693227
0.117 98.4362549800796
0.1185 98.4501992031872
0.12 98.4641434262947
0.1215 98.4860557768924
0.123 98.4980079681274
0.1245 98.5059760956175
0.126 98.5278884462151
0.1275 98.5378486055776
0.129 98.5478087649402
0.1305 98.5717131474103
0.132 98.5776892430278
0.1335 98.5896414342629
0.135 98.6055776892429
0.1365 98.6235059760955
0.138 98.6314741035856
0.1395 98.6374501992031
0.141 98.6414342629481
0.1425 98.6494023904382
0.144 98.6633466135457
0.1455 98.6673306772908
0.147 98.6752988047808
0.1485 98.6832669322708
0.15 98.6832669322708
};
\addplot [ultra thick, crimson2143940]
table {%
0 58.4123505976096
0.0015 58.4123505976096
0.003 58.4282868525896
0.0045 58.4721115537848
0.006 58.5876494023904
0.0075 58.9541832669323
0.009 59.5836653386454
0.0105 60.4840637450199
0.012 61.4820717131474
0.0135 62.7171314741036
0.015 64.0637450199203
0.0165 65.2749003984064
0.018 66.601593625498
0.0195 68.1215139442231
0.021 69.6254980079681
0.0225 71.0478087649402
0.024 72.609561752988
0.0255 74.0418326693227
0.027 75.4003984063745
0.0285 76.8665338645418
0.03 78.292828685259
0.0315 79.6573705179283
0.033 80.8705179282868
0.0345 82.1573705179283
0.036 83.3247011952191
0.0375 84.4501992031873
0.039 85.3745019920319
0.0405 86.2808764940239
0.042 87.1035856573705
0.0435 87.7828685258964
0.045 88.5119521912351
0.0465 89.1852589641434
0.048 89.7390438247012
0.0495 90.308764940239
0.051 90.8167330677291
0.0525 91.2749003984063
0.054 91.7250996015936
0.0555 92.1115537848605
0.057 92.4183266932271
0.0585 92.7270916334661
0.06 93.0239043824701
0.0615 93.2649402390438
0.063 93.5278884462151
0.0645 93.7629482071713
0.066 93.9840637450199
0.0675 94.1474103585657
0.069 94.3247011952191
0.0705 94.4860557768924
0.072 94.6414342629482
0.0735 94.7729083665338
0.075 94.8904382470119
0.0765 94.9920318725099
0.078 95.1035856573704
0.0795 95.1872509960159
0.081 95.2609561752987
0.0825 95.3227091633465
0.084 95.4003984063744
0.0855 95.4760956175298
0.087 95.5358565737051
0.0885 95.5796812749003
0.09 95.6434262948206
0.0915 95.715139442231
0.093 95.7589641434262
0.0945 95.7828685258963
0.096 95.816733067729
0.0975 95.8505976095617
0.099 95.8984063745019
0.1005 95.9163346613545
0.102 95.9482071713147
0.1035 95.9661354581673
0.105 95.9880478087649
0.1065 96.00796812749
0.108 96.0258964143426
0.1095 96.0577689243027
0.111 96.0737051792828
0.1125 96.0956175298804
0.114 96.121513944223
0.1155 96.1414342629481
0.117 96.1553784860557
0.1185 96.1812749003983
0.12 96.1992031872509
0.1215 96.2231075697211
0.123 96.2390438247011
0.1245 96.2609561752987
0.126 96.2808764940238
0.1275 96.2908366533864
0.129 96.3007968127489
0.1305 96.3346613545816
0.132 96.3446215139441
0.1335 96.3565737051792
0.135 96.3784860557768
0.1365 96.4043824701195
0.138 96.4262948207171
0.1395 96.4402390438246
0.141 96.4482071713147
0.1425 96.4661354581673
0.144 96.4820717131473
0.1455 96.4900398406374
0.147 96.503984063745
0.1485 96.51593625498
0.15 96.51593625498
};

\addplot [ultra thick,dashed, darkorange25512714]
table {%
0 91.6682988047811
0.0015 91.8573559096949
0.003 92.1534072598499
0.0045 92.2699092518816
0.006 92.4444147853034
0.0075 92.6561049136786
0.009 92.9077742363879
0.0105 93.1210084108013
0.012 93.3243311199647
0.0135 93.5868308986276
0.015 93.820584329349
0.0165 93.9781297034086
0.018 94.1964696768483
0.0195 94.4627162461269
0.021 94.6034532979197
0.0225 94.7307100486944
0.024 94.866814962373
0.0255 95.0051606905713
0.027 95.1204501992035
0.0285 95.2372828685262
0.03 95.3412611775124
0.0315 95.4511212926075
0.033 95.5294431164233
0.0345 95.6038857901726
0.036 95.6777923860115
0.0375 95.7374249667994
0.039 95.7899269588312
0.0405 95.8569738822487
0.042 95.9315214696767
0.0435 96.0127498893314
0.045 96.1053359893757
0.0465 96.1861797255421
0.048 96.2939172200086
0.0495 96.4090708277998
0.051 96.5369482071712
0.0525 96.667708277999
0.054 96.7856184152279
0.0555 96.8869433377599
0.057 96.9861580345284
0.0585 97.0896980965028
0.06 97.2038640991587
0.0615 97.3278587870737
0.063 97.4450053120849
0.0645 97.5617963700753
0.066 97.6993984063747
0.0675 97.8068924302792
0.069 97.9534856131036
0.0705 98.0742872952638
0.072 98.1945254537411
0.0735 98.304020362993
0.075 98.389926516158
0.0765 98.4715276671097
0.078 98.5557777777782
0.0795 98.6202328463926
0.081 98.6725170429397
0.0825 98.7215692784422
0.084 98.7778260292168
0.0855 98.8353948649849
0.087 98.8738255865432
0.0885 98.9146892430284
0.09 98.9640730411692
0.0915 99.0067091633472
0.093 99.0378357680395
0.0945 99.0886843736173
0.096 99.1400743691906
0.0975 99.1785024347063
0.099 99.2314125719351
0.1005 99.2930987162467
0.102 99.3449207613994
0.1035 99.395184594954
0.105 99.457755201417
0.1065 99.503370075255
0.108 99.5598401947769
0.1095 99.5897419212044
0.111 99.6271522797702
0.1125 99.6498583444005
0.114 99.6764285081898
0.1155 99.6943700752548
0.117 99.7095489154496
0.1185 99.7231677733513
0.12 99.737783089863
0.1215 99.7508472775567
0.123 99.7730004426739
0.1245 99.786519698982
0.126 99.7996024789731
0.1275 99.8113652058434
0.129 99.8235599822932
0.1305 99.8379654714476
0.132 99.8480455953963
0.1335 99.8579132359452
0.135 99.8658946436477
0.1365 99.8769853917663
0.138 99.8854218680833
0.1395 99.8899667994688
0.141 99.8975090748119
0.1425 99.907606020363
0.144 99.9174860557769
0.1455 99.920591854803
0.147 99.924586542718
0.1485 99.9319065958389
0.15 99.9319065958389
};
\addplot [ultra thick,dashed, forestgreen4416044]
table {%
0 90.8100929614874
0.0015 90.8100929614874
0.003 90.8100929614874
0.0045 90.8123063302346
0.006 90.8277999114652
0.0075 90.9074811863656
0.009 90.9473218238158
0.0105 91.1598052235503
0.012 91.3877822045153
0.0135 91.6046923417441
0.015 91.8946436476317
0.0165 92.2000885347499
0.018 92.4833997343958
0.0195 92.921646746348
0.021 93.149623727313
0.0225 93.3532536520584
0.024 93.5502434705622
0.0255 93.7782204515272
0.027 94.0283311199645
0.0285 94.3470562195661
0.03 94.5861000442673
0.0315 94.9933598937582
0.033 95.2324037184595
0.0345 95.4493138556883
0.036 95.6086764054891
0.0375 95.7259849490926
0.039 95.8676405489155
0.0405 96.091190792386
0.042 96.2571934484285
0.0435 96.4696768481629
0.045 96.6865869853918
0.0465 96.9389110225763
0.048 97.0451527224436
0.0495 97.2244355909695
0.051 97.3417441345728
0.0525 97.476759628154
0.054 97.6361221779548
0.0555 97.7711376715361
0.057 97.8950863213812
0.0585 97.9814077025233
0.06 98.0544488711819
0.0615 98.1319167773351
0.063 98.2115980522355
0.0645 98.273572377158
0.066 98.3554670208056
0.0675 98.4484285081895
0.069 98.4882691456396
0.0705 98.5369632580787
0.072 98.6033643204957
0.0735 98.6321381142098
0.075 98.6963258078795
0.0765 98.7450199203187
0.078 98.8069942452412
0.0795 98.8778220451527
0.081 98.9243027888446
0.0825 98.9420097388224
0.084 98.9707835325365
0.0855 99.0128375387339
0.087 99.0548915449313
0.0885 99.1367861885789
0.09 99.1899070385125
0.0915 99.2386011509517
0.093 99.2961487383797
0.0945 99.327135900841
0.096 99.3714032757857
0.0975 99.422310756972
0.099 99.4532979194333
0.1005 99.4975652943779
0.102 99.5130588756086
0.1035 99.5396193005754
0.105 99.5706064630366
0.1065 99.6170872067285
0.108 99.6436476316953
0.1095 99.6702080566622
0.111 99.6857016378928
0.1125 99.7100486941123
0.114 99.7100486941123
0.1155 99.7211155378485
0.117 99.7410358565736
0.1185 99.7521027003098
0.12 99.7609561752987
0.1215 99.7675962815404
0.123 99.7720230190348
0.1245 99.7786631252766
0.126 99.7919433377599
0.1275 99.8007968127489
0.129 99.8140770252323
0.1305 99.8185037627268
0.132 99.8339973439574
0.1335 99.8428508189463
0.135 99.8517042939353
0.1365 99.8517042939353
0.138 99.8561310314297
0.1395 99.8561310314297
0.141 99.8627711376714
0.1425 99.8716246126604
0.144 99.8782647189021
0.1455 99.887118193891
0.147 99.8893315626383
0.1485 99.8937583001328
0.15 99.8937583001328
};

\addplot [ultra thick,dashed, sienna1408675]
table {%
0 49.5334218680832
0.0015 49.5356352368305
0.003 49.5374059318282
0.0045 49.5542275343072
0.006 49.622399291722
0.0075 49.852810978309
0.009 50.1963258078796
0.0105 50.6945551128818
0.012 51.315183709606
0.0135 52.1086764054892
0.015 52.9134572819832
0.0165 53.6142098273572
0.018 54.3846834882691
0.0195 55.2233289065958
0.021 56.0903054448871
0.0225 56.9612660469234
0.024 57.7932713590084
0.0255 58.6648959716689
0.027 59.566401062417
0.0285 60.4922532093847
0.03 61.4444444444444
0.0315 62.4065958388667
0.033 63.3114209827357
0.0345 64.199203187251
0.036 65.0396193005755
0.0375 65.8306772908366
0.039 66.6102257636122
0.0405 67.3461708720673
0.042 68.0745905267818
0.0435 68.7605135015494
0.045 69.4428950863214
0.0465 70.083222664896
0.048 70.6945551128818
0.0495 71.2870739265162
0.051 71.8893315626383
0.0525 72.4814077025232
0.054 73.0004426737494
0.0555 73.535192563081
0.057 74.037627268703
0.0585 74.5216910137229
0.06 74.9820717131474
0.0615 75.4342629482072
0.063 75.8875608676405
0.0645 76.2915006640106
0.066 76.6540504648074
0.0675 77.0644090305445
0.069 77.4891544931385
0.0705 77.8853474988933
0.072 78.3317839752103
0.0735 78.720008853475
0.075 79.1108897742364
0.0765 79.4692341744135
0.078 79.8530323151837
0.0795 80.196768481629
0.081 80.5013280212483
0.0825 80.8123063302346
0.084 81.0967242142541
0.0855 81.3773793714033
0.087 81.6624612660469
0.0885 81.9654714475432
0.09 82.2633908809208
0.0915 82.5190349712262
0.093 82.7983621071271
0.0945 83.0542275343072
0.096 83.3162903939796
0.0975 83.5564409030545
0.099 83.7937140327579
0.1005 84.006418769367
0.102 84.1907923860115
0.1035 84.359229747676
0.105 84.5805666223993
0.1065 84.7746790615317
0.108 84.9643647631696
0.1095 85.1562638335547
0.111 85.3417441345728
0.1125 85.5017706949978
0.114 85.6648959716689
0.1155 85.8083222664896
0.117 85.967242142541
0.1185 86.1157591854803
0.12 86.2540947321824
0.1215 86.421646746348
0.123 86.566401062417
0.1245 86.7144754316069
0.126 86.8446215139442
0.1275 86.9805223550244
0.129 87.0969455511288
0.1305 87.2149181053564
0.132 87.3339973439575
0.1335 87.4619300575476
0.135 87.55201416556
0.1365 87.6613545816733
0.138 87.7543160690571
0.1395 87.859229747676
0.141 87.9619300575476
0.1425 88.0537848605578
0.144 88.1436476316954
0.1455 88.2392651615759
0.147 88.3286852589641
0.1485 88.4163346613546
0.15 88.4163346613546
};
\end{axis}

\end{tikzpicture}

%% file: figures/PCK_TOSCA.tex
\begin{tikzpicture}

\definecolor{crimson2143940}{RGB}{214,39,40}
\definecolor{darkorange25512714}{RGB}{255,127,14}
\definecolor{darkslategray38}{RGB}{38,38,38}
\definecolor{forestgreen4416044}{RGB}{44,160,44}
\definecolor{lightgray204}{RGB}{204,204,204}
\definecolor{mediumpurple148103189}{RGB}{148,103,189}
\definecolor{sienna1408675}{RGB}{140,86,75}
\definecolor{steelblue31119180}{RGB}{31,119,180}

\Large
\begin{axis}[
axis line style={lightgray204},
legend cell align={left},
legend style={
  fill opacity=0.8,
  draw opacity=1,
  text opacity=1,
  at={(0.97,0.03)},
  anchor=south east,
  draw=lightgray204
},
xtick = {0,0.05,0.1,0.15},
ytick = {60,70,80,90,100},
xticklabel style={yshift= 5pt},
yticklabel style={xshift= 5pt},
tick align=outside,
tick pos=left,
x grid style={lightgray204},
xlabel=\textcolor{darkslategray38}{Geodesic Error},
xmajorgrids,
xmin=0, xmax=0.15,
xtick style={color=darkslategray38},
y grid style={lightgray204},
ylabel=\textcolor{darkslategray38}{\% Correspondences},
ymajorgrids,
ymin= 50, ymax=100,
ytick style={color=darkslategray38}
]
\addplot [ultra thick, steelblue31119180]
table {%
0 79.259936244157
0.0015 80.2546927190107
0.003 81.3942429041723
0.0045 82.6937871953191
0.006 84.1289444123709
0.0075 85.2643642610836
0.009 86.2596260731504
0.0105 87.2244456984884
0.012 88.2287142779153
0.0135 89.1263145056084
0.015 89.912063593532
0.0165 90.6263850703459
0.018 91.2802914141005
0.0195 91.9036651613743
0.021 92.4202388987404
0.0225 92.9678455102713
0.024 93.4733600323674
0.0255 93.9332163410713
0.027 94.3509251641448
0.0285 94.7296816403578
0.03 95.0920428723022
0.0315 95.394873598893
0.033 95.6826766799424
0.0345 95.9700026576032
0.036 96.2270097894923
0.0375 96.4681458009723
0.039 96.6872170004125
0.0405 96.9175424967315
0.042 97.0989982292176
0.0435 97.2983474598748
0.045 97.4591159297782
0.0465 97.6116270031478
0.048 97.7619371684585
0.0495 97.881912065686
0.051 97.9980282236704
0.0525 98.1076311552515
0.054 98.1994460634578
0.0555 98.2892551731614
0.057 98.3701331971687
0.0585 98.4542627317001
0.06 98.5189970370577
0.0615 98.5771012635562
0.063 98.6394390304672
0.0645 98.6932273876752
0.066 98.7369489205214
0.0675 98.7854239862609
0.069 98.8245478455438
0.0705 98.873118699513
0.072 98.9066737102316
0.0735 98.9384553344556
0.075 98.9697487905415
0.0765 98.9962479562827
0.078 99.0352358102711
0.0795 99.0656844316171
0.081 99.0961993562425
0.0825 99.1302661592989
0.084 99.1566718196572
0.0855 99.1827486340621
0.087 99.2016446313698
0.0885 99.2217466013697
0.09 99.2412888966924
0.0915 99.2582537266746
0.093 99.2753107983538
0.0945 99.2906504576653
0.096 99.3013580714801
0.0975 99.3145625006333
0.099 99.3279364112699
0.1005 99.340164586834
0.102 99.3506096501869
0.1035 99.3590519104526
0.105 99.3715491809408
0.1065 99.3830608925325
0.108 99.3928520604707
0.1095 99.4037430114886
0.111 99.4114862177474
0.1125 99.4159483437188
0.114 99.421770265621
0.1155 99.4336487819992
0.117 99.4420542960291
0.1185 99.4531100864898
0.12 99.4606389515544
0.1215 99.4708922999636
0.123 99.4788171906107
0.1245 99.4875362269393
0.126 99.4942635500861
0.1275 99.497921276865
0.129 99.5011325668775
0.1305 99.5081798075421
0.132 99.5127887076305
0.1335 99.5188497580933
0.135 99.5284871150723
0.1365 99.5314209716611
0.138 99.5338169090898
0.1395 99.5387926599573
0.141 99.5420072671322
0.1425 99.5451322261887
0.144 99.5488951860057
0.1455 99.5515102278895
0.147 99.5550279355013
0.1485 99.5580227067586
0.15 99.5580227067586
};
\addplot [ultra thick, mediumpurple148103189]
table {%
0 64.2973839234925
0.0015 65.775856259235
0.003 67.4445688326788
0.0045 69.240256672839
0.006 71.218405996147
0.0075 73.0099456313324
0.009 74.7577205369642
0.0105 76.1034586951531
0.012 77.5804455410408
0.0135 78.8752247739303
0.015 80.0427272468615
0.0165 81.1601935596876
0.018 82.1668562930586
0.0195 83.1986083789964
0.021 84.0813298801088
0.0225 84.9313787912834
0.024 85.7679438649629
0.0255 86.418133856639
0.027 87.1161054511011
0.0285 87.7619766875625
0.03 88.3654140175797
0.0315 88.9982345456972
0.033 89.5384827926116
0.0345 90.0897950218278
0.036 90.6299579985702
0.0375 91.1704466056157
0.039 91.6558211080978
0.0405 92.0704572123684
0.042 92.4971177263102
0.0435 92.9221230315245
0.045 93.3532955181705
0.0465 93.6874532865856
0.048 94.0568555431156
0.0495 94.327971266388
0.051 94.6345545228461
0.0525 94.918910763848
0.054 95.1463321490961
0.0555 95.3585042637245
0.057 95.578584604219
0.0585 95.7750375151779
0.06 95.997314057124
0.0615 96.1791731239714
0.063 96.3961663411008
0.0645 96.5589010622758
0.066 96.7066394940833
0.0675 96.8105090168199
0.069 96.9631729428548
0.0705 97.0733198161283
0.072 97.2137271868706
0.0735 97.3418646625049
0.075 97.47759555696
0.0765 97.5830864909731
0.078 97.7097410187181
0.0795 97.8123107583382
0.081 97.8946482056779
0.0825 97.9797663604946
0.084 98.0544784639896
0.0855 98.1361837005699
0.087 98.2038394226105
0.0885 98.2544927160267
0.09 98.3383160771282
0.0915 98.395941783012
0.093 98.4679471422518
0.0945 98.5160039002204
0.096 98.5747795144134
0.0975 98.6209396067986
0.099 98.6480119094453
0.1005 98.6962975033298
0.102 98.7415254935252
0.1035 98.7554199101652
0.105 98.7724814665277
0.1065 98.814751122364
0.108 98.8291167240539
0.1095 98.8407758240957
0.111 98.8533369304132
0.1125 98.8698280336025
0.114 98.8839084087058
0.1155 98.8977307649677
0.117 98.922614077532
0.1185 98.9430891629158
0.12 98.9642831842809
0.1215 98.9792741374552
0.123 99.0048073248352
0.1245 99.0126396280128
0.126 99.0252103498407
0.1275 99.0470556076543
0.129 99.057051167052
0.1305 99.0841168546073
0.132 99.0886173046523
0.1335 99.0988603359158
0.135 99.1081927891084
0.1365 99.1268849032386
0.138 99.1313853532836
0.1395 99.1382502048986
0.141 99.1475826580911
0.1425 99.1584152612987
0.144 99.1629157113437
0.1455 99.1689163114037
0.147 99.1860705207981
0.1485 99.2013361535833
0.15 99.2013361535833
};
\addplot [ultra thick, crimson2143940]
table {%
0 71.9555080336812
0.0015 73.3461282282205
0.003 74.7162496823725
0.0045 76.3362245256527
0.006 78.0435202292405
0.0075 79.4588567979661
0.009 80.7581533252066
0.0105 81.9156592308489
0.012 83.1814283170838
0.0135 84.2980192483088
0.015 85.2501057891439
0.0165 86.0751049669761
0.018 86.8551375865903
0.0195 87.6219820072188
0.021 88.2189553383751
0.0225 88.7793040159473
0.024 89.3792166636788
0.0255 89.9779357751569
0.027 90.5521427541607
0.0285 91.0490835468233
0.03 91.5317224396663
0.0315 91.9532700479189
0.033 92.3195796568492
0.0345 92.7283862560603
0.036 93.1134978398052
0.0375 93.5329404688984
0.039 93.8571728370075
0.0405 94.1330362829009
0.042 94.414990946628
0.0435 94.6553178985164
0.045 94.9022791480717
0.0465 95.184861150193
0.048 95.4252388856226
0.0495 95.5680285630558
0.051 95.7503485416768
0.0525 95.9393156046452
0.054 96.1367149577772
0.0555 96.3171030682387
0.057 96.4907278192132
0.0585 96.5853763426426
0.06 96.7458275656927
0.0615 96.8828280369983
0.063 97.0229057787075
0.0645 97.1550278328141
0.066 97.2573526288493
0.0675 97.3431923056129
0.069 97.4364980154593
0.0705 97.5241451710938
0.072 97.6068090660603
0.0735 97.6711315733357
0.075 97.7522467532388
0.0765 97.828739666823
0.078 97.9230787801829
0.0795 97.9996544686281
0.081 98.0611643872854
0.0825 98.1451600452932
0.084 98.2151635081345
0.0855 98.2808632311865
0.087 98.3337536687339
0.0885 98.3725472560571
0.09 98.445696446752
0.0915 98.4873342859075
0.093 98.5177017306813
0.0945 98.5753017027028
0.096 98.6176208353692
0.0975 98.6601876067169
0.099 98.6813921750578
0.1005 98.7039216330279
0.102 98.7403156750758
0.1035 98.7723344182538
0.105 98.8225705488822
0.1065 98.8530267517432
0.108 98.8673461391673
0.1095 98.8891652329511
0.111 98.9146722563012
0.1125 98.9266734564213
0.114 98.9607361920256
0.1155 98.9806204949343
0.117 98.9907688979853
0.1185 99.0186525105797
0.12 99.0365501792868
0.1215 99.0469374824085
0.123 99.0532696355711
0.1245 99.0637978934212
0.126 99.0706260340306
0.1275 99.0780331996085
0.129 99.0888658028161
0.1305 99.1052643652235
0.132 99.1172550183679
0.1335 99.1376011698817
0.135 99.1500888541883
0.1365 99.1635796573476
0.138 99.1764083980171
0.1395 99.2013248057078
0.141 99.2106572589004
0.1425 99.2214329886163
0.144 99.238425596063
0.1455 99.2450052210913
0.147 99.2500846961047
0.1485 99.2539955742056
0.15 99.2539955742056
};

\addplot [ultra thick,dashed, darkorange25512714]
table {%
0 85.123828125
0.0015 85.3377604166666
0.003 85.68046875
0.0045 85.9401041666666
0.006 86.3386718749999
0.0075 86.7875000000001
0.009 87.1759114583331
0.0105 87.5686197916667
0.012 87.9473958333333
0.0135 88.301171875
0.015 88.6307291666667
0.0165 89.0481770833334
0.018 89.5158854166668
0.0195 89.8457031250001
0.021 90.1723958333336
0.0225 90.5069010416667
0.024 90.837109375
0.0255 91.1389322916668
0.027 91.4626302083334
0.0285 91.7174479166666
0.03 91.9463541666668
0.0315 92.1565104166666
0.033 92.3970052083333
0.0345 92.6528645833332
0.036 92.9052083333333
0.0375 93.1994791666666
0.039 93.4334635416666
0.0405 93.6726562500001
0.042 93.9173177083334
0.0435 94.1441406250001
0.045 94.3638020833334
0.0465 94.5752604166668
0.048 94.7721354166669
0.0495 94.9174479166668
0.051 95.0666666666669
0.0525 95.2266927083335
0.054 95.3842447916669
0.0555 95.5476562500003
0.057 95.6774739583336
0.0585 95.8122395833336
0.06 95.9528645833337
0.0615 96.0838541666669
0.063 96.2000000000002
0.0645 96.3355468750001
0.066 96.4453125000002
0.0675 96.5483072916669
0.069 96.6395833333334
0.0705 96.7199218750003
0.072 96.8005208333336
0.0735 96.8700520833335
0.075 96.9248697916668
0.0765 96.9923177083335
0.078 97.0390625000002
0.0795 97.0920572916669
0.081 97.1570312500002
0.0825 97.2023437500002
0.084 97.2593750000002
0.0855 97.3160156250002
0.087 97.3694010416669
0.0885 97.4510416666669
0.09 97.5055989583335
0.0915 97.5403645833334
0.093 97.5828125000002
0.0945 97.6290364583335
0.096 97.6615885416669
0.0975 97.7045572916669
0.099 97.7490885416669
0.1005 97.8096354166669
0.102 97.8697916666668
0.1035 97.9264322916669
0.105 97.9727864583335
0.1065 98.0082031250002
0.108 98.0454427083335
0.1095 98.1085937500002
0.111 98.1389322916669
0.1125 98.1907552083336
0.114 98.2378906250003
0.1155 98.2936197916669
0.117 98.3322916666669
0.1185 98.3809895833336
0.12 98.4281250000003
0.1215 98.4679687500002
0.123 98.534244791667
0.1245 98.5799479166669
0.126 98.6312500000003
0.1275 98.6825520833336
0.129 98.7494791666668
0.1305 98.8075520833335
0.132 98.8449218750002
0.1335 98.8994791666669
0.135 98.9377604166669
0.1365 98.9830729166669
0.138 99.0096354166669
0.1395 99.0575520833335
0.141 99.1031250000002
0.1425 99.1386718750001
0.144 99.1703125
0.1455 99.2026041666666
0.147 99.2296874999999
0.1485 99.2611979166666
0.15 99.2611979166666
};
\addplot [ultra thick,dashed, forestgreen4416044]
table {%
0 74.7643166456427
0.0015 75.008347922503
0.003 75.3637980889881
0.0045 75.630675820511
0.006 76.1012510310256
0.0075 76.5476168801857
0.009 76.9854094615546
0.0105 77.4082449074712
0.012 77.7993144643011
0.0135 78.1542444000659
0.015 78.5206045696653
0.0165 79.000909983016
0.018 79.5257573872974
0.0195 79.8862360367844
0.021 80.2310707517499
0.0225 80.5816167432914
0.024 80.9599036709756
0.0255 81.3052485909865
0.027 81.6564730283616
0.0285 81.945484382457
0.03 82.1891354302316
0.0315 82.4327668055065
0.033 82.6890137257714
0.0345 82.9708130414499
0.036 83.2536195772854
0.0375 83.5870151929037
0.039 83.8607459365852
0.0405 84.1378263435052
0.042 84.3967534755136
0.0435 84.6457606272329
0.045 84.8872102863127
0.0465 85.1513468430875
0.048 85.3785092824993
0.0495 85.5538869714606
0.051 85.7368467403475
0.0525 85.9208103980754
0.054 86.1027525742844
0.0555 86.3043485987722
0.057 86.4871274242689
0.0585 86.6668983989156
0.06 86.8527253097632
0.0615 87.0264537051999
0.063 87.1867280486982
0.0645 87.3664851790221
0.066 87.5225418212161
0.0675 87.6688537485765
0.069 87.8015380225747
0.0705 87.9041656427131
0.072 88.0133380510278
0.0735 88.1050247850468
0.075 88.1876428852833
0.0765 88.2811738326182
0.078 88.3515195087544
0.0795 88.4249052573476
0.081 88.5164384277301
0.0825 88.5859546576853
0.084 88.6653874850971
0.0855 88.7518758030961
0.087 88.8279434456289
0.0885 88.9381327591567
0.09 89.0224388752548
0.0915 89.0808704826219
0.093 89.1445085172768
0.0945 89.2185505290573
0.096 89.2697457273924
0.0975 89.3360718245056
0.099 89.4032350867129
0.1005 89.4897174178852
0.102 89.5750356306956
0.1035 89.6584980809691
0.105 89.7305468154058
0.1065 89.7866086297617
0.108 89.8458680591349
0.1095 89.9378979257034
0.111 89.9880877249371
0.1125 90.0648418327017
0.114 90.1395587653714
0.1155 90.2206644208417
0.117 90.2851349306057
0.1185 90.3586840765079
0.12 90.4295466601786
0.1215 90.492842522036
0.123 90.5926061398045
0.1245 90.6604341654247
0.126 90.7326532903877
0.1275 90.811746134597
0.129 90.9068145731186
0.1305 90.9918009509509
0.132 91.0534259858713
0.1335 91.1353728135574
0.135 91.1939766555487
0.1365 91.2645024156747
0.138 91.3083181792932
0.1395 91.382382046387
0.141 91.4520809974018
0.1425 91.5059727768617
0.144 91.555335091179
0.1455 91.6115702818799
0.147 91.6577379786583
0.1485 91.7129612876918
0.15 91.7129612876918
};

\addplot [ultra thick, dashed, sienna1408675]
table {%
0 43.6130373799488
0.0015 44.6148162751932
0.003 45.7109159879115
0.0045 46.9807421446604
0.006 48.1499627074764
0.0075 49.3765065468152
0.009 50.5040758886611
0.0105 51.5599879310967
0.012 52.7251828393354
0.0135 53.7589678185619
0.015 54.771198803517
0.0165 55.7579758740588
0.018 56.8242880037723
0.0195 57.8717939488922
0.021 58.8393195955498
0.0225 59.7353749125895
0.024 60.6309462480339
0.0255 61.5555671479234
0.027 62.3867935182587
0.0285 63.2461439801024
0.03 64.0760999123542
0.0315 64.8136096749319
0.033 65.57249491899
0.0345 66.3143897956781
0.036 67.0561147850053
0.0375 67.7467784664929
0.039 68.3902121726997
0.0405 69.0537628876897
0.042 69.6719362060956
0.0435 70.3074927007582
0.045 70.9363374996573
0.0465 71.5524402396114
0.048 72.1051569867894
0.0495 72.6355975911602
0.051 73.1230528509921
0.0525 73.6228333133068
0.054 74.0679925129959
0.0555 74.5258999337628
0.057 74.9356392823911
0.0585 75.3497367797745
0.06 75.7120356922039
0.0615 76.0777286508637
0.063 76.4456619668701
0.0645 76.7835445766962
0.066 77.1048347601376
0.0675 77.3967235692166
0.069 77.6911065095804
0.0705 77.9582985992723
0.072 78.2090090617876
0.0735 78.4900872400918
0.075 78.7497106662457
0.0765 79.0322812357082
0.078 79.3017346235903
0.0795 79.563523136256
0.081 79.7896020120149
0.0825 80.0366145454736
0.084 80.2780221835561
0.0855 80.4881857459239
0.087 80.7207730965887
0.0885 80.9210915445103
0.09 81.1115130434409
0.0915 81.311385921948
0.093 81.4845004734133
0.0945 81.6847695011114
0.096 81.8819553980631
0.0975 82.0723484490901
0.099 82.2400351185308
0.1005 82.4096744550579
0.102 82.5760585562586
0.1035 82.7235665485631
0.105 82.8868890098849
0.1065 83.0317356093127
0.108 83.172395470983
0.1095 83.3054544159084
0.111 83.4507820710679
0.1125 83.5953031578734
0.114 83.7315208261266
0.1155 83.873405796865
0.117 84.0117373764452
0.1185 84.1490815873192
0.12 84.2831923583845
0.1215 84.4129586099825
0.123 84.5294847268995
0.1245 84.6483647125933
0.126 84.7696304216271
0.1275 84.8770206129783
0.129 84.9947612114426
0.1305 85.1145583679046
0.132 85.2276250615453
0.1335 85.3357718663863
0.135 85.4416856718801
0.1365 85.5366205516846
0.138 85.6434139784624
0.1395 85.7483345438565
0.141 85.853553333764
0.1425 85.9596101920305
0.144 86.0626757734223
0.1455 86.1529310101213
0.147 86.2469296735544
0.1485 86.3482586256212
0.15 86.3482586256212
};
\end{axis}

\end{tikzpicture}

%% file: figures/PCK_SMAL.tex
\begin{tikzpicture}

\definecolor{crimson2143940}{RGB}{214,39,40}
\definecolor{darkorange25512714}{RGB}{255,127,14}
\definecolor{darkslategray38}{RGB}{38,38,38}
\definecolor{forestgreen4416044}{RGB}{44,160,44}
\definecolor{lightgray204}{RGB}{204,204,204}
\definecolor{mediumpurple148103189}{RGB}{148,103,189}
\definecolor{sienna1408675}{RGB}{140,86,75}
\definecolor{steelblue31119180}{RGB}{31,119,180}

\Large
\begin{axis}[
axis line style={lightgray204},
legend cell align={left},
legend style={
  fill opacity=0.8,
  draw opacity=1,
  text opacity=1,
  at={(0.97,0.03)},
  anchor=south east,
  draw=lightgray204
},
xtick = {0,0.05,0.1,0.15},
ytick = {20,40,60,80,100},
xticklabel style={yshift= 5pt},
yticklabel style={xshift= 5pt},
tick align=outside,
tick pos=left,
x grid style={lightgray204},
xlabel=\textcolor{darkslategray38}{Geodesic Error},
xmajorgrids,
xmin=0, xmax=0.15,
xtick style={color=darkslategray38},
y grid style={lightgray204},
ylabel=\textcolor{darkslategray38}{\% Correspondences},
ymajorgrids,
ymin= 10, ymax=100,
ytick style={color=darkslategray38}
]
\addplot [ultra thick, steelblue31119180]
table {%
0 36.5777690650445
0.0015 36.5963509488959
0.003 36.6836089725311
0.0045 36.9177121946583
0.006 37.4088846117289
0.0075 38.2860429933783
0.009 39.6013964135222
0.0105 41.3659506384057
0.012 43.4547096283623
0.0135 45.7648710515477
0.015 48.1782308399075
0.0165 50.6061408928175
0.018 53.1527381744448
0.0195 55.750039603333
0.021 58.5415676583341
0.0225 61.2850370687197
0.024 64.0470249976238
0.0255 66.6662151886703
0.027 69.1682111966543
0.0285 71.6048419034946
0.03 73.8612378417768
0.0315 75.9606224059817
0.033 78.011619617907
0.0345 79.8930908025219
0.036 81.5825610683395
0.0375 83.1676646706587
0.039 84.681165446884
0.0405 85.974293476539
0.042 87.2300082374933
0.0435 88.3052585305579
0.045 89.2717818331591
0.0465 90.162955834363
0.048 90.9701826505719
0.0495 91.7000562367329
0.051 92.4020411557837
0.0525 93.0377736590312
0.054 93.6170238887305
0.0555 94.1523223394481
0.057 94.6379384088965
0.0585 95.0536941989038
0.06 95.4530225263758
0.0615 95.8099515255204
0.063 96.1319503849444
0.0645 96.4382782054938
0.066 96.7010304787251
0.0675 96.9536957830371
0.069 97.1668805246649
0.0705 97.3724614263536
0.072 97.5411082596711
0.0735 97.7063214840161
0.075 97.8479390425498
0.0765 98.0028316383107
0.078 98.1122873301017
0.0795 98.2322339448088
0.081 98.3330521496689
0.0825 98.4225319202864
0.084 98.4989584323417
0.0855 98.5666128061338
0.087 98.6392770015525
0.0885 98.6888603744891
0.09 98.7352754807845
0.0915 98.8000110889332
0.093 98.8420777492634
0.0945 98.8865918955739
0.096 98.9252328675981
0.0975 98.9616006083072
0.099 98.9866536767734
0.1005 99.013286918227
0.102 99.0519516522511
0.1035 99.0824660203403
0.105 99.1008182048601
0.1065 99.1194317713779
0.108 99.1413086525362
0.1095 99.1587460000634
0.111 99.1736328929443
0.1125 99.1899890694801
0.114 99.2010146373919
0.1155 99.2159292526059
0.117 99.2290379558344
0.1185 99.236162595444
0.12 99.2507405823274
0.1215 99.2604156765833
0.123 99.2727204321516
0.1245 99.2815321737477
0.126 99.2903953996768
0.1275 99.3003714792637
0.129 99.3091832208599
0.1305 99.3156979691411
0.132 99.3266680923867
0.1335 99.3321848366759
0.135 99.3382797896271
0.1365 99.3431034755885
0.138 99.3481489402148
0.1395 99.3512261191902
0.141 99.3581567024681
0.1425 99.363000190096
0.144 99.3655506447423
0.1455 99.3674912080601
0.147 99.3693961283781
0.1485 99.371336691696
0.15 99.371336691696
};
 \addlegendentry{Ours}

\addplot [ultra thick, mediumpurple148103189]
table {%
0 19.2065591040142
0.0015 19.2294023064981
0.003 19.3066367265469
0.0045 19.5028276779774
0.006 19.9180805056554
0.0075 20.6714903526281
0.009 21.8117099135063
0.0105 23.2729263694833
0.012 24.9774617431803
0.0135 26.8233532934132
0.015 28.7103016189842
0.0165 30.6700487913063
0.018 32.8595309381237
0.0195 35.1870980261699
0.021 37.8935185185185
0.0225 40.4378188068308
0.024 43.057163450876
0.0255 45.6969948990907
0.027 48.2943280106454
0.0285 50.7436238633843
0.03 53.1798902195609
0.0315 55.3515469061876
0.033 57.5362608117099
0.0345 59.7042304280328
0.036 61.7371645597693
0.0375 63.6530272787758
0.039 65.575543357729
0.0405 67.4892437347527
0.042 69.2753659347971
0.0435 70.9244011976048
0.045 72.4991128853404
0.0465 73.9672599245953
0.048 75.3996451541362
0.0495 76.6540252827678
0.051 77.8541805278332
0.0525 78.9901585717454
0.054 80.0780660900421
0.0555 81.1485362608117
0.057 82.2283488578399
0.0585 83.147122421823
0.06 84.0774839210468
0.0615 84.9687569305833
0.063 85.7261865158572
0.0645 86.4248170326015
0.066 87.2027334220448
0.0675 87.9280328232424
0.069 88.5690563317809
0.0705 89.1962186737636
0.072 89.7854291417166
0.0735 90.3724218230206
0.075 90.8174761587935
0.0765 91.2993457529386
0.078 91.7334220447993
0.0795 92.1617598137059
0.081 92.5918440895986
0.0825 92.9268130405855
0.084 93.2727323131515
0.0855 93.5833610556664
0.087 93.8990075404746
0.0885 94.178171434908
0.09 94.4513195830561
0.0915 94.7005156353959
0.093 94.8835107562652
0.0945 95.0979984475493
0.096 95.2684353515192
0.0975 95.4466345087603
0.099 95.6338434242626
0.1005 95.8247116877356
0.102 95.9929585273896
0.1035 96.1597360833888
0.105 96.3124029718341
0.1065 96.4520126413839
0.108 96.5875194056331
0.1095 96.7215014415613
0.111 96.8491905078731
0.1125 96.9556442670215
0.114 97.0588545131958
0.1155 97.1496728764693
0.117 97.2372477267687
0.1185 97.3203038367709
0.12 97.3845087602573
0.1215 97.4485196274119
0.123 97.5185462408516
0.1245 97.594921268574
0.126 97.6762585939232
0.1275 97.7438733643823
0.129 97.8021734309159
0.1305 97.8653526280772
0.132 97.9353792415169
0.1335 97.989326901752
0.135 98.0426646706587
0.1365 98.0825848303393
0.138 98.136199822577
0.1395 98.1593479707252
0.141 98.2050066533599
0.1425 98.2377744510978
0.144 98.2670215125305
0.1455 98.2967952982923
0.147 98.3332224440009
0.1485 98.3551785318253
0.15 98.3551785318253
};
 \addlegendentry{Q-MatchV2-cc }
\addplot [ultra thick, crimson2143940]
table {%
0 27.8302284320248
0.0015 27.8444777112442
0.003 27.9158904413396
0.0045 28.1185684187181
0.006 28.5204868041694
0.0075 29.1949157241073
0.009 30.1281049013085
0.0105 31.4784597471723
0.012 33.2379962297627
0.0135 35.1693834553116
0.015 37.4038589487691
0.0165 39.6816090042138
0.018 42.1834664005323
0.0195 44.48708139277
0.021 46.8973442004879
0.0225 49.3324184963407
0.024 51.724412286538
0.0255 54.3097139055223
0.027 56.9114548680417
0.0285 59.5429418939898
0.03 61.8442559325793
0.0315 64.0368429807052
0.033 66.2665779552007
0.0345 68.2605067642493
0.036 70.201042359725
0.0375 72.0177422931914
0.039 73.5849412286538
0.0405 75.075044355733
0.042 76.6185129740519
0.0435 77.9096529163894
0.045 79.2156797516079
0.0465 80.4944832557108
0.048 81.6387779995565
0.0495 82.6590984697272
0.051 83.7134342426258
0.0525 84.6318751386117
0.054 85.5865214016411
0.0555 86.5301341760922
0.057 87.2957972943003
0.0585 88.0463517409625
0.06 88.8542636948326
0.0615 89.526835218452
0.063 90.1345919272566
0.0645 90.5986083388778
0.066 91.1258593923264
0.0675 91.6982424040807
0.069 92.2055611000222
0.0705 92.6197050343757
0.072 93.0529773785762
0.0735 93.5000554446662
0.075 93.869538700377
0.0765 94.2226380572189
0.078 94.5210135284986
0.0795 94.8142603681526
0.081 95.0976935018851
0.0825 95.3621922821025
0.084 95.6688290086494
0.0855 95.8555666444888
0.087 96.0421656686626
0.0885 96.2426258593923
0.09 96.4196329563096
0.0915 96.5798403193613
0.093 96.7222776668885
0.0945 96.838711465957
0.096 96.9504324683966
0.0975 97.0782324240408
0.099 97.2185905965847
0.1005 97.3516023508539
0.102 97.4332446218674
0.1035 97.563872255489
0.105 97.6631182080284
0.1065 97.7260756265248
0.108 97.7968230206254
0.1095 97.8883067198936
0.111 97.9449157241073
0.1125 98.0114770459082
0.114 98.076790862719
0.1155 98.1336770902639
0.117 98.1902860944777
0.1185 98.2451485917055
0.12 98.2889498780218
0.1215 98.3524894655135
0.123 98.3939343535152
0.1245 98.4335772898647
0.126 98.4647649146152
0.1275 98.5215125304946
0.129 98.5670048791306
0.1305 98.6147427367487
0.132 98.6522233311156
0.1335 98.6856841871812
0.135 98.7146817476159
0.1365 98.7457584830339
0.138 98.7727600354846
0.1395 98.8091871811931
0.141 98.8303947660235
0.1425 98.8590596584608
0.144 98.8695941450432
0.1455 98.8855899312486
0.147 98.8979818141495
0.1485 98.9161953870037
0.15 98.9161953870037
};
 \addlegendentry{Q-MatchV2-nc }
\addplot [ultra thick, dashed,darkorange25512714]
table {%
0 75.1717213114754
0.0015 75.1848360655738
0.003 75.2213114754098
0.0045 75.3204918032787
0.006 75.5282786885246
0.0075 75.8803278688524
0.009 76.3905737704918
0.0105 77.1647540983607
0.012 78.0086065573771
0.0135 78.8729508196721
0.015 79.769262295082
0.0165 80.6327868852459
0.018 81.6598360655738
0.0195 82.5729508196721
0.021 83.5094262295082
0.0225 84.4545081967213
0.024 85.3159836065574
0.0255 86.1225409836065
0.027 86.9110655737705
0.0285 87.6680327868853
0.03 88.3774590163934
0.0315 88.9975409836065
0.033 89.6409836065574
0.0345 90.1131147540983
0.036 90.6225409836066
0.0375 91.0606557377049
0.039 91.4426229508197
0.0405 91.8102459016394
0.042 92.1102459016393
0.0435 92.4508196721312
0.045 92.7454918032787
0.0465 93.0008196721312
0.048 93.2770491803279
0.0495 93.5270491803279
0.051 93.7569672131148
0.0525 93.9795081967213
0.054 94.2045081967213
0.0555 94.4233606557378
0.057 94.6446721311475
0.0585 94.8434426229508
0.06 95.0127049180327
0.0615 95.177868852459
0.063 95.3303278688525
0.0645 95.4713114754098
0.066 95.6377049180327
0.0675 95.7803278688524
0.069 95.9151639344262
0.0705 96.0459016393443
0.072 96.1565573770492
0.0735 96.2446721311475
0.075 96.3487704918033
0.0765 96.4213114754099
0.078 96.5147540983607
0.0795 96.5790983606558
0.081 96.630737704918
0.0825 96.6889344262296
0.084 96.7331967213115
0.0855 96.7840163934427
0.087 96.8393442622951
0.0885 96.8704918032787
0.09 96.9204918032787
0.0915 96.9545081967214
0.093 96.991393442623
0.0945 97.0241803278689
0.096 97.0520491803279
0.0975 97.0852459016394
0.099 97.1176229508197
0.1005 97.15
0.102 97.1700819672131
0.1035 97.1934426229509
0.105 97.2094262295083
0.1065 97.2303278688525
0.108 97.2565573770492
0.1095 97.2733606557378
0.111 97.2868852459017
0.1125 97.2991803278689
0.114 97.316393442623
0.1155 97.3315573770492
0.117 97.3450819672132
0.1185 97.3573770491804
0.12 97.3680327868853
0.1215 97.3852459016394
0.123 97.4012295081968
0.1245 97.4114754098361
0.126 97.419262295082
0.1275 97.4237704918033
0.129 97.4323770491804
0.1305 97.4491803278689
0.132 97.4668032786886
0.1335 97.4741803278689
0.135 97.4819672131148
0.1365 97.488524590164
0.138 97.497131147541
0.1395 97.5081967213115
0.141 97.5176229508197
0.1425 97.5311475409836
0.144 97.5368852459017
0.1455 97.5475409836066
0.147 97.5569672131147
0.1485 97.5647540983607
0.15 97.5647540983607
};
 \addlegendentry{IsoMuSh }
\addplot [ultra thick, dashed, forestgreen4416044]
table {%
0 50.0203703703704
0.0015 50.0412037037037
0.003 50.1050925925926
0.0045 50.300462962963
0.006 50.625462962963
0.0075 51.2467592592593
0.009 52.1282407407407
0.0105 53.512037037037
0.012 55.1847222222222
0.0135 56.8986111111111
0.015 58.5921296296296
0.0165 60.3773148148148
0.018 62.3430555555556
0.0195 64.2759259259259
0.021 66.1138888888889
0.0225 68.0226851851852
0.024 69.7754629629629
0.0255 71.5365740740741
0.027 73.2675925925926
0.0285 74.8259259259259
0.03 76.2851851851852
0.0315 77.5912037037037
0.033 78.9013888888889
0.0345 80.0898148148148
0.036 81.1685185185185
0.0375 82.0888888888889
0.039 82.8958333333334
0.0405 83.6787037037037
0.042 84.3606481481482
0.0435 85.0481481481482
0.045 85.7
0.0465 86.2736111111111
0.048 86.8407407407407
0.0495 87.4037037037037
0.051 87.8379629629629
0.0525 88.3458333333333
0.054 88.7916666666666
0.0555 89.2018518518518
0.057 89.5712962962963
0.0585 89.9379629629629
0.06 90.287037037037
0.0615 90.637037037037
0.063 90.9587962962963
0.0645 91.2462962962964
0.066 91.5162037037037
0.0675 91.7634259259259
0.069 91.9930555555556
0.0705 92.1842592592593
0.072 92.3916666666667
0.0735 92.5824074074074
0.075 92.7305555555555
0.0765 92.8587962962963
0.078 92.9967592592593
0.0795 93.1250000000001
0.081 93.2199074074074
0.0825 93.3185185185185
0.084 93.3879629629629
0.0855 93.4763888888888
0.087 93.5560185185185
0.0885 93.625
0.09 93.6875
0.0915 93.7486111111111
0.093 93.8087962962963
0.0945 93.8615740740741
0.096 93.9055555555556
0.0975 93.9648148148148
0.099 94.0087962962963
0.1005 94.0495370370371
0.102 94.0921296296296
0.1035 94.1319444444444
0.105 94.1675925925926
0.1065 94.1944444444445
0.108 94.2324074074075
0.1095 94.2643518518519
0.111 94.2921296296297
0.1125 94.3115740740741
0.114 94.3356481481482
0.1155 94.3597222222223
0.117 94.3782407407408
0.1185 94.3981481481483
0.12 94.4203703703704
0.1215 94.4342592592593
0.123 94.4537037037038
0.1245 94.4717592592593
0.126 94.500462962963
0.1275 94.5199074074075
0.129 94.5347222222223
0.1305 94.5532407407408
0.132 94.575462962963
0.1335 94.5916666666667
0.135 94.6106481481482
0.1365 94.625462962963
0.138 94.6398148148149
0.1395 94.6527777777778
0.141 94.6694444444445
0.1425 94.6925925925927
0.144 94.7087962962964
0.1455 94.7296296296297
0.147 94.7435185185186
0.1485 94.7569444444445
0.15 94.7569444444445
};
 \addlegendentry{ZoomOut }

\addplot [ultra thick, dashed, sienna1408675]
table {%
0 14.7003215790641
0.0015 14.7177312042581
0.003 14.7653146088775
0.0045 14.9225754839527
0.006 15.2381902860945
0.0075 15.7476515223521
0.009 16.5300510090929
0.0105 17.592295567595
0.012 18.7731125051484
0.0135 20.0321420650762
0.015 21.3487984348763
0.0165 22.6572410734087
0.018 24.021872920825
0.0195 25.462622374299
0.021 26.9911446947375
0.0225 28.5729453790831
0.024 30.10583198682
0.0255 31.6135150334252
0.027 33.0302054620917
0.0285 34.5102414219181
0.03 35.8817008839464
0.0315 37.2104402306498
0.033 38.5074414662738
0.0345 39.7634057282261
0.036 40.9564402940151
0.0375 42.1758467192599
0.039 43.3739029876754
0.0405 44.5212075848303
0.042 45.6407660868739
0.0435 46.6848921205209
0.045 47.7170341855971
0.0465 48.7497623800019
0.048 49.744669391376
0.0495 50.6341602509267
0.051 51.5000198016665
0.0525 52.3723663783544
0.054 53.2258459271932
0.0555 54.0045306212971
0.057 54.7779520324431
0.0585 55.5109701232456
0.06 56.1934107974527
0.0615 56.8701644330387
0.063 57.5383954313595
0.0645 58.1772328359155
0.066 58.7896033330165
0.0675 59.3724693470202
0.069 59.9479651807496
0.0705 60.5078612616038
0.072 61.0340667870608
0.0735 61.5364152647087
0.075 62.0323836454076
0.0765 62.5030415359757
0.078 62.973663783544
0.0795 63.4149360010138
0.081 63.8453846275703
0.0825 64.2442337547128
0.084 64.6715655989608
0.0855 65.0280629217755
0.087 65.3833166999335
0.0885 65.7259092925261
0.09 66.0562486138834
0.0915 66.3756732566613
0.093 66.7035730127047
0.0945 66.9865863511073
0.096 67.2997497069353
0.0975 67.5869966416374
0.099 67.8644853150841
0.1005 68.1304335772899
0.102 68.3978273611507
0.1035 68.6452333428381
0.105 68.8697010740424
0.1065 69.116045686405
0.108 69.3493608022051
0.1095 69.5627237588315
0.111 69.8074921585401
0.1125 70.0179680321896
0.114 70.2320319678104
0.1155 70.4181240693216
0.117 70.6218317333587
0.1185 70.8050407122264
0.12 70.9926654627253
0.1215 71.1703220543041
0.123 71.35707949181
0.1245 71.5251639577987
0.126 71.7105511199823
0.1275 71.8865720939074
0.129 72.0521021449165
0.1305 72.2246815892025
0.132 72.4001243544657
0.1335 72.5688424737826
0.135 72.7278062921776
0.1365 72.8904571808763
0.138 73.0399043183474
0.1395 73.2029433197098
0.141 73.3425727909261
0.1425 73.4937149510503
0.144 73.6461085764978
0.1455 73.8056783258879
0.147 73.9477909260843
0.1485 74.08875106929
0.15 74.08875106929
};
\addlegendentry{HKS}
\end{axis}

\end{tikzpicture}

%% file: figures/PCK_Geo_Ablation.tex
\begin{tikzpicture}

\definecolor{darkorange25512714}{RGB}{255,127,14}
\definecolor{darkslategray38}{RGB}{38,38,38}
\definecolor{lightgray204}{RGB}{204,204,204}
\definecolor{steelblue31119180}{RGB}{31,119,180}

\Large 
\begin{axis}[
axis line style={lightgray204},
legend cell align={left},
legend style={
  fill opacity=0.8,
  draw opacity=1,
  text opacity=1,
  at={(0.97,0.03)},
  anchor=south east,
  draw=none
},
tick align=outside,
tick pos=left,
x grid style={lightgray204},
xlabel=\textcolor{darkslategray38}{Geodesic Error},
xmajorgrids,
xmin=0, xmax=0.15,
xtick style={color=darkslategray38},
xtick ={0,0.05,0.1,0.15},
y grid style={lightgray204},
ylabel=\textcolor{darkslategray38}{\% Correspondences},
ymin=62.3818061088978, ymax=100,
ymajorgrids,
xticklabel style={yshift= 5pt},
yticklabel style={xshift= 5pt},
ytick style={color=darkslategray38}
]
\addplot [ultra thick, steelblue31119180]
table {%
0 93.4019477644976
0.0015 93.4021691013723
0.003 93.4074811863656
0.0045 93.4229747675963
0.006 93.4645861000443
0.0075 93.6702080566622
0.009 93.9803010181496
0.0105 94.4156706507304
0.012 94.7726870296591
0.0135 95.1188579017264
0.015 95.4712262062859
0.0165 95.8612217795485
0.018 96.199203187251
0.0195 96.4645861000443
0.021 96.8025675077468
0.0225 97.1790615316512
0.024 97.4922532093847
0.0255 97.7833111996458
0.027 98.0336432049579
0.0285 98.2386011509517
0.03 98.4420097388225
0.0315 98.6250553342187
0.033 98.7990261177512
0.0345 98.9389110225763
0.036 99.0894200973882
0.0375 99.1963258078796
0.039 99.2682602921646
0.0405 99.3530323151836
0.042 99.4156706507303
0.0435 99.4873837981407
0.045 99.5559982293049
0.0465 99.6177512173528
0.048 99.6757414785303
0.0495 99.7153607791058
0.051 99.7563081009296
0.0525 99.7886232846392
0.054 99.8065515714918
0.0555 99.8247011952191
0.057 99.839973439575
0.0585 99.847941567065
0.06 99.8601150951748
0.0615 99.8634351482957
0.063 99.8691899070385
0.0645 99.8747233289066
0.066 99.881584772023
0.0675 99.8851261620185
0.069 99.8860115095174
0.0705 99.8906595838866
0.072 99.893536963258
0.0735 99.8968570163789
0.075 99.9026117751217
0.0765 99.9052678176184
0.078 99.9061531651173
0.0795 99.9065958388667
0.081 99.9096945551128
0.0825 99.9163346613546
0.084 99.9196547144754
0.0855 99.9225320938468
0.087 99.9260734838424
0.0885 99.9267374944666
0.09 99.9280655157149
0.0915 99.9285081894643
0.093 99.9285081894643
0.0945 99.9293935369632
0.096 99.9318282425852
0.0975 99.9322709163346
0.099 99.9324922532093
0.1005 99.9329349269588
0.102 99.935148295706
0.1035 99.9393536963258
0.105 99.9393536963258
0.1065 99.9424524125719
0.108 99.9424524125719
0.1095 99.9431164231961
0.111 99.9446657813191
0.1125 99.945551128818
0.114 99.9459938025675
0.1155 99.9464364763169
0.117 99.9475431606906
0.1185 99.9510845506861
0.12 99.951969898185
0.1215 99.9528552456839
0.123 99.9532979194334
0.1245 99.954404603807
0.126 99.9552899513059
0.1275 99.95595396193
0.129 99.9561752988048
0.1305 99.9572819831784
0.132 99.9581673306773
0.1335 99.9586100044267
0.135 99.9586100044267
0.1365 99.9586100044267
0.138 99.9586100044267
0.1395 99.9588313413014
0.141 99.9590526781762
0.1425 99.9610447100487
0.144 99.9612660469234
0.1455 99.9612660469234
0.147 99.9612660469234
0.1485 99.9614873837981
0.15 99.9614873837981
};
\addlegendentry{With Gaussians}
\addplot [ultra thick, darkorange25512714]
table {%
0 64.1713147410359
0.0015 64.17197875166
0.003 64.1812749003984
0.0045 64.2184594953519
0.006 64.3326693227092
0.0075 64.6423196104471
0.009 65.1117751217353
0.0105 65.736830455954
0.012 66.5245683930943
0.0135 67.4889331562638
0.015 68.5444887118194
0.0165 69.6580345285525
0.018 70.8089862771138
0.0195 71.9194333776007
0.021 73.1677733510403
0.0225 74.4402390438247
0.024 75.7297476759628
0.0255 76.9541832669323
0.027 78.1250553342187
0.0285 79.3076582558654
0.03 80.4721115537849
0.0315 81.5360779105799
0.033 82.6022576361222
0.0345 83.5792386011509
0.036 84.5179282868526
0.0375 85.3802567507747
0.039 86.2202301903497
0.0405 86.9760956175299
0.042 87.6834882691456
0.0435 88.3430721558212
0.045 88.9725542275343
0.0465 89.6091190792386
0.048 90.1892430278884
0.0495 90.7476759628154
0.051 91.259185480301
0.0525 91.7392651615759
0.054 92.1549358123063
0.0555 92.5336432049579
0.057 92.9183266932271
0.0585 93.2718016821602
0.06 93.6288180610889
0.0615 93.9760956175299
0.063 94.3173970783533
0.0645 94.5900841080124
0.066 94.8373173970784
0.0675 95.0925188136343
0.069 95.3541389995573
0.0705 95.5672864099159
0.072 95.8123063302346
0.0735 96.0345285524568
0.075 96.2636122177955
0.0765 96.4679061531651
0.078 96.6586985391766
0.0795 96.8459495351925
0.081 96.9962372731297
0.0825 97.1699867197875
0.084 97.3324479858344
0.0855 97.4811863656485
0.087 97.6206285967242
0.0885 97.7496679946879
0.09 97.893536963258
0.0915 98.0227976980965
0.093 98.1336874723328
0.0945 98.235945108455
0.096 98.3426294820717
0.0975 98.4289508632138
0.099 98.5150509074811
0.1005 98.5962815405046
0.102 98.6496237273129
0.1035 98.6978751660026
0.105 98.773572377158
0.1065 98.827135900841
0.108 98.883134130146
0.1095 98.939132359451
0.111 98.9820717131473
0.1125 99.0325365205843
0.114 99.082337317397
0.1155 99.1206285967241
0.117 99.1584772023018
0.1185 99.1901283753873
0.12 99.2186808322266
0.1215 99.2569721115537
0.123 99.2813191677733
0.1245 99.3116423196104
0.126 99.3428508189463
0.1275 99.3658698539176
0.129 99.3930942895086
0.1305 99.4218680832226
0.132 99.4402390438246
0.1335 99.4555112881805
0.135 99.4707835325365
0.1365 99.4900398406374
0.138 99.5070827799911
0.1395 99.5261177512173
0.141 99.5462594068171
0.1425 99.5610889774236
0.144 99.5770252324037
0.1455 99.5914121292606
0.147 99.6060203629924
0.1485 99.6210712704736
0.15 99.6210712704736
};
\addlegendentry{Without Gaussians}
\end{axis}

\end{tikzpicture}

%% file: figures/PCK_Shapes_Ablation.tex
\begin{tikzpicture}

\definecolor{crimson2143940}{RGB}{214,39,40}
\definecolor{darkorange25512714}{RGB}{255,127,14}
\definecolor{darkslategray38}{RGB}{38,38,38}
\definecolor{forestgreen4416044}{RGB}{44,160,44}
\definecolor{gray127}{RGB}{127,127,127}
\definecolor{lightgray204}{RGB}{204,204,204}
\definecolor{mediumpurple148103189}{RGB}{148,103,189}
\definecolor{orchid227119194}{RGB}{227,119,194}
\definecolor{sienna1408675}{RGB}{140,86,75}
\definecolor{steelblue31119180}{RGB}{31,119,180}

\Large 
\begin{axis}[
axis line style={lightgray204},
legend cell align={left},
legend style={
  fill opacity=0.8,
  draw opacity=1,
  text opacity=1,
  at={(0.97,0.03)},
  anchor=south east,
  draw=none
},
tick align=outside,
tick pos=left,
x grid style={lightgray204},
xlabel=\textcolor{darkslategray38}{Geodesic Error},
xmin=0, xmax=0.15,
xtick style={color=darkslategray38},
xmajorgrids,
xtick ={0,0.05,0.1,0.15},
xticklabel style={yshift= 5pt},
yticklabel style={xshift= 5pt},
y grid style={lightgray204},
ylabel=\textcolor{darkslategray38}{\% Correspondences},
ymin=80, ymax=100,
ymajorgrids,
ytick style={color=darkslategray38}
]
\addplot [ultra thick, steelblue31119180]
table {%
0 81.8208646893906
0.0015 81.8208646893906
0.003 81.8208646893906
0.0045 81.8688210122473
0.006 81.9684226058728
0.0075 82.3742068761989
0.009 82.9054153755349
0.0105 83.4661354581673
0.012 84.2666371550834
0.0135 85.0671388519994
0.015 85.8270621218828
0.0165 86.5390290689095
0.018 87.3063302346171
0.0195 88.1510993064778
0.021 88.9405341596577
0.0225 89.7816142836063
0.024 90.6448280950273
0.0255 91.489597166888
0.027 92.3159214991884
0.0285 93.0758447690719
0.03 93.7250996015936
0.0315 94.3227091633466
0.033 94.9350745167478
0.0345 95.5363730264129
0.036 95.9532241404751
0.0375 96.3737642024495
0.039 96.8016821602479
0.0405 97.07466430574
0.042 97.3550243470562
0.0435 97.5689833259554
0.045 97.7497417736462
0.0465 97.9637007525454
0.048 98.140770252324
0.0495 98.2846392208942
0.051 98.4211302936402
0.0525 98.5502434705622
0.054 98.7236240224288
0.0555 98.8084698244061
0.057 98.9228272096798
0.0585 98.9966061679209
0.06 99.070385126162
0.0615 99.1072746052826
0.063 99.1404751364911
0.0645 99.1736756676996
0.066 99.2216319905563
0.0675 99.2806551571492
0.069 99.332300427918
0.0705 99.3691899070385
0.072 99.3913235945109
0.0735 99.4171462298952
0.075 99.4540357090158
0.0765 99.4872362402243
0.078 99.5204367714328
0.0795 99.531503615169
0.081 99.5462594068172
0.0825 99.5647041463775
0.084 99.5831488859377
0.0855 99.5942157296739
0.087 99.6052825734101
0.0885 99.6200383650583
0.09 99.6421720525306
0.0915 99.6458610004427
0.093 99.6458610004427
0.0945 99.6495499483547
0.096 99.6532388962668
0.0975 99.6606167920909
0.099 99.664305740003
0.1005 99.6716836358271
0.102 99.6753725837391
0.1035 99.6753725837391
0.105 99.6790615316512
0.1065 99.6790615316512
0.108 99.6827504795633
0.1095 99.6827504795633
0.111 99.6864394274753
0.1125 99.6938173232994
0.114 99.6938173232994
0.1155 99.6938173232994
0.117 99.6975062712115
0.1185 99.6975062712115
0.12 99.6975062712115
0.1215 99.7011952191235
0.123 99.7011952191235
0.1245 99.7011952191235
0.126 99.7048841670356
0.1275 99.7048841670356
0.129 99.7048841670356
0.1305 99.7048841670356
0.132 99.7085731149476
0.1335 99.7122620628597
0.135 99.7122620628597
0.1365 99.7196399586838
0.138 99.7196399586838
0.1395 99.7196399586838
0.141 99.73070680242
0.1425 99.73070680242
0.144 99.734395750332
0.1455 99.734395750332
0.147 99.7380846982441
0.1485 99.7380846982441
0.15 99.7380846982441
};
\addlegendentry{3}
\addplot [ultra thick, darkorange25512714]
table {%
0 91.3457281983178
0.0015 91.3457281983178
0.003 91.3457281983178
0.0045 91.3899955732625
0.006 91.5098863804043
0.0075 91.8806256455659
0.009 92.2015641139147
0.0105 92.6368599675372
0.012 93.1201121440165
0.0135 93.6273424819242
0.015 94.1825291426885
0.0165 94.6584034233437
0.018 95.1545669175151
0.0195 95.6341301460823
0.021 96.0288475726723
0.0225 96.4918105356352
0.024 96.8939058580493
0.0255 97.2388224878265
0.027 97.5597609561753
0.0285 97.8401209974915
0.03 98.1278589346318
0.0315 98.2846392208942
0.033 98.5281097830899
0.0345 98.7051792828685
0.036 98.8840932566032
0.0375 98.9984506418769
0.039 99.1312527667109
0.0405 99.2492990998967
0.042 99.3267670060499
0.0435 99.4355909694555
0.045 99.5204367714328
0.0465 99.599749151542
0.048 99.667994687915
0.0495 99.7067286409916
0.051 99.7436181201121
0.0525 99.7546849638483
0.054 99.7675962815405
0.0555 99.7786631252767
0.057 99.7915744429689
0.0585 99.800796812749
0.06 99.8081747085731
0.0615 99.8081747085731
0.063 99.8118636564852
0.0645 99.8247749741774
0.066 99.8284639220894
0.0675 99.8284639220894
0.069 99.8358418179135
0.0705 99.8505976095618
0.072 99.8727312970341
0.0735 99.8782647189022
0.075 99.8856426147263
0.0765 99.8911760365944
0.078 99.8985539324185
0.0795 99.9022428803305
0.081 99.9059318282426
0.0825 99.9059318282426
0.084 99.9096207761546
0.0855 99.9096207761546
0.087 99.9096207761546
0.0885 99.9096207761546
0.09 99.9133097240667
0.0915 99.9133097240667
0.093 99.9133097240667
0.0945 99.9151541980227
0.096 99.9206876198908
0.0975 99.9206876198908
0.099 99.9206876198908
0.1005 99.9262210417589
0.102 99.9262210417589
0.1035 99.9262210417589
0.105 99.9262210417589
0.1065 99.9262210417589
0.108 99.9262210417589
0.1095 99.9262210417589
0.111 99.9280655157149
0.1125 99.9280655157149
0.114 99.9280655157149
0.1155 99.9280655157149
0.117 99.929909989671
0.1185 99.929909989671
0.12 99.929909989671
0.1215 99.933598937583
0.123 99.933598937583
0.1245 99.935443411539
0.126 99.9428213073631
0.1275 99.9428213073631
0.129 99.9428213073631
0.1305 99.9428213073631
0.132 99.9446657813192
0.1335 99.9465102552752
0.135 99.9465102552752
0.1365 99.9501992031873
0.138 99.9501992031873
0.1395 99.9538881510993
0.141 99.9538881510993
0.1425 99.9557326250553
0.144 99.9594215729674
0.1455 99.9594215729674
0.147 99.9594215729674
0.1485 99.9631105208794
0.15 99.9631105208794
};
\addlegendentry{4}
\addplot [ultra thick, forestgreen4416044]
table {%
0 92.0805666223993
0.0015 92.0816733067729
0.003 92.0894200973882
0.0045 92.130367419212
0.006 92.2321823815848
0.0075 92.471226206286
0.009 92.6870296591412
0.0105 93.031208499336
0.012 93.3908809207614
0.0135 93.8335546702081
0.015 94.2275343072156
0.0165 94.7310756972111
0.018 95.2456839309429
0.0195 95.6850376272687
0.021 96.1454183266932
0.0225 96.6257193448428
0.024 97.0075254537406
0.0255 97.3218238158477
0.027 97.6195219123506
0.0285 97.9183266932271
0.03 98.2281983178398
0.0315 98.4395750332006
0.033 98.6752988047809
0.0345 98.8512616201859
0.036 98.9796370075255
0.0375 99.0770252324037
0.039 99.2308543603364
0.0405 99.2994687915007
0.042 99.3813634351483
0.0435 99.4743249225321
0.045 99.5230190349712
0.0465 99.5805666223993
0.048 99.5993802567508
0.0495 99.6403275785746
0.051 99.6635679504206
0.0525 99.6834882691457
0.054 99.7045152722444
0.0555 99.7255422753431
0.057 99.7399291722001
0.0585 99.7521027003099
0.06 99.7598494909252
0.0615 99.7687029659141
0.063 99.7897299690129
0.0645 99.7930500221337
0.066 99.8030101814962
0.0675 99.8074369189907
0.069 99.815183709606
0.0705 99.8196104471005
0.072 99.8207171314741
0.0735 99.8229305002213
0.075 99.8251438689686
0.0765 99.8295706064631
0.078 99.8295706064631
0.0795 99.8328906595839
0.081 99.8328906595839
0.0825 99.8328906595839
0.084 99.8328906595839
0.0855 99.8328906595839
0.087 99.8362107127047
0.0885 99.838424081452
0.09 99.838424081452
0.0915 99.8428508189464
0.093 99.8461708720673
0.0945 99.8472775564409
0.096 99.8550243470562
0.0975 99.8561310314298
0.099 99.8561310314298
0.1005 99.8561310314298
0.102 99.8572377158035
0.1035 99.8594510845507
0.105 99.8616644532979
0.1065 99.8660911907924
0.108 99.8694112439132
0.1095 99.8694112439132
0.111 99.8694112439132
0.1125 99.8694112439132
0.114 99.8694112439132
0.1155 99.8694112439132
0.117 99.8694112439132
0.1185 99.8694112439132
0.12 99.8705179282869
0.1215 99.8705179282869
0.123 99.8705179282869
0.1245 99.8705179282869
0.126 99.8705179282869
0.1275 99.8705179282869
0.129 99.8705179282869
0.1305 99.8705179282869
0.132 99.8716246126605
0.1335 99.8716246126605
0.135 99.8716246126605
0.1365 99.8716246126605
0.138 99.8716246126605
0.1395 99.8716246126605
0.141 99.8716246126605
0.1425 99.8716246126605
0.144 99.8716246126605
0.1455 99.8716246126605
0.147 99.8716246126605
0.1485 99.8716246126605
0.15 99.8716246126605
};
\addlegendentry{5}
\addplot [ultra thick, crimson2143940]
table {%
0 93.6918990703851
0.0015 93.6918990703851
0.003 93.7066548620334
0.0045 93.7302641286705
0.006 93.7848605577689
0.0075 94.0113619595691
0.009 94.2813929467316
0.0105 94.5720820422016
0.012 94.9867197875166
0.0135 95.2914268850524
0.015 95.6529437804338
0.0165 96.0476612070238
0.018 96.4246716836358
0.0195 96.7374944665781
0.021 97.070975357828
0.0225 97.435443411539
0.024 97.6951453445477
0.0255 97.979194333776
0.027 98.2248782647189
0.0285 98.4491662977719
0.03 98.6498450641877
0.0315 98.8128965619006
0.033 98.958978899218
0.0345 99.0630072303379
0.036 99.1899070385126
0.0375 99.2725394717427
0.039 99.3500073778958
0.0405 99.4171462298953
0.042 99.485022871477
0.0435 99.5654419359599
0.045 99.6111848900694
0.0465 99.6716836358271
0.048 99.718164379519
0.0495 99.7616939648812
0.051 99.7919433377601
0.0525 99.8199793418917
0.054 99.8339973439575
0.0555 99.8443263981113
0.057 99.8539176626826
0.0585 99.8583444001771
0.06 99.8701490334957
0.0615 99.8804780876494
0.063 99.8871181938911
0.0645 99.8974472480449
0.066 99.9018739855393
0.0675 99.9092518813634
0.069 99.9181053563524
0.0705 99.9225320938468
0.072 99.9328611480006
0.0735 99.9350745167478
0.075 99.9424524125719
0.0765 99.9431902021543
0.078 99.9431902021543
0.0795 99.9439279917368
0.081 99.9476169396488
0.0825 99.9483547292312
0.084 99.949830308396
0.0855 99.9513058875609
0.087 99.9513058875609
0.0885 99.9542570458905
0.09 99.9549948354729
0.0915 99.9564704146377
0.093 99.9645861000443
0.0945 99.9645861000443
0.096 99.9660616792091
0.0975 99.9667994687915
0.099 99.9704884167035
0.1005 99.9704884167035
0.102 99.9727017854508
0.1035 99.9727017854508
0.105 99.9734395750332
0.1065 99.9741773646156
0.108 99.9741773646156
0.1095 99.9741773646156
0.111 99.974915154198
0.1125 99.9756529437804
0.114 99.9786041021101
0.1155 99.9786041021101
0.117 99.9786041021101
0.1185 99.9786041021101
0.12 99.9800796812749
0.1215 99.9800796812749
0.123 99.9800796812749
0.1245 99.9800796812749
0.126 99.9800796812749
0.1275 99.9800796812749
0.129 99.9800796812749
0.1305 99.9815552604397
0.132 99.9815552604397
0.1335 99.9815552604397
0.135 99.9815552604397
0.1365 99.9822930500221
0.138 99.9822930500221
0.1395 99.9830308396046
0.141 99.9830308396046
0.1425 99.983768629187
0.144 99.983768629187
0.1455 99.983768629187
0.147 99.983768629187
0.1485 99.983768629187
0.15 99.983768629187
};
\addlegendentry{6}
\addplot [ultra thick, mediumpurple148103189]
table {%
0 93.672927338266
0.0015 93.672927338266
0.003 93.6734543308249
0.0045 93.7045468917979
0.006 93.7793798351567
0.0075 93.9912308438205
0.009 94.2167836590148
0.0105 94.5545858892472
0.012 94.9203187250996
0.0135 95.2892135163052
0.015 95.6206918358313
0.0165 95.981154746095
0.018 96.3416176563587
0.0195 96.6615021395898
0.021 97.019857079618
0.0225 97.3956027740888
0.024 97.7086363540547
0.0255 98.0021712093425
0.027 98.2456417715382
0.0285 98.441683003436
0.03 98.6508990493055
0.0315 98.8089968169649
0.033 98.9818503762727
0.0345 99.0909378359578
0.036 99.1937013849364
0.0375 99.257994477118
0.039 99.3212335841818
0.0405 99.3892156242754
0.042 99.4566706718101
0.0435 99.5278146672569
0.045 99.5789329454668
0.0465 99.6258352832058
0.048 99.6764265688569
0.0495 99.7175319884483
0.051 99.7433546238327
0.0525 99.7834060583065
0.054 99.8044857606611
0.0555 99.8260924555745
0.057 99.8345243365164
0.0585 99.8461181728114
0.06 99.8619279495774
0.0615 99.8645629123717
0.063 99.8798456965788
0.0645 99.8888045700795
0.066 99.8993444212568
0.0675 99.9077763021986
0.069 99.9193701384937
0.0705 99.9230590864057
0.072 99.9267480343178
0.0735 99.9283290119944
0.075 99.9314909673475
0.0765 99.9320179599064
0.078 99.9325449524653
0.0795 99.9362339003773
0.081 99.9362339003773
0.0825 99.941503825966
0.084 99.945192773878
0.0855 99.9473007441135
0.087 99.9488817217901
0.0885 99.9520436771433
0.09 99.953097662261
0.0915 99.9562596176142
0.093 99.961002550644
0.0945 99.9625835283206
0.096 99.9641645059972
0.0975 99.9657454836738
0.099 99.9657454836738
0.1005 99.9662724762326
0.102 99.9667994687915
0.1035 99.9667994687915
0.105 99.9678534539092
0.1065 99.9683804464681
0.108 99.972596386939
0.1095 99.972596386939
0.111 99.9736503720567
0.1125 99.9736503720567
0.114 99.9736503720567
0.1155 99.9736503720567
0.117 99.9736503720567
0.1185 99.9778663125277
0.12 99.9778663125277
0.1215 99.9778663125277
0.123 99.9783933050865
0.1245 99.9799742827631
0.126 99.9815552604397
0.1275 99.9815552604397
0.129 99.9826092455574
0.1305 99.9836632306752
0.132 99.984190223234
0.1335 99.9857712009106
0.135 99.9857712009106
0.1365 99.9857712009106
0.138 99.9857712009106
0.1395 99.9868251860284
0.141 99.9868251860284
0.1425 99.9868251860284
0.144 99.9868251860284
0.1455 99.9868251860284
0.147 99.9868251860284
0.1485 99.9873521785872
0.15 99.9873521785872
};
\addlegendentry{7}
\addplot [ultra thick, sienna1408675]
table {%
0 93.2602921646746
0.0015 93.2602921646746
0.003 93.2626636311895
0.0045 93.2994213621704
0.006 93.3776797571618
0.0075 93.5903212546639
0.009 93.8760829697085
0.0105 94.299784987036
0.012 94.6966103838614
0.0135 95.1131980016442
0.015 95.5282046417505
0.0165 95.9155441725163
0.018 96.3060456586353
0.0195 96.6882470119522
0.021 97.0388288117372
0.0225 97.386248656169
0.024 97.7079776133561
0.0255 97.9984822614305
0.027 98.2755485992538
0.0285 98.4850281414026
0.03 98.6557737304749
0.0315 98.823357364194
0.033 98.9846170872067
0.0345 99.1241383671663
0.036 99.2525928033896
0.0375 99.3474514639853
0.039 99.42887181433
0.0405 99.4909251881363
0.042 99.5363782963384
0.0435 99.6067318029469
0.045 99.660089799532
0.0465 99.7098905963448
0.048 99.7490197938405
0.0495 99.7834060583064
0.051 99.8043540125213
0.0525 99.8347878327958
0.054 99.8450641876937
0.0555 99.8660121419085
0.057 99.8711503193575
0.0585 99.8739170302915
0.06 99.8818219186745
0.0615 99.8845886296085
0.063 99.8881458293809
0.0645 99.8932840068298
0.066 99.9000031619553
0.0675 99.9007936507936
0.069 99.9019793840511
0.0705 99.9075128059192
0.072 99.9126509833681
0.0735 99.9138367166256
0.075 99.9173939163979
0.0765 99.9209511161702
0.078 99.9237178271043
0.0795 99.9260892936192
0.081 99.9284607601341
0.0825 99.9332036931638
0.084 99.9375513817745
0.0855 99.9395276038702
0.087 99.9430848036425
0.0885 99.9458515145766
0.09 99.9474324922532
0.0915 99.9474324922532
0.093 99.9482229810915
0.0945 99.9498039587681
0.096 99.9525706697021
0.0975 99.9537564029596
0.099 99.9581040915702
0.1005 99.9581040915702
0.102 99.960080313666
0.1035 99.9604755580851
0.105 99.9608708025042
0.1065 99.9608708025042
0.108 99.9636375134383
0.1095 99.9636375134383
0.111 99.9636375134383
0.1125 99.9636375134383
0.114 99.9636375134383
0.1155 99.9664042243723
0.117 99.9664042243723
0.1185 99.9664042243723
0.12 99.9664042243723
0.1215 99.9664042243723
0.123 99.9664042243723
0.1245 99.9664042243723
0.126 99.9667994687915
0.1275 99.9671947132106
0.129 99.9671947132106
0.1305 99.9675899576298
0.132 99.973518623917
0.1335 99.973518623917
0.135 99.973518623917
0.1365 99.973518623917
0.138 99.973518623917
0.1395 99.973518623917
0.141 99.973518623917
0.1425 99.973518623917
0.144 99.973518623917
0.1455 99.973518623917
0.147 99.973518623917
0.1485 99.973518623917
0.15 99.973518623917
};
\addlegendentry{8}
\addplot [ultra thick, orchid227119194]
table {%
0 94.9240076730117
0.0015 94.9240076730117
0.003 94.9249299099897
0.0045 94.946756185136
0.006 95.0359057596773
0.0075 95.2280384634302
0.009 95.4463012148935
0.0105 95.7226648959717
0.012 95.9381609365009
0.0135 96.141975308642
0.015 96.4063499090059
0.0165 96.7349736855049
0.018 97.0015001721509
0.0195 97.2326742413064
0.021 97.4346441394914
0.0225 97.6796517633171
0.024 97.8905366189563
0.0255 98.0636097584969
0.027 98.1807338547046
0.0285 98.3123063302346
0.03 98.4195932320102
0.0315 98.5201170626137
0.033 98.6221779548472
0.0345 98.6981087993704
0.036 98.7949436820618
0.0375 98.8533520240027
0.039 98.9040750577935
0.0405 98.9461905464561
0.042 98.9806207269687
0.0435 99.0159731444592
0.045 99.0590108700998
0.0465 99.1029708327185
0.048 99.1423196104471
0.0495 99.1629162362894
0.051 99.1896611086518
0.0525 99.2173282179922
0.054 99.2366951945305
0.0555 99.2529880478087
0.057 99.267129014805
0.0585 99.2935664748414
0.06 99.3120112144016
0.0615 99.334144901874
0.063 99.3685750823865
0.0645 99.3925532438149
0.066 99.4048497368551
0.0675 99.4143795189612
0.069 99.4254463626974
0.0705 99.4315946092175
0.072 99.4383576803895
0.0735 99.4571098322758
0.075 99.46602478973
0.0765 99.4727878609021
0.078 99.4826250553342
0.0795 99.4900029511583
0.081 99.4955363730264
0.0825 99.5044513304805
0.084 99.5072180414146
0.0855 99.5130588756087
0.087 99.5241257193448
0.0885 99.532425852147
0.09 99.5407259849491
0.0915 99.5477964684472
0.093 99.5551743642713
0.0945 99.5625522600954
0.096 99.5665486203335
0.0975 99.5803821750037
0.099 99.5871452461758
0.1005 99.5948305543259
0.102 99.5966750282819
0.1035 99.6071270473661
0.105 99.6102011706261
0.1065 99.6194235404063
0.108 99.6264940239044
0.1095 99.6317200334465
0.111 99.6470906497467
0.1125 99.6498573606807
0.114 99.6563130195268
0.1155 99.6633835030249
0.117 99.667994687915
0.1185 99.6805985932812
0.12 99.6907432000393
0.1215 99.7088805272736
0.123 99.7177954847278
0.1245 99.7254807928779
0.126 99.732858688702
0.1275 99.736547636614
0.129 99.7442329447642
0.1305 99.7497663666322
0.132 99.7528404898923
0.1335 99.7605257980424
0.135 99.7669814568885
0.1365 99.7712852294526
0.138 99.7740519403866
0.1395 99.7863484334268
0.141 99.7909596183169
0.1425 99.7943411539029
0.144 99.799567163445
0.1455 99.804178348335
0.147 99.8069450592691
0.1485 99.8137081304412
0.15 99.8137081304412
};
\addlegendentry{9}
\addplot [ultra thick, gray127]
table {%
0 94.5142885249127
0.0015 94.5142885249127
0.003 94.5194530519896
0.0045 94.5492105651468
0.006 94.6092174511829
0.0075 94.8022723919138
0.009 95.0241011263587
0.0105 95.3118390634991
0.012 95.6332693915695
0.0135 95.9414195071565
0.015 96.2316167429049
0.0165 96.5771481973341
0.018 96.8980866656829
0.0195 97.1437705966258
0.021 97.4551178004033
0.0225 97.7888446215139
0.024 98.0246913580247
0.0255 98.2856229403374
0.027 98.5017952879838
0.0285 98.6478776253012
0.03 98.8202744577246
0.0315 98.9535684422802
0.033 99.0819438296197
0.0345 99.1849884412965
0.036 99.2781958585411
0.0375 99.3605823619103
0.039 99.4368206187595
0.0405 99.5118292263046
0.042 99.5696227435935
0.0435 99.6229895233879
0.045 99.6711917761054
0.0465 99.7166888003541
0.048 99.7498893315626
0.0495 99.7948944960897
0.051 99.8261275884118
0.0525 99.8416211696424
0.054 99.8549013821258
0.0555 99.864738576558
0.057 99.8733461216861
0.0585 99.8814618070926
0.06 99.8866263341695
0.0615 99.8932664404112
0.063 99.8994146869313
0.0645 99.9026117751217
0.066 99.9058088633122
0.0675 99.9070385126162
0.069 99.9139245487187
0.0705 99.9144164084403
0.072 99.9163838473267
0.0735 99.9203187250996
0.075 99.921056514682
0.0765 99.9227780237076
0.078 99.9227780237076
0.0795 99.9316314986966
0.081 99.9372878854951
0.0825 99.9446657813192
0.084 99.946141360484
0.0855 99.9478628695096
0.087 99.9508140278393
0.0885 99.9530273965865
0.09 99.9542570458905
0.0915 99.9562244847769
0.093 99.9562244847769
0.0945 99.9567163444985
0.096 99.9572082042202
0.0975 99.9584378535242
0.099 99.9633564507402
0.1005 99.9643401701834
0.102 99.9650779597659
0.1035 99.9655698194875
0.105 99.9655698194875
0.1065 99.9655698194875
0.108 99.9655698194875
0.1095 99.9655698194875
0.111 99.9677831882347
0.1125 99.9680291180955
0.114 99.9680291180955
0.1155 99.9687669076779
0.117 99.9709802764252
0.1185 99.971226206286
0.12 99.971226206286
0.1215 99.971226206286
0.123 99.9714721361468
0.1245 99.97245585559
0.126 99.9727017854508
0.1275 99.9734395750332
0.129 99.9741773646156
0.1305 99.9763907333628
0.132 99.9766366632236
0.1335 99.9768825930844
0.135 99.977374452806
0.1365 99.977374452806
0.138 99.9776203826669
0.1395 99.9776203826669
0.141 99.9776203826669
0.1425 99.9776203826669
0.144 99.9776203826669
0.1455 99.9778663125277
0.147 99.9778663125277
0.1485 99.9778663125277
0.15 99.9778663125277
};
\addlegendentry{10}
\end{axis}

\end{tikzpicture}

%% file: sections/6_discussion_conclusion.tex
\section{Conclusion}\label{sec:conclusion} 
\label{sec:discussion}

The proposed method achieves our main goal: improving mesh alignment w.r.t.~the quantum state of the art. 
Furthermore, it is even highly competitive among classical state-of-the-art methods. %
This suggests that the proposed approach can be used as a reference for comparisons and extensions of classical mesh-alignment works in the future. 
(For such cases, classical SA is a viable alternative when access to quantum computers is lacking.) 
Our results show that ignoring certain higher-order terms still allows for high-quality matchings, which is promising for future quantum approaches that could use similar approximations. 
Finally, unlike classical work, we designed our method within the constraints of contemporary quantum hardware. 
We found that iteratively considering shape triplets is highly effective, perhaps even for classical methods. 

{\small 
\noindent\textbf{Acknowledgements}. 
This work was partially supported by the ERC Consolidator Grant 4DReply (770784). 
ZL is funded by the Ministry of Culture and Science of the State of NRW. %
}

%% file: SupplimentFigures/PCK_Evolution.tex
\begin{tikzpicture}

\definecolor{black1000}{RGB}{10,0,0}
\definecolor{darkred14400}{RGB}{144,0,0}
\definecolor{darkslategray38}{RGB}{38,38,38}
\definecolor{gold2552250}{RGB}{255,225,0}
\definecolor{lightgray204}{RGB}{204,204,204}
\definecolor{maroon7600}{RGB}{76,0,0}
\definecolor{orange2551570}{RGB}{255,157,0}
\definecolor{orangered255910}{RGB}{255,91,0}
\definecolor{palegoldenrod255255156}{RGB}{255,255,156}
\definecolor{red21000}{RGB}{210,0,0}
\definecolor{red255230}{RGB}{255,23,0}
\definecolor{yellow25525554}{RGB}{255,255,54}

\Large
\begin{axis}[
axis line style={lightgray204},
legend style={
  fill opacity=0.8,
  draw opacity=1,
  text opacity=1,
  at={(0.03,0.97)},
  anchor=north west,
  draw=lightgray204
},
tick align=outside,
tick pos=left,
x grid style={lightgray204},
xtick = {0,0.05,0.1,0.15},
xticklabel style={yshift= 5pt},
xlabel=\textcolor{darkslategray38}{Geodesic Error},
xmajorgrids,
xmin=0, xmax=0.15,
xtick style={color=darkslategray38},
y grid style={lightgray204},
ylabel=\textcolor{darkslategray38}{\% Correspondences},
ymajorgrids,
ymin=50.5604249667995, ymax=100,
ytick style={color=darkslategray38}
]
\addplot [semithick, black1000, forget plot]
table {%
0 52.912793271359
0.0015 52.912793271359
0.003 52.912793271359
0.0045 52.9194333776007
0.006 52.9769809650288
0.0075 53.1297034085879
0.009 53.5347498893316
0.0105 53.8468348826915
0.012 54.5396193005755
0.0135 55.2146967684816
0.015 56.0823373173971
0.0165 56.5692784417884
0.018 57.2111553784861
0.0195 57.8242585214697
0.021 58.5059760956175
0.0225 59.2120407259849
0.024 60.0442673749446
0.0255 60.6817175741478
0.027 61.5958388667552
0.0285 62.4324922532094
0.03 63.2979194333776
0.0315 64.1102257636122
0.033 64.8450641876937
0.0345 65.4493138556884
0.036 66.1775121735281
0.0375 66.896857016379
0.039 67.6339088092076
0.0405 68.3532536520584
0.042 69.0172642762284
0.0435 69.6591412129261
0.045 70.2678176184152
0.0465 70.7481186365649
0.048 71.2129260734838
0.0495 71.7662682602921
0.051 72.2155821159805
0.0525 72.7822045152722
0.054 73.3709606020363
0.0555 73.8866755201416
0.057 74.4378043382028
0.0585 74.9690128375387
0.06 75.4471004869411
0.0615 75.8565737051793
0.063 76.3235945108455
0.0645 76.6710934041612
0.066 77.0252324037185
0.0675 77.4435590969456
0.069 77.7844178840195
0.0705 78.2204515272244
0.072 78.6299247454626
0.0735 79.1323594510846
0.075 79.5440460380699
0.0765 80.0154935812306
0.078 80.360779105799
0.0795 80.7658255865427
0.081 81.0424966799469
0.0825 81.3014608233732
0.084 81.6046923417442
0.0855 82.0385126162019
0.087 82.3616644532979
0.0885 82.6826029216468
0.09 83.0256750774679
0.0915 83.2713590084108
0.093 83.576803895529
0.0945 83.8291279327136
0.096 84.1212926073484
0.0975 84.4090305444887
0.099 84.6746347941567
0.1005 84.8295706064631
0.102 85.0420540061974
0.1035 85.2434705621957
0.105 85.4803010181496
0.1065 85.6972111553785
0.108 85.8764940239044
0.1095 86.0579902611775
0.111 86.2483399734396
0.1125 86.3700752545374
0.114 86.518370960602
0.1155 86.6356795042054
0.117 86.7706949977867
0.1185 86.8503762726871
0.12 86.9234174413458
0.1215 87.0916334661355
0.123 87.1757414785303
0.1245 87.315183709606
0.126 87.4059318282426
0.1275 87.5387339530766
0.129 87.5940681717574
0.1305 87.6958831341301
0.132 87.7866312527667
0.1335 87.8884462151394
0.135 88.0212483399734
0.1365 88.0942895086321
0.138 88.1651173085436
0.1395 88.2957060646304
0.141 88.3709606020363
0.1425 88.4749889331563
0.144 88.5502434705622
0.1455 88.6122177954847
0.147 88.7051792828685
0.1485 88.7782204515272
0.15 88.7782204515272
};
\addplot [semithick, maroon7600, forget plot]
table {%
0 58.0301018149624
0.0015 58.0301018149624
0.003 58.0301018149624
0.0045 58.0433820274458
0.006 58.1230633023462
0.0075 58.3466135458167
0.009 58.833554670208
0.0105 59.1522797698096
0.012 59.933598937583
0.0135 60.7104913678619
0.015 61.7485613103143
0.0165 62.5055334218681
0.018 63.406374501992
0.0195 64.402390438247
0.021 65.369632580788
0.0225 66.2151394422311
0.024 67.4103585657371
0.0255 68.4108012394865
0.027 69.3846834882692
0.0285 70.4249667994688
0.03 71.5914121292607
0.0315 72.8574590526781
0.033 73.9353696325808
0.0345 74.6702080566622
0.036 75.6772908366534
0.0375 76.6954404603807
0.039 77.5365205843294
0.0405 78.4484285081895
0.042 79.2917220008854
0.0435 80.2257636122178
0.045 80.9451084550686
0.0465 81.6533864541833
0.048 82.2797698096503
0.0495 82.8065515714918
0.051 83.3820274457725
0.0525 84.0283311199646
0.054 84.5551128818061
0.0555 85.0841080123949
0.057 85.692784417884
0.0585 86.1775121735281
0.06 86.6998671978751
0.0615 87.2266489597167
0.063 87.6516157591855
0.0645 88.0146082337318
0.066 88.4395750332006
0.0675 88.9043824701195
0.069 89.3050022133688
0.0705 89.8738379814077
0.072 90.3740593182824
0.0735 90.7547587428066
0.075 91.0491367861886
0.0765 91.4364763169544
0.078 91.7662682602921
0.0795 92.1646746347941
0.081 92.4701195219123
0.0825 92.7999114652501
0.084 93.0256750774679
0.0855 93.3023461708721
0.087 93.5590969455511
0.0885 93.8269145639663
0.09 94.0969455511288
0.0915 94.373616644533
0.093 94.5794599380257
0.0945 94.8207171314742
0.096 95.0442673749447
0.0975 95.3010181496238
0.099 95.5444887118194
0.1005 95.7436918990704
0.102 95.9119079238601
0.1035 96.1044710048694
0.105 96.2859672421426
0.1065 96.4298362107127
0.108 96.555998229305
0.1095 96.6445329791943
0.111 96.746347941567
0.1125 96.8282425852147
0.114 96.9101372288623
0.1155 97.0097388224878
0.117 97.1845949535192
0.1185 97.3041168658699
0.12 97.4037184594953
0.1215 97.5409473218238
0.123 97.6117751217353
0.1245 97.6914563966356
0.126 97.7755644090305
0.1275 97.846392208942
0.129 97.9061531651172
0.1305 97.9637007525453
0.132 97.999114652501
0.1335 98.0411686586985
0.135 98.0898627711376
0.1365 98.140770252324
0.138 98.1850376272686
0.1395 98.2093846834882
0.141 98.2492253209384
0.1425 98.2957060646303
0.144 98.3532536520584
0.1455 98.4218680832226
0.147 98.4506418769367
0.1485 98.4949092518813
0.15 98.4949092518813
};
\addplot [semithick, darkred14400, forget plot]
table {%
0 62.2045152722444
0.0015 62.2045152722444
0.003 62.2177954847278
0.0045 62.2377158034529
0.006 62.2974767596282
0.0075 62.5453740593183
0.009 63.1540504648074
0.0105 63.6254980079681
0.012 64.5019920318725
0.0135 65.3674192120407
0.015 66.5382912793271
0.0165 67.3727312970341
0.018 68.2337317397078
0.0195 69.1854803010182
0.021 70.0929614873838
0.0225 71.1266046923418
0.024 72.4391323594512
0.0255 73.4949092518814
0.027 74.5949535192563
0.0285 75.7171314741036
0.03 76.8016821602479
0.0315 77.9393536963258
0.033 79.0792386011509
0.0345 79.89597166888
0.036 80.8455068614431
0.0375 81.7042939353696
0.039 82.419212040726
0.0405 83.045595396193
0.042 83.6874723328906
0.0435 84.3315626383355
0.045 84.8605577689243
0.0465 85.3895528995131
0.048 85.8410801239487
0.0495 86.3368747233289
0.051 86.8083222664896
0.0525 87.3683045595396
0.054 87.8707392651616
0.0555 88.3886675520142
0.057 88.9774236387782
0.0585 89.402390438247
0.06 89.8959716688801
0.0615 90.2899513058876
0.063 90.6972111553785
0.0645 91.0048694112439
0.066 91.3501549358123
0.0675 91.7419212040726
0.069 92.0429393536963
0.0705 92.5785745905268
0.072 92.9570606463036
0.0735 93.2536520584329
0.075 93.4971226206285
0.0765 93.7516600265604
0.078 93.9774236387782
0.0795 94.2164674634794
0.081 94.4245241257193
0.0825 94.6746347941567
0.084 94.8694112439132
0.0855 95.0464807436919
0.087 95.227976980965
0.0885 95.4780876494024
0.09 95.6905710491368
0.0915 95.8366533864542
0.093 95.9849490925189
0.0945 96.128818061089
0.096 96.2682602921647
0.0975 96.4032757857459
0.099 96.6334661354581
0.1005 96.7640548915449
0.102 96.9101372288623
0.1035 96.9853917662682
0.105 97.1890216910137
0.1065 97.3019034971226
0.108 97.4546259406817
0.1095 97.5453740593182
0.111 97.580787959274
0.1125 97.6516157591854
0.114 97.7268702965914
0.1155 97.8264718902169
0.117 97.9526339088092
0.1185 98.0389552899512
0.12 98.1119964586099
0.1215 98.1961044710048
0.123 98.2846392208941
0.1245 98.3333333333333
0.126 98.4041611332447
0.1275 98.4440017706949
0.129 98.5081894643647
0.1305 98.5502434705621
0.132 98.5922974767595
0.1335 98.6454183266931
0.135 98.6963258078795
0.1365 98.7250996015936
0.138 98.7782204515272
0.1395 98.7915006640106
0.141 98.8092076139884
0.1425 98.8778220451526
0.144 98.8933156263832
0.1455 98.9353696325807
0.147 98.9552899513058
0.1485 98.9862771137671
0.15 98.9862771137671
};
\addplot [semithick, red21000, forget plot]
table {%
0 63.377600708278
0.0015 63.377600708278
0.003 63.3908809207614
0.0045 63.4019477644976
0.006 63.4617087206728
0.0075 63.6985391766268
0.009 64.240814519699
0.0105 64.6281540504648
0.012 65.4692341744135
0.0135 66.4254094732183
0.015 67.6826029216468
0.0165 68.5989375830013
0.018 69.535192563081
0.0195 70.5555555555555
0.021 71.5626383355467
0.0225 72.5630810092961
0.024 73.908809207614
0.0255 74.9291722000885
0.027 76.1420982735724
0.0285 77.2797698096503
0.03 78.3576803895529
0.0315 79.5285524568393
0.033 80.626383355467
0.0345 81.3523683045595
0.036 82.220008853475
0.0375 83.1961044710048
0.039 83.7915006640106
0.0405 84.3957503320053
0.042 85.0774679061532
0.0435 85.6772908366534
0.045 86.1133244798584
0.0465 86.651173085436
0.048 87.0628596724214
0.0495 87.5542275343072
0.051 88.0057547587428
0.0525 88.5635236830456
0.054 89.0659583886675
0.0555 89.4931385568836
0.057 89.9977866312528
0.0585 90.4138999557326
0.06 90.9274015050908
0.0615 91.3634351482957
0.063 91.7795484727755
0.0645 92.0385126162019
0.066 92.3505976095617
0.0675 92.6604692341744
0.069 92.9437804338203
0.0705 93.4196547144754
0.072 93.7737937140327
0.0735 94.0637450199203
0.075 94.2740150509075
0.0765 94.5374059318282
0.078 94.7830898627711
0.0795 94.9800796812749
0.081 95.2102700309872
0.0825 95.4515272244357
0.084 95.6285967242143
0.0855 95.7901726427623
0.087 95.9561752988047
0.0885 96.1155378486056
0.09 96.2970340858787
0.0915 96.4453297919433
0.093 96.5869853917663
0.0945 96.6976538291279
0.096 96.8149623727313
0.0975 96.9433377600707
0.099 97.1403275785746
0.1005 97.2045152722443
0.102 97.3063302346171
0.1035 97.381584772023
0.105 97.5475874280655
0.1065 97.6560424966799
0.108 97.7689243027888
0.1095 97.8685258964143
0.111 97.9172200088534
0.1125 97.9902611775121
0.114 98.0566622399291
0.1155 98.140770252324
0.117 98.2558654271801
0.1185 98.3178397521027
0.12 98.3731739707835
0.1215 98.4484285081894
0.123 98.5126162018592
0.1245 98.5568835768039
0.126 98.6277113767153
0.1275 98.6542718016821
0.129 98.7051792828685
0.1305 98.7317397078353
0.132 98.7583001328021
0.1335 98.8114209827357
0.135 98.8424081451969
0.1365 98.8933156263833
0.138 98.9707835325364
0.1395 98.9884904825143
0.141 99.0150509074811
0.1425 99.1102257636121
0.144 99.1389995573262
0.1455 99.1611332447985
0.147 99.1788401947763
0.1485 99.1965471447542
0.15 99.1965471447542
};
\addplot [semithick, red255230, forget plot]
table {%
0 65.0730411686587
0.0015 65.0730411686587
0.003 65.0863213811421
0.0045 65.1173085436034
0.006 65.1726427622842
0.0075 65.4205400619743
0.009 66.0314298362107
0.0105 66.3877822045152
0.012 67.1469676848163
0.0135 67.912793271359
0.015 69.1478530323152
0.0165 70.0531208499336
0.018 71.0159362549801
0.0195 71.9632580787959
0.021 72.9172200088535
0.0225 73.8512616201859
0.024 75.1128818061089
0.0255 76.1752988047809
0.027 77.3505976095617
0.0285 78.5590969455511
0.03 79.4864984506419
0.0315 80.5090748118637
0.033 81.5914121292608
0.0345 82.2952633908809
0.036 83.1274900398406
0.0375 83.9796370075255
0.039 84.630367419212
0.0405 85.2788844621514
0.042 85.9849490925188
0.0435 86.6157591854803
0.045 87.0982735723771
0.0465 87.5918548030102
0.048 87.9659141212926
0.0495 88.406374501992
0.051 88.8180610889774
0.0525 89.3492695883134
0.054 89.8716246126605
0.0555 90.3917662682603
0.057 90.8964143426295
0.0585 91.2749003984064
0.06 91.7330677290837
0.0615 92.1292607348384
0.063 92.4634794156706
0.0645 92.675962815405
0.066 92.9791943337759
0.0675 93.2691456396635
0.069 93.5413899955732
0.0705 93.8888888888889
0.072 94.1677733510402
0.0735 94.4599380256751
0.075 94.6458610004427
0.0765 94.9003984063745
0.078 95.0774679061532
0.0795 95.2766710934042
0.081 95.5090748118637
0.0825 95.7171314741036
0.084 95.8654271801682
0.0855 95.9871624612661
0.087 96.1487383798141
0.0885 96.3103142983621
0.09 96.5250110668437
0.0915 96.6843736166445
0.093 96.8193891102258
0.0945 96.954404603807
0.096 97.0916334661355
0.0975 97.1978751660026
0.099 97.4236387782205
0.1005 97.5830013280212
0.102 97.6826029216467
0.1035 97.742363877822
0.105 97.8685258964143
0.1065 97.9614873837981
0.108 98.0588756086764
0.1095 98.1208499335989
0.111 98.1761841522797
0.1125 98.2138114209827
0.114 98.2691456396635
0.1155 98.2824258521469
0.117 98.3377600708278
0.1185 98.3709606020363
0.12 98.4152279769809
0.1215 98.4594953519256
0.123 98.5126162018592
0.1245 98.5834440017707
0.126 98.6609119079238
0.1275 98.6808322266489
0.129 98.7494466578131
0.1305 98.7782204515272
0.132 98.8025675077467
0.1335 98.833554670208
0.135 98.862328463922
0.1365 98.891102257636
0.138 98.966356795042
0.1395 98.9884904825143
0.141 99.0349712262062
0.1425 99.0947321823815
0.144 99.1412129260734
0.1455 99.1677733510402
0.147 99.1832669322708
0.1485 99.2142540947321
0.15 99.2142540947321
};
\addplot [semithick, orangered255910, forget plot]
table {%
0 66.2062859672422
0.0015 66.2107127047366
0.003 66.2284196547145
0.0045 66.2638335546702
0.006 66.3258078795928
0.0075 66.6113324479859
0.009 67.2664895971669
0.0105 67.713590084108
0.012 68.6232846392209
0.0135 69.5573262505534
0.015 70.8787073926516
0.0165 71.7751217352811
0.018 72.7600708277999
0.0195 73.6675520141656
0.021 74.6591412129261
0.0225 75.6595838866755
0.024 76.983178397521
0.0255 78.0743691899071
0.027 79.1589198760514
0.0285 80.2700309871625
0.03 81.2372731297034
0.0315 82.1890216910137
0.033 83.3178397521027
0.0345 84.0748118636564
0.036 84.838424081452
0.0375 85.5843293492696
0.039 86.1930057547588
0.0405 86.7463479415671
0.042 87.4324922532094
0.0435 88.036741921204
0.045 88.4860557768924
0.0465 88.9774236387782
0.048 89.3957503320053
0.0495 89.89597166888
0.051 90.294378043382
0.0525 90.7857459052678
0.054 91.2439132359451
0.0555 91.7131474103586
0.057 92.1336874723329
0.0585 92.4789729969013
0.06 92.8773793714033
0.0615 93.2425852146968
0.063 93.5790172642762
0.0645 93.7870739265162
0.066 94.0836653386454
0.0675 94.3559096945551
0.069 94.5927401505091
0.0705 94.9026117751217
0.072 95.132802124834
0.0735 95.3342186808322
0.075 95.5046480743692
0.0765 95.7038512616202
0.078 95.8610004426738
0.0795 96.0336432049579
0.081 96.2262062859673
0.0825 96.3922089420097
0.084 96.5227976980965
0.0855 96.6157591854803
0.087 96.8061088977424
0.0885 96.983178397521
0.09 97.1779548472776
0.0915 97.3373173970783
0.093 97.4811863656485
0.0945 97.5675077467906
0.096 97.6803895528995
0.0975 97.7733510402833
0.099 97.9592740150509
0.1005 98.083222664896
0.102 98.1673306772908
0.1035 98.2027445772465
0.105 98.3266932270916
0.1065 98.3731739707835
0.108 98.4506418769366
0.1095 98.5104028331119
0.111 98.5258964143425
0.1125 98.5568835768038
0.114 98.6166445329791
0.1155 98.634351482957
0.117 98.6697653829127
0.1185 98.6808322266489
0.12 98.7029659141212
0.1215 98.758300132802
0.123 98.7892872952633
0.1245 98.8446215139441
0.126 98.8933156263833
0.1275 98.9176626826029
0.129 98.9685701637892
0.1305 99.0061974324922
0.132 99.0239043824701
0.1335 99.0416113324479
0.135 99.0593182824258
0.1365 99.0814519698981
0.138 99.156706507304
0.1395 99.1899070385126
0.141 99.2253209384683
0.1425 99.2673749446657
0.144 99.3072155821159
0.1455 99.3204957945993
0.147 99.3492695883133
0.1485 99.3846834882691
0.15 99.3846834882691
};
\addplot [semithick, orange2551570, forget plot]
table {%
0 83.0278884462151
0.0015 83.0278884462151
0.003 83.0278884462151
0.0045 83.0522355024347
0.006 83.1053563523683
0.0075 83.4838424081452
0.009 84.1057990261178
0.0105 84.5595396193006
0.012 85.2810978308986
0.0135 86.1708720672864
0.015 87.3837981407702
0.0165 88.0190349712262
0.018 88.7605135015493
0.0195 89.3979637007525
0.021 90.1438689685702
0.0225 90.834440017707
0.024 91.8525896414343
0.0255 92.4656927844179
0.027 93.0610889774237
0.0285 93.6122177954847
0.03 94.1013722886233
0.0315 94.7476759628154
0.033 95.3142983621071
0.0345 95.7392651615759
0.036 96.1199645861001
0.0375 96.4586100044267
0.039 96.6976538291279
0.0405 96.9212040725985
0.042 97.1513944223108
0.0435 97.3594510845507
0.045 97.6250553342187
0.0465 97.9349269588313
0.048 98.1097830898627
0.0495 98.3266932270916
0.051 98.3975210270031
0.0525 98.5325365205842
0.054 98.6122177954846
0.0555 98.6941124391323
0.057 98.7538733953076
0.0585 98.824701195219
0.06 98.9530765825586
0.0615 99.0349712262062
0.063 99.1522797698096
0.0645 99.2319610447099
0.066 99.3426294820716
0.0675 99.3780433820273
0.069 99.4090305444886
0.0705 99.4665781319167
0.072 99.5086321381141
0.0735 99.5152722443558
0.075 99.5683930942894
0.0765 99.6060203629924
0.078 99.6480743691898
0.0795 99.6746347941566
0.081 99.7122620628595
0.0825 99.7432492253208
0.084 99.754316069057
0.0855 99.7587428065515
0.087 99.7609561752987
0.0885 99.7631695440459
0.09 99.7653829127932
0.0915 99.7698096502876
0.093 99.7742363877821
0.0945 99.7742363877821
0.096 99.783089862771
0.0975 99.7897299690127
0.099 99.8162903939796
0.1005 99.820717131474
0.102 99.8251438689685
0.1035 99.8251438689685
0.105 99.8251438689685
0.1065 99.8251438689685
0.108 99.8251438689685
0.1095 99.8251438689685
0.111 99.8317839752102
0.1125 99.8317839752102
0.114 99.8384240814519
0.1155 99.8384240814519
0.117 99.8406374501992
0.1185 99.8428508189464
0.12 99.8450641876937
0.1215 99.8450641876937
0.123 99.8494909251881
0.1245 99.8494909251881
0.126 99.8517042939353
0.1275 99.8539176626826
0.129 99.8826914563966
0.1305 99.8826914563966
0.132 99.8826914563966
0.1335 99.8826914563966
0.135 99.8826914563966
0.1365 99.8826914563966
0.138 99.8849048251439
0.1395 99.8849048251439
0.141 99.9026117751217
0.1425 99.9026117751217
0.144 99.9026117751217
0.1455 99.9048251438689
0.147 99.9070385126162
0.1485 99.9070385126162
0.15 99.9070385126162
};
\addplot [semithick, gold2552250, forget plot]
table {%
0 89.0261177512173
0.0015 89.0283311199646
0.003 89.0283311199646
0.0045 89.0571049136786
0.006 89.1190792386012
0.0075 89.3182824258522
0.009 89.8649845064188
0.0105 90.1350154935812
0.012 90.8056662239929
0.0135 91.6002656042497
0.015 92.485613103143
0.0165 93.0212483399735
0.018 93.6586985391766
0.0195 94.1279327135901
0.021 94.6193005754759
0.0225 95.1837096060203
0.024 95.8410801239486
0.0255 96.2262062859672
0.027 96.6378928729526
0.0285 96.9189907038512
0.03 97.3262505533422
0.0315 97.6936697653829
0.033 98.0544488711819
0.0345 98.3089862771137
0.036 98.5436033643205
0.0375 98.7649402390438
0.039 98.9110225763612
0.0405 98.9818503762726
0.042 99.1124391323594
0.0435 99.2186808322266
0.045 99.3448428508189
0.0465 99.4577246569278
0.048 99.5064187693669
0.0495 99.5506861443116
0.051 99.5861000442673
0.0525 99.6414342629481
0.054 99.6569278441788
0.0555 99.6746347941566
0.057 99.6812749003983
0.0585 99.68791500664
0.06 99.7277556440902
0.0615 99.7786631252766
0.063 99.7941567065073
0.0645 99.7985834440017
0.066 99.8273572377157
0.0675 99.8339973439574
0.069 99.8362107127047
0.0705 99.8384240814519
0.072 99.8406374501992
0.0735 99.8406374501992
0.075 99.8561310314298
0.0765 99.8605577689242
0.078 99.8605577689242
0.0795 99.8627711376715
0.081 99.89597166888
0.0825 99.89597166888
0.084 99.89597166888
0.0855 99.89597166888
0.087 99.89597166888
0.0885 99.89597166888
0.09 99.89597166888
0.0915 99.89597166888
0.093 99.89597166888
0.0945 99.89597166888
0.096 99.89597166888
0.0975 99.89597166888
0.099 99.8981850376272
0.1005 99.8981850376272
0.102 99.9003984063744
0.1035 99.9003984063744
0.105 99.9003984063744
0.1065 99.9003984063744
0.108 99.9003984063744
0.1095 99.9003984063744
0.111 99.9003984063744
0.1125 99.9003984063744
0.114 99.9003984063744
0.1155 99.9003984063744
0.117 99.9003984063744
0.1185 99.9003984063744
0.12 99.9026117751217
0.1215 99.9026117751217
0.123 99.9026117751217
0.1245 99.9026117751217
0.126 99.9026117751217
0.1275 99.9026117751217
0.129 99.9203187250995
0.1305 99.9203187250995
0.132 99.9203187250995
0.1335 99.9203187250995
0.135 99.9203187250995
0.1365 99.9203187250995
0.138 99.9203187250995
0.1395 99.9203187250995
0.141 99.9203187250995
0.1425 99.9203187250995
0.144 99.9203187250995
0.1455 99.9203187250995
0.147 99.9203187250995
0.1485 99.9203187250995
0.15 99.9203187250995
};
\addplot [semithick, yellow25525554, forget plot]
table {%
0 91.9964586100044
0.0015 91.9986719787517
0.003 91.9986719787517
0.0045 92.0274457724657
0.006 92.0805666223993
0.0075 92.3085436033643
0.009 92.8818061088977
0.0105 93.1983178397521
0.012 93.7184594953519
0.0135 94.1434262948207
0.015 94.829570606463
0.0165 95.2368304559539
0.018 95.7193448428508
0.0195 96.0779105799026
0.021 96.4497565294378
0.0225 96.8459495351925
0.024 97.2664895971669
0.0255 97.5719344842851
0.027 97.817618415228
0.0285 97.9836210712705
0.03 98.2248782647189
0.0315 98.4528552456839
0.033 98.6564851704293
0.0345 98.8490482514387
0.036 98.9907038512616
0.0375 99.1611332447985
0.039 99.2629482071712
0.0405 99.3315626383355
0.042 99.4555112881806
0.0435 99.5462594068171
0.045 99.6347941567064
0.0465 99.734395750332
0.048 99.7742363877821
0.0495 99.8007968127489
0.051 99.8162903939796
0.0525 99.8317839752102
0.054 99.8339973439574
0.0555 99.8362107127046
0.057 99.8384240814519
0.0585 99.8428508189463
0.06 99.8539176626825
0.0615 99.8738379814076
0.063 99.8738379814076
0.0645 99.8738379814076
0.066 99.8937583001327
0.0675 99.8937583001327
0.069 99.8937583001327
0.0705 99.8937583001327
0.072 99.89597166888
0.0735 99.89597166888
0.075 99.9136786188578
0.0765 99.9136786188578
0.078 99.9136786188578
0.0795 99.9136786188578
0.081 99.9158919876051
0.0825 99.9181053563523
0.084 99.9181053563523
0.0855 99.9181053563523
0.087 99.9203187250995
0.0885 99.9203187250995
0.09 99.9203187250995
0.0915 99.9203187250995
0.093 99.9203187250995
0.0945 99.9203187250995
0.096 99.9203187250995
0.0975 99.9203187250995
0.099 99.9203187250995
0.1005 99.9203187250995
0.102 99.9203187250995
0.1035 99.9203187250995
0.105 99.9203187250995
0.1065 99.9203187250995
0.108 99.9203187250995
0.1095 99.9203187250995
0.111 99.9203187250995
0.1125 99.9203187250995
0.114 99.9203187250995
0.1155 99.9203187250995
0.117 99.9203187250995
0.1185 99.9203187250995
0.12 99.9203187250995
0.1215 99.9203187250995
0.123 99.9203187250995
0.1245 99.9203187250995
0.126 99.9203187250995
0.1275 99.9203187250995
0.129 99.9203187250995
0.1305 99.9203187250995
0.132 99.9203187250995
0.1335 99.9203187250995
0.135 99.9203187250995
0.1365 99.9203187250995
0.138 99.9203187250995
0.1395 99.9203187250995
0.141 99.9203187250995
0.1425 99.9203187250995
0.144 99.9203187250995
0.1455 99.9203187250995
0.147 99.9203187250995
0.1485 99.9203187250995
0.15 99.9203187250995
};
\addplot [semithick, palegoldenrod255255156, forget plot]
table {%
0 93.8733953076583
0.0015 93.8756086764055
0.003 93.8756086764055
0.0045 93.8866755201417
0.006 93.9353696325809
0.0075 94.1235059760957
0.009 94.6193005754758
0.0105 94.9092518813635
0.012 95.2899513058876
0.0135 95.5621956617973
0.015 96.1066843736166
0.0165 96.394422310757
0.018 96.7330677290836
0.0195 96.9433377600708
0.021 97.211155378486
0.0225 97.3948649845064
0.024 97.6914563966357
0.0255 97.788844621514
0.027 98.0699424524126
0.0285 98.2492253209384
0.03 98.501549358123
0.0315 98.7782204515272
0.033 99.0548915449314
0.0345 99.2363877822045
0.036 99.355909694555
0.0375 99.4710048694112
0.039 99.5219123505975
0.0405 99.6015936254979
0.042 99.7011952191234
0.0435 99.7587428065515
0.045 99.7875166002655
0.0465 99.7985834440017
0.048 99.8273572377158
0.0495 99.8384240814519
0.051 99.8539176626826
0.0525 99.8671978751659
0.054 99.8782647189021
0.0555 99.8871181938911
0.057 99.9092518813634
0.0585 99.9092518813634
0.06 99.9114652501107
0.0615 99.9136786188579
0.063 99.9136786188579
0.0645 99.9136786188579
0.066 99.933598937583
0.0675 99.933598937583
0.069 99.933598937583
0.0705 99.933598937583
0.072 99.9358123063302
0.0735 99.9358123063302
0.075 99.9535192563081
0.0765 99.9535192563081
0.078 99.9535192563081
0.0795 99.9535192563081
0.081 99.9557326250553
0.0825 99.9579459938026
0.084 99.9579459938026
0.0855 99.9579459938026
0.087 99.9601593625498
0.0885 99.9601593625498
0.09 99.9601593625498
0.0915 99.9601593625498
0.093 99.9601593625498
0.0945 99.9601593625498
0.096 99.9601593625498
0.0975 99.9601593625498
0.099 99.9601593625498
0.1005 99.9601593625498
0.102 99.9601593625498
0.1035 99.9601593625498
0.105 99.9601593625498
0.1065 99.9601593625498
0.108 99.9601593625498
0.1095 99.9601593625498
0.111 99.9601593625498
0.1125 99.9601593625498
0.114 99.9601593625498
0.1155 99.9601593625498
0.117 99.9601593625498
0.1185 99.9601593625498
0.12 99.9601593625498
0.1215 99.9601593625498
0.123 99.9601593625498
0.1245 99.9601593625498
0.126 99.9601593625498
0.1275 99.9601593625498
0.129 99.9601593625498
0.1305 99.9601593625498
0.132 99.9601593625498
0.1335 99.9601593625498
0.135 99.9601593625498
0.1365 99.9601593625498
0.138 99.9601593625498
0.1395 99.9601593625498
0.141 99.9601593625498
0.1425 99.9601593625498
0.144 99.9601593625498
0.1455 99.9601593625498
0.147 99.9601593625498
0.1485 99.9601593625498
0.15 99.9601593625498
};
\addplot [semithick, white, forget plot]
table {%
0 94.1921204072599
0.0015 94.1943337760071
0.003 94.1943337760071
0.0045 94.2054006197433
0.006 94.2540947321824
0.0075 94.4422310756972
0.009 94.9468791500665
0.0105 95.2567507746791
0.012 95.6662239929172
0.0135 95.9096945551128
0.015 96.385568835768
0.0165 96.6954404603807
0.018 96.99203187251
0.0195 97.1691013722886
0.021 97.4656927844178
0.0225 97.6980965028773
0.024 98.0256750774679
0.0255 98.1363435148295
0.027 98.4639220894201
0.0285 98.5967242142541
0.03 98.8092076139885
0.0315 99.0106241699867
0.033 99.1810535635236
0.0345 99.4046038069942
0.036 99.5285524568393
0.0375 99.6458610004426
0.039 99.696768481629
0.0405 99.7388224878264
0.042 99.8229305002213
0.0435 99.8317839752102
0.045 99.8517042939353
0.0465 99.858344400177
0.048 99.8826914563966
0.0495 99.8871181938911
0.051 99.8937583001328
0.0525 99.9048251438689
0.054 99.9070385126162
0.0555 99.9092518813634
0.057 99.9092518813634
0.0585 99.9092518813634
0.06 99.9114652501107
0.0615 99.9136786188579
0.063 99.9158919876051
0.0645 99.9158919876051
0.066 99.9380256750775
0.0675 99.9380256750775
0.069 99.9380256750775
0.0705 99.9380256750775
0.072 99.9402390438247
0.0735 99.9402390438247
0.075 99.9579459938025
0.0765 99.9579459938025
0.078 99.9579459938025
0.0795 99.9579459938025
0.081 99.9601593625498
0.0825 99.9601593625498
0.084 99.9601593625498
0.0855 99.9601593625498
0.087 99.9601593625498
0.0885 99.9601593625498
0.09 99.9601593625498
0.0915 99.9601593625498
0.093 99.9601593625498
0.0945 99.9601593625498
0.096 99.9601593625498
0.0975 99.9601593625498
0.099 99.9601593625498
0.1005 99.9601593625498
0.102 99.9601593625498
0.1035 99.9601593625498
0.105 99.9601593625498
0.1065 99.9601593625498
0.108 99.9601593625498
0.1095 99.9601593625498
0.111 99.9601593625498
0.1125 99.9601593625498
0.114 99.9601593625498
0.1155 99.9601593625498
0.117 99.9601593625498
0.1185 99.9601593625498
0.12 99.9601593625498
0.1215 99.9601593625498
0.123 99.9601593625498
0.1245 99.9601593625498
0.126 99.9601593625498
0.1275 99.9601593625498
0.129 99.9601593625498
0.1305 99.9601593625498
0.132 99.9601593625498
0.1335 99.9601593625498
0.135 99.9601593625498
0.1365 99.9601593625498
0.138 99.9601593625498
0.1395 99.9601593625498
0.141 99.9601593625498
0.1425 99.9601593625498
0.144 99.9601593625498
0.1455 99.9601593625498
0.147 99.9601593625498
0.1485 99.9601593625498
0.15 99.9601593625498
};
\end{axis}

\end{tikzpicture}

%% file: SupplimentFigures/Energy_Change.tex
\begin{tikzpicture}

\definecolor{darkorange25512714}{RGB}{255,127,14}
\definecolor{darkslategray38}{RGB}{38,38,38}
\definecolor{lightgray204}{RGB}{204,204,204}
\definecolor{steelblue31119180}{RGB}{31,119,180}

\Large
\begin{axis}[
axis line style={lightgray204},
legend cell align={left},
legend style={fill opacity=0.8, draw opacity=1, text opacity=1, draw=none},
tick align=outside,
tick pos=left,
x grid style={lightgray204},
xlabel=\textcolor{darkslategray38}{Iteration},
xmajorgrids,
xmin=0, xmax=200,
xtick style={color=darkslategray38},
y grid style={lightgray204},
ylabel=\textcolor{darkslategray38}{Energy},
ymajorgrids,
xticklabel style={yshift= 5pt},
yticklabel style={xshift= 5pt},
ymin=406793.434551037, ymax=1460495.06139121,
ytick style={color=darkslategray38},
ytick={400000,600000,800000,1000000,1200000,1400000,1600000},
yticklabels={0.4,0.6,0.8,1.0,1.2,1.4,1.6}
]
\addplot [ultra thick, steelblue31119180, opacity=0.7]
table {%
0 1412599.53289847
1 1183293.77062305
2 1072454.6975754
3 898482.753120199
4 895513.465773958
5 861165.121138548
6 745710.510909138
7 725822.194337174
8 711194.094283449
9 702845.217672286
10 699956.366642935
11 690091.676691849
12 683512.468663089
13 670394.810387172
14 662531.853289157
15 661283.563884537
16 658912.712604156
17 657761.436602939
18 656099.394278964
19 655082.938715587
20 653025.401642406
21 649695.31989293
22 647076.057273849
23 645285.878078882
24 645012.821138859
25 642893.777253901
26 641544.632198263
27 640242.039463242
28 633228.297661631
29 632648.501476486
30 629989.009194106
31 627303.857346239
32 627013.617889443
33 623208.217561243
34 621941.868290954
35 621202.415953106
36 620380.197402601
37 617807.079543611
38 616974.375787247
39 613343.771662781
40 613021.882862556
41 612794.081019129
42 612439.243610229
43 610628.68898855
44 609848.354258433
45 610443.81083058
46 609693.324223531
47 610074.206729221
48 608031.165399297
49 607629.705308602
50 606638.107575965
51 605984.900534482
52 605916.949109419
53 605813.03797345
54 604262.320120821
55 603764.1751954
56 602837.757346848
57 601391.782918322
58 601157.596220475
59 601123.542646458
60 600474.238052119
61 599996.925854989
62 599157.437784079
63 596513.444950086
64 596381.941649309
65 595510.394673576
66 595202.489398315
67 595165.722495269
68 594367.110504942
69 594055.9940645
70 592900.371104277
71 592885.2828775
72 590598.814343286
73 590661.779999414
74 590002.082943428
75 589938.29574777
76 589121.071902582
77 589957.143187013
78 591111.174671788
79 589686.549271123
80 587873.885021586
81 586808.061773484
82 586143.282542341
83 586030.072096886
84 585893.271688835
85 587086.748757131
86 586913.379231967
87 585955.48156708
88 586348.192112311
89 586333.303944257
90 586330.755202951
91 585882.080217232
92 585713.905784788
93 585141.496080228
94 584504.096818426
95 584993.277509551
96 584807.718793883
97 584845.291712538
98 585615.28282306
99 578886.853490675
};
\addlegendentry{Geodesic Distances}
\addplot [ultra thick, darkorange25512714, opacity=0.7]
table {%
100 576840.021402506
101 573530.952997623
102 567924.947849655
103 564379.118272291
104 554017.619681238
105 548102.527810449
106 544845.820683948
107 539614.350585595
108 536378.545716367
109 536261.087739481
110 534273.19758171
111 520882.404063444
112 514231.523811859
113 509932.280156059
114 507244.337288239
115 497030.195362117
116 496029.794819013
117 494256.945363337
118 492368.307323401
119 489450.59238252
120 487972.472939164
121 485781.882601781
122 483384.576253682
123 481120.692590119
124 480479.449344539
125 480347.378035902
126 480474.779354289
127 480272.831157448
128 480136.821383942
129 477092.45274907
130 475298.750603847
131 474461.412423961
132 472837.595204226
133 470872.134176471
134 470779.83527647
135 470557.038687321
136 470425.952850023
137 470114.277416967
138 466733.061625208
139 466038.781682035
140 465997.785615338
141 465702.627932939
142 463677.236469406
143 463677.236469406
144 463526.968502686
145 463613.353369596
146 463756.321559119
147 463398.60316162
148 463398.60316162
149 463538.801183693
150 461463.1988533
151 461284.919335796
152 461285.389410617
153 461298.629044072
154 459127.055741884
155 459218.819948569
156 459126.164127098
157 459126.164127098
158 459132.474202224
159 459132.474202224
160 459067.376111775
161 459058.244597648
162 459018.754495017
163 458908.856551856
164 458834.306698186
165 458251.07859838
166 458132.308423048
167 458132.308423048
168 457586.333582372
169 457149.95343917
170 457807.711916646
171 457388.451009342
172 457321.730705592
173 457321.730705592
174 457321.730705592
175 457321.730705592
176 457036.656321891
177 456517.127315651
178 457203.6052235
179 457203.6052235
180 456992.743887026
181 456992.743887026
182 456992.743887026
183 457003.555488247
184 456628.392774066
185 456371.173240389
186 456371.173240389
187 455206.561356197
188 455062.504217137
189 454871.926787803
190 454939.016781403
191 454979.948351088
192 454765.030021894
193 454688.963043772
194 454754.987391642
195 454761.298556856
196 454736.419213906
};
\addlegendentry{Gaussian Kernel}
\path [draw=red, draw opacity=0.5, very thick, dash pattern=on 11.1pt off 4.8pt]
(axis cs:0,455693.978434343)
--(axis cs:200,455693.978434343);

\end{axis}

\end{tikzpicture}

%% file: SupplimentFigures/PCK_Evolution_TOSCA.tex
\begin{tikzpicture}

\definecolor{black1000}{RGB}{10,0,0}
\definecolor{darkorange2551200}{RGB}{255,120,0}
\definecolor{darkslategray38}{RGB}{38,38,38}
\definecolor{firebrick19100}{RGB}{191,0,0}
\definecolor{lightgray204}{RGB}{204,204,204}
\definecolor{maroon13100}{RGB}{131,0,0}
\definecolor{maroon7000}{RGB}{70,0,0}
\definecolor{orange2551800}{RGB}{255,180,0}
\definecolor{orangered255600}{RGB}{255,60,0}
\definecolor{palegoldenrod255255164}{RGB}{255,255,164}
\definecolor{red25400}{RGB}{254,0,0}
\definecolor{yellow2552430}{RGB}{255,243,0}
\definecolor{yellow25525573}{RGB}{255,255,73}

\Large
\begin{axis}[
axis line style={lightgray204},
tick align=outside,
tick pos=left,
x grid style={lightgray204},
xtick = {0,0.05,0.1,0.15},
xticklabel style={yshift= 5pt},
xlabel=\textcolor{darkslategray38}{Geodesic Error},
xmajorgrids,
xmin=0, xmax=0.15,
xtick style={color=darkslategray38},
y grid style={lightgray204},
ylabel=\textcolor{darkslategray38}{\% Correspondences},
ymajorgrids,
ymin=20.5604249667995, ymax=100,
ytick style={color=darkslategray38}
]
\addplot [semithick, black1000]
table {%
0 23.2664266426643
0.0015 23.6561656165617
0.003 24.4023402340234
0.0045 25.4194419441944
0.006 26.5616561656166
0.0075 27.8631863186319
0.009 29.4041404140414
0.0105 30.8406840684068
0.012 32.3645364536453
0.0135 33.9414941494149
0.015 35.3231323132313
0.0165 36.5679567956796
0.018 37.8838883888389
0.0195 39.1863186318632
0.021 40.4410441044105
0.0225 41.6417641764176
0.024 42.7713771377138
0.0255 43.8109810981098
0.027 44.7956795679568
0.0285 45.7848784878488
0.03 46.8199819981998
0.0315 47.7146714671467
0.033 48.7164716471647
0.0345 49.6165616561656
0.036 50.6948694869487
0.0375 51.6084608460846
0.039 52.5283528352835
0.0405 53.4149414941494
0.042 54.1629162916292
0.0435 54.8775877587759
0.045 55.6183618361836
0.0465 56.3348334833483
0.048 56.9585958595859
0.0495 57.6129612961296
0.051 58.2628262826283
0.0525 58.8730873087309
0.054 59.4869486948695
0.0555 60.0522052205221
0.057 60.5949594959496
0.0585 61.1134113411341
0.06 61.6309630963096
0.0615 62.1161116111611
0.063 62.6129612961296
0.0645 63.0315031503151
0.066 63.4338433843384
0.0675 63.7776777677768
0.069 64.1170117011701
0.0705 64.4770477047705
0.072 64.8145814581458
0.0735 65.1611161116112
0.075 65.4455445544554
0.0765 65.8145814581458
0.078 66.1611161116112
0.0795 66.4986498649865
0.081 66.8082808280828
0.0825 67.1521152115211
0.084 67.4545454545454
0.0855 67.7137713771377
0.087 68.041404140414
0.0885 68.2745274527453
0.09 68.5301530153015
0.0915 68.7983798379838
0.093 69.03600360036
0.0945 69.3051305130513
0.096 69.5508550855085
0.0975 69.7794779477947
0.099 70.026102610261
0.1005 70.2610261026103
0.102 70.4968496849685
0.1035 70.7335733573357
0.105 70.972097209721
0.1065 71.1890189018902
0.108 71.4023402340234
0.1095 71.6228622862287
0.111 71.8253825382538
0.1125 72.036903690369
0.114 72.2412241224122
0.1155 72.4554455445544
0.117 72.6777677767777
0.1185 72.8784878487849
0.12 73.0612061206121
0.1215 73.2664266426643
0.123 73.4347434743474
0.1245 73.6021602160216
0.126 73.7929792979298
0.1275 73.9585958595859
0.129 74.1692169216922
0.1305 74.3564356435643
0.132 74.5220522052205
0.1335 74.7083708370837
0.135 74.8550855085508
0.1365 75.019801980198
0.138 75.1818181818182
0.1395 75.3240324032403
0.141 75.4977497749775
0.1425 75.6561656165616
0.144 75.8244824482448
0.1455 75.98199819982
0.147 76.1395139513951
0.1485 76.2763276327633
0.15 76.2763276327633
};
\addplot [semithick, maroon7000]
table {%
0 21.7371737173717
0.0015 22.1566156615662
0.003 22.9243924392439
0.0045 24.1773177317732
0.006 25.6579657965797
0.0075 27.4185418541854
0.009 29.2268226822682
0.0105 30.997299729973
0.012 32.8793879387939
0.0135 34.8955895589559
0.015 36.7830783078308
0.0165 38.5094509450945
0.018 40.2754275427543
0.0195 41.9702970297029
0.021 43.7200720072007
0.0225 45.3618361836184
0.024 46.8802880288029
0.0255 48.3186318631863
0.027 49.6867686768677
0.0285 51.028802880288
0.03 52.4545454545454
0.0315 53.7272727272728
0.033 54.990099009901
0.0345 56.2880288028803
0.036 57.6732673267326
0.0375 58.8856885688569
0.039 60.030603060306
0.0405 61.1044104410441
0.042 62.0918091809181
0.0435 63.1269126912691
0.045 64.1827182718272
0.0465 65.1944194419442
0.048 66.1539153915391
0.0495 67.0891089108911
0.051 67.944194419442
0.0525 68.8172817281728
0.054 69.6228622862286
0.0555 70.4770477047705
0.057 71.3114311431143
0.0585 72.0747074707471
0.06 72.9243924392439
0.0615 73.7065706570657
0.063 74.4410441044105
0.0645 75.2412241224122
0.066 75.960396039604
0.0675 76.6525652565256
0.069 77.3069306930693
0.0705 77.907290729073
0.072 78.5049504950495
0.0735 79.0765076507651
0.075 79.6102610261026
0.0765 80.1917191719172
0.078 80.7947794779478
0.0795 81.2772277227723
0.081 81.7659765976598
0.0825 82.2547254725472
0.084 82.7668766876688
0.0855 83.1629162916292
0.087 83.6552655265526
0.0885 84.0702070207021
0.09 84.4797479747975
0.0915 84.8370837083708
0.093 85.2520252025202
0.0945 85.6327632763276
0.096 85.967596759676
0.0975 86.3105310531053
0.099 86.7182718271827
0.1005 87.0738073807381
0.102 87.4599459945994
0.1035 87.7794779477948
0.105 88.1638163816382
0.1065 88.4752475247525
0.108 88.7803780378038
0.1095 89.1071107110711
0.111 89.4086408640864
0.1125 89.7011701170117
0.114 90
0.1155 90.2808280828083
0.117 90.5733573357336
0.1185 90.8397839783978
0.12 91.1233123312332
0.1215 91.3807380738074
0.123 91.5976597659766
0.1245 91.8388838883888
0.126 92.0558055805581
0.1275 92.2763276327633
0.129 92.5112511251126
0.1305 92.7578757875788
0.132 92.9738973897389
0.1335 93.1863186318631
0.135 93.3789378937893
0.1365 93.5571557155716
0.138 93.7164716471647
0.1395 93.9099909990999
0.141 94.0828082808281
0.1425 94.2520252025203
0.144 94.4401440144014
0.1455 94.5940594059406
0.147 94.7713771377138
0.1485 94.9162916291629
0.15 94.9162916291629
};
\addplot [semithick, maroon13100]
table {%
0 24.5391539153915
0.0015 25.001800180018
0.003 25.9045904590459
0.0045 27.2628262826283
0.006 28.8343834383439
0.0075 30.7686768676868
0.009 32.7812781278128
0.0105 34.6642664266427
0.012 36.6795679567957
0.0135 38.8829882988299
0.015 40.8649864986499
0.0165 42.7227722772277
0.018 44.5355535553555
0.0195 46.3501350135014
0.021 48.033303330333
0.0225 49.7362736273628
0.024 51.2430243024302
0.0255 52.7020702070207
0.027 54.2682268226823
0.0285 55.6318631863186
0.03 57.026102610261
0.0315 58.2925292529253
0.033 59.4599459945995
0.0345 60.6201620162016
0.036 61.9189918991899
0.0375 63.041404140414
0.039 64.1809180918092
0.0405 65.2700270027003
0.042 66.2475247524752
0.0435 67.2601260126013
0.045 68.2376237623763
0.0465 69.1953195319532
0.048 70.1665166516652
0.0495 71.1512151215122
0.051 72.001800180018
0.0525 72.8523852385239
0.054 73.6948694869487
0.0555 74.5211521152115
0.057 75.3348334833484
0.0585 76.0810081008101
0.06 76.8613861386139
0.0615 77.6228622862286
0.063 78.3771377137714
0.0645 79.1314131413141
0.066 79.8505850585058
0.0675 80.5220522052205
0.069 81.1080108010801
0.0705 81.6759675967597
0.072 82.2556255625563
0.0735 82.8208820882088
0.075 83.3411341134113
0.0765 83.8928892889289
0.078 84.4149414941494
0.0795 84.8748874887489
0.081 85.3492349234924
0.0825 85.8676867686768
0.084 86.3519351935194
0.0855 86.7695769576957
0.087 87.2583258325833
0.0885 87.6228622862286
0.09 88.020702070207
0.0915 88.3978397839784
0.093 88.7704770477048
0.0945 89.0954095409541
0.096 89.4059405940594
0.0975 89.7425742574257
0.099 90.0756075607561
0.1005 90.3735373537354
0.102 90.6795679567957
0.1035 90.931593159316
0.105 91.2304230423043
0.1065 91.5049504950495
0.108 91.7704770477047
0.1095 92.038703870387
0.111 92.2889288928893
0.1125 92.5436543654365
0.114 92.7848784878488
0.1155 93.040504050405
0.117 93.2745274527453
0.1185 93.5202520252025
0.12 93.7443744374438
0.1215 93.9441944194419
0.123 94.1251125112511
0.1245 94.3087308730873
0.126 94.4851485148515
0.1275 94.6462646264627
0.129 94.8262826282628
0.1305 95.002700270027
0.132 95.1512151215122
0.1335 95.3114311431143
0.135 95.4482448244825
0.1365 95.5922592259226
0.138 95.7299729972997
0.1395 95.8577857785778
0.141 95.9936993699369
0.1425 96.1197119711971
0.144 96.2682268226823
0.1455 96.3924392439244
0.147 96.5022502250225
0.1485 96.6228622862287
0.15 96.6228622862287
};
\addplot [semithick, firebrick19100]
table {%
0 26.5616561656166
0.0015 27.019801980198
0.003 28.0612061206121
0.0045 29.4707470747075
0.006 31.2520252025203
0.0075 33.3150315031503
0.009 35.2430243024302
0.0105 37.1494149414942
0.012 39.1737173717372
0.0135 41.4491449144915
0.015 43.4428442844284
0.0165 45.1890189018902
0.018 47.029702970297
0.0195 48.8586858685869
0.021 50.5058505850585
0.0225 52.2457245724573
0.024 53.7785778577858
0.0255 55.2367236723672
0.027 56.6471647164717
0.0285 57.957695769577
0.03 59.2844284428443
0.0315 60.5103510351035
0.033 61.6192619261927
0.0345 62.7614761476147
0.036 63.9108910891089
0.0375 64.976597659766
0.039 66.023402340234
0.0405 67.038703870387
0.042 67.976597659766
0.0435 68.9531953195319
0.045 69.8919891989199
0.0465 70.8388838883888
0.048 71.7569756975697
0.0495 72.7047704770477
0.051 73.4698469846985
0.0525 74.2709270927093
0.054 74.999099909991
0.0555 75.8028802880288
0.057 76.5499549954996
0.0585 77.2556255625562
0.06 78.0378037803781
0.0615 78.7740774077408
0.063 79.4959495949595
0.0645 80.2205220522052
0.066 80.8613861386139
0.0675 81.4473447344735
0.069 82.026102610261
0.0705 82.5895589558956
0.072 83.1647164716472
0.0735 83.6993699369937
0.075 84.2331233123312
0.0765 84.7542754275428
0.078 85.2313231323133
0.0795 85.6588658865886
0.081 86.1170117011701
0.0825 86.6120612061206
0.084 87.0630063006301
0.0855 87.4815481548155
0.087 87.934293429343
0.0885 88.2943294329433
0.09 88.6714671467147
0.0915 89.0216021602161
0.093 89.3834383438344
0.0945 89.7119711971197
0.096 90.026102610261
0.0975 90.3528352835284
0.099 90.7029702970297
0.1005 91.015301530153
0.102 91.3186318631863
0.1035 91.6039603960396
0.105 91.8901890189019
0.1065 92.1557155715572
0.108 92.4140414041404
0.1095 92.6822682268227
0.111 92.9333933393339
0.1125 93.1872187218722
0.114 93.4563456345635
0.1155 93.7119711971197
0.117 93.9369936993699
0.1185 94.1620162016201
0.12 94.3636363636364
0.1215 94.5508550855086
0.123 94.7155715571557
0.1245 94.9090909090909
0.126 95.0684068406841
0.1275 95.2061206120612
0.129 95.3546354635463
0.1305 95.5328532853285
0.132 95.6849684968497
0.1335 95.8298829882988
0.135 95.950495049505
0.1365 96.0783078307831
0.138 96.2070207020702
0.1395 96.3222322232223
0.141 96.4347434743474
0.1425 96.5517551755175
0.144 96.6804680468047
0.1455 96.7884788478848
0.147 96.8991899189919
0.1485 97
0.15 97
};
\addplot [semithick, red25400]
table {%
0 28.0594059405941
0.0015 28.4842484248425
0.003 29.4626462646265
0.0045 30.8244824482448
0.006 32.6345634563456
0.0075 34.7011701170117
0.009 36.7047704770477
0.0105 38.6111611161116
0.012 40.6813681368137
0.0135 43.0495049504951
0.015 45.041404140414
0.0165 46.7911791179118
0.018 48.5562556255625
0.0195 50.4401440144014
0.021 52.0864086408641
0.0225 53.7776777677768
0.024 55.2925292529253
0.0255 56.7668766876688
0.027 58.1863186318632
0.0285 59.5247524752475
0.03 60.8073807380738
0.0315 61.990099009901
0.033 63.0810081008101
0.0345 64.2196219621962
0.036 65.3519351935194
0.0375 66.4095409540954
0.039 67.4896489648965
0.0405 68.5391539153915
0.042 69.4680468046804
0.0435 70.4275427542754
0.045 71.3294329432943
0.0465 72.2376237623762
0.048 73.0936093609361
0.0495 73.9612961296129
0.051 74.7308730873088
0.0525 75.4905490549055
0.054 76.1683168316832
0.0555 76.9252925292529
0.057 77.5814581458146
0.0585 78.2529252925293
0.06 78.968496849685
0.0615 79.7335733573357
0.063 80.4230423042304
0.0645 81.0990099009901
0.066 81.7038703870387
0.0675 82.3060306030603
0.069 82.8838883888389
0.0705 83.4302430243024
0.072 83.975697569757
0.0735 84.4905490549055
0.075 85.012601260126
0.0765 85.5274527452746
0.078 85.999099909991
0.0795 86.4203420342034
0.081 86.8523852385239
0.0825 87.3420342034204
0.084 87.7524752475248
0.0855 88.1647164716471
0.087 88.6039603960396
0.0885 88.980198019802
0.09 89.3447344734474
0.0915 89.6714671467147
0.093 90.031503150315
0.0945 90.3312331233123
0.096 90.6480648064806
0.0975 90.998199819982
0.099 91.3249324932493
0.1005 91.6156615661566
0.102 91.8964896489649
0.1035 92.1449144914491
0.105 92.4176417641764
0.1065 92.6678667866787
0.108 92.930693069307
0.1095 93.1602160216022
0.111 93.3636363636364
0.1125 93.5742574257426
0.114 93.8325832583258
0.1155 94.0630063006301
0.117 94.2664266426643
0.1185 94.4779477947795
0.12 94.6696669666966
0.1215 94.8307830783077
0.123 95.006300630063
0.1245 95.1962196219622
0.126 95.3636363636364
0.1275 95.5031503150315
0.129 95.6462646264627
0.1305 95.8208820882088
0.132 95.964896489649
0.1335 96.0891089108911
0.135 96.2394239423943
0.1365 96.3681368136814
0.138 96.4833483348335
0.1395 96.5913591359136
0.141 96.7029702970297
0.1425 96.8235823582358
0.144 96.934293429343
0.1455 97.029702970297
0.147 97.1323132313231
0.1485 97.2142214221422
0.15 97.2142214221422
};
\addplot [semithick, orangered255600]
table {%
0 28.8316831683168
0.0015 29.2520252025202
0.003 30.2349234923492
0.0045 31.6597659765977
0.006 33.5886588658866
0.0075 35.6156615661566
0.009 37.6723672367237
0.0105 39.5823582358236
0.012 41.7110711071107
0.0135 44.021602160216
0.015 46.0621062106211
0.0165 47.9135913591359
0.018 49.7236723672367
0.0195 51.4986498649865
0.021 53.1485148514852
0.0225 54.8847884788479
0.024 56.4320432043205
0.0255 57.8613861386139
0.027 59.2151215121512
0.0285 60.4806480648065
0.03 61.6984698469847
0.0315 62.8523852385239
0.033 63.9108910891089
0.0345 64.950495049505
0.036 66.1548154815481
0.0375 67.1827182718272
0.039 68.2826282628263
0.0405 69.3078307830783
0.042 70.2322232223222
0.0435 71.2133213321332
0.045 72.1611161116112
0.0465 73.068406840684
0.048 73.9198919891989
0.0495 74.8343834383438
0.051 75.6336633663366
0.0525 76.3528352835284
0.054 77.015301530153
0.0555 77.7902790279028
0.057 78.4509450945094
0.0585 79.1314131413141
0.06 79.8010801080108
0.0615 80.4671467146715
0.063 81.1458145814582
0.0645 81.8028802880288
0.066 82.3969396939693
0.0675 82.960396039604
0.069 83.5229522952295
0.0705 84.030603060306
0.072 84.5589558955895
0.0735 85.0675067506751
0.075 85.6093609360936
0.0765 86.1188118811881
0.078 86.5589558955896
0.0795 86.9558955895589
0.081 87.3708370837084
0.0825 87.8703870387039
0.084 88.2655265526552
0.0855 88.6624662466246
0.087 89.075607560756
0.0885 89.4635463546355
0.09 89.8343834383439
0.0915 90.1440144014401
0.093 90.4797479747975
0.0945 90.7686768676868
0.096 91.0405040504051
0.0975 91.3555355535553
0.099 91.6903690369037
0.1005 91.990099009901
0.102 92.2700270027003
0.1035 92.5076507650765
0.105 92.7722772277228
0.1065 93.026102610261
0.108 93.2907290729073
0.1095 93.5238523852385
0.111 93.7335733573357
0.1125 93.9468946894689
0.114 94.1980198019802
0.1155 94.4122412241224
0.117 94.6282628262826
0.1185 94.8487848784879
0.12 95.0414041404141
0.1215 95.2214221422142
0.123 95.3798379837983
0.1245 95.5553555355535
0.126 95.6993699369937
0.1275 95.8208820882089
0.129 95.96399639964
0.1305 96.1287128712872
0.132 96.2655265526553
0.1335 96.3969396939694
0.135 96.5400540054006
0.1365 96.6687668766877
0.138 96.7767776777678
0.1395 96.8829882988299
0.141 96.977497749775
0.1425 97.0873087308731
0.144 97.1917191719171
0.1455 97.2754275427542
0.147 97.3636363636363
0.1485 97.4437443744374
0.15 97.4437443744374
};
\addplot [semithick, darkorange2551200]
table {%
0 31.4644464446445
0.0015 31.8631863186319
0.003 32.8100810081008
0.0045 34.1638163816382
0.006 36.1854185418542
0.0075 38.4617461746175
0.009 40.5823582358236
0.0105 42.6201620162016
0.012 44.8838883888389
0.0135 47.2709270927093
0.015 49.3924392439244
0.0165 51.2817281728173
0.018 53.0999099909991
0.0195 54.97299729973
0.021 56.6957695769577
0.0225 58.4860486048605
0.024 60.0891089108911
0.0255 61.5337533753376
0.027 62.9198919891989
0.0285 64.2025202520252
0.03 65.4464446444644
0.0315 66.7056705670567
0.033 67.7875787578758
0.0345 68.971197119712
0.036 70.2403240324032
0.0375 71.2952295229523
0.039 72.4113411341134
0.0405 73.4707470747074
0.042 74.3798379837984
0.0435 75.4068406840684
0.045 76.3177317731773
0.0465 77.2304230423042
0.048 78.1026102610261
0.0495 79
0.051 79.7677767776778
0.0525 80.5274527452746
0.054 81.1863186318632
0.0555 81.957695769577
0.057 82.6048604860486
0.0585 83.2466246624663
0.06 83.955895589559
0.0615 84.5580558055806
0.063 85.1503150315032
0.0645 85.7182718271827
0.066 86.2367236723673
0.0675 86.7803780378037
0.069 87.2709270927093
0.0705 87.7515751575158
0.072 88.2394239423942
0.0735 88.7299729972997
0.075 89.1629162916291
0.0765 89.5841584158416
0.078 89.956795679568
0.0795 90.3141314131413
0.081 90.6822682268227
0.0825 91.1017101710171
0.084 91.4887488748874
0.0855 91.8235823582359
0.087 92.2124212421242
0.0885 92.4707470747074
0.09 92.8019801980198
0.0915 93.0927092709271
0.093 93.3429342934293
0.0945 93.5814581458146
0.096 93.7938793879388
0.0975 94.068406840684
0.099 94.3231323132313
0.1005 94.5454545454546
0.102 94.7794779477948
0.1035 94.989198919892
0.105 95.2349234923492
0.1065 95.4113411341134
0.108 95.6237623762376
0.1095 95.7821782178218
0.111 95.945094509451
0.1125 96.090909090909
0.114 96.2763276327633
0.1155 96.4356435643565
0.117 96.6192619261926
0.1185 96.7785778577858
0.12 96.8919891989199
0.1215 96.978397839784
0.123 97.1044104410441
0.1245 97.2241224122413
0.126 97.3267326732674
0.1275 97.4113411341134
0.129 97.4986498649864
0.1305 97.6102610261026
0.132 97.7065706570657
0.1335 97.7794779477948
0.135 97.8748874887489
0.1365 97.962196219622
0.138 98.044104410441
0.1395 98.1125112511251
0.141 98.1800180018001
0.1425 98.2547254725473
0.144 98.3375337533754
0.1455 98.4005400540054
0.147 98.4617461746175
0.1485 98.5202520252025
0.15 98.5202520252025
};
\addplot [semithick, orange2551800]
table {%
0 38.9153915391539
0.0015 39.3816381638164
0.003 40.5715571557156
0.0045 42.1080108010801
0.006 44.4644464446445
0.0075 47.0999099909991
0.009 49.6057605760576
0.0105 51.973897389739
0.012 54.4194419441944
0.0135 56.953195319532
0.015 59.2754275427543
0.0165 61.1998199819982
0.018 63.0999099909991
0.0195 65.0693069306931
0.021 66.7434743474347
0.0225 68.4311431143114
0.024 69.9153915391539
0.0255 71.2511251125112
0.027 72.5454545454545
0.0285 73.7893789378938
0.03 75.015301530153
0.0315 76.2394239423942
0.033 77.2520252025203
0.0345 78.2880288028803
0.036 79.4221422142214
0.0375 80.4455445544555
0.039 81.4059405940594
0.0405 82.3141314131413
0.042 83.1458145814582
0.0435 83.939693969397
0.045 84.7020702070207
0.0465 85.5013501350135
0.048 86.2439243924393
0.0495 86.9621962196219
0.051 87.6327632763276
0.0525 88.2889288928893
0.054 88.8172817281728
0.0555 89.3906390639064
0.057 89.8190819081908
0.0585 90.3411341134114
0.06 90.8073807380738
0.0615 91.2655265526553
0.063 91.6642664266427
0.0645 92.028802880288
0.066 92.4068406840684
0.0675 92.7209720972097
0.069 93.0558055805581
0.0705 93.3861386138615
0.072 93.7038703870387
0.0735 93.993699369937
0.075 94.2916291629163
0.0765 94.5130513051305
0.078 94.8001800180018
0.0795 95.0207020702071
0.081 95.2358235823583
0.0825 95.4752475247525
0.084 95.6984698469847
0.0855 95.8874887488749
0.087 96.1278127812782
0.0885 96.3195319531954
0.09 96.5031503150315
0.0915 96.6885688568857
0.093 96.8406840684068
0.0945 96.964896489649
0.096 97.0693069306931
0.0975 97.1962196219622
0.099 97.3060306030603
0.1005 97.4302430243024
0.102 97.5400540054006
0.1035 97.6300630063006
0.105 97.7308730873088
0.1065 97.8289828982898
0.108 97.9288928892889
0.1095 98.006300630063
0.111 98.0882088208821
0.1125 98.1593159315931
0.114 98.2349234923492
0.1155 98.3114311431142
0.117 98.3924392439244
0.1185 98.4419441944194
0.12 98.4761476147615
0.1215 98.5310531053105
0.123 98.5922592259226
0.1245 98.6498649864987
0.126 98.7038703870387
0.1275 98.7353735373538
0.129 98.7713771377138
0.1305 98.8046804680468
0.132 98.8514851485149
0.1335 98.8892889288929
0.135 98.935193519352
0.1365 98.964896489649
0.138 98.985598559856
0.1395 99.00900090009
0.141 99.027902790279
0.1425 99.045904590459
0.144 99.0756075607561
0.1455 99.0972097209721
0.147 99.1350135013501
0.1485 99.1575157515752
0.15 99.1575157515752
};
\addplot [semithick, yellow2552430]
table {%
0 44.1674167416741
0.0015 44.6885688568857
0.003 45.9360936093609
0.0045 47.5571557155716
0.006 50.0972097209721
0.0075 52.8046804680468
0.009 55.3168316831684
0.0105 57.6732673267327
0.012 60
0.0135 62.3843384338434
0.015 64.4689468946895
0.0165 66.3348334833483
0.018 68.0648064806481
0.0195 69.8064806480648
0.021 71.3096309630963
0.0225 73.024302430243
0.024 74.4185418541854
0.0255 75.7686768676868
0.027 77.027902790279
0.0285 78.1845184518452
0.03 79.2673267326733
0.0315 80.2718271827183
0.033 81.2223222322232
0.0345 82.1962196219622
0.036 83.2628262826283
0.0375 84.1296129612961
0.039 85.026102610261
0.0405 85.8100810081008
0.042 86.5769576957696
0.0435 87.2916291629163
0.045 87.962196219622
0.0465 88.6255625562556
0.048 89.2826282628263
0.0495 89.9342934293429
0.051 90.4698469846984
0.0525 91.026102610261
0.054 91.4806480648065
0.0555 91.9234923492349
0.057 92.2997299729973
0.0585 92.7128712871287
0.06 93.0909090909091
0.0615 93.4563456345635
0.063 93.8226822682268
0.0645 94.1683168316832
0.066 94.4518451845185
0.0675 94.6858685868587
0.069 94.93699369937
0.0705 95.1611161116112
0.072 95.3897389738974
0.0735 95.5922592259226
0.075 95.7911791179118
0.0765 95.975697569757
0.078 96.1737173717371
0.0795 96.3312331233123
0.081 96.4671467146714
0.0825 96.6453645364536
0.084 96.7983798379838
0.0855 96.9261926192619
0.087 97.1143114311431
0.0885 97.2286228622862
0.09 97.3609360936094
0.0915 97.4734473447345
0.093 97.5886588658866
0.0945 97.6813681368137
0.096 97.7677767776778
0.0975 97.8631863186319
0.099 97.9603960396039
0.1005 98.053105310531
0.102 98.1368136813681
0.1035 98.2007200720072
0.105 98.2745274527453
0.1065 98.3420342034203
0.108 98.4221422142214
0.1095 98.4761476147615
0.111 98.5391539153915
0.1125 98.5967596759676
0.114 98.6381638163817
0.1155 98.7047704770477
0.117 98.7551755175518
0.1185 98.7965796579658
0.12 98.8190819081908
0.1215 98.8568856885689
0.123 98.8901890189019
0.1245 98.922592259226
0.126 98.945094509451
0.1275 98.972097209721
0.129 98.988298829883
0.1305 99.018901890189
0.132 99.0504050405041
0.1335 99.0693069306931
0.135 99.1017101710171
0.1365 99.1215121512152
0.138 99.1422142214221
0.1395 99.1620162016202
0.141 99.1872187218722
0.1425 99.2061206120612
0.144 99.2277227722772
0.1455 99.2493249324932
0.147 99.2772277227723
0.1485 99.3069306930693
0.15 99.3069306930693
};
\addplot [semithick, yellow25525573]
table {%
0 46.7965796579658
0.0015 47.3843384338434
0.003 48.6579657965797
0.0045 50.3627362736274
0.006 53.032403240324
0.0075 55.6120612061206
0.009 57.9216921692169
0.0105 60.1818181818182
0.012 62.3708370837084
0.0135 64.5886588658866
0.015 66.5859585958596
0.0165 68.2997299729973
0.018 69.8415841584158
0.0195 71.5202520252025
0.021 72.949594959496
0.0225 74.5832583258326
0.024 75.9027902790279
0.0255 77.1719171917192
0.027 78.3060306030603
0.0285 79.3879387938794
0.03 80.3645364536454
0.0315 81.3222322232223
0.033 82.1305130513051
0.0345 83.044104410441
0.036 84.045904590459
0.0375 84.9180918091809
0.039 85.7677767776778
0.0405 86.5670567056706
0.042 87.3258325832583
0.0435 88.066606660666
0.045 88.7146714671467
0.0465 89.3888388838884
0.048 90.001800180018
0.0495 90.6597659765976
0.051 91.1755175517551
0.0525 91.6705670567057
0.054 92.077407740774
0.0555 92.4941494149415
0.057 92.8604860486049
0.0585 93.2727272727273
0.06 93.5769576957696
0.0615 93.8892889288929
0.063 94.2079207920792
0.0645 94.4878487848785
0.066 94.7596759675968
0.0675 94.956795679568
0.069 95.1836183618362
0.0705 95.3987398739874
0.072 95.5958595859586
0.0735 95.7812781278128
0.075 95.9522952295229
0.0765 96.1476147614761
0.078 96.3357335733574
0.0795 96.5103510351036
0.081 96.6813681368137
0.0825 96.8550855085509
0.084 97.0297029702971
0.0855 97.1764176417641
0.087 97.3276327632763
0.0885 97.4320432043204
0.09 97.5508550855086
0.0915 97.6543654365437
0.093 97.7605760576058
0.0945 97.8523852385239
0.096 97.931593159316
0.0975 98.007200720072
0.099 98.0936093609361
0.1005 98.1638163816381
0.102 98.2205220522052
0.1035 98.2700270027002
0.105 98.3492349234923
0.1065 98.4032403240324
0.108 98.4869486948695
0.1095 98.5571557155716
0.111 98.6147614761476
0.1125 98.6543654365437
0.114 98.6984698469847
0.1155 98.7551755175518
0.117 98.8082808280828
0.1185 98.8379837983798
0.12 98.8595859585958
0.1215 98.8793879387939
0.123 98.9090909090909
0.1245 98.962196219622
0.126 98.993699369937
0.1275 99.0144014401441
0.129 99.0333033303331
0.1305 99.0639063906391
0.132 99.0945094509452
0.1335 99.1152115211521
0.135 99.1521152115212
0.1365 99.1782178217822
0.138 99.1980198019802
0.1395 99.2196219621963
0.141 99.2403240324032
0.1425 99.2619261926193
0.144 99.2844284428443
0.1455 99.3051305130513
0.147 99.3222322232223
0.1485 99.3483348334833
0.15 99.3483348334833
};
\addplot [semithick, palegoldenrod255255164]
table {%
0 48.4869486948695
0.0015 49.0909090909091
0.003 50.3861386138614
0.0045 52.1341134113411
0.006 54.7425742574258
0.0075 57.3573357335734
0.009 59.6057605760576
0.0105 61.9117911791179
0.012 64.1701170117012
0.0135 66.4095409540954
0.015 68.4239423942394
0.0165 70.1116111611161
0.018 71.5733573357336
0.0195 73.1368136813681
0.021 74.5409540954096
0.0225 76.1314131413142
0.024 77.4041404140414
0.0255 78.6606660666066
0.027 79.7191719171917
0.0285 80.7596759675967
0.03 81.7488748874888
0.0315 82.6552655265527
0.033 83.4707470747074
0.0345 84.2790279027903
0.036 85.2583258325832
0.0375 86.1035103510351
0.039 86.983798379838
0.0405 87.7695769576957
0.042 88.4500450045004
0.0435 89.1539153915391
0.045 89.7884788478848
0.0465 90.3717371737174
0.048 90.963096309631
0.0495 91.4932493249325
0.051 91.978397839784
0.0525 92.4653465346535
0.054 92.8775877587759
0.0555 93.2304230423043
0.057 93.5985598559856
0.0585 93.9549954995499
0.06 94.2502250225022
0.0615 94.5364536453645
0.063 94.8361836183618
0.0645 95.0954095409541
0.066 95.3096309630963
0.0675 95.5400540054005
0.069 95.7713771377137
0.0705 95.988298829883
0.072 96.1530153015302
0.0735 96.3015301530153
0.075 96.4518451845184
0.0765 96.5949594959496
0.078 96.7452745274527
0.0795 96.9018901890189
0.081 97.024302430243
0.0825 97.1584158415842
0.084 97.3204320432043
0.0855 97.4482448244824
0.087 97.5769576957695
0.0885 97.6768676867686
0.09 97.7857785778578
0.0915 97.8748874887488
0.093 97.9639963996399
0.0945 98.0486048604861
0.096 98.1017101710171
0.0975 98.1719171917192
0.099 98.2412241224123
0.1005 98.2979297929793
0.102 98.3564356435644
0.1035 98.4113411341134
0.105 98.4815481548154
0.1065 98.5220522052205
0.108 98.5913591359136
0.1095 98.6354635463545
0.111 98.6750675067506
0.1125 98.7065706570657
0.114 98.7515751575157
0.1155 98.8001800180018
0.117 98.8460846084608
0.1185 98.8676867686769
0.12 98.9009900990099
0.1215 98.9333933393339
0.123 98.9531953195319
0.1245 98.979297929793
0.126 98.995499549955
0.1275 99.015301530153
0.129 99.033303330333
0.1305 99.054905490549
0.132 99.0801080108011
0.1335 99.1008100810081
0.135 99.1350135013502
0.1365 99.1566156615662
0.138 99.1719171917192
0.1395 99.1881188118812
0.141 99.2133213321332
0.1425 99.2241224122413
0.144 99.2421242124212
0.1455 99.2592259225923
0.147 99.2862286228623
0.1485 99.3051305130514
0.15 99.3051305130514
};
\addplot [semithick, white]
table {%
0 50.8865886588659
0.0015 51.4635463546355
0.003 52.8028802880288
0.0045 54.5544554455446
0.006 57.0675067506751
0.0075 59.5391539153915
0.009 61.7371737173717
0.0105 63.8973897389739
0.012 66.037803780378
0.0135 68.1863186318632
0.015 70.038703870387
0.0165 71.6966696669667
0.018 73.1800180018002
0.0195 74.6381638163816
0.021 75.980198019802
0.0225 77.4743474347434
0.024 78.7272727272727
0.0255 79.9225922592259
0.027 80.8937893789379
0.0285 81.8208820882088
0.03 82.7632763276328
0.0315 83.6066606660666
0.033 84.3807380738074
0.0345 85.2016201620162
0.036 86.1323132313231
0.0375 86.9252925292529
0.039 87.7101710171017
0.0405 88.4545454545455
0.042 89.0873087308731
0.0435 89.8208820882088
0.045 90.4158415841584
0.0465 91.015301530153
0.048 91.5661566156616
0.0495 92.0891089108911
0.051 92.5364536453646
0.0525 93.0072007200721
0.054 93.4050405040504
0.0555 93.7560756075608
0.057 94.0729072907291
0.0585 94.3870387038704
0.06 94.6336633663366
0.0615 94.8865886588659
0.063 95.1467146714671
0.0645 95.4068406840684
0.066 95.5859585958595
0.0675 95.7407740774077
0.069 95.9414941494149
0.0705 96.1233123312331
0.072 96.2619261926192
0.0735 96.4041404140414
0.075 96.5589558955896
0.0765 96.6849684968497
0.078 96.8496849684969
0.0795 96.972097209721
0.081 97.078307830783
0.0825 97.2349234923492
0.084 97.3780378037804
0.0855 97.4896489648965
0.087 97.6102610261026
0.0885 97.6858685868587
0.09 97.7803780378038
0.0915 97.8667866786678
0.093 97.9486948694869
0.0945 98.0126012601259
0.096 98.0738073807381
0.0975 98.1422142214222
0.099 98.2115211521152
0.1005 98.2646264626463
0.102 98.3132313231323
0.1035 98.3609360936094
0.105 98.4239423942394
0.1065 98.4689468946895
0.108 98.5301530153015
0.1095 98.5805580558056
0.111 98.6120612061206
0.1125 98.6381638163817
0.114 98.6660666066607
0.1155 98.7038703870387
0.117 98.7506750675068
0.1185 98.7668766876688
0.12 98.7974797479748
0.1215 98.8343834383438
0.123 98.8496849684968
0.1245 98.8892889288928
0.126 98.9090909090909
0.1275 98.9261926192619
0.129 98.9387938793879
0.1305 98.9558955895589
0.132 98.977497749775
0.1335 98.99099909991
0.135 99.024302430243
0.1365 99.0432043204321
0.138 99.0567056705671
0.1395 99.0792079207921
0.141 99.1017101710171
0.1425 99.1170117011701
0.144 99.1332133213321
0.1455 99.1440144014402
0.147 99.1629162916292
0.1485 99.1836183618362
0.15 99.1836183618362
};
\end{axis}

\end{tikzpicture}

%% file: SupplimentFigures/EnergyTOSCA.tex
\begin{tikzpicture}

\definecolor{darkorange25512714}{RGB}{255,127,14}
\definecolor{darkslategray38}{RGB}{38,38,38}
\definecolor{lightgray204}{RGB}{204,204,204}
\definecolor{steelblue31119180}{RGB}{31,119,180}

\Large
\begin{axis}[
axis line style={lightgray204},
legend cell align={left},
legend style={fill opacity=0.8, draw opacity=1, text opacity=1, draw=none},
tick align=outside,
tick pos=left,
x grid style={lightgray204},
xlabel=\textcolor{darkslategray38}{Iteration},
xmajorgrids,
xmin=0, xmax=219,
xtick style={color=darkslategray38},
y grid style={lightgray204},
ylabel=\textcolor{darkslategray38}{Energy},
ymajorgrids,
xticklabel style={yshift= 5pt},
yticklabel style={xshift= 5pt},
ymin=550542683.434551037, ymax=1358694122.06139121,
ytick style={color=darkslategray38},
yticklabels={0.4,0.6,0.8,1.0,1.2,1.4,1.6}
]
\addplot [ultra thick, steelblue31119180, opacity=0.7]
table {%
0 1308694122.67322
1 1255134039.19642
2 1223143574.89468
3 1145212342.00702
4 1071660562.52114
5 1010549281.51583
6 909540279.799554
7 825072742.727202
8 722517705.220701
9 706707776.639338
10 703998240.263056
11 698214941.603546
12 693503206.207628
13 688529891.538741
14 685062361.313832
15 682052952.801207
16 678021369.660316
17 670123568.643231
18 669012199.949173
19 667011616.611918
20 663466895.396668
21 661757055.446414
22 660006286.159989
23 658125855.838093
24 651537664.648434
25 650230631.206639
26 648924271.43899
27 647794608.278477
28 645434591.941106
29 645362324.969734
30 644781869.689846
31 643597448.709508
32 641287076.672148
33 639770366.346507
34 639481415.206297
35 638968123.474803
36 638415874.334982
37 637457143.07313
38 636514349.138247
39 636063523.988704
40 634598240.270878
41 634432043.831609
42 633797571.9302
43 633562946.208826
44 633206379.507686
45 633081450.620621
46 632822632.084459
47 632740719.207734
48 632016723.514857
49 631955056.037509
50 631164507.449628
51 630592635.274753
52 630015274.851763
53 629220774.094387
54 629063618.911333
55 628450327.107812
56 628045840.125723
57 627711061.261458
58 627707323.134207
59 627400731.708286
60 627201869.660048
61 627021194.235807
62 625561087.166252
63 625305693.449827
64 624574218.597225
65 624262614.068763
66 623819569.673078
67 623390783.397586
68 623033637.777877
69 622800922.444446
70 622555846.918175
71 622486827.896623
72 622312483.468501
73 622284957.070023
74 622001706.063426
75 621836798.133723
76 621477827.785391
77 621120777.491069
78 621011652.318446
79 620915309.40067
80 620833917.009068
81 620730591.594267
82 620741111.799099
83 619926102.450784
84 619705930.775391
85 619782868.090446
86 619343207.157681
87 619078080.636396
88 618909041.878258
89 618798975.189423
90 618044745.826577
91 617988358.746567
92 617839875.140302
93 617455184.674866
94 617308766.294549
95 617273807.952901
96 617125941.460096
97 616802811.886803
98 616770392.156628
99 616535878.351132
100 616456569.517101
101 616345619.596019
102 616300969.571058
103 616118124.685929
104 615937019.478624
105 615839606.125607
106 615535318.192135
107 615370770.214879
108 615436458.25241
109 615077928.885213
};
\addlegendentry{Geodesic Distances}
\addplot [ultra thick, darkorange25512714, opacity=0.7]
table {%
111 612648372.301906
112 611662205.674109
113 610015170.898197
114 609765111.367364
115 608643620.962938
116 607782890.182881
117 605572518.678166
118 603830782.870329
119 603021851.230176
120 602411377.009876
121 601515259.825287
122 600680106.548735
123 600688145.09389
124 598979294.307838
125 597796863.870596
126 596565066.633388
127 596284372.450001
128 595566341.213237
129 594185609.280201
130 593463916.118725
131 592041568.039415
132 591695059.577726
133 591677157.769964
134 591755925.50642
135 590810755.829279
136 590596877.475266
137 590011242.175887
138 590072535.336405
139 589071941.418743
140 588686442.998376
141 588411728.738422
142 588171047.344407
143 589125737.717016
144 589443632.784902
145 589122400.739687
146 589333318.325826
147 589193788.536528
148 588803128.622151
149 588482574.038744
150 588251629.499683
151 588377677.653596
152 587789202.390581
153 586828313.777811
154 587071861.781051
155 586448515.020599
156 585765021.699341
157 585823464.309016
158 586018138.37902
159 585328574.751973
160 584767511.773227
161 584771092.066519
162 584694431.837749
163 584769694.690591
164 584338693.530078
165 584515805.993903
166 584848676.036812
167 584815992.26678
168 584371937.38755
169 584292312.542714
170 584200089.400307
171 583973664.353263
172 584033204.300175
173 583786553.297186
174 583394360.977737
175 583892638.028153
176 583404843.291737
177 583310188.494599
178 584088049.15557
179 583276014.086446
180 583592940.641809
181 583652702.931611
182 583601291.329494
183 583653176.0634
184 583330123.066812
185 583445946.241169
186 584049156.11105
187 583998672.576763
188 583730160.652931
189 582931472.104415
190 583078961.65941
191 583089180.407408
192 586754756.01036
193 586024426.939826
194 585908209.980571
195 586005037.50462
196 586023762.914037
197 585687205.144002
198 585871595.059091
199 585607281.273647
200 585264662.775361
201 587754137.024738
202 584961281.160602
203 583726839.491549
204 583361301.77597
205 584731186.68361
206 586391797.788864
207 585833866.085215
208 586016205.03599
209 586248894.349355
210 586808952.581457
211 585594315.315987
212 585226482.009306
213 584991869.32544
214 584999533.516353
215 584669305.210516
216 585563585.038832
217 586482905.161003
218 586197157.422245
219 590542683.701471
};
\addlegendentry{Gaussian Kernel}
\path [draw=red, draw opacity=0.5, very thick, dash pattern=on 11.1pt off 4.8pt]
(axis cs:0,561958324.8750211)
--(axis cs:219,561958324.8750211);

\end{axis}

\end{tikzpicture}

%% file: SupplimentFigures/PCK_Evolution_SMAL.tex
\begin{tikzpicture}

\definecolor{black1000}{RGB}{10,0,0}
\definecolor{darkred14400}{RGB}{144,0,0}
\definecolor{darkslategray38}{RGB}{38,38,38}
\definecolor{gold2552250}{RGB}{255,225,0}
\definecolor{lightgray204}{RGB}{204,204,204}
\definecolor{maroon7600}{RGB}{76,0,0}
\definecolor{orange2551570}{RGB}{255,157,0}
\definecolor{orangered255910}{RGB}{255,91,0}
\definecolor{palegoldenrod255255156}{RGB}{255,255,156}
\definecolor{red21000}{RGB}{210,0,0}
\definecolor{red255230}{RGB}{255,23,0}
\definecolor{yellow25525554}{RGB}{255,255,54}

\Large
\begin{axis}[
axis line style={lightgray204},
tick align=outside,
tick pos=left,
x grid style={lightgray204},
xtick = {0,0.05,0.1,0.15},
xticklabel style={yshift= 5pt},
xlabel=\textcolor{darkslategray38}{Geodesic Error},
xmajorgrids,
xmin=0, xmax=0.15,
xtick style={color=darkslategray38},
y grid style={lightgray204},
ylabel=\textcolor{darkslategray38}{\% Correspondences},
ymajorgrids,
ymin=5.5604249667995, ymax=100,
ytick style={color=darkslategray38}
]
\addplot [semithick, black1000]
table {%
0 11.1111111111111
0.0015 11.1319028609448
0.003 11.1984364604125
0.0045 11.3398203592814
0.006 11.7334774894655
0.0075 12.3281215347084
0.009 13.3025615435795
0.0105 14.6831337325349
0.012 16.4337990685296
0.0135 18.090208471945
0.015 19.8062208915502
0.0165 21.2505544466622
0.018 22.7877578176979
0.0195 24.1946662231093
0.021 25.7027611443779
0.0225 27.2940230649811
0.024 28.8312264360169
0.0255 30.4612996229763
0.027 31.8293967620315
0.0285 33.3264027500554
0.03 34.7762807717897
0.0315 35.9960634286982
0.033 37.3184187181193
0.0345 38.6366156575737
0.036 39.9076846307385
0.0375 41.1219228210246
0.039 42.4123974273675
0.0405 43.6294078509647
0.042 44.8297848746951
0.0435 45.9719449988911
0.045 47.1820248392105
0.0465 48.279829230428
0.048 49.3526835218452
0.0495 50.3021734309159
0.051 51.3140385894877
0.0525 52.2995675316035
0.054 53.1852960745176
0.0555 54.1597360833888
0.057 55.0094255932579
0.0585 55.8632734530938
0.06 56.6713794632956
0.0615 57.4365158571746
0.063 58.1642271013529
0.0645 58.8364936793081
0.066 59.4560878243513
0.0675 60.0576624528721
0.069 60.55666444888
0.0705 61.0722998447549
0.072 61.5061543579508
0.0735 61.9524839210468
0.075 62.3863384342426
0.0765 62.7827677977379
0.078 63.1736526946108
0.0795 63.534043025061
0.081 63.8708693723664
0.0825 64.1813595032158
0.084 64.4835329341317
0.0855 64.738578398758
0.087 65.0005544466622
0.0885 65.273619427811
0.09 65.5411399423375
0.0915 65.789254823686
0.093 66.0318252384121
0.0945 66.2730095364826
0.096 66.5252827677977
0.0975 66.7623087159015
0.099 66.9619095143048
0.1005 67.1975493457529
0.102 67.4123974273675
0.1035 67.6702151253049
0.105 67.8462519405633
0.1065 68.0708028387669
0.108 68.2579285872699
0.1095 68.4575293856731
0.111 68.6654468840097
0.1125 68.8664337990685
0.114 69.071579064094
0.1155 69.2351408294522
0.117 69.419494344644
0.1185 69.5858283433134
0.12 69.767409625194
0.1215 69.9517631403859
0.123 70.1236416056775
0.1245 70.2844311377246
0.126 70.4521512530495
0.1275 70.6073963184742
0.129 70.7460079840319
0.1305 70.8984808161455
0.132 71.067587048126
0.1335 71.2145154136172
0.135 71.3836216455977
0.1365 71.5513417609226
0.138 71.701042359725
0.1395 71.8576735418052
0.141 72.0087602572632
0.1425 72.164005322688
0.144 72.3289532047017
0.1455 72.4675648702595
0.147 72.6422155688623
0.1485 72.7752827677977
0.15 72.7752827677977
};
\addplot [semithick, maroon7600]
table {%
0 9.40064315812819
0.0015 9.42836549123975
0.003 9.49351297405189
0.0045 9.66261920603237
0.006 9.9911288534043
0.0075 10.7022066977157
0.009 11.7958527389665
0.0105 13.3621645597694
0.012 15.1405522288756
0.0135 17.0506209802617
0.015 18.9579174983367
0.0165 20.7972943002883
0.018 22.6311266356177
0.0195 24.4413949878022
0.021 26.3500776225327
0.0225 28.4043025060989
0.024 30.3836770902639
0.0255 32.2410734087381
0.027 34.0485695276114
0.0285 36.0445775116434
0.03 38.1085052117986
0.0315 39.8453093812375
0.033 41.5446884009758
0.0345 43.2620869372366
0.036 44.9032490574407
0.0375 46.5665890441339
0.039 48.2784431137724
0.0405 49.7976269682857
0.042 51.4235418052783
0.0435 52.9080727434021
0.045 54.4300288312265
0.0465 55.839709469949
0.048 57.2299844754935
0.0495 58.7269904635174
0.051 60.0756819693945
0.0525 61.5200155245065
0.054 62.8562319804835
0.0555 64.105123087159
0.057 65.3803504102905
0.0585 66.5446884009758
0.06 67.7284320248392
0.0615 68.7943557329785
0.063 69.9198824573076
0.0645 70.9220447992903
0.066 71.9463850077623
0.0675 72.8418163672655
0.069 73.6235861610113
0.0705 74.5259481037924
0.072 75.2772233311156
0.0735 76.0118651585718
0.075 76.7700709691727
0.0765 77.4492681304059
0.078 78.1326236416056
0.0795 78.7882568196939
0.081 79.4078509647372
0.0825 80.0205145265026
0.084 80.5680306054557
0.0855 81.089210467953
0.087 81.571579064094
0.0885 82.0816699933466
0.09 82.5418607229984
0.0915 82.9992792193391
0.093 83.4677866489244
0.0945 83.9071856287425
0.096 84.3451984919051
0.0975 84.7735085384786
0.099 85.1228099356842
0.1005 85.4610223996452
0.102 85.8020070969173
0.1035 86.1998225770681
0.105 86.5200155245065
0.1065 86.8374362386339
0.108 87.1590153027279
0.1095 87.4320802838767
0.111 87.7439565313817
0.1125 88.0475160789532
0.114 88.3427589265913
0.1155 88.6172100243956
0.117 88.8791860722998
0.1185 89.1453204701707
0.12 89.416999334664
0.1215 89.656797516079
0.123 89.8674872477267
0.1245 90.0823353293413
0.126 90.3318363273453
0.1275 90.5425260589931
0.129 90.7573741406077
0.1305 90.9528165890441
0.132 91.163506320692
0.1335 91.350632069195
0.135 91.5654801508095
0.1365 91.7262696828565
0.138 91.9133954313595
0.1395 92.08250166334
0.141 92.2668551785319
0.1425 92.4123974273675
0.144 92.5787314260368
0.1455 92.7312042581504
0.147 92.9155577733422
0.1485 93.0583277888667
0.15 93.0583277888667
};
\addplot [semithick, darkred14400]
table {%
0 10.9004213794633
0.0015 10.9323020625416
0.003 11.0445775116434
0.0045 11.2275449101796
0.006 11.5962519405633
0.0075 12.4722776668884
0.009 13.6865158571745
0.0105 15.2209469948991
0.012 17.0977489465513
0.0135 19.0854402306498
0.015 21.1327345309381
0.0165 23.0663672654691
0.018 25.1053448658239
0.0195 27.0833333333333
0.021 29.1735972499446
0.0225 31.2846529163894
0.024 33.3000665335995
0.0255 35.3127079174983
0.027 37.1839654025283
0.0285 39.2589820359281
0.03 41.3936016855179
0.0315 43.3715901530273
0.033 45.2705699711688
0.0345 47.0586604568641
0.036 48.8439787092482
0.0375 50.5267243291195
0.039 52.2981814149479
0.0405 53.9337990685296
0.042 55.6650587713462
0.0435 57.1551341760923
0.045 58.7089709469949
0.0465 60.1519183854513
0.048 61.6461521401641
0.0495 63.071080062098
0.051 64.416999334664
0.0525 65.7698491905079
0.054 67.0395320470171
0.0555 68.3050565535595
0.057 69.5470170769572
0.0585 70.6822466178754
0.06 71.8465846085606
0.0615 72.9208250166334
0.063 74.0338766910623
0.0645 74.9722776668884
0.066 75.9924595253936
0.0675 76.8878908848969
0.069 77.7070858283433
0.0705 78.5290530051009
0.072 79.3052783322244
0.0735 80.0745730760701
0.075 80.7717897538257
0.0765 81.4412840984697
0.078 82.0830561100022
0.0795 82.7165114216013
0.081 83.2889776003548
0.0825 83.836493679308
0.084 84.3826236416057
0.0855 84.8885562208915
0.087 85.3778554003105
0.0885 85.8227988467509
0.09 86.2621978265691
0.0915 86.7223885562209
0.093 87.1312929696163
0.0945 87.5443557329785
0.096 87.9338545131958
0.0975 88.3607784431138
0.099 88.7031492570415
0.1005 88.9928476380572
0.102 89.3296739853625
0.1035 89.6706586826347
0.105 89.9686737635839
0.1065 90.2472832113551
0.108 90.5508427589266
0.1095 90.798957640275
0.111 91.1039033045021
0.1125 91.3838988689288
0.114 91.6375582168995
0.1155 91.8759702816589
0.117 92.1310157462852
0.1185 92.3444777112442
0.12 92.5759591927257
0.1215 92.7714016411621
0.123 92.9571412730096
0.1245 93.1345642049235
0.126 93.3369372366378
0.1275 93.500499001996
0.129 93.6654468840098
0.1305 93.8470281658904
0.132 94.061876247505
0.1335 94.2060323796851
0.135 94.3792969616323
0.1365 94.5123641605678
0.138 94.6703814593036
0.1395 94.7868152583722
0.141 94.9489909070748
0.1425 95.0806719893546
0.144 95.1915613218008
0.1455 95.3066090042138
0.147 95.4576957196718
0.1485 95.5810601020182
0.15 95.5810601020182
};
\addplot [semithick, red21000]
table {%
0 10.9905189620758
0.0015 11.0126968285651
0.003 11.1152694610778
0.0045 11.2816034597472
0.006 11.7196163229097
0.0075 12.6164337990685
0.009 13.9290862719006
0.0105 15.6270791749834
0.012 17.5205145265025
0.0135 19.5068196939454
0.015 21.7329230428033
0.0165 23.6831891772011
0.018 25.7900864936793
0.0195 27.763916611222
0.021 29.983089376802
0.0225 32.211964958971
0.024 34.3840097582612
0.0255 36.4313040585496
0.027 38.4231536926148
0.0285 40.8322244400089
0.03 42.9668440895986
0.0315 44.9365158571745
0.033 46.8576735418053
0.0345 48.8439787092482
0.036 50.7554335772899
0.0375 52.4062985140829
0.039 54.2359724994456
0.0405 55.8993124861388
0.042 57.5155245065425
0.0435 59.0915391439344
0.045 60.6412175648703
0.0465 62.1229762696829
0.048 63.6643379906853
0.0495 65.1668884453315
0.051 66.521124417831
0.0525 68.0070414726104
0.054 69.2656353958749
0.0555 70.5574961188734
0.057 71.7675759591927
0.0585 72.9485473497449
0.06 74.1475382568197
0.0615 75.2453426480373
0.063 76.2863162563762
0.0645 77.2205588822355
0.066 78.1783654912397
0.0675 79.0433022843202
0.069 79.8971501441561
0.0705 80.6456531381681
0.072 81.4052450654247
0.0735 82.1412730095365
0.075 82.8149257041473
0.0765 83.4830339321357
0.078 84.1553005100909
0.0795 84.7485584386782
0.081 85.2586493679308
0.0825 85.7839875803948
0.084 86.2608117099135
0.0855 86.7015968063872
0.087 87.0938678199158
0.0885 87.4847527167886
0.09 87.8659347970726
0.0915 88.2568196939454
0.093 88.6282989576403
0.0945 89.0136393878909
0.096 89.354624085163
0.0975 89.6886781991572
0.099 89.9534264803726
0.1005 90.2140164116212
0.102 90.4953980927035
0.1035 90.7864825903748
0.105 91.0443002883123
0.1065 91.3021179862497
0.108 91.5363717010423
0.1095 91.7955755156354
0.111 92.0575515635396
0.1125 92.3001219782657
0.114 92.4983366600133
0.1155 92.689620758483
0.117 92.9044688400976
0.1185 93.135950321579
0.12 93.3244621867376
0.1215 93.526835218452
0.123 93.6903969838102
0.1245 93.874750499002
0.126 94.0341539143934
0.1275 94.1893989798182
0.129 94.37375249501
0.1305 94.4971168773564
0.132 94.6537480594367
0.1335 94.7812707917498
0.135 94.9462186737636
0.1365 95.0917609225992
0.138 95.2192836549124
0.1395 95.3468063872255
0.141 95.4951208693723
0.1425 95.6365047682413
0.144 95.7709580838323
0.1455 95.8776890663118
0.147 96.0065979152805
0.1485 96.1271900643158
0.15 96.1271900643158
};
\addplot [semithick, red255230]
table {%
0 11.8512974051896
0.0015 11.8693169217121
0.003 11.9580283876691
0.0045 12.1645597693502
0.006 12.5970281658904
0.0075 13.4896872920825
0.009 14.7676868485252
0.0105 16.7318141494788
0.012 18.9551452650255
0.0135 21.1327345309381
0.015 23.2673541805278
0.0165 25.3451430472389
0.018 27.6350077622533
0.0195 29.6864604125083
0.021 31.9250388112664
0.0225 34.1774783765802
0.024 36.3911066755378
0.0255 38.5922599245953
0.027 40.6409403415391
0.0285 42.9737746728765
0.03 45.1693834553116
0.0315 47.2305389221557
0.033 49.2459525393657
0.0345 51.1269128409847
0.036 52.9468840097583
0.0375 54.7252716788645
0.039 56.5327677977379
0.0405 58.2238301175427
0.042 59.8927145708583
0.0435 61.5255599911289
0.045 63.0267243291196
0.0465 64.568086050122
0.048 65.9833111554669
0.0495 67.3624972277667
0.051 68.6585163007318
0.0525 70.023841206476
0.054 71.2131292969617
0.0555 72.4107340873808
0.057 73.5223996451542
0.0585 74.6950543357729
0.06 75.7651363938789
0.0615 76.7284874695054
0.063 77.7417387447328
0.0645 78.6011310711909
0.066 79.4646817476159
0.0675 80.3199157241073
0.069 81.1751497005988
0.0705 81.8987025948103
0.072 82.5903748059437
0.0735 83.3097693501885
0.075 83.9695608782435
0.0765 84.5794522066977
0.078 85.2156797516079
0.0795 85.7853736970504
0.081 86.3037813262364
0.0825 86.813872255489
0.084 87.2338656021291
0.0855 87.6552450654247
0.087 88.0544466622311
0.0885 88.409292526059
0.09 88.7752273231316
0.0915 89.1453204701708
0.093 89.4724440008871
0.0945 89.8577844311377
0.096 90.165502328676
0.0975 90.4538145930362
0.099 90.736582390774
0.1005 90.9791528055001
0.102 91.2480594366822
0.1035 91.5225105344866
0.105 91.7692392991794
0.1065 92.0062652472832
0.108 92.1947771124418
0.1095 92.4318030605456
0.111 92.6605123087159
0.1125 92.8878354402306
0.114 93.0680306054558
0.1155 93.2482257706808
0.117 93.4353515191838
0.1185 93.6474273674872
0.12 93.8359392326458
0.1215 94.0161343978709
0.123 94.1741516966068
0.1245 94.3460301618984
0.126 94.5026613439787
0.1275 94.6357285429141
0.129 94.8173098247948
0.1305 94.9462186737636
0.132 95.095919272566
0.1335 95.2165114216012
0.135 95.3634397870924
0.1365 95.4965069860279
0.138 95.6004657351962
0.1395 95.7168995342648
0.141 95.8444222665779
0.1425 95.9719449988911
0.144 96.0939232645819
0.1455 96.1964958970947
0.147 96.2976824129519
0.1485 96.4196606786427
0.15 96.4196606786427
};
\addplot [semithick, orangered255910]
table {%
0 12.0051563539587
0.0015 12.0328786870703
0.003 12.1202040363717
0.0045 12.3544577511643
0.006 12.785540031049
0.0075 13.7378021734309
0.009 15.0532268795742
0.0105 16.9023064981149
0.012 19.1367265469062
0.0135 21.4321357285429
0.015 23.7455644267021
0.0165 25.9314703925482
0.018 28.138168108228
0.0195 30.2852628077179
0.021 32.6014637391883
0.0225 35.0202373031714
0.024 37.3378243512974
0.0255 39.6914504324684
0.027 41.8939897981814
0.0285 44.3002883122644
0.03 46.5624306941672
0.0315 48.6776447105788
0.033 50.7692947438456
0.0345 52.7722333111555
0.036 54.8098247948547
0.0375 56.6852406298514
0.039 58.5662009314704
0.0405 60.3057773342204
0.042 62.0342648037259
0.0435 63.7142381902861
0.045 65.3041139942337
0.0465 66.8011199822577
0.048 68.2690175205145
0.0495 69.6884009758261
0.051 70.9747172322022
0.0525 72.2818252384121
0.054 73.5279441117764
0.0555 74.7352517187846
0.057 75.9079064094034
0.0585 77.1166001330672
0.06 78.1353958749169
0.0615 79.1278554003105
0.063 80.1480372588157
0.0645 81.1030716345088
0.066 81.9444444444444
0.0675 82.7969061876247
0.069 83.6285761809714
0.0705 84.4186626746507
0.072 85.083998669328
0.0735 85.7465624306942
0.075 86.3980372588157
0.0765 87.0453537369705
0.078 87.562375249501
0.0795 88.0918718119317
0.081 88.6185961410513
0.0825 89.0981925038811
0.084 89.5084830339322
0.0855 89.8924373475272
0.087 90.3027278775782
0.0885 90.6700487913062
0.09 91.022122421823
0.0915 91.3741960523398
0.093 91.6763694832557
0.0945 91.9840873807939
0.096 92.2779441117764
0.0975 92.5773453093812
0.099 92.831004657352
0.1005 93.0486249722777
0.102 93.2939676203149
0.1035 93.5393102683522
0.105 93.7416833000666
0.1065 93.9870259481038
0.108 94.1810822798847
0.1095 94.4042470614327
0.111 94.6149367930805
0.1125 94.7951319583056
0.114 94.9753271235307
0.1155 95.1485917054779
0.117 95.3010645375915
0.1185 95.4826458194722
0.12 95.6420492348636
0.1215 95.7903637170104
0.123 95.9123419827013
0.1245 96.0398647150144
0.126 96.1812486138833
0.1275 96.3558993124862
0.129 96.4598580616544
0.1305 96.5707473941007
0.132 96.7162896429363
0.1335 96.8188622754491
0.135 96.9117320913728
0.1365 97.0087602572633
0.138 97.0946994899091
0.1395 97.1986582390774
0.141 97.2748946551342
0.1425 97.3552894211577
0.144 97.4453870037702
0.1455 97.5424151696607
0.147 97.646373918829
0.1485 97.7461743180306
0.15 97.7461743180306
};
\addplot [semithick, orange2551570]
table {%
0 16.2743956531382
0.0015 16.297959636283
0.003 16.38667110224
0.0045 16.6250831669993
0.006 17.0838877799956
0.0075 18.1179308050566
0.009 19.5594921268574
0.0105 21.540252827678
0.012 24.001996007984
0.0135 26.7104679529829
0.015 29.5076513639388
0.0165 32.2729540918164
0.018 35.0049900199601
0.0195 37.6898979818141
0.021 40.5785650920381
0.0225 43.4603016189842
0.024 46.3642160124196
0.0255 49.2764471057884
0.027 51.926702151253
0.0285 54.7987358616101
0.03 57.516910623198
0.0315 60.057662452872
0.033 62.6663339986693
0.0345 65.0296628964294
0.036 67.2710135284986
0.0375 69.3695941450432
0.039 71.4654025282768
0.0405 73.3643823464183
0.042 75.2619760479042
0.0435 76.8684852517188
0.045 78.4209359059658
0.0465 79.9110113107119
0.048 81.3026724329119
0.0495 82.6513639387891
0.051 83.9085717453981
0.0525 85.1477600354846
0.054 86.2829895764028
0.0555 87.3641605677534
0.057 88.3982035928144
0.0585 89.318585052118
0.06 90.1876801951652
0.0615 91.0443002883123
0.063 91.7609225992459
0.0645 92.4858616101131
0.066 93.1207030383677
0.0675 93.661288534043
0.069 94.2157351962741
0.0705 94.7396872920825
0.072 95.1763140385895
0.0735 95.5866045686405
0.075 95.9650144156132
0.0765 96.3059991128853
0.078 96.566589044134
0.0795 96.8604457751164
0.081 97.1141051230872
0.0825 97.3358837879796
0.084 97.5202373031714
0.0855 97.6976602350854
0.087 97.8556775338213
0.0885 98.0067642492792
0.09 98.1523064981149
0.0915 98.2784431137725
0.093 98.3810157462852
0.0945 98.4711133288978
0.096 98.5598247948547
0.0975 98.6429917941894
0.099 98.6970503437569
0.1005 98.778831226436
0.102 98.8384342426259
0.1035 98.913284542027
0.105 98.9534819250388
0.1065 99.0130849412287
0.108 99.0615990241739
0.1095 99.117043690397
0.111 99.1558549567532
0.1125 99.1877356398315
0.114 99.2251607895321
0.1155 99.2570414726104
0.117 99.2847638057219
0.1185 99.3221889554225
0.12 99.3526835218452
0.1215 99.3679308050565
0.123 99.3845642049234
0.1245 99.4053559547571
0.126 99.4344644045242
0.1275 99.4497116877356
0.129 99.4635728542914
0.1305 99.4871368374362
0.132 99.5051563539587
0.1335 99.5176314038589
0.135 99.5356509203814
0.1365 99.5411953870037
0.138 99.5453537369705
0.1395 99.5550565535595
0.141 99.5578287868707
0.1425 99.5675316034597
0.144 99.5703038367709
0.1455 99.5786205367043
0.147 99.5800066533599
0.1485 99.5855511199822
0.15 99.5855511199822
};
\addplot [semithick, gold2552250]
table {%
0 18.447826569084
0.0015 18.4727766688844
0.003 18.5739631847416
0.0045 18.8359392326458
0.006 19.3169217121313
0.0075 20.4050232867598
0.009 22.0447992903083
0.0105 24.1821911732091
0.012 26.9807607008206
0.0135 30.0329895764027
0.015 33.1392770015525
0.0165 36.1554668440896
0.018 39.0316589044134
0.0195 41.9383455311599
0.021 45.016910623198
0.0225 47.9873586161011
0.024 50.8885007762253
0.0255 53.7771678864493
0.027 56.4218784652916
0.0285 59.3299512086937
0.03 62.1895098691506
0.0315 64.7773896651142
0.033 67.315369261477
0.0345 69.8089931248614
0.036 72.0170769571967
0.0375 74.1281326236416
0.039 76.1102794411178
0.0405 77.8110445775116
0.042 79.6490352628077
0.0435 81.1903969838102
0.045 82.6430472388556
0.0465 83.9958970946995
0.048 85.2808272344201
0.0495 86.581004657352
0.051 87.7009869150588
0.0525 88.7170104235972
0.054 89.7261033488578
0.0555 90.6423264581947
0.057 91.4545908183633
0.0585 92.3070525615436
0.06 93.0250609891328
0.0615 93.6585163007319
0.063 94.2351408294522
0.0645 94.8186959414504
0.066 95.3371035706365
0.0675 95.7376912840985
0.069 96.1798624972277
0.0705 96.5804502106897
0.072 96.9158904413395
0.0735 97.2222222222222
0.075 97.5521179862497
0.0765 97.7794411177644
0.078 97.9776557995121
0.0795 98.1564648480816
0.081 98.3214127300954
0.0825 98.4766577955201
0.084 98.5861610113107
0.0855 98.7039809270348
0.087 98.7857618097139
0.0885 98.8481370592149
0.09 98.9326901752051
0.0915 99.0172432911954
0.093 99.064371257485
0.0945 99.1239742736748
0.096 99.1572410734087
0.0975 99.189121756487
0.099 99.2223885562209
0.1005 99.2598137059215
0.102 99.3013972055888
0.1035 99.3332778886671
0.105 99.3554557551564
0.1065 99.3748613883345
0.108 99.3970392548237
0.1095 99.4095143047239
0.111 99.419217121313
0.1125 99.4372366378355
0.114 99.4497116877357
0.1155 99.4608006209803
0.117 99.4760479041916
0.1185 99.4829784874695
0.12 99.4982257706809
0.1215 99.5093147039255
0.123 99.5120869372366
0.1245 99.5176314038589
0.126 99.5342648037259
0.1275 99.5384231536926
0.129 99.5453537369705
0.1305 99.5522843202484
0.132 99.5564426702151
0.1335 99.5578287868707
0.135 99.5606010201818
0.1365 99.5606010201818
0.138 99.5647593701486
0.1395 99.5772344200488
0.141 99.5786205367044
0.1425 99.5827788866711
0.144 99.5869372366378
0.1455 99.589709469949
0.147 99.5924817032601
0.1485 99.5952539365713
0.15 99.5952539365713
};
\addplot [semithick, yellow25525554]
table {%
0 19.3959303614992
0.0015 19.4167221113329
0.003 19.5123641605677
0.0045 19.7784985584387
0.006 20.3010645375915
0.0075 21.4099578620537
0.009 23.1412175648703
0.0105 25.383954313595
0.012 28.4167775559991
0.0135 31.6672211133289
0.015 34.887170104236
0.0165 38.0322687957419
0.018 41.0831115546684
0.0195 44.1422710135285
0.021 47.3275670880461
0.0225 50.4962297626968
0.024 53.6454868041694
0.0255 56.5993013972056
0.027 59.4491572410734
0.0285 62.3253493013972
0.03 65.2098580616545
0.0315 67.8032823242404
0.033 70.3523508538479
0.0345 72.8043912175649
0.036 75.0180195165225
0.0375 77.0500665335995
0.039 79.0044910179641
0.0405 80.6207030383677
0.042 82.3533488578399
0.0435 83.7436238633844
0.045 85.1269682856509
0.0465 86.367542692393
0.048 87.615047682413
0.0495 88.757207806609
0.051 89.7801618984254
0.0525 90.6922266577955
0.054 91.5557773342204
0.0555 92.3541805278332
0.057 93.0791195387004
0.0585 93.7957418496341
0.06 94.4139498780217
0.0615 94.9794854734974
0.063 95.506209802617
0.0645 96.0093701485917
0.066 96.4141162120204
0.0675 96.7703481925039
0.069 97.1390552228876
0.0705 97.4717232202262
0.072 97.7503326679973
0.0735 97.9637946329563
0.075 98.1925038811266
0.0765 98.3837879795963
0.078 98.5168551785319
0.0795 98.6526946107785
0.081 98.7566533599468
0.0825 98.8619982257706
0.084 98.9423929917942
0.0855 99.0130849412286
0.087 99.0615990241739
0.0885 99.1017964071856
0.09 99.1572410734087
0.0915 99.2126857396319
0.093 99.268130405855
0.0945 99.3041694389
0.096 99.3429807052562
0.0975 99.3776336216456
0.099 99.402583721446
0.1005 99.419217121313
0.102 99.4538700377024
0.1035 99.4788201375028
0.105 99.491295187403
0.1065 99.4982257706808
0.108 99.5065424706143
0.1095 99.5162452872033
0.111 99.5190175205145
0.1125 99.527334220448
0.114 99.5342648037259
0.1155 99.5425815036593
0.117 99.5495120869372
0.1185 99.5550565535595
0.12 99.5606010201819
0.1215 99.5689177201153
0.123 99.5716899534265
0.1245 99.5772344200488
0.126 99.5841650033266
0.1275 99.5883233532934
0.129 99.589709469949
0.1305 99.599412286538
0.132 99.599412286538
0.1335 99.6007984031936
0.135 99.6007984031936
0.1365 99.6021845198492
0.138 99.6035706365048
0.1395 99.6035706365048
0.141 99.6049567531603
0.1425 99.6049567531603
0.144 99.6049567531603
0.1455 99.6063428698159
0.147 99.6091151031271
0.1485 99.6091151031271
0.15 99.6091151031271
};
\addplot [semithick, palegoldenrod255255156]
table {%
0 20.3357174539809
0.0015 20.355123087159
0.003 20.4452206697716
0.0045 20.7044244843646
0.006 21.2172876469284
0.0075 22.3372699046352
0.009 24.0962519405633
0.0105 26.408294522067
0.012 29.4688400975826
0.0135 32.8107673541805
0.015 36.0501219782657
0.0165 39.09542027057
0.018 42.171213129297
0.0195 45.1763140385895
0.021 48.4142825460191
0.0225 51.5316589044134
0.024 54.6878465291639
0.0255 57.7068086050122
0.027 60.5012197826569
0.0285 63.3164227101353
0.03 66.2078620536704
0.0315 68.7694056331781
0.033 71.3143158128188
0.0345 73.75249500998
0.036 75.9633510756265
0.0375 77.9995564426702
0.039 79.9484364604125
0.0405 81.4870259481038
0.042 83.19333555112
0.0435 84.5545021068973
0.045 85.8824018629408
0.0465 87.0508982035928
0.048 88.1972166777556
0.0495 89.3158128188068
0.051 90.3193612774451
0.0525 91.2064759370149
0.054 92.0256708804613
0.0555 92.7838766910623
0.057 93.5240629851408
0.0585 94.2074184963407
0.06 94.8186959414504
0.0615 95.3592814371258
0.063 95.8250166333999
0.0645 96.300454646263
0.066 96.6968840097582
0.0675 97.0018296739854
0.069 97.334497671324
0.0705 97.6297405189621
0.072 97.8903304502107
0.0735 98.1065646484808
0.075 98.3172543801286
0.0765 98.4822022621424
0.078 98.6000221778665
0.0795 98.7289310268352
0.081 98.8287314260368
0.0825 98.913284542027
0.084 98.9978376580173
0.0855 99.0463517409625
0.087 99.0962519405633
0.0885 99.1309048569528
0.09 99.1932801064538
0.0915 99.2320913728099
0.093 99.2681304058549
0.0945 99.3152583721446
0.096 99.3415945886006
0.0975 99.366544688401
0.099 99.392880904857
0.1005 99.4067420714127
0.102 99.4413949878021
0.1035 99.4663450876025
0.105 99.4802062541583
0.1065 99.4940674207141
0.108 99.4996118873364
0.1095 99.5093147039254
0.111 99.5120869372366
0.1125 99.5162452872033
0.114 99.527334220448
0.1155 99.5356509203814
0.117 99.5508982035928
0.1185 99.5564426702151
0.12 99.5592149035262
0.1215 99.5689177201153
0.123 99.573076070082
0.1245 99.5786205367043
0.126 99.5869372366378
0.1275 99.5910955866046
0.129 99.5938678199157
0.1305 99.6021845198492
0.132 99.6021845198492
0.1335 99.6105012197826
0.135 99.6118873364382
0.1365 99.6118873364382
0.138 99.6118873364382
0.1395 99.6132734530938
0.141 99.6132734530938
0.1425 99.6132734530938
0.144 99.6174318030605
0.1455 99.6174318030605
0.147 99.6188179197161
0.1485 99.6188179197161
0.15 99.6188179197161
};
\addplot [semithick, white]
table {%
0 21.2699600798403
0.0015 21.2893657130184
0.003 21.38916611222
0.0045 21.6372809935684
0.006 22.1432135728543
0.0075 23.3075515635396
0.009 25.1136615657574
0.0105 27.4728321135507
0.012 30.6082279884675
0.0135 33.9640164116212
0.015 37.2491128853404
0.0165 40.3456974939011
0.018 43.4214903526281
0.0195 46.4390663118208
0.021 49.7699046351741
0.0225 52.9219339099578
0.024 56.0808937680195
0.0255 59.0097582612553
0.027 61.8263473053892
0.0285 64.6997671324019
0.03 67.5399201596806
0.0315 69.9656243069417
0.033 72.4675648702595
0.0345 74.9057440674207
0.036 77.051452650255
0.0375 78.9989465513418
0.039 80.9256487025948
0.0405 82.4725548902196
0.042 84.1317365269461
0.0435 85.4624085163007
0.045 86.7501108893324
0.0465 87.8964293634952
0.048 88.9457196717676
0.0495 90.0116433799068
0.051 90.9930139720559
0.0525 91.8510201818585
0.054 92.6203149257042
0.0555 93.3785207363052
0.057 94.0272233311155
0.0585 94.6620647593702
0.06 95.2442337547128
0.0615 95.7515524506542
0.063 96.1854069638501
0.0645 96.5721335107563
0.066 96.9560878243513
0.0675 97.2263805721889
0.069 97.5340984697272
0.0705 97.7988467509425
0.072 98.0233976491462
0.0735 98.207751164338
0.075 98.3782435129741
0.0765 98.5431913949878
0.078 98.6416056775338
0.0795 98.7552672432912
0.081 98.8536815258372
0.0825 98.9340762918607
0.084 99.0033821246396
0.0855 99.0588267908627
0.087 99.1045686404968
0.0885 99.1309048569528
0.09 99.1780328232424
0.0915 99.225160789532
0.093 99.2598137059215
0.0945 99.2972388556221
0.096 99.3194167221113
0.0975 99.349911288534
0.099 99.3748613883344
0.1005 99.3928809048569
0.102 99.4164448880017
0.1035 99.4469394544245
0.105 99.464958970947
0.1065 99.4774340208472
0.108 99.4843646041251
0.1095 99.4926813040585
0.111 99.500998003992
0.1125 99.5134730538922
0.114 99.5231758704812
0.1155 99.5328786870703
0.117 99.5425815036593
0.1185 99.5453537369705
0.12 99.5522843202483
0.1215 99.5578287868707
0.123 99.5675316034597
0.1245 99.5703038367709
0.126 99.5800066533599
0.1275 99.5827788866711
0.129 99.589709469949
0.1305 99.5952539365713
0.132 99.5980261698824
0.1335 99.6021845198492
0.135 99.6021845198492
0.1365 99.6021845198492
0.138 99.6021845198492
0.1395 99.6035706365047
0.141 99.6049567531603
0.1425 99.6049567531603
0.144 99.6049567531603
0.1455 99.6077289864715
0.147 99.6077289864715
0.1485 99.6091151031271
0.15 99.6091151031271
};
\end{axis}

\end{tikzpicture}

%% file: SupplimentFigures/EnergySMAL.tex
\begin{tikzpicture}

\definecolor{darkorange25512714}{RGB}{255,127,14}
\definecolor{darkslategray38}{RGB}{38,38,38}
\definecolor{lightgray204}{RGB}{204,204,204}
\definecolor{steelblue31119180}{RGB}{31,119,180}

\Large
\begin{axis}[
axis line style={lightgray204},
legend cell align={left},
legend style={fill opacity=0.8, draw opacity=1, text opacity=1, draw=none},
tick align=outside,
tick pos=left,
x grid style={lightgray204},
xlabel=\textcolor{darkslategray38}{Iteration},
xmajorgrids,
xmin=0, xmax=174,
xtick style={color=darkslategray38},
xtick={0,50,100,150},
y grid style={lightgray204},
ylabel=\textcolor{darkslategray38}{Energy},
ymajorgrids,
xticklabel style={yshift= 5pt},
yticklabel style={xshift= 5pt},
ymin=11899175.434551037, ymax=18512179.06139121,
ytick style={color=darkslategray38},
yticklabels={0.4,0.6,0.8,1.0,1.2,1.4,1.6}
]
\addplot [ultra thick, steelblue31119180, opacity=0.7]
table {%
0 18122179.8669987
1 17798598.3730782
2 17336005.1639099
3 16565909.8601921
4 15887572.0343383
5 15060570.0062235
6 14418579.5084637
7 14352494.8399955
8 14304678.1541237
9 14125914.9140919
10 13962793.9564897
11 13702388.304692
12 13570296.7785852
13 13487029.0765126
14 13473959.336156
15 13450187.8127192
16 13386065.2981522
17 13374884.0466379
18 13360185.2965563
19 13138836.069291
20 13113510.5345848
21 13103944.2723291
22 13098748.4271311
23 13098189.4844467
24 13059058.5748893
25 13044477.6776274
26 13013300.9280487
27 12997610.0498212
28 12990455.8789019
29 12983247.5432979
30 12976661.7930118
31 12969046.3110918
32 12953556.3152338
33 12946624.421873
34 12936175.2850095
35 12919247.2335172
36 12912272.2179775
37 12904967.2091222
38 12903576.5460348
39 12901286.8466065
40 12878229.5909736
41 12876444.4885278
42 12851921.5565257
43 12848209.7258882
44 12847648.6106095
45 12836721.3261999
46 12769279.6355737
47 12775831.4618156
48 12773918.4500608
49 12765782.4750709
50 12757603.4901589
51 12746119.2546517
52 12730557.7945106
53 12726601.1047852
54 12715806.5118059
55 12687134.5579642
56 12682547.0364317
57 12676958.5786945
58 12672506.4689028
59 12664773.8833204
60 12665280.4519888
61 12672748.575727
62 12664385.8700152
63 12659464.1963306
64 12664280.8741793
65 12660394.6615701
66 12656404.1785754
67 12661052.2751021
68 12657224.4543779
69 12652347.1963262
70 12652090.7819975
71 12648035.6510842
72 12644967.5809122
73 12644268.8623104
74 12643161.6593385
75 12635126.8284286
76 12634425.0718222
77 12627546.7113692
78 12621055.805236
79 12619516.5889837
80 12615320.2847639
81 12608362.5930857
82 12603123.1146949
83 12591840.9369272
84 12588683.0306296
85 12586683.6923936
86 12590342.6964495
};
\addlegendentry{Geodesic Distances}
\addplot [ultra thick, darkorange25512714, opacity=0.7]
table {%
87 12583387.277718
88 12542999.4465372
89 12518474.5297786
90 12524955.0759853
91 12519876.5330193
92 12472696.2763353
93 12456960.2232831
94 12448596.19386
95 12443230.7195219
96 12430795.5521528
97 12397623.5076284
98 12368280.3529401
99 12365248.0556883
100 12343671.023693
101 12303197.4275818
102 12289659.6883343
103 12288461.7572656
104 12276455.6970974
105 12268666.3932064
106 12262701.0959159
107 12259737.4622653
108 12252427.017184
109 12245204.588397
110 12237265.4480939
111 12232832.818362
112 12232059.8829914
113 12231748.8493207
114 12226514.9965066
115 12223080.8072794
116 12218360.1126516
117 12213862.4189998
118 12210566.1962907
119 12206843.1582104
120 12200425.6031285
121 12200254.8458254
122 12200100.6554202
123 12199639.7221578
124 12199562.069871
125 12197324.7456874
126 12204193.3926257
127 12200508.4215749
128 12199846.3280101
129 12197228.3270102
130 12188020.4954249
131 12186942.3049936
132 12178355.7806965
133 12176187.89913
134 12185860.036086
135 12185755.2171826
136 12184769.1192064
137 12186384.7033895
138 12184634.3986512
139 12184876.188495
140 12183592.6457171
141 12181556.2298388
142 12185036.2546158
143 12185017.2513453
144 12176382.2081102
145 12177028.749105
146 12175184.8693114
147 12173797.7727082
148 12173678.6278561
149 12172058.3039014
150 12166621.8888097
151 12166822.8893195
152 12166654.3670695
153 12166344.8847359
154 12166489.7970455
155 12167219.3556666
156 12169358.5142826
157 12169368.6820421
158 12169864.7808969
159 12169244.87127
160 12167426.742234
161 12166656.7228009
162 12166626.7213735
163 12166270.2589766
164 12164082.2880854
165 12157667.3197123
166 12162859.6349452
167 12169564.4856691
168 12170444.0432437
169 12169781.648779
170 12181184.3464855
171 12183398.99473
172 12181501.4735145
173 12187109.7056383
174 12172671.1344483
};
\addlegendentry{Gaussian Kernel}
\path [draw=red, draw opacity=0.5, very thick, dash pattern=on 11.1pt off 4.8pt]
(axis cs:0,11999175.853289634)
--(axis cs:174,11999175.853289634);

\end{axis}

\end{tikzpicture}

%% file: SupplimentFigures/PCK_vm.tex
\begin{tikzpicture}

\definecolor{darkorange25512714}{RGB}{255,127,14}
\definecolor{darkslategray38}{RGB}{38,38,38}
\definecolor{lightgray204}{RGB}{204,204,204}
\definecolor{steelblue31119180}{RGB}{31,119,180}

\Large
\begin{axis}[
axis line style={lightgray204},
legend cell align={left},
legend style={fill opacity=0.8, draw opacity=1, text opacity=1, draw=lightgray204, anchor=south east,at={(0.97,0.03)}},
tick align=outside,
tick pos=left,
xtick = {0,0.05,0.1,0.15},
x grid style={lightgray204},
xlabel=\textcolor{darkslategray38}{Geodesic Error},
xmajorgrids,
xmin=0, xmax=0.15,
xtick style={color=darkslategray38},
xticklabel style={yshift= 5pt},
y grid style={lightgray204},
ylabel=\textcolor{darkslategray38}{\% Correspondences},
ymajorgrids,
ymin=90, ymax=100,
ytick style={color=darkslategray38}
]
\addplot [ultra thick, steelblue31119180]
table {%
0 93.4019477644976
0.0015 93.4021691013723
0.003 93.4074811863656
0.0045 93.4229747675963
0.006 93.4645861000443
0.0075 93.6702080566622
0.009 93.9803010181496
0.0105 94.4156706507304
0.012 94.7726870296591
0.0135 95.1188579017264
0.015 95.4712262062859
0.0165 95.8612217795485
0.018 96.199203187251
0.0195 96.4645861000443
0.021 96.8025675077468
0.0225 97.1790615316512
0.024 97.4922532093847
0.0255 97.7833111996458
0.027 98.0336432049579
0.0285 98.2386011509517
0.03 98.4420097388225
0.0315 98.6250553342187
0.033 98.7990261177512
0.0345 98.9389110225763
0.036 99.0894200973882
0.0375 99.1963258078796
0.039 99.2682602921646
0.0405 99.3530323151836
0.042 99.4156706507303
0.0435 99.4873837981407
0.045 99.5559982293049
0.0465 99.6177512173528
0.048 99.6757414785303
0.0495 99.7153607791058
0.051 99.7563081009296
0.0525 99.7886232846392
0.054 99.8065515714918
0.0555 99.8247011952191
0.057 99.839973439575
0.0585 99.847941567065
0.06 99.8601150951748
0.0615 99.8634351482957
0.063 99.8691899070385
0.0645 99.8747233289066
0.066 99.881584772023
0.0675 99.8851261620185
0.069 99.8860115095174
0.0705 99.8906595838866
0.072 99.893536963258
0.0735 99.8968570163789
0.075 99.9026117751217
0.0765 99.9052678176184
0.078 99.9061531651173
0.0795 99.9065958388667
0.081 99.9096945551128
0.0825 99.9163346613546
0.084 99.9196547144754
0.0855 99.9225320938468
0.087 99.9260734838424
0.0885 99.9267374944666
0.09 99.9280655157149
0.0915 99.9285081894643
0.093 99.9285081894643
0.0945 99.9293935369632
0.096 99.9318282425852
0.0975 99.9322709163346
0.099 99.9324922532093
0.1005 99.9329349269588
0.102 99.935148295706
0.1035 99.9393536963258
0.105 99.9393536963258
0.1065 99.9424524125719
0.108 99.9424524125719
0.1095 99.9431164231961
0.111 99.9446657813191
0.1125 99.945551128818
0.114 99.9459938025675
0.1155 99.9464364763169
0.117 99.9475431606906
0.1185 99.9510845506861
0.12 99.951969898185
0.1215 99.9528552456839
0.123 99.9532979194334
0.1245 99.954404603807
0.126 99.9552899513059
0.1275 99.95595396193
0.129 99.9561752988048
0.1305 99.9572819831784
0.132 99.9581673306773
0.1335 99.9586100044267
0.135 99.9586100044267
0.1365 99.9586100044267
0.138 99.9586100044267
0.1395 99.9588313413014
0.141 99.9590526781762
0.1425 99.9610447100487
0.144 99.9612660469234
0.1455 99.9612660469234
0.147 99.9612660469234
0.1485 99.9614873837981
0.15 99.9614873837981
};
\addlegendentry{Fixed Anchor (Ours)}
\addplot [ultra thick, darkorange25512714]
table {%
0 92.1204072598495
0.0015 92.1205179282868
0.003 92.1233953076583
0.0045 92.1414342629482
0.006 92.2014165559982
0.0075 92.4837317397078
0.009 92.8562416998672
0.0105 93.3673085436033
0.012 93.8514829570606
0.0135 94.3527003098716
0.015 94.8806994245241
0.0165 95.3092076139885
0.018 95.7250996015936
0.0195 96.1220672864099
0.021 96.5141655599823
0.0225 96.9173306772908
0.024 97.2471226206286
0.0255 97.550796812749
0.027 97.8229305002213
0.0285 98.0624169986719
0.03 98.2751217352811
0.0315 98.480411686587
0.033 98.6510624169987
0.0345 98.789176626826
0.036 98.9171093404161
0.0375 99.0163789287295
0.039 99.0973882248782
0.0405 99.1697653829127
0.042 99.2395971668879
0.0435 99.3108676405488
0.045 99.3849048251438
0.0465 99.4542939353696
0.048 99.5174856131031
0.0495 99.56795042054
0.051 99.6148738379814
0.0525 99.649955732625
0.054 99.6751881363435
0.0555 99.696768481629
0.057 99.7156927844178
0.0585 99.7252102700309
0.06 99.7380478087649
0.0615 99.7449092518813
0.063 99.7526560424966
0.0645 99.7617308543603
0.066 99.7702523240371
0.0675 99.7764497565294
0.069 99.7820938468348
0.0705 99.7878486055776
0.072 99.7905046480743
0.0735 99.7954847277556
0.075 99.7999114652501
0.0765 99.804780876494
0.078 99.807547587428
0.0795 99.8124169986719
0.081 99.8147410358565
0.0825 99.8198317839752
0.084 99.8248118636564
0.0855 99.8273572377158
0.087 99.8303452855245
0.0885 99.8325586542718
0.09 99.8346613545816
0.0915 99.8365427180168
0.093 99.8382027445772
0.0945 99.8422974767596
0.096 99.8442895086321
0.0975 99.8461708720672
0.099 99.847941567065
0.1005 99.8493802567507
0.102 99.8520362992474
0.1035 99.8560203629924
0.105 99.8580123948649
0.1065 99.8602257636122
0.108 99.863545816733
0.1095 99.8644311642319
0.111 99.8660911907923
0.1125 99.8677512173528
0.114 99.868747233289
0.1155 99.8705179282868
0.117 99.8727312970341
0.1185 99.8750553342186
0.12 99.8756086764054
0.1215 99.876383355467
0.123 99.8766046923417
0.1245 99.8776007082779
0.126 99.8789287295263
0.1275 99.8797034085878
0.129 99.880367419212
0.1305 99.8819167773351
0.132 99.8841301460823
0.1335 99.8851261620186
0.135 99.8871181938911
0.1365 99.8877822045152
0.138 99.8884462151394
0.1395 99.8885568835768
0.141 99.8897742363877
0.1425 99.8909915891987
0.144 99.8914342629482
0.1455 99.8924302788844
0.147 99.8928729526339
0.1485 99.8930942895086
0.15 99.8930942895086
};
\addlegendentry{Random Triplets}
\end{axis}

\end{tikzpicture}

%% file: SupplimentFigures/PCK_Zephyr.tex
\begin{tikzpicture}

\definecolor{darkorange25512714}{RGB}{255,127,14}
\definecolor{darkslategray38}{RGB}{38,38,38}
\definecolor{lightgray204}{RGB}{204,204,204}
\definecolor{steelblue31119180}{RGB}{31,119,180}

\Large
\begin{axis}[
axis line style={lightgray204},
legend cell align={left},
legend style={
  fill opacity=0.8,
  draw opacity=1,
  text opacity=1,
  at={(0.97,0.03)},
  anchor=south east,
  draw=lightgray204
},
tick align=outside,
tick pos=left,
x grid style={lightgray204},
xlabel=\textcolor{darkslategray38}{Geodesic Error},
xmajorgrids,
xtick = {0,0.05,0.1,0.15},
xmin=0, xmax=0.15,
xtick style={color=darkslategray38},
xticklabel style={yshift= 5pt},
y grid style={lightgray204},
ylabel=\textcolor{darkslategray38}{\% Correspondences},
ymajorgrids,
ymin=47.1115537848606, ymax=100,
ytick style={color=darkslategray38}
]
\addplot [ultra thick, steelblue31119180]
table {%
0 49.601593625498
0.0015 49.601593625498
0.003 49.6347941567065
0.0045 49.6347941567065
0.006 49.6347941567065
0.0075 49.7675962815405
0.009 50.0996015936255
0.0105 50.597609561753
0.012 51.4940239043825
0.0135 52.8552456839309
0.015 54.3824701195219
0.0165 55.4448871181939
0.018 56.5073041168659
0.0195 57.9349269588313
0.021 59.3625498007968
0.0225 60.9893758300133
0.024 62.4501992031873
0.0255 63.9442231075697
0.027 65.7370517928287
0.0285 67.6626826029216
0.03 69.9203187250996
0.0315 72.0451527224436
0.033 73.140770252324
0.0345 74.601593625498
0.036 75.7636122177955
0.0375 77.5232403718459
0.039 78.4196547144754
0.0405 79.6480743691899
0.042 80.7104913678619
0.0435 81.9057104913679
0.045 82.7025232403718
0.0465 83.6321381142098
0.048 84.7277556440903
0.0495 85.7901726427623
0.051 86.4873837981408
0.0525 87.4169986719788
0.054 88.2470119521912
0.0555 88.9442231075697
0.057 89.5418326693227
0.0585 90.1394422310757
0.06 90.5710491367862
0.0615 90.8034528552457
0.063 91.0690571049137
0.0645 91.4674634794157
0.066 91.8990703851262
0.0675 92.3638778220452
0.069 92.7954847277556
0.0705 93.1938911022576
0.072 93.5258964143426
0.0735 94.0239043824701
0.075 94.4555112881806
0.0765 94.9203187250996
0.078 95.2523240371846
0.0795 95.5511288180611
0.081 96.0823373173971
0.0825 96.2151394422311
0.084 96.3811420982736
0.0855 96.6135458167331
0.087 96.9455511288181
0.0885 97.0783532536521
0.09 97.2443559096945
0.0915 97.4767596281541
0.093 97.7091633466136
0.0945 97.9747675962815
0.096 98.00796812749
0.0975 98.0411686586986
0.099 98.074369189907
0.1005 98.1075697211156
0.102 98.1075697211156
0.1035 98.207171314741
0.105 98.207171314741
0.1065 98.2403718459495
0.108 98.3067729083665
0.1095 98.406374501992
0.111 98.4395750332005
0.1125 98.539176626826
0.114 98.539176626826
0.1155 98.67197875166
0.117 98.67197875166
0.1185 98.7051792828685
0.12 98.738379814077
0.1215 98.738379814077
0.123 98.7715803452855
0.1245 98.7715803452855
0.126 98.871181938911
0.1275 98.937583001328
0.129 99.0371845949535
0.1305 99.0371845949535
0.132 99.1035856573705
0.1335 99.1035856573705
0.135 99.1035856573705
0.1365 99.1699867197875
0.138 99.203187250996
0.1395 99.203187250996
0.141 99.2363877822045
0.1425 99.269588313413
0.144 99.269588313413
0.1455 99.3027888446215
0.147 99.33598937583
0.1485 99.402390438247
0.15 99.402390438247
};
\addlegendentry{Pegasus (Main Paper)}
\addplot [ultra thick, darkorange25512714]
table {%
0 50.199203187251
0.0015 50.199203187251
0.003 50.199203187251
0.0045 50.199203187251
0.006 50.2324037184595
0.0075 50.531208499336
0.009 51.1620185922975
0.0105 51.726427622842
0.012 52.4900398406375
0.0135 53.8844621513944
0.015 54.8804780876494
0.0165 55.7768924302789
0.018 56.8393094289509
0.0195 57.9017264276228
0.021 58.9309428950863
0.0225 60.5245683930943
0.024 61.7529880478088
0.0255 63.1142098273572
0.027 65.3054448871182
0.0285 67.1978751660027
0.03 68.9243027888446
0.0315 70.6839309428951
0.033 71.8459495351926
0.0345 73.207171314741
0.036 74.7675962815405
0.0375 76.3612217795485
0.039 77.9880478087649
0.0405 79.5816733067729
0.042 80.4116865869854
0.0435 81.4741035856574
0.045 82.2045152722444
0.0465 83.1673306772908
0.048 84.0969455511288
0.0495 85.0929614873838
0.051 86.0557768924303
0.0525 86.7197875166003
0.054 87.5830013280212
0.0555 88.3134130146082
0.057 88.8446215139442
0.0585 89.5086321381142
0.06 89.9734395750332
0.0615 90.4382470119522
0.063 90.8698539176627
0.0645 91.0690571049137
0.066 91.4342629482072
0.0675 92.0650730411687
0.069 92.5630810092961
0.0705 93.0278884462151
0.072 93.4926958831341
0.0735 93.9907038512616
0.075 94.3227091633466
0.0765 94.6215139442231
0.078 95.0199203187251
0.0795 95.0863213811421
0.081 95.3851261620186
0.0825 95.6175298804781
0.084 95.7503320053121
0.0855 95.8831341301461
0.087 96.2151394422311
0.0885 96.2815405046481
0.09 96.5803452855246
0.0915 96.7795484727756
0.093 96.9123505976096
0.0945 97.0451527224436
0.096 97.1447543160691
0.0975 97.2775564409031
0.099 97.4435590969455
0.1005 97.543160690571
0.102 97.543160690571
0.1035 97.6427622841965
0.105 97.7755644090305
0.1065 97.808764940239
0.108 97.9415670650731
0.1095 98.140770252324
0.111 98.273572377158
0.1125 98.3731739707835
0.114 98.539176626826
0.1155 98.539176626826
0.117 98.5723771580345
0.1185 98.5723771580345
0.12 98.5723771580345
0.1215 98.6387782204515
0.123 98.6387782204515
0.1245 98.6387782204515
0.126 98.804780876494
0.1275 98.871181938911
0.129 98.9707835325365
0.1305 98.9707835325365
0.132 99.136786188579
0.1335 99.1699867197875
0.135 99.1699867197875
0.1365 99.1699867197875
0.138 99.1699867197875
0.1395 99.1699867197875
0.141 99.203187250996
0.1425 99.203187250996
0.144 99.203187250996
0.1455 99.203187250996
0.147 99.2363877822045
0.1485 99.2363877822045
0.15 99.2363877822045
};
\addlegendentry{Zephyr Prototype}
\end{axis}

\end{tikzpicture}

%% file: SupplimentFigures/PCK_10_Shapes_Zephyr.tex
\begin{tikzpicture}

\definecolor{darkorange25512714}{RGB}{255,127,14}
\definecolor{darkslategray38}{RGB}{38,38,38}
\definecolor{forestgreen4416044}{RGB}{44,160,44}
\definecolor{lightgray204}{RGB}{204,204,204}
\definecolor{steelblue31119180}{RGB}{31,119,180}

\Large
\begin{axis}[
axis line style={lightgray204},
legend cell align={left},
legend style={
  fill opacity=0.8,
  draw opacity=1,
  text opacity=1,
  at={(0.97,0.03)},
  anchor=south east,
  draw=lightgray204
},
tick align=outside,
tick pos=left,
xtick = {0,0.05,0.1,0.15},
x grid style={lightgray204},
xlabel=\textcolor{darkslategray38}{Geodesic Error},
xmajorgrids,
xmin=0, xmax=0.15,
xtick style={color=darkslategray38},
xticklabel style={yshift= 5pt},
y grid style={lightgray204},
ylabel=\textcolor{darkslategray38}{\% Correspondences},
ymajorgrids,
ymin=52.1674413457282, ymax=100,
ytick style={color=darkslategray38}
]

\addplot [ultra thick, steelblue31119180]
table {%
0 54.763169544046
0.0015 54.7941567065073
0.003 54.7941567065073
0.0045 54.8317839752102
0.006 54.8605577689243
0.0075 55.3497122620629
0.009 56.3789287295263
0.0105 57.153607791058
0.012 58.0787959274015
0.0135 59.4510845506861
0.015 60.522355024347
0.0165 61.5316511730855
0.018 62.9681274900398
0.0195 64.6790615316512
0.021 66.4763169544046
0.0225 68.1474103585657
0.024 70.0177069499778
0.0255 71.7020805666224
0.027 73.481629039398
0.0285 75.1947764497565
0.03 76.8260292164674
0.0315 78.444001770695
0.033 79.8162903939796
0.0345 81.3036741921204
0.036 82.6715360779106
0.0375 83.8490482514387
0.039 85.0265604249668
0.0405 86.0358565737052
0.042 86.9853917662683
0.0435 87.8331119964586
0.045 88.5613103142984
0.0465 89.3426294820717
0.048 89.9513058875609
0.0495 90.5953961930057
0.051 91.3457281983178
0.0525 91.9344842850819
0.054 92.6228419654714
0.0555 93.0035413899956
0.057 93.4373616644533
0.0585 93.8291279327136
0.06 94.1722000885348
0.0615 94.572819831784
0.063 94.9513058875609
0.0645 95.2102700309872
0.066 95.4448871181939
0.0675 95.6883576803895
0.069 96.0292164674634
0.0705 96.281540504648
0.072 96.5073041168658
0.0735 96.6888003541389
0.075 96.8348826914563
0.0765 96.9654714475431
0.078 97.1536077910579
0.0795 97.3660911907924
0.081 97.5276671093404
0.0825 97.6029216467463
0.084 97.7467906153165
0.0855 97.846392208942
0.087 97.9570606463037
0.0885 98.0168216024789
0.09 98.074369189907
0.0915 98.1363435148295
0.093 98.1584772023019
0.0945 98.235945108455
0.096 98.3045595396193
0.0975 98.3554670208056
0.099 98.3820274457724
0.1005 98.3997343957503
0.102 98.435148295706
0.1035 98.4550686144311
0.105 98.4772023019035
0.1065 98.4904825143869
0.108 98.5413899955732
0.1095 98.5723771580345
0.111 98.5945108455068
0.1125 98.6033643204958
0.114 98.6188579017264
0.1155 98.6321381142098
0.117 98.6852589641434
0.1185 98.7118193891101
0.12 98.729526339088
0.1215 98.7472332890659
0.123 98.7649402390438
0.1245 98.7782204515272
0.126 98.8069942452412
0.1275 98.8424081451969
0.129 98.8733953076582
0.1305 98.8888888888889
0.132 98.9243027888446
0.1335 98.9397963700752
0.135 98.9508632138114
0.1365 98.9840637450199
0.138 99.0061974324922
0.1395 99.0239043824701
0.141 99.0393979637007
0.1425 99.0593182824258
0.144 99.0681717574147
0.1455 99.0836653386454
0.147 99.1146525011067
0.1485 99.1212926073484
0.15 99.1212926073484
};
\addplot [ultra thick, darkorange25512714]
table {%
0 54.4134572819831
0.0015 54.4422310756972
0.003 54.4422310756972
0.0045 54.4732182381585
0.006 54.5042054006197
0.0075 54.9822930500221
0.009 55.920761398849
0.0105 56.7352810978309
0.012 57.8795927401505
0.0135 59.6082337317397
0.015 60.6795042054006
0.0165 62.1292607348384
0.018 63.9065958388668
0.0195 65.9650287737937
0.021 67.7999114652501
0.0225 69.2164674634794
0.024 71.0469234174414
0.0255 72.8087649402391
0.027 74.7853032315184
0.0285 76.850376272687
0.03 78.6321381142099
0.0315 80.2633908809208
0.033 81.651173085436
0.0345 83.0190349712262
0.036 84.3116423196105
0.0375 85.6153165117309
0.039 86.6733067729084
0.0405 87.722443559097
0.042 88.7162461266047
0.0435 89.6525011066844
0.045 90.3829127932713
0.0465 91.2549800796812
0.048 91.8149623727313
0.0495 92.6383355467021
0.051 93.1850376272687
0.0525 93.6631252766711
0.054 94.0858787073927
0.0555 94.4643647631695
0.057 94.8030101814962
0.0585 95.0553342186808
0.06 95.2943780433821
0.0615 95.588756086764
0.063 95.796812749004
0.0645 96.0004426737494
0.066 96.2018592297476
0.0675 96.3346613545816
0.069 96.5161575918548
0.0705 96.6910137228862
0.072 96.9411243913236
0.0735 97.0628596724214
0.075 97.1956617972554
0.0765 97.3505976095618
0.078 97.4236387782204
0.0795 97.485613103143
0.081 97.5586542718016
0.0825 97.6626826029216
0.084 97.7379371403276
0.0855 97.817618415228
0.087 97.8906595838866
0.0885 97.9902611775121
0.09 98.1319167773351
0.0915 98.2270916334661
0.093 98.3089862771137
0.0945 98.3643204957945
0.096 98.4152279769809
0.0975 98.4838424081452
0.099 98.5281097830898
0.1005 98.5524568393094
0.102 98.5834440017706
0.1035 98.605577689243
0.105 98.6166445329791
0.1065 98.6321381142097
0.108 98.6586985391765
0.1095 98.6963258078795
0.111 98.7250996015935
0.1125 98.7339530765825
0.114 98.7494466578131
0.1155 98.758300132802
0.117 98.8025675077466
0.1185 98.8202744577246
0.12 98.8313413014607
0.1215 98.8601150951748
0.123 98.8667552014165
0.1245 98.8822487826471
0.126 98.891102257636
0.1275 98.9176626826029
0.129 98.9265161575918
0.1305 98.9309428950862
0.132 98.937583001328
0.1335 98.9530765825586
0.135 98.9884904825143
0.1365 99.0084108012394
0.138 99.0305444887117
0.1395 99.0460380699424
0.141 99.0703851261619
0.1425 99.0792386011509
0.144 99.0947321823815
0.1455 99.1035856573704
0.147 99.1257193448428
0.1485 99.1589198760513
0.15 99.1589198760513
};

\end{axis}

\end{tikzpicture}

%% file: SupplimentFigures/qpuworstvertices.tex
\begin{tikzpicture}

\definecolor{darkorange25512714}{RGB}{255,127,14}
\definecolor{darkslategray38}{RGB}{38,38,38}
\definecolor{forestgreen4416044}{RGB}{44,160,44}
\definecolor{gray}{RGB}{128,128,128}
\definecolor{lightgray204}{RGB}{204,204,204}
\definecolor{steelblue31119180}{RGB}{31,119,180}

\Large
\begin{axis}[
axis line style={lightgray204},
legend cell align={left},
legend style={fill opacity=0.8, draw opacity=1, text opacity=1, draw=none},
tick align=outside,
tick pos=left,
x grid style={lightgray204},
xlabel=\textcolor{darkslategray38}{Iteration},
xmajorgrids,
xmin=0, xmax=9,
xtick style={color=darkslategray38},
xticklabel style={yshift= 5pt},
y grid style={lightgray204},
ylabel=\textcolor{darkslategray38}{Energy},
ymajorgrids,
ymin=34035.05, ymax=124695.95,
ytick style={color=darkslategray38}
]

\addplot [ultra thick, steelblue31119180]
table {%
0 120575
1 93379.7859163443
2 65856.5540745175
3 56859.9989051177
4 55788.5351135226
5 55966.5265769375
6 55667.9724943428
7 55642.2407912453
8 56251.7817966032
9 55310.1550710056
};
\addlegendentry{Pegasus (Main Paper)}
\addplot [ultra thick, steelblue31119180, dashed, forget plot]
table {%
0 120575
1 69943.4553888252
2 66342.7915206047
3 64016.3755693446
4 65876.1730058255
5 62550.7472240753
6 62120.6832089319
7 65168.2282910738
8 64517.2823635984
9 66055.0810633431
};
\addplot [ultra thick, darkorange25512714]
table {%
0 120575
1 93138.6592221228
2 64395.8443103471
3 55764.3192470843
4 55063.0161003932
5 53960.8348714902
6 54082.2253346483
7 53797.6614173209
8 53843.8312559086
9 54235.8511564518
};
\addlegendentry{Zephyr Prototype}
\addplot [ultra thick, darkorange25512714, dashed, forget plot]
table {%
0 120575
1 67345.7011906687
2 62116.4510778603
3 59744.6481853321
4 60555.3274782254
5 60718.3256542676
6 64160.1567897874
7 63162.0652870842
8 60797.5553337758
9 60221.6952612643
};
\path [draw=gray, semithick, dash pattern=on 5.55pt off 2.4pt]
(axis cs:0,38156)
--(axis cs:10,38156);

\addplot [ultra thick, forestgreen4416044]
table {%
0 120575
1 91142.904325786
2 62211.7790507435
3 55623.5223354449
4 53529.4107996722
5 53374.1822750181
6 53214.7761427653
7 53197.9737940531
8 53197.9737940531
9 53197.9737940531
};
\addlegendentry{SA}
\addplot [ultra thick, forestgreen4416044, dashed, forget plot]
table {%
0 120575
1 58960.4301751069
2 53646.1820547435
3 52543.3119711551
4 51678.8554060219
5 51631.1356493409
6 51515.6549776222
7 51515.6549776222
8 51515.6549776222
9 51515.6549776222
};

\end{axis}

\end{tikzpicture}

%% file: SupplimentFigures/PCK_worst_vertices.tex
\begin{tikzpicture}

\definecolor{darkorange25512714}{RGB}{255,127,14}
\definecolor{darkslategray38}{RGB}{38,38,38}
\definecolor{forestgreen4416044}{RGB}{44,160,44}
\definecolor{lightgray204}{RGB}{204,204,204}
\definecolor{steelblue31119180}{RGB}{31,119,180}

\Large
\begin{axis}[
axis line style={lightgray204},
legend cell align={left},
legend style={
  fill opacity=0.8,
  draw opacity=1,
  text opacity=1,
  at={(0.65,0.15)},
  anchor=center,
  draw=lightgray204
},
tick align=outside,
tick pos=left,
xtick = {0,0.05,0.1,0.15},
x grid style={lightgray204},
xlabel=\textcolor{darkslategray38}{Geodesic Error},
xmajorgrids,
xmin=0, xmax=0.15,
xtick style={color=darkslategray38},
xticklabel style={yshift= 5pt},
y grid style={lightgray204},
ylabel=\textcolor{darkslategray38}{\% Correspondences},
ymajorgrids,
ymin=38.1158698539177, ymax=100,
ytick style={color=darkslategray38}
]
\addplot [ultra thick, steelblue31119180]
table {%
0 49.269588313413
0.0015 49.269588313413
0.003 49.269588313413
0.0045 49.269588313413
0.006 49.269588313413
0.0075 49.269588313413
0.009 49.734395750332
0.0105 50.2324037184595
0.012 50.8964143426295
0.0135 51.792828685259
0.015 52.7224435590969
0.0165 53.1540504648074
0.018 53.6520584329349
0.0195 54.3824701195219
0.021 55.5444887118194
0.0225 56.5737051792829
0.024 56.9721115537849
0.0255 58.1009296148738
0.027 59.9933598937583
0.0285 61.2217795484728
0.03 62.6162018592298
0.0315 64.1434262948207
0.033 65.1394422310757
0.0345 66.2682602921647
0.036 67.2310756972112
0.0375 68.5258964143426
0.039 69.6215139442231
0.0405 71.0491367861886
0.042 71.7795484727756
0.0435 72.410358565737
0.045 73.2403718459495
0.0465 74.070385126162
0.048 75.066401062417
0.0495 75.996015936255
0.051 76.6932270916335
0.0525 77.4236387782205
0.054 78.3532536520584
0.0555 79.2164674634794
0.057 79.6812749003984
0.0585 80.2788844621514
0.06 80.9428950863214
0.0615 81.1420982735724
0.063 81.8393094289509
0.0645 82.2377158034528
0.066 83.1341301460823
0.0675 83.7981407702523
0.069 84.2629482071713
0.0705 84.8605577689243
0.072 85.3253652058433
0.0735 86.1885790172643
0.075 86.7197875166003
0.0765 87.2509960159363
0.078 87.6826029216467
0.0795 87.9150066401063
0.081 88.4462151394422
0.0825 88.7782204515272
0.084 88.9774236387782
0.0855 89.3426294820717
0.087 89.9070385126162
0.0885 90.1394422310757
0.09 90.4382470119522
0.0915 90.8034528552457
0.093 91.3014608233732
0.0945 91.4674634794157
0.096 91.6334661354582
0.0975 91.9986719787517
0.099 92.2974767596282
0.1005 92.6294820717132
0.102 92.9282868525896
0.1035 93.0610889774236
0.105 93.3266932270916
0.1065 93.5922974767596
0.108 93.8579017264276
0.1095 94.0571049136786
0.111 94.2895086321381
0.1125 94.4887118193891
0.114 94.7211155378486
0.1155 94.8539176626826
0.117 94.9867197875166
0.1185 95.1527224435591
0.12 95.3187250996016
0.1215 95.6175298804781
0.123 95.7171314741036
0.1245 95.9163346613546
0.126 96.0823373173971
0.1275 96.2151394422311
0.129 96.4807436918991
0.1305 96.4807436918991
0.132 96.5803452855246
0.1335 96.6135458167331
0.135 96.6135458167331
0.1365 96.8459495351926
0.138 96.9455511288181
0.1395 97.1115537848606
0.141 97.2775564409031
0.1425 97.4103585657371
0.144 97.4435590969455
0.1455 97.543160690571
0.147 97.675962815405
0.1485 97.675962815405
0.15 97.675962815405
};
\addplot [ultra thick, steelblue31119180, dashed, forget plot]
table {%
0 49.136786188579
0.0015 49.136786188579
0.003 49.136786188579
0.0045 49.136786188579
0.006 49.136786188579
0.0075 49.203187250996
0.009 49.535192563081
0.0105 50
0.012 50.6972111553785
0.0135 51.6932270916335
0.015 52.9216467463479
0.0165 53.5192563081009
0.018 54.3824701195219
0.0195 55.2456839309429
0.021 56.3081009296149
0.0225 57.5033200531209
0.024 58.0345285524568
0.0255 58.8977423638778
0.027 60.3253652058433
0.0285 61.8525896414343
0.03 63.6122177954847
0.0315 65.3054448871182
0.033 66.3014608233732
0.0345 67.6294820717132
0.036 68.5922974767596
0.0375 69.7543160690571
0.039 71.1155378486056
0.0405 72.2443559096946
0.042 73.140770252324
0.0435 74.1699867197875
0.045 74.933598937583
0.0465 76.0956175298805
0.048 77.0916334661355
0.0495 77.8220451527224
0.051 78.5524568393094
0.0525 79.6148738379814
0.054 80.6440903054449
0.0555 81.5737051792829
0.057 82.1381142098274
0.0585 82.7357237715804
0.06 83.6321381142098
0.0615 84.0305444887118
0.063 84.4621513944223
0.0645 84.8605577689243
0.066 85.5245683930943
0.0675 86.0889774236388
0.069 86.5869853917663
0.0705 87.1513944223108
0.072 87.6494023904383
0.0735 88.2470119521912
0.075 88.6454183266932
0.0765 89.2762284196547
0.078 89.7078353253652
0.0795 90.0398406374502
0.081 90.6042496679947
0.0825 90.7370517928287
0.084 90.9694555112882
0.0855 91.4010624169987
0.087 92.2310756972112
0.0885 92.4634794156706
0.09 92.6294820717132
0.0915 93.0278884462152
0.093 93.3930942895086
0.0945 93.5590969455511
0.096 93.6254980079681
0.0975 93.9907038512616
0.099 94.1567065073041
0.1005 94.2231075697211
0.102 94.4887118193891
0.1035 94.6879150066401
0.105 94.7875166002656
0.1065 94.9535192563081
0.108 95.1859229747676
0.1095 95.5179282868526
0.111 95.7835325365206
0.1125 96.1487383798141
0.114 96.3811420982736
0.1155 96.5139442231076
0.117 96.6135458167331
0.1185 96.6799468791501
0.12 96.7795484727756
0.1215 96.9123505976096
0.123 96.9787516600266
0.1245 97.1447543160691
0.126 97.2443559096945
0.1275 97.3107569721115
0.129 97.3771580345285
0.1305 97.476759628154
0.132 97.476759628154
0.1335 97.5099601593625
0.135 97.5099601593625
0.1365 97.543160690571
0.138 97.5763612217796
0.1395 97.6759628154051
0.141 97.8419654714476
0.1425 97.875166002656
0.144 97.9083665338645
0.1455 97.9747675962815
0.147 98.00796812749
0.1485 98.0411686586985
0.15 98.0411686586985
};
\addplot [ultra thick, darkorange25512714]
table {%
0 45.2191235059761
0.0015 45.2191235059761
0.003 45.2191235059761
0.0045 45.2191235059761
0.006 45.2191235059761
0.0075 45.2855245683931
0.009 45.5511288180611
0.0105 45.9495351925631
0.012 46.8127490039841
0.0135 47.7755644090305
0.015 48.7051792828685
0.0165 49.2363877822045
0.018 49.8339973439575
0.0195 50.6308100929615
0.021 51.7596281540505
0.0225 53.1208499335989
0.024 53.9176626826029
0.0255 54.7808764940239
0.027 56.3745019920319
0.0285 57.7689243027889
0.03 59.1965471447543
0.0315 60.4581673306773
0.033 61.3545816733068
0.0345 62.4501992031872
0.036 63.4462151394422
0.0375 64.8074369189907
0.039 65.9030544488712
0.0405 67.0650730411687
0.042 68.0278884462151
0.0435 68.9907038512616
0.045 69.8871181938911
0.0465 70.8831341301461
0.048 71.8459495351926
0.0495 72.875166002656
0.051 73.7051792828685
0.0525 74.3691899070385
0.054 75.332005312085
0.0555 76.1620185922975
0.057 76.8260292164675
0.0585 77.5896414342629
0.06 78.2868525896414
0.0615 78.7184594953519
0.063 79.2164674634794
0.0645 79.8804780876494
0.066 80.8100929614874
0.0675 81.6733067729084
0.069 82.4369189907038
0.0705 83.1009296148738
0.072 83.6985391766268
0.0735 84.4289508632138
0.075 84.9601593625498
0.0765 85.6241699867198
0.078 86.1221779548473
0.0795 86.3877822045153
0.081 86.9853917662683
0.0825 87.1845949535193
0.084 87.4169986719788
0.0855 87.8486055776892
0.087 88.5790172642762
0.0885 88.9774236387782
0.09 89.2098273572377
0.0915 89.6414342629482
0.093 90.0730411686587
0.0945 90.3386454183267
0.096 90.7038512616202
0.0975 90.9694555112882
0.099 91.2350597609562
0.1005 91.4010624169987
0.102 91.6666666666667
0.1035 91.9986719787517
0.105 92.1646746347942
0.1065 92.5298804780876
0.108 92.7954847277556
0.1095 93.0610889774236
0.111 93.4926958831341
0.1125 93.7250996015936
0.114 94.0571049136786
0.1155 94.3227091633466
0.117 94.4223107569721
0.1185 94.4887118193891
0.12 94.6547144754316
0.1215 94.8871181938911
0.123 94.9203187250996
0.1245 95.1527224435591
0.126 95.3851261620186
0.1275 95.4847277556441
0.129 95.8167330677291
0.1305 95.8499335989376
0.132 95.8831341301461
0.1335 96.0823373173971
0.135 96.1819389110226
0.1365 96.3811420982736
0.138 96.4475431606906
0.1395 96.5803452855246
0.141 96.6467463479416
0.1425 96.8459495351926
0.144 96.8459495351926
0.1455 96.9455511288181
0.147 97.011952191235
0.1485 97.011952191235
0.15 97.011952191235
};
\addplot [ultra thick, darkorange25512714, dashed, forget plot]
table {%
0 41.8990703851262
0.0015 41.8990703851262
0.003 41.8990703851262
0.0045 41.8990703851262
0.006 41.8990703851262
0.0075 42.0318725099602
0.009 42.3306772908367
0.0105 42.7622841965471
0.012 43.4262948207171
0.0135 44.4555112881806
0.015 45.5511288180611
0.0165 46.0159362549801
0.018 46.6467463479416
0.0195 47.3771580345286
0.021 48.273572377158
0.0225 49.33598937583
0.024 49.800796812749
0.0255 50.4316069057105
0.027 51.9588313413015
0.0285 53.2868525896414
0.03 54.7144754316069
0.0315 56.5073041168659
0.033 57.5365205843293
0.0345 58.4661354581673
0.036 59.2629482071713
0.0375 60.3585657370518
0.039 61.4541832669323
0.0405 62.5830013280213
0.042 63.4794156706507
0.0435 64.3094289508632
0.045 65.3386454183267
0.0465 66.3014608233732
0.048 67.3970783532537
0.0495 68.2270916334661
0.051 69.0571049136786
0.0525 69.8207171314741
0.054 70.8167330677291
0.0555 71.9787516600266
0.057 72.609561752988
0.0585 73.273572377158
0.06 74.070385126162
0.0615 74.33598937583
0.063 75.2324037184595
0.0645 75.863213811421
0.066 76.8260292164675
0.0675 77.6560424966799
0.069 78.4528552456839
0.0705 79.2828685258964
0.072 79.8140770252324
0.0735 80.8100929614874
0.075 81.4409030544489
0.0765 82.0053120849934
0.078 82.5033200531209
0.0795 82.9017264276228
0.081 83.4329349269588
0.0825 83.7317397078353
0.084 84.1301460823373
0.0855 84.5285524568393
0.087 85.1261620185923
0.0885 85.5245683930943
0.09 85.7569721115538
0.0915 86.1885790172643
0.093 86.5205843293493
0.0945 86.7529880478088
0.096 86.9853917662683
0.0975 87.3505976095617
0.099 87.7822045152722
0.1005 88.0146082337317
0.102 88.4130146082337
0.1035 88.5790172642762
0.105 88.7782204515272
0.1065 89.0770252324037
0.108 89.3426294820717
0.1095 89.6746347941567
0.111 90.1394422310757
0.1125 90.4382470119522
0.114 90.8034528552457
0.1155 90.9694555112882
0.117 91.1686586985392
0.1185 91.2350597609562
0.12 91.4342629482072
0.1215 91.5670650730412
0.123 91.6998671978752
0.1245 91.8990703851262
0.126 92.0650730411687
0.1275 92.1978751660027
0.129 92.3638778220452
0.1305 92.5298804780876
0.132 92.5630810092961
0.1335 92.5962815405046
0.135 92.7290836653386
0.1365 92.8950863213811
0.138 93.0278884462151
0.1395 93.0942895086321
0.141 93.2270916334661
0.1425 93.3598937583001
0.144 93.4262948207171
0.1455 93.4594953519256
0.147 93.5590969455511
0.1485 93.6254980079681
0.15 93.6254980079681
};
\addplot [ultra thick, forestgreen4416044]
table {%
0 46.9455511288181
0.0015 46.9455511288181
0.003 46.9455511288181
0.0045 46.9455511288181
0.006 46.9455511288181
0.0075 47.0783532536521
0.009 47.4103585657371
0.0105 47.9083665338645
0.012 48.6387782204515
0.0135 49.6347941567065
0.015 50.6308100929615
0.0165 51.1620185922975
0.018 51.726427622842
0.0195 52.589641434263
0.021 53.8180610889774
0.0225 55.0796812749004
0.024 55.8100929614874
0.0255 56.8725099601594
0.027 58.7649402390438
0.0285 60.4249667994688
0.03 61.8193891102258
0.0315 63.3134130146082
0.033 64.2430278884462
0.0345 65.6374501992032
0.036 66.5670650730412
0.0375 68.0610889774237
0.039 69.0571049136786
0.0405 70.2855245683931
0.042 71.0159362549801
0.0435 71.7795484727756
0.045 72.6427622841966
0.0465 73.5723771580345
0.048 74.5683930942895
0.0495 75.4316069057105
0.051 76.593625498008
0.0525 77.4236387782205
0.054 78.3532536520584
0.0555 79.0836653386454
0.057 79.7144754316069
0.0585 80.4448871181939
0.06 81.1752988047809
0.0615 81.4741035856574
0.063 82.0717131474104
0.0645 82.4037184594954
0.066 83.0345285524568
0.0675 83.8313413014608
0.069 84.4953519256308
0.0705 85.2257636122178
0.072 85.7569721115538
0.0735 86.4209827357238
0.075 86.9853917662683
0.0765 87.5498007968127
0.078 88.0478087649402
0.0795 88.3466135458167
0.081 88.8446215139442
0.0825 89.1434262948207
0.084 89.3094289508632
0.0855 89.6414342629482
0.087 90.5046480743692
0.0885 90.7038512616202
0.09 91.1354581673307
0.0915 91.5006640106242
0.093 91.8990703851262
0.0945 92.0318725099601
0.096 92.1978751660026
0.0975 92.3638778220452
0.099 92.6626826029216
0.1005 92.8286852589641
0.102 92.9946879150066
0.1035 93.1274900398407
0.105 93.3930942895086
0.1065 93.6918990703851
0.108 93.9243027888446
0.1095 94.2231075697211
0.111 94.5551128818061
0.1125 94.7543160690571
0.114 94.9535192563081
0.1155 95.0863213811421
0.117 95.2855245683931
0.1185 95.3519256308101
0.12 95.4847277556441
0.1215 95.8499335989376
0.123 96.2151394422311
0.1245 96.2815405046481
0.126 96.4143426294821
0.1275 96.4807436918991
0.129 96.6799468791501
0.1305 96.7463479415671
0.132 96.7463479415671
0.1335 96.8459495351926
0.135 96.8791500664011
0.1365 96.9787516600266
0.138 97.1115537848606
0.1395 97.1779548472775
0.141 97.3439575033201
0.1425 97.4103585657371
0.144 97.5099601593625
0.1455 97.609561752988
0.147 97.675962815405
0.1485 97.675962815405
0.15 97.675962815405
};
\addplot [ultra thick, forestgreen4416044, dashed, forget plot]
table {%
0 40.9694555112882
0.0015 40.9694555112882
0.003 40.9694555112882
0.0045 40.9694555112882
0.006 40.9694555112882
0.0075 41.0690571049137
0.009 41.2018592297477
0.0105 41.5670650730412
0.012 42.2642762284197
0.0135 43.3266932270916
0.015 44.3891102257636
0.0165 44.9867197875166
0.018 45.7835325365206
0.0195 46.6799468791501
0.021 47.742363877822
0.0225 48.937583001328
0.024 49.402390438247
0.0255 50.597609561753
0.027 52.058432934927
0.0285 53.3864541832669
0.03 54.8140770252324
0.0315 56.2416998671979
0.033 57.3041168658698
0.0345 58.4661354581673
0.036 59.2629482071713
0.0375 60.6241699867198
0.039 62.0517928286853
0.0405 63.5790172642762
0.042 64.8738379814077
0.0435 66.0026560424967
0.045 66.8326693227092
0.0465 67.9282868525896
0.048 69.0571049136786
0.0495 70.3187250996016
0.051 70.9495351925631
0.0525 71.7131474103586
0.054 72.8419654714476
0.0555 73.8379814077025
0.057 74.468791500664
0.0585 75.2988047808765
0.06 76.128818061089
0.0615 76.6268260292165
0.063 77.2244355909695
0.0645 77.6560424966799
0.066 78.3200531208499
0.0675 79.2496679946879
0.069 80.1460823373174
0.0705 80.8764940239044
0.072 81.4077025232404
0.0735 82.3041168658699
0.075 82.8353253652058
0.0765 83.3997343957503
0.078 83.9641434262948
0.0795 84.3293492695883
0.081 84.8273572377158
0.0825 85.1925630810093
0.084 85.5245683930943
0.0855 85.9893758300133
0.087 86.6865869853918
0.0885 87.0185922974768
0.09 87.4833997343957
0.0915 87.8818061088977
0.093 88.3466135458167
0.0945 88.6122177954847
0.096 88.8778220451527
0.0975 89.0770252324037
0.099 89.2430278884462
0.1005 89.5086321381142
0.102 89.8074369189907
0.1035 89.9734395750332
0.105 90.0730411686587
0.1065 90.5710491367862
0.108 90.9362549800797
0.1095 91.3346613545817
0.111 91.6334661354582
0.1125 91.9986719787517
0.114 92.2974767596281
0.1155 92.3970783532536
0.117 92.5962815405047
0.1185 92.6958831341302
0.12 92.7954847277556
0.1215 93.0942895086321
0.123 93.1274900398406
0.1245 93.3266932270916
0.126 93.5590969455511
0.1275 93.8247011952191
0.129 94.1235059760956
0.1305 94.1899070385126
0.132 94.2231075697211
0.1335 94.3227091633466
0.135 94.3891102257636
0.1365 94.6215139442231
0.138 94.7543160690571
0.1395 94.8871181938911
0.141 94.9535192563081
0.1425 95.0863213811421
0.144 95.1527224435591
0.1455 95.1859229747676
0.147 95.2855245683931
0.1485 95.3187250996016
0.15 95.3187250996016
};
\end{axis}

\end{tikzpicture}

%% file: SupplimentFigures/qubits.tex
\begin{tikzpicture}

\definecolor{darkorange25512714}{RGB}{255,127,14}
\definecolor{darkslategray38}{RGB}{38,38,38}
\definecolor{gray}{RGB}{128,128,128}
\definecolor{lightgray204}{RGB}{204,204,204}
\definecolor{steelblue31119180}{RGB}{31,119,180}

\Large
\begin{axis}[
axis line style={lightgray204},
legend cell align={left},
legend style={
  fill opacity=0.8,
  draw opacity=1,
  text opacity=1,
  at={(0.03,0.97)},
  anchor=north west,
  draw=none
},
tick align=outside,
tick pos=left,
xtick = {20,30,40,50},
x grid style={lightgray204},
xmajorgrids,
xmin=20, xmax=50,
xtick style={color=darkslategray38},
xlabel=\textcolor{darkslategray38}{Worst Vertices},
xticklabel style={yshift= 5pt},
y grid style={lightgray204},
ymajorgrids,
ymin=5.5, ymax=280.5,
ytick style={color=darkslategray38},
ylabel=\textcolor{darkslategray38}{Physical Qubits},
]
\addplot [ultra thick, gray, dashed]
table {%
20 18
30 28
40 36
50 44
};
\addlegendentry{Logical Qubits}
\addplot [ultra thick, steelblue31119180]
table {%
20 47
30 104
40 196
50 268
};
\addlegendentry{Pegasus (Main Paper)}
\addplot [ultra thick, darkorange25512714]
table {%
20 41
30 86
40 135
50 221
};
\addlegendentry{Zephyr Prototype}
\end{axis}

\end{tikzpicture}

%% file: SupplimentFigures/chains.tex
\begin{tikzpicture}

\definecolor{darkorange25512714}{RGB}{255,127,14}
\definecolor{darkslategray38}{RGB}{38,38,38}
\definecolor{lightgray204}{RGB}{204,204,204}
\definecolor{steelblue31119180}{RGB}{31,119,180}

\Large
\begin{axis}[
axis line style={lightgray204},
legend cell align={left},
legend style={
  fill opacity=0.8,
  draw opacity=1,
  text opacity=1,
  at={(0.03,0.97)},
  anchor=north west,
  draw=none
},
tick align=outside,
tick pos=left,
xtick = {20,30,40,50},
x grid style={lightgray204},
xmajorgrids,
xmin=18.5, xmax=51.5,
xtick style={color=darkslategray38},
xticklabel style={yshift= 5pt},
xlabel=\textcolor{darkslategray38}{Worst Vertices},
y grid style={lightgray204},
ymajorgrids,
ymin=1.22424242424242, ymax=8.51313131313131,
ytick style={color=darkslategray38},
ylabel=\textcolor{darkslategray38}{Average Chain Length},
]

\addplot [ultra thick, darkorange25512714, mark=-, mark size=5, mark options={solid}, only marks]
table {%
20 1.55555555555556
30 2.14285714285714
40 2.5
50 3.04545454545454
};

\path [draw=steelblue31119180, thick]
(axis cs:20,2.22222222222222)
--(axis cs:20,3.22222222222222);

\path [draw=steelblue31119180, thick]
(axis cs:30,3.42857142857143)
--(axis cs:30,4.42857142857143);

\path [draw=steelblue31119180, thick]
(axis cs:40,3.88888888888889)
--(axis cs:40,6.88888888888889);

\path [draw=steelblue31119180, thick]
(axis cs:50,4.18181818181818)
--(axis cs:50,8.18181818181818);

\addplot [ultra thick, steelblue31119180, mark=-, mark size=5, mark options={solid}, only marks]
table {%
20 2.22222222222222
30 3.42857142857143
40 3.88888888888889
50 4.18181818181818
};
\addplot [ultra thick, steelblue31119180, mark=-, mark size=5, mark options={solid}, only marks]
table {%
20 3.22222222222222
30 4.42857142857143
40 6.88888888888889
50 8.18181818181818
};

\path [draw=darkorange25512714, thick]
(axis cs:20,1.55555555555556)
--(axis cs:20,2.55555555555556);

\path [draw=darkorange25512714, thick]
(axis cs:30,2.14285714285714)
--(axis cs:30,4.14285714285714);

\path [draw=darkorange25512714, thick]
(axis cs:40,2.5)
--(axis cs:40,4.5);

\path [draw=darkorange25512714, thick]
(axis cs:50,3.04545454545454)
--(axis cs:50,7.04545454545454);

\addplot [ultra thick, darkorange25512714, mark=-, mark size=5, mark options={solid}, only marks]
table {%
20 2.55555555555556
30 4.14285714285714
40 4.5
50 7.04545454545454
};

\addplot [ultra thick, steelblue31119180, mark=*, mark size=3, mark options={solid}]
table {%
20 2.61111111111111
30 3.71428571428571
40 5.44444444444444
50 6.09090909090909
};

\addplot [ultra thick, darkorange25512714, mark=*, mark size= 3 , mark options={solid}]
table {%
20 2.27777777777778
30 3.07142857142857
40 3.75
50 5.02272727272727
};
\end{axis}

\end{tikzpicture}

%% file: SupplimentFigures/PCK_TOSCA_class.tex
\begin{tikzpicture}

\definecolor{crimson2143940}{RGB}{214,39,40}
\definecolor{darkgray176}{RGB}{176,176,176}
\definecolor{darkorange25512714}{RGB}{255,127,14}
\definecolor{darkturquoise23190207}{RGB}{23,190,207}
\definecolor{forestgreen4416044}{RGB}{44,160,44}
\definecolor{goldenrod18818934}{RGB}{188,189,34}
\definecolor{gray127}{RGB}{127,127,127}
\definecolor{mediumpurple148103189}{RGB}{148,103,189}
\definecolor{orchid227119194}{RGB}{227,119,194}
\definecolor{sienna1408675}{RGB}{140,86,75}
\definecolor{steelblue31119180}{RGB}{31,119,180}
\definecolor{lightgray204}{RGB}{204,204,204}

\begin{axis}[
tick align=outside,
tick pos=left,
legend cell align={left},
legend style={
  fill opacity=0.8,
  draw opacity=1,
  text opacity=1,
  at={(0.8,0.3)},
  anchor=center,
  draw=lightgray204
},
xticklabel style={yshift= 5pt},
yticklabel style={xshift= 5pt},
x grid style={darkgray176},
xlabel={Geodesic Error},
xmin=0, xmax=0.10,
xtick style={color=black},
xtick = {0,0.05,0.1,0.15},
xmajorgrids,
ymajorgrids,
ytick = {40,60,70,80,90,100},
y grid style={darkgray176},
ylabel={\% Correspondences},
ymin=75, ymax=100,
ytick style={color=black}
]
\addplot [ultra thick, steelblue31119180]
table {%
0 73.3570930855583
0.0015 73.592603164312
0.003 74.3112220386222
0.0045 75.7228575135922
0.006 77.461471379082
0.0075 78.8864058221451
0.009 80.029911746404
0.0105 81.1831873160213
0.012 82.4126695477934
0.0135 83.6123376478342
0.015 84.6826935740389
0.0165 85.5444660364339
0.018 86.3928193013573
0.0195 87.0983122611407
0.021 87.82401502229
0.0225 88.6988208285774
0.024 89.4775462751514
0.0255 90.2532273932116
0.027 91.0417705547129
0.0285 91.6579873652031
0.03 92.3276077505469
0.0315 92.9472467300254
0.033 93.5036695158651
0.0345 94.089936722167
0.036 94.4577943125488
0.0375 94.9147870652092
0.039 95.2394710515672
0.0405 95.5874841544399
0.042 95.8857457242795
0.0435 96.2998015954398
0.045 96.6940356595158
0.0465 97.0417985161075
0.048 97.3960737331824
0.0495 97.5981128989376
0.051 97.8962196144792
0.0525 98.1114965300571
0.054 98.3035791278116
0.0555 98.482444129456
0.057 98.6083066000835
0.0585 98.7508515001469
0.06 98.8667866044262
0.0615 98.9660557258359
0.063 99.0488858396455
0.0645 99.0885916538088
0.066 99.1813589006121
0.0675 99.2708258633568
0.069 99.2940299291512
0.0705 99.3767941880481
0.072 99.4099485353589
0.0735 99.4198626805889
0.075 99.4364300089635
0.0765 99.4596209457223
0.078 99.5126430955028
0.0795 99.5358537881497
0.081 99.5590810051202
0.0825 99.6285585818376
0.084 99.6551025220441
0.0855 99.7446485610952
0.087 99.7512820668337
0.0885 99.7711398408861
0.09 99.7810605018436
0.0915 99.8109279561762
0.093 99.8208716109969
0.0945 99.8573988617421
0.096 99.8640290639451
0.0975 99.8706658666513
0.099 99.8706658666513
0.1005 99.8939062582789
0.102 99.903863114106
0.1035 99.9105065436649
0.105 99.9105065436649
0.1065 99.9171433463711
0.108 99.920450224678
0.1095 99.9370505100639
0.111 99.9403672596493
0.1125 99.9403672596493
0.114 99.9436807122669
0.1155 99.9436807122669
0.117 99.9470073922403
0.1185 99.9470073922403
0.12 99.9470073922403
0.1215 99.9503044578519
0.123 99.9536311378253
0.1245 99.9536311378253
0.126 99.9536311378253
0.1275 99.9536311378253
0.129 99.9635485343614
0.1305 99.9801488197473
0.132 99.9834655693327
0.1335 99.996696399075
0.135 99.996696399075
0.1365 99.996696399075
0.138 100
0.1395 100
0.141 100
0.1425 100
0.144 100
0.1455 100
0.147 100
0.1485 100
0.15 100
};
\addlegendentry{Dog}
\addplot [ultra thick, darkorange25512714]
table {%
0 52.7714062657999
0.0015 53.2133817377019
0.003 54.6202148493812
0.0045 56.6261837070809
0.006 59.1574863284341
0.0075 61.4600581901197
0.009 63.7252229712705
0.0105 66.1541015636357
0.012 68.2539976676214
0.0135 70.5114527386252
0.015 72.4194205898474
0.0165 74.2061833204971
0.018 75.7991464112943
0.0195 77.3567697377323
0.021 78.7200925567779
0.0225 80.176717847444
0.024 81.5250313619298
0.0255 82.693627258145
0.027 83.5418022255023
0.0285 84.3504758115997
0.03 85.2321215170503
0.0315 85.9041364278426
0.033 86.6429455190787
0.0345 87.498262708699
0.036 88.4398629739118
0.0375 89.2516395872194
0.039 89.8972662873064
0.0405 90.6258523175926
0.042 91.327939525673
0.0435 92.0901785711135
0.045 92.6761499864056
0.0465 93.128455133223
0.048 93.6143162692293
0.0495 94.123790544866
0.051 94.5466968614042
0.0525 94.9931609129751
0.054 95.3861235089858
0.0555 95.6560146223765
0.057 95.9822863291845
0.0585 96.2318903373849
0.06 96.4517801585735
0.0615 96.681654048647
0.063 96.9149580454908
0.0645 97.085007225017
0.066 97.2015055699254
0.0675 97.3514905114559
0.069 97.5178636720453
0.0705 97.6643990740097
0.072 97.7909078030738
0.0735 97.8675423746434
0.075 98.0040712817145
0.0765 98.0907021031679
0.078 98.1939735184826
0.0795 98.313742271881
0.081 98.4136704618421
0.0825 98.5271185903115
0.084 98.6404103378317
0.0855 98.7469752915751
0.087 98.8101830739466
0.0885 98.8534939434447
0.09 98.903444665306
0.0915 98.9566888659706
0.093 98.9966393807799
0.0945 99.0299596798736
0.096 99.0798769215151
0.0975 99.1232074377674
0.099 99.1632080042675
0.1005 99.1831384322889
0.102 99.233125798537
0.1035 99.2797799178011
0.105 99.3332473665983
0.1065 99.3632844435713
0.108 99.4198686179562
0.1095 99.4664956906352
0.111 99.4864227222041
0.1125 99.4997863595079
0.114 99.5231432100142
0.1155 99.5598036161829
0.117 99.6165444573932
0.1185 99.6365814081955
0.12 99.6665684344939
0.1215 99.6966555559891
0.123 99.7166224086323
0.1245 99.7667562908553
0.126 99.7767430571308
0.1275 99.7933865801636
0.129 99.8034166905953
0.1305 99.8100834639303
0.132 99.8333572757159
0.1335 99.8400040616669
0.135 99.8799246194816
0.1365 99.8899315192749
0.138 99.8899315192749
0.1395 99.8965782457621
0.141 99.9400556183993
0.1425 99.9500691186958
0.144 99.9567191751467
0.1455 99.963412615575
0.147 99.9700793889101
0.1485 100
0.15 100
};

\addlegendentry{Cat}
\addplot [ultra thick, forestgreen4416044]
table {%
0 94.033043572761
0.0015 94.2433889572111
0.003 94.7471356396499
0.0045 95.1309448763415
0.006 95.6916229597299
0.0075 96.1285858752191
0.009 96.4587714114379
0.0105 96.8458710722858
0.012 97.2630709058786
0.0135 97.4232255371046
0.015 97.6536176086494
0.0165 97.8704922463109
0.018 98.000893933363
0.0195 98.1409447663513
0.021 98.1843551042203
0.0225 98.3146866002387
0.024 98.3914369779126
0.0255 98.4048105523721
0.027 98.5082078438832
0.0285 98.6417187245278
0.03 98.7518929740204
0.0315 98.8087237874003
0.033 98.8720407378067
0.0345 99.0654954563503
0.036 99.095508889878
0.0375 99.1021822900656
0.039 99.12549574348
0.0405 99.165502690979
0.042 99.1688494111932
0.0435 99.1855227711799
0.045 99.2089565802594
0.0465 99.2156232735929
0.048 99.2589301269809
0.0495 99.2589301269809
0.051 99.265610153701
0.0525 99.2756168337277
0.054 99.2856536375797
0.0555 99.298986970913
0.057 99.3490607188385
0.0585 99.4391414678178
0.06 99.4692452264172
0.0615 99.4725919466314
0.063 99.4759186266048
0.0645 99.4992387333792
0.066 99.5125720667125
0.0675 99.545972501842
0.069 99.5727128096267
0.0705 99.5794062500551
0.072 99.5794062500551
0.0735 99.5994130301825
0.075 99.6295402796904
0.0765 99.6295402796904
0.078 99.6328869999045
0.0795 99.6462437972989
0.081 99.6462437972989
0.0825 99.6729774049566
0.084 99.6729774049566
0.0855 99.6729774049566
0.087 99.6729774049566
0.0885 99.6763140749599
0.09 99.6796540883199
0.0915 99.6829874216533
0.093 99.6863307850769
0.0945 99.6996775319578
0.096 99.703024252172
0.0975 99.703024252172
0.099 99.7063509321454
0.1005 99.7163510123064
0.102 99.7230110256131
0.1035 99.7230110256131
0.105 99.7563312395378
0.1065 99.7563312395378
0.108 99.7596712528979
0.1095 99.7629979328713
0.111 99.7629979328713
0.1125 99.7629979328713
0.114 99.7729846261514
0.1155 99.7897115271155
0.117 99.7897115271155
0.1185 99.8030182470091
0.12 99.8097184212809
0.1215 99.8297485115519
0.123 99.8799292144443
0.1245 99.8832725778679
0.126 99.9065859777882
0.1275 99.9132593244815
0.129 99.9132593244815
0.1305 99.9266462120856
0.132 99.9400063996642
0.1335 99.9633365031352
0.135 99.99665998664
0.1365 99.99665998664
0.138 99.99665998664
0.1395 99.99665998664
0.141 99.99665998664
0.1425 99.99665998664
0.144 99.99665998664
0.1455 99.99665998664
0.147 100
0.1485 100
0.15 100
};

\addlegendentry{Horse}
\addplot [ultra thick, crimson2143940]
table {%
0 91.3751182434072
0.0015 92.631605344031
0.003 93.9510478123857
0.0045 94.9208271990571
0.006 95.5936558497487
0.0075 96.0133657494823
0.009 96.676172163793
0.0105 97.0826387571173
0.012 97.5624913147168
0.0135 97.9854802222938
0.015 98.2057051442043
0.0165 98.3922849621175
0.018 98.5123935508633
0.0195 98.6157620042005
0.021 98.759261901114
0.0225 98.8327466923997
0.024 98.9461719719585
0.0255 99.0431646064024
0.027 99.1698939193518
0.0285 99.2565388201949
0.03 99.3299403611472
0.0315 99.3665707655599
0.033 99.3766514107212
0.0345 99.3867051269297
0.036 99.4100088910907
0.0375 99.4332992006329
0.039 99.466586367931
0.0405 99.5099137105306
0.042 99.5565042270737
0.0435 99.5765247767297
0.045 99.6031819215711
0.0465 99.6198289351967
0.048 99.6465328082779
0.0495 99.6665265027013
0.051 99.6832402861155
0.0525 99.6899138668523
0.054 99.706650871864
0.0555 99.7100110869178
0.057 99.72342501818
0.0585 99.743391764634
0.06 99.7600520992362
0.0615 99.7834058078393
0.063 99.8000627585391
0.0645 99.80338281166
0.066 99.8100295447542
0.0675 99.8133662147575
0.069 99.8133662147575
0.0705 99.8233731827267
0.072 99.82670985273
0.0735 99.8334268989352
0.075 99.8500839035341
0.0765 99.8534441185878
0.078 99.87010445319
0.0795 99.8734377865233
0.081 99.8734377865233
0.0825 99.8867445263172
0.084 99.8934649564248
0.0855 99.8934649564248
0.087 99.8967949597548
0.0885 99.9167517559995
0.09 99.9200718091204
0.0915 99.9367321437226
0.093 99.9367321437226
0.0945 99.9367321437226
0.096 99.9367321437226
0.0975 99.9434124118972
0.099 99.9600727464994
0.1005 99.9633994264728
0.102 99.9633994264728
0.1035 99.9633994264728
0.105 99.9633994264728
0.1065 99.970036229179
0.108 99.970036229179
0.1095 99.970036229179
0.111 99.9866965637812
0.1125 99.9900232437546
0.114 99.9933532470846
0.1155 99.9933532470846
0.117 99.9933532470846
0.1185 99.9933532470846
0.12 99.99666999667
0.1215 99.99666999667
0.123 99.99666999667
0.1245 99.99666999667
0.126 99.99666999667
0.1275 99.99666999667
0.129 99.99666999667
0.1305 99.99666999667
0.132 99.99666999667
0.1335 99.99666999667
0.135 99.99666999667
0.1365 100
0.138 100
0.1395 100
0.141 100
0.1425 100
0.144 100
0.1455 100
0.147 100
0.1485 100
0.15 100
};

\addlegendentry{Michael}
\addplot [ultra thick, mediumpurple148103189]
table {%
0 54.2807807797069
0.0015 56.5078104790393
0.003 58.6749368561178
0.0045 60.7560700783497
0.006 62.651172973637
0.0075 64.4946894795725
0.009 66.1715716172846
0.0105 67.766174608909
0.012 69.238770313953
0.0135 70.7454760322188
0.015 72.2617546339672
0.0165 73.7511129356525
0.018 75.1224463094033
0.0195 76.4227239795392
0.021 77.5108223707943
0.0225 78.6091317816005
0.024 79.4981461956224
0.0255 80.5682217353292
0.027 81.5663864325901
0.0285 82.6150905430054
0.03 83.5661130619301
0.0315 84.495563194923
0.033 85.4576267845771
0.0345 86.2154412905837
0.036 87.0418689634206
0.0375 87.7150825629677
0.039 88.4450993961376
0.0405 89.0310703449647
0.042 89.5718781000831
0.0435 90.1240723560339
0.045 90.6787250501776
0.0465 91.1926292818278
0.048 91.7172434801713
0.0495 92.2177627049593
0.051 92.6405293362419
0.0525 93.1134289083821
0.054 93.505681180562
0.0555 93.9251457678326
0.057 94.2696532800442
0.0585 94.5837260528373
0.06 94.8617853934031
0.0615 95.1205599860415
0.063 95.3654122741856
0.0645 95.6200442569251
0.066 95.8528463248074
0.0675 96.0388332839371
0.069 96.1917251131656
0.0705 96.3819332774377
0.072 96.5095598060885
0.0735 96.7277606710106
0.075 96.9420908965312
0.0765 97.0831575853632
0.078 97.215125166925
0.0795 97.3530755741291
0.081 97.49098756252
0.0825 97.6220421323528
0.084 97.77461246225
0.0855 97.9029728932905
0.087 97.9865297224949
0.0885 98.1355940315509
0.09 98.2188161286933
0.0915 98.3194427697131
0.093 98.3712629304263
0.0945 98.4744472995029
0.096 98.5499096757795
0.0975 98.6844396434185
0.099 98.7254871927189
0.1005 98.8260520054517
0.102 98.8881089370827
0.1035 98.9573609919338
0.105 99.0442840467842
0.1065 99.0929630968289
0.108 99.1729339852717
0.1095 99.2426815857341
0.111 99.2874883014851
0.1125 99.3187609101314
0.114 99.3604751357944
0.1155 99.4116684086412
0.117 99.45312375304
0.1185 99.4946817549838
0.12 99.5188349980588
0.1215 99.5703773817715
0.123 99.6121941445651
0.1245 99.6572048271626
0.126 99.6887141295252
0.1275 99.7094665762384
0.129 99.72342844323
0.1305 99.7337927772856
0.132 99.7713494151334
0.1335 99.7994200250625
0.135 99.827399184817
0.1365 99.8481958682777
0.138 99.8621392414916
0.1395 99.8828668698961
0.141 99.9106583679034
0.1425 99.9346254457308
0.144 99.952004032231
0.1455 99.9723222056416
0.147 99.9930222979753
0.1485 100
0.15 100
};

\addlegendentry{Victora}
\addplot [ultra thick, sienna1408675]
table {%
0 85.9880834160874
0.0015 87.2426348891096
0.003 89.11287653095
0.0045 90.4402515723271
0.006 92.0191989407481
0.0075 92.9824561403509
0.009 93.8000662032439
0.0105 94.7004303210857
0.012 95.7100297914598
0.0135 96.1966236345581
0.015 96.7957629923866
0.0165 97.2393247269116
0.018 97.6530950016551
0.0195 98.1529294935452
0.021 98.3747103608077
0.0225 98.6891757696127
0.024 98.9043363124793
0.0255 98.9606090698444
0.027 99.0996358821582
0.0285 99.3346573982125
0.03 99.3909301555776
0.0315 99.4273419397551
0.033 99.5067858325058
0.0345 99.556438265475
0.036 99.6259516716319
0.0375 99.6954650777888
0.039 99.7384971863621
0.0405 99.7980801059252
0.042 99.8047004303211
0.0435 99.8344918901026
0.045 99.8477325388944
0.0465 99.8642833498842
0.048 99.8709036742801
0.0495 99.8841443230719
0.051 99.8874544852698
0.0525 99.8874544852698
0.054 99.8874544852698
0.0555 99.8874544852698
0.057 99.8874544852698
0.0585 99.8907646474677
0.06 99.8940748096657
0.0615 99.8940748096657
0.063 99.8940748096657
0.0645 99.8940748096657
0.066 99.8940748096657
0.0675 99.8940748096657
0.069 99.8940748096657
0.0705 99.8940748096657
0.072 99.8940748096657
0.0735 99.8940748096657
0.075 99.8940748096657
0.0765 99.8940748096657
0.078 99.8940748096657
0.0795 99.8940748096657
0.081 99.8940748096657
0.0825 99.8940748096657
0.084 99.8940748096657
0.0855 99.8940748096657
0.087 99.8940748096657
0.0885 99.8940748096657
0.09 99.8940748096657
0.0915 99.8940748096657
0.093 99.8940748096657
0.0945 99.8940748096657
0.096 99.8940748096657
0.0975 99.8940748096657
0.099 99.8940748096657
0.1005 99.8940748096657
0.102 99.8940748096657
0.1035 99.8940748096657
0.105 99.8940748096657
0.1065 99.8940748096657
0.108 99.8940748096657
0.1095 99.8940748096657
0.111 99.8940748096657
0.1125 99.8940748096657
0.114 99.8940748096657
0.1155 99.9205561072493
0.117 99.9205561072493
0.1185 99.9602780536246
0.12 99.9735187024164
0.1215 99.9867593512082
0.123 99.9867593512082
0.1245 100
0.126 100
0.1275 100
0.129 100
0.1305 100
0.132 100
0.1335 100
0.135 100
0.1365 100
0.138 100
0.1395 100
0.141 100
0.1425 100
0.144 100
0.1455 100
0.147 100
0.1485 100
0.15 100
};

\addlegendentry{David}
\addplot [ultra thick, orchid227119194]
table {%
0 85.6686626746507
0.0015 86.4604125083167
0.003 87.0592149035263
0.0045 88.6560212907518
0.006 90.0099800399202
0.0075 90.7651363938789
0.009 91.2408516300732
0.0105 91.8396540252827
0.012 92.4517631403859
0.0135 93.1902860944777
0.015 93.75249500998
0.0165 94.16500332668
0.018 94.6506986027944
0.0195 95.0565535595476
0.021 95.5089820359281
0.0225 95.9115103127079
0.024 96.4204923486361
0.0255 96.8928809048569
0.027 97.272122421823
0.0285 97.5848303393214
0.03 97.9008649367931
0.0315 98.1437125748503
0.033 98.3965402528277
0.0345 98.5462408516301
0.036 98.6460412508317
0.0375 98.7658017298736
0.039 98.8656021290752
0.0405 99.0818363273453
0.042 99.2714570858283
0.0435 99.3745841650033
0.045 99.4178310046574
0.0465 99.4311377245509
0.048 99.4411177644711
0.0495 99.4710578842316
0.051 99.4777112441783
0.0525 99.5076513639388
0.054 99.5242847638057
0.0555 99.5409181636727
0.057 99.5575515635396
0.0585 99.620758483034
0.06 99.6407185628743
0.0615 99.6506986027944
0.063 99.6540252827678
0.0645 99.6773120425815
0.066 99.7039254823686
0.0675 99.7604790419162
0.069 99.7638057218896
0.0705 99.8270126413839
0.072 99.8669328010646
0.0735 99.8935462408516
0.075 99.9068529607452
0.0765 99.9168330006653
0.078 99.9600798403194
0.0795 99.9634065202928
0.081 99.9933466400532
0.0825 99.9933466400532
0.084 99.9966733200266
0.0855 99.9966733200266
0.087 99.9966733200266
0.0885 99.9966733200266
0.09 100
0.0915 100
0.093 100
0.0945 100
0.096 100
0.0975 100
0.099 100
0.1005 100
0.102 100
0.1035 100
0.105 100
0.1065 100
0.108 100
0.1095 100
0.111 100
0.1125 100
0.114 100
0.1155 100
0.117 100
0.1185 100
0.12 100
0.1215 100
0.123 100
0.1245 100
0.126 100
0.1275 100
0.129 100
0.1305 100
0.132 100
0.1335 100
0.135 100
0.1365 100
0.138 100
0.1395 100
0.141 100
0.1425 100
0.144 100
0.1455 100
0.147 100
0.1485 100
0.15 100
};

\addlegendentry{Centaur}

\addplot [ultra thick, dashed, steelblue31119180]
table {%
0 81.9310924205011
0.0015 81.9310924205011
0.003 82.0045829187697
0.0045 82.2249607859724
0.006 82.6159272328364
0.0075 83.2108089193954
0.009 83.6283616713594
0.0105 84.2795988500978
0.012 84.9209365311463
0.0135 85.418716698959
0.015 85.9499472446791
0.0165 86.4509637614316
0.018 87.0388544213631
0.0195 87.7102889659608
0.021 88.1745051486454
0.0225 88.7456656448596
0.024 89.2935094305675
0.0255 89.8247199161071
0.027 90.3625034407103
0.0285 90.8268064568538
0.03 91.2176759193243
0.0315 91.4848709448291
0.033 91.7921896882079
0.0345 92.1729755531655
0.036 92.4702843067355
0.0375 92.7373723177565
0.039 93.1082150064652
0.0405 93.5557610851295
0.042 93.8631333357502
0.0435 94.2606359775445
0.045 94.6480550621074
0.0465 95.1291819098527
0.048 95.5601181456014
0.0495 95.9075070996121
0.051 96.1780253880786
0.0525 96.6024255094601
0.054 97.0769126254265
0.0555 97.384224655224
0.057 97.6180896639047
0.0585 97.8820015870906
0.06 98.0991063015016
0.0615 98.332927873312
0.063 98.5200459746038
0.0645 98.7505843635324
0.066 98.9410391549481
0.0675 99.1280402519473
0.069 99.308327888619
0.0705 99.4820123153933
0.072 99.5755898485282
0.0735 99.6859142890519
0.075 99.7794583348439
0.0765 99.8128585623
0.078 99.8462621062653
0.0795 99.8562821664458
0.081 99.8629488331125
0.0825 99.8729789233833
0.084 99.8729789233833
0.0855 99.8830090136541
0.087 99.8963824673485
0.0885 99.9064125576193
0.09 99.9231227014377
0.0915 99.9231227014377
0.093 99.9231227014377
0.0945 99.9264660648613
0.096 99.9331427616182
0.0975 99.9331427616182
0.099 99.943172851889
0.1005 99.9532029421598
0.102 99.9532029421598
0.1035 99.9532029421598
0.105 99.9532029421598
0.1065 99.9532029421598
0.108 99.9532029421598
0.1095 99.9532029421598
0.111 99.9532029421598
0.1125 99.9632330324306
0.114 99.9632330324306
0.1155 99.9632330324306
0.117 99.9699097291876
0.1185 99.9799398194584
0.12 99.9799398194584
0.1215 99.9799398194584
0.123 99.9799398194584
0.1245 99.9799398194584
0.126 99.9799398194584
0.1275 99.9799398194584
0.129 99.9799398194584
0.1305 99.9799398194584
0.132 99.9899699097292
0.1335 99.9899699097292
0.135 99.9899699097292
0.1365 99.9899699097292
0.138 100
0.1395 100
0.141 100
0.1425 100
0.144 100
0.1455 100
0.147 100
0.1485 100
0.15 100
};
\addplot [ultra thick, dashed, darkorange25512714]
table {%
0 79.0695104752439
0.0015 79.1437265034712
0.003 79.220680976213
0.0045 79.6018207508755
0.006 79.7336546179349
0.0075 80.2928414629182
0.009 80.9329827983725
0.0105 81.766424259137
0.012 82.4313792553213
0.0135 83.3764296003184
0.015 84.1304684321768
0.0165 84.8027411973464
0.018 85.747064358888
0.0195 86.607267319321
0.021 87.6442076661863
0.0225 88.67200358882
0.024 89.4849457666955
0.0255 90.0532111761397
0.027 90.6791757170906
0.0285 91.171825272922
0.03 91.745102023647
0.0315 92.3568772378027
0.033 92.9542418732756
0.0345 93.4673066909861
0.036 93.906965219519
0.0375 94.3265968907763
0.039 94.7498045556795
0.0405 95.2606433782864
0.042 95.862431523376
0.0435 96.3571079009398
0.045 96.8051679098262
0.0465 97.108567197513
0.048 97.4190697871032
0.0495 97.6773203339521
0.051 97.9795391864349
0.0525 98.2809863724242
0.054 98.5422668517115
0.0555 98.7403035356363
0.057 98.8907691016851
0.0585 99.0227395205273
0.06 99.1309474959176
0.0615 99.2107379312389
0.063 99.2703501591244
0.0645 99.3078969156981
0.066 99.3528241391603
0.0675 99.4105489920868
0.069 99.4353536309481
0.0705 99.4683774048647
0.072 99.4913289014169
0.0735 99.5133185597697
0.075 99.5280026900593
0.0765 99.563735964218
0.078 99.6004291783194
0.0795 99.6187716315584
0.081 99.6417083524643
0.0825 99.6527092120827
0.084 99.6774620421857
0.0855 99.689381147376
0.087 99.7058953574878
0.0885 99.7223772010335
0.09 99.7324542236016
0.0915 99.7407076249052
0.093 99.7654410780219
0.0945 99.7727771452916
0.096 99.7993397474061
0.0975 99.8112551947513
0.099 99.8176775804651
0.1005 99.8259226645443
0.102 99.8479187477535
0.1035 99.8524999753316
0.105 99.8570959897534
0.1065 99.8616725992672
0.108 99.8690068428425
0.1095 99.8754227571137
0.111 99.879091289588
0.1125 99.8809140323746
0.114 99.8836642314319
0.1155 99.8909911400997
0.117 99.8937404237247
0.1185 99.9001600276351
0.12 99.912088378808
0.1215 99.912088378808
0.123 99.9139184603006
0.1245 99.9194068793889
0.126 99.9221570970436
0.1275 99.9230707562487
0.129 99.9249026816605
0.1305 99.9304003372148
0.132 99.9358942936708
0.1335 99.9377243714551
0.135 99.9432090989968
0.1365 99.9450456537833
0.138 99.9569573647778
0.1395 99.9569573647778
0.141 99.9633603882818
0.1425 99.9715934951637
0.144 99.9761682328199
0.1455 99.9844087210448
0.147 99.9899128181845
0.1485 100
0.15 100
};
\addplot [ultra thick, dashed, forestgreen4416044]
table {%
0 90.7035356785357
0.0015 90.7356803231803
0.003 90.8249713999714
0.0045 90.9339124839125
0.006 91.1589357214357
0.0075 91.5446875446875
0.009 91.7108036608036
0.0105 91.8376287001287
0.012 92.2341323466324
0.0135 92.5770556270556
0.015 92.8806574431575
0.0165 93.3539611039611
0.018 93.6397468897469
0.0195 94.0327005577005
0.021 94.3309488059488
0.0225 94.6970935220935
0.024 94.9846828971829
0.0255 95.3793901043901
0.027 95.8348187473187
0.0285 96.1938080938081
0.03 96.408117045617
0.0315 96.5135081510082
0.033 96.6903153153153
0.0345 96.9582028457028
0.036 97.2582707707707
0.0375 97.551169026169
0.039 97.7101065351065
0.0405 97.8279708279708
0.042 97.9315726440726
0.0435 98.1012333762334
0.045 98.2494637494637
0.0465 98.3905709280709
0.048 98.5370245245245
0.0495 98.635270985271
0.051 98.7692156442157
0.0525 98.9245978120978
0.054 99.0710943085943
0.0555 99.238988988989
0.057 99.3318497068497
0.0585 99.3550657800658
0.06 99.3675657800658
0.0615 99.3693514943515
0.063 99.3747086372086
0.0645 99.39078006578
0.066 99.4104354354354
0.0675 99.4157925782926
0.069 99.4336622336622
0.0705 99.460472972973
0.072 99.4854854854855
0.0735 99.5819909194909
0.075 99.6177302302302
0.0765 99.6999106249106
0.078 99.7320534820535
0.0795 99.7516963391963
0.081 99.7695534820535
0.0825 99.7766963391964
0.084 99.794565994566
0.0855 99.8249374374375
0.087 99.8463785213786
0.0885 99.84994994995
0.09 99.8570928070928
0.0915 99.8642356642357
0.093 99.8713785213785
0.0945 99.8731642356642
0.096 99.8731642356642
0.0975 99.8767356642357
0.099 99.8803070928071
0.1005 99.8946053196053
0.102 99.896391033891
0.1035 99.9160464035464
0.105 99.9196178321179
0.1065 99.9446303446303
0.108 99.9517732017732
0.1095 99.9571303446304
0.111 99.9571303446304
0.1125 99.9732142857143
0.114 99.975
0.1155 99.9785714285714
0.117 99.9839285714286
0.1185 99.9875
0.12 99.9910714285714
0.1215 99.9910714285714
0.123 99.9928571428571
0.1245 99.9928571428571
0.126 99.9982142857143
0.1275 100
0.129 100
0.1305 100
0.132 100
0.1335 100
0.135 100
0.1365 100
0.138 100
0.1395 100
0.141 100
0.1425 100
0.144 100
0.1455 100
0.147 100
0.1485 100
0.15 100
};
\addplot [ultra thick, dashed, crimson2143940]
table {%
0 82.8216899292996
0.0015 83.2058394473594
0.003 83.6241160340079
0.0045 83.8153282752127
0.006 84.3435341747518
0.0075 84.6945801199768
0.009 85.1019422696843
0.0105 85.4776385293923
0.012 85.8323347560135
0.0135 86.0999983071133
0.015 86.372754478914
0.0165 86.8311720315121
0.018 87.3051467173633
0.0195 87.5423483562087
0.021 87.7442964189842
0.0225 87.9658468876969
0.024 88.2457563593707
0.0255 88.5335067407806
0.027 88.8372708896843
0.0285 89.0692421408393
0.03 89.2450205894764
0.0315 89.4187573295854
0.033 89.6373367799273
0.0345 89.9011457828938
0.036 90.1757467024618
0.0375 90.540222093595
0.039 90.8068613222536
0.0405 91.0708337530166
0.042 91.3245588398071
0.0435 91.5418079401854
0.045 91.7829945610806
0.0465 92.0371320450529
0.048 92.2553188179808
0.0495 92.4103124487021
0.051 92.562430095293
0.0525 92.7204954623873
0.054 92.8971937748744
0.0555 93.1141147889398
0.057 93.2939555689015
0.0585 93.503370086244
0.06 93.732227276726
0.0615 93.9537003983255
0.063 94.1530898947933
0.0645 94.3963530995321
0.066 94.5844161103526
0.0675 94.7541848431315
0.069 94.9181957641153
0.0705 95.0502264989765
0.072 95.1903220264438
0.0735 95.2999433559307
0.075 95.3945130470747
0.0765 95.5083786137908
0.078 95.5849261570319
0.0795 95.6802537641237
0.081 95.7989916606853
0.0825 95.8878260769159
0.084 95.9958437392715
0.0855 96.103537578074
0.087 96.2055351080913
0.0885 96.3685708871454
0.09 96.4764443733926
0.0915 96.5453311873151
0.093 96.6253778304054
0.0945 96.7186166359395
0.096 96.7771950523909
0.0975 96.8620153476561
0.099 96.9506445446387
0.1005 97.0705058929194
0.102 97.1879193050659
0.1035 97.2994334014879
0.105 97.3908132307092
0.1065 97.4593108408014
0.108 97.5337487876734
0.1095 97.6616365805801
0.111 97.7236651882648
0.1125 97.8276476529156
0.114 97.9241782003679
0.1155 98.0333447666767
0.117 98.1114418389967
0.1185 98.2097158416412
0.12 98.3021340581478
0.1215 98.3846035576397
0.123 98.5216641093732
0.1245 98.6120837724141
0.126 98.717261566227
0.1275 98.8233727805865
0.129 98.961503137701
0.1305 99.0801511840256
0.132 99.1555695002671
0.1335 99.2687912083751
0.135 99.3469167697517
0.1365 99.4406476415933
0.138 99.4913721400845
0.1395 99.59125427067
0.141 99.6843965205802
0.1425 99.7529300937787
0.144 99.8174950282333
0.1455 99.8823754196331
0.147 99.9371821367986
0.1485 100
0.15 100
};
\addplot [ultra thick, dashed, mediumpurple148103189]
table {%
0 94.3223481056814
0.0015 94.3973587223587
0.003 94.9103012103013
0.0045 95.2914323414324
0.006 95.5876892043559
0.0075 96.2802241635575
0.009 96.4908977158977
0.0105 96.6060689477356
0.012 96.7409507992842
0.0135 96.864460672794
0.015 97.0144750811417
0.0165 97.1948076864743
0.018 97.50091000091
0.0195 97.6623183789851
0.021 97.8267472017472
0.0225 97.9729926896594
0.024 98.0957472624139
0.0255 98.2230556313889
0.027 98.3564230897564
0.0285 98.4806875056875
0.03 98.6307246640579
0.0315 98.7292239208906
0.033 98.850455000455
0.0345 98.9322921406255
0.036 99.02549670883
0.0375 99.1300785634119
0.039 99.1944785694785
0.0405 99.2209967543301
0.042 99.2808566141899
0.0435 99.3717960384627
0.045 99.42407331574
0.0465 99.4725642308976
0.048 99.5604513604514
0.0495 99.6104543937877
0.051 99.6513680347014
0.0525 99.6953210786544
0.054 99.7021415354749
0.0555 99.7286612369946
0.057 99.7544218460885
0.0585 99.7642756392756
0.06 99.7756438256438
0.0615 99.7870135286802
0.063 99.7961074711074
0.0645 99.8074748991415
0.066 99.8233862650529
0.0675 99.8468877968878
0.069 99.8499180999181
0.0705 99.8688658355325
0.072 99.8900794734128
0.0735 99.8999302332635
0.075 99.9006885673552
0.0765 99.9014469014469
0.078 99.9029635696302
0.0795 99.9112976612976
0.081 99.9279658446325
0.0825 99.9287234203901
0.084 99.9302385719053
0.0855 99.9309961476628
0.087 99.9309961476628
0.0885 99.9332696332697
0.09 99.934786301453
0.0915 99.9355446355447
0.093 99.9355446355447
0.0945 99.9363029696363
0.096 99.9378196378197
0.0975 99.9385779719113
0.099 99.9408529741864
0.1005 99.9423696423697
0.102 99.9454029787363
0.1035 99.9476779810113
0.105 99.9567772317773
0.1065 99.9575355658689
0.108 99.9575355658689
0.1095 99.9590522340523
0.111 99.9590522340523
0.1125 99.9598105681439
0.114 99.9613272363273
0.1155 99.9727014893681
0.117 99.9734598234598
0.1185 99.9734598234598
0.12 99.9742181575514
0.1215 99.9757348257348
0.123 99.9757348257348
0.1245 99.9848340765007
0.126 99.9848340765007
0.1275 99.9855924105924
0.129 99.9878674128674
0.1305 99.9901424151424
0.132 99.9901424151424
0.1335 99.9901424151424
0.135 99.9901424151424
0.1365 99.990900749234
0.138 99.9916590833257
0.1395 99.9916590833257
0.141 99.9916590833257
0.1425 100
0.144 100
0.1455 100
0.147 100
0.1485 100
0.15 100
};
\addplot [ultra thick, dashed, sienna1408675]
table {%
0 96.4633333333333
0.0015 96.4633333333333
0.003 96.89
0.0045 96.9633333333333
0.006 97.3966666666666
0.0075 97.6299999999999
0.009 97.8133333333333
0.0105 98.2566666666667
0.012 98.4833333333333
0.0135 98.69
0.015 98.8066666666667
0.0165 98.91
0.018 98.9966666666667
0.0195 99.1166666666667
0.021 99.1866666666667
0.0225 99.2566666666667
0.024 99.3166666666667
0.0255 99.4366666666667
0.027 99.4733333333333
0.0285 99.54
0.03 99.56
0.0315 99.5866666666667
0.033 99.6133333333333
0.0345 99.6433333333333
0.036 99.6866666666667
0.0375 99.6966666666667
0.039 99.71
0.0405 99.7133333333333
0.042 99.7133333333333
0.0435 99.74
0.045 99.7466666666667
0.0465 99.76
0.048 99.77
0.0495 99.79
0.051 99.8033333333333
0.0525 99.83
0.054 99.83
0.0555 99.8466666666667
0.057 99.87
0.0585 99.88
0.06 99.8933333333333
0.0615 99.9
0.063 99.9033333333333
0.0645 99.9166666666667
0.066 99.9266666666667
0.0675 99.9666666666667
0.069 99.9666666666667
0.0705 99.9666666666667
0.072 99.9833333333333
0.0735 100
0.075 100
0.0765 100
0.078 100
0.0795 100
0.081 100
0.0825 100
0.084 100
0.0855 100
0.087 100
0.0885 100
0.09 100
0.0915 100
0.093 100
0.0945 100
0.096 100
0.0975 100
0.099 100
0.1005 100
0.102 100
0.1035 100
0.105 100
0.1065 100
0.108 100
0.1095 100
0.111 100
0.1125 100
0.114 100
0.1155 100
0.117 100
0.1185 100
0.12 100
0.1215 100
0.123 100
0.1245 100
0.126 100
0.1275 100
0.129 100
0.1305 100
0.132 100
0.1335 100
0.135 100
0.1365 100
0.138 100
0.1395 100
0.141 100
0.1425 100
0.144 100
0.1455 100
0.147 100
0.1485 100
0.15 100
};
\addplot [ultra thick, dashed, orchid227119194]
table {%
0 90.14
0.0015 90.1466666666667
0.003 90.49
0.0045 91.1866666666667
0.006 91.7866666666666
0.0075 92.2766666666666
0.009 92.9933333333333
0.0105 93.52
0.012 94.2066666666667
0.0135 94.61
0.015 95.0366666666666
0.0165 95.31
0.018 95.4466666666667
0.0195 95.5833333333333
0.021 95.8466666666667
0.0225 95.9633333333333
0.024 96.32
0.0255 96.48
0.027 96.7266666666666
0.0285 96.8533333333334
0.03 96.9866666666667
0.0315 97.07
0.033 97.1566666666667
0.0345 97.3166666666666
0.036 97.4766666666666
0.0375 97.7033333333333
0.039 97.9033333333333
0.0405 98.13
0.042 98.3166666666667
0.0435 98.51
0.045 98.6333333333334
0.0465 98.8466666666667
0.048 98.9733333333333
0.0495 99.0766666666667
0.051 99.2133333333333
0.0525 99.3366666666667
0.054 99.47
0.0555 99.51
0.057 99.5333333333333
0.0585 99.5666666666667
0.06 99.6533333333333
0.0615 99.7
0.063 99.7633333333333
0.0645 99.78
0.066 99.81
0.0675 99.8133333333333
0.069 99.82
0.0705 99.8266666666667
0.072 99.8333333333333
0.0735 99.8366666666667
0.075 99.86
0.0765 99.8666666666667
0.078 99.89
0.0795 99.9266666666667
0.081 99.9333333333333
0.0825 99.9366666666667
0.084 99.9366666666667
0.0855 99.95
0.087 99.95
0.0885 99.95
0.09 99.95
0.0915 99.95
0.093 99.95
0.0945 99.95
0.096 99.9533333333333
0.0975 99.9566666666667
0.099 99.9566666666667
0.1005 99.96
0.102 99.96
0.1035 99.9766666666667
0.105 99.98
0.1065 99.98
0.108 99.98
0.1095 99.9833333333333
0.111 99.9833333333333
0.1125 99.9833333333333
0.114 99.9833333333333
0.1155 99.9833333333333
0.117 99.9833333333333
0.1185 99.9833333333333
0.12 100
0.1215 100
0.123 100
0.1245 100
0.126 100
0.1275 100
0.129 100
0.1305 100
0.132 100
0.1335 100
0.135 100
0.1365 100
0.138 100
0.1395 100
0.141 100
0.1425 100
0.144 100
0.1455 100
0.147 100
0.1485 100
0.15 100
};
\end{axis}

\end{tikzpicture}

%% file: SupplimentFigures/LRdesc.tex
\begin{tikzpicture}

\definecolor{crimson2143940}{RGB}{214,39,40}
\definecolor{darkgray176}{RGB}{176,176,176}
\definecolor{darkorange25512714}{RGB}{255,127,14}
\definecolor{forestgreen4416044}{RGB}{44,160,44}
\definecolor{steelblue31119180}{RGB}{31,119,180}
\definecolor{lightgray204}{RGB}{204,204,204}

\Large
\begin{axis}[
legend cell align={left},
legend style={
  fill opacity=0.8,
  draw opacity=1,
  text opacity=1,
  at={(0.97,0.03)},
  anchor=south east,
  draw=lightgray204
},
tick align=outside,
tick pos=left,
x grid style={darkgray176},
xtick = {0,0.05,0.1,0.15},
xlabel={Geodesic Error},
xmin=0, xmax=0.15,
xtick style={color=black},
xticklabel style={yshift= 5pt},
xmajorgrids,
ymajorgrids,
yticklabel style={xshift= 5pt},
y grid style={darkgray176},
ylabel={\% Correspondences},
ymin=30, ymax=100,
ytick style={color=black}
]
\addplot [ultra thick, steelblue31119180]
table {%
0 51.2827555390533
0.0015 51.8641372208934
0.003 53.2146159655153
0.0045 54.9806382405994
0.006 57.5146025317151
0.0075 60.0077856895293
0.009 62.2242898780877
0.0105 64.4028120915058
0.012 66.5613921234063
0.0135 68.7289465041221
0.015 70.5978661899041
0.0165 72.2701509840696
0.018 73.7657422983166
0.0195 75.2364019588098
0.021 76.5895894445652
0.0225 78.0961042728416
0.024 79.3599204131689
0.0255 80.565712515973
0.027 81.5449937514256
0.0285 82.4800868600155
0.03 83.4303835051003
0.0315 84.2809246820583
0.033 85.0615693434214
0.0345 85.8897366138886
0.036 86.8285148833775
0.0375 87.6282134975959
0.039 88.4200562405174
0.0405 89.1711195013369
0.042 89.8093719119894
0.0435 90.549537342114
0.045 91.1498265445945
0.0465 91.7545031478986
0.048 92.3102448944225
0.0495 92.8378606567253
0.051 93.2893909734178
0.0525 93.7646146635959
0.054 94.1660987558939
0.0555 94.5203149535873
0.057 94.8399191615949
0.0585 95.1568627220012
0.06 95.4059253437614
0.0615 95.6612266571517
0.063 95.9238159659513
0.0645 96.1864891910602
0.066 96.3673276048486
0.0675 96.5236294751415
0.069 96.7261918561814
0.0705 96.9096410762237
0.072 97.0495961686294
0.0735 97.1931381524204
0.075 97.3493913029419
0.0765 97.4765878627066
0.078 97.6429356159051
0.0795 97.7664438653261
0.081 97.8736574395139
0.0825 98.0318578047599
0.084 98.1764065379189
0.0855 98.2891758744252
0.087 98.4108905798865
0.0885 98.4872071879577
0.09 98.5826375265531
0.0915 98.6698075988052
0.093 98.7524639437443
0.0945 98.8170101170457
0.096 98.8787447655987
0.0975 98.947842144327
0.099 99.0178790993535
0.1005 99.0714371161864
0.102 99.1204759060856
0.1035 99.1687286116406
0.105 99.2324616730183
0.1065 99.2779569428163
0.108 99.3397469513088
0.1095 99.3906720609429
0.111 99.4224652531733
0.1125 99.448871149192
0.114 99.4770505330181
0.1155 99.5152549382668
0.117 99.5625720820572
0.1185 99.5789450218972
0.12 99.6098882483429
0.1215 99.6471732513815
0.123 99.6626280416896
0.1245 99.7026838798648
0.126 99.7226868241501
0.1275 99.7399697710862
0.129 99.7527210027383
0.1305 99.769995827692
0.132 99.7917734316924
0.1335 99.8053954570241
0.135 99.8390259623233
0.1365 99.8581309500769
0.138 99.8717538435083
0.1395 99.8944714971738
0.141 99.9172683508007
0.1425 99.9327394912886
0.144 99.9490933303313
0.1455 99.9600254218557
0.147 99.9791076411799
0.1485 100
0.15 100
};
\addlegendentry{With Left-Right}
\addplot [ultra thick, darkorange25512714]
table {%
0 30.8864013818146
0.0015 31.3228911230655
0.003 32.0042701802558
0.0045 32.9805270466876
0.006 34.2026321595216
0.0075 35.6839164914589
0.009 37.458313192461
0.0105 39.1481346636783
0.012 40.6454369335862
0.0135 42.2248943899164
0.015 43.7578306157743
0.0165 45.0573616792517
0.018 46.3436839854523
0.0195 47.6560882537004
0.021 48.9596869155032
0.0225 50.3027046752211
0.024 51.5686465903498
0.0255 52.845318758874
0.027 54.0125607781556
0.0285 55.1522410601754
0.03 56.3677777391881
0.0315 57.5128841148021
0.033 58.5092758727202
0.0345 59.5225796506218
0.036 60.5602142256785
0.0375 61.5492043458269
0.039 62.4530560742832
0.0405 63.4029900967566
0.042 64.2865342945899
0.0435 65.1603898250192
0.045 66.0505618738915
0.0465 66.9746949643089
0.048 67.9183410572697
0.0495 68.8227379810735
0.051 69.630824422822
0.0525 70.4582982844795
0.054 71.2523511597591
0.0555 71.9386383338911
0.057 72.5835809059024
0.0585 73.2080315641417
0.06 73.9086890462388
0.0615 74.4850139083331
0.063 75.1435008856221
0.0645 75.7774064462289
0.066 76.3155157209705
0.0675 76.8545341832699
0.069 77.2977061542876
0.0705 77.7457450908665
0.072 78.2199403626026
0.0735 78.6411115702198
0.075 79.055747303738
0.0765 79.5081981374898
0.078 79.9671784227686
0.0795 80.4757214040484
0.081 80.9153083785992
0.0825 81.4810116792612
0.084 82.0090490594881
0.0855 82.502068654266
0.087 83.0063415817128
0.0885 83.6430205669229
0.09 84.1199631074617
0.0915 84.6432107119071
0.093 85.1495485773543
0.0945 85.635644569429
0.096 86.0389197230268
0.0975 86.4800179249469
0.099 86.9305216703729
0.1005 87.2947793066458
0.102 87.7891482385605
0.1035 88.1929931961544
0.105 88.5325875885285
0.1065 88.9946512590125
0.108 89.3933716224593
0.1095 89.8074862512947
0.111 90.1721705785302
0.1125 90.4940091516515
0.114 90.8751571754104
0.1155 91.2588840625922
0.117 91.6793207261531
0.1185 92.0633832265236
0.12 92.4236134653785
0.1215 92.7943290348873
0.123 93.202902040884
0.1245 93.5982003073131
0.126 93.9812169742051
0.1275 94.3146090377232
0.129 94.6918770245565
0.1305 95.0519308509003
0.132 95.4613384889146
0.1335 95.880425196154
0.135 96.294433426717
0.1365 96.6239159470051
0.138 96.9984225089091
0.1395 97.419678444136
0.141 97.8565808702181
0.1425 98.297323639062
0.144 98.7522214469978
0.1455 99.1577983152942
0.147 99.6219074422467
0.1485 100
0.15 100
};
\addlegendentry{Without Left-Right}

\addplot [ultra thick,dashed, steelblue31119180]
table {%
0 79.0695104752439
0.0015 79.1437265034712
0.003 79.220680976213
0.0045 79.6018207508755
0.006 79.7336546179349
0.0075 80.2928414629182
0.009 80.9329827983725
0.0105 81.766424259137
0.012 82.4313792553213
0.0135 83.3764296003184
0.015 84.1304684321768
0.0165 84.8027411973464
0.018 85.747064358888
0.0195 86.607267319321
0.021 87.6442076661863
0.0225 88.67200358882
0.024 89.4849457666955
0.0255 90.0532111761397
0.027 90.6791757170906
0.0285 91.171825272922
0.03 91.745102023647
0.0315 92.3568772378027
0.033 92.9542418732756
0.0345 93.4673066909861
0.036 93.906965219519
0.0375 94.3265968907763
0.039 94.7498045556795
0.0405 95.2606433782864
0.042 95.862431523376
0.0435 96.3571079009398
0.045 96.8051679098262
0.0465 97.108567197513
0.048 97.4190697871032
0.0495 97.6773203339521
0.051 97.9795391864349
0.0525 98.2809863724242
0.054 98.5422668517115
0.0555 98.7403035356363
0.057 98.8907691016851
0.0585 99.0227395205273
0.06 99.1309474959176
0.0615 99.2107379312389
0.063 99.2703501591244
0.0645 99.3078969156981
0.066 99.3528241391603
0.0675 99.4105489920868
0.069 99.4353536309481
0.0705 99.4683774048647
0.072 99.4913289014169
0.0735 99.5133185597697
0.075 99.5280026900593
0.0765 99.563735964218
0.078 99.6004291783194
0.0795 99.6187716315584
0.081 99.6417083524643
0.0825 99.6527092120827
0.084 99.6774620421857
0.0855 99.689381147376
0.087 99.7058953574878
0.0885 99.7223772010335
0.09 99.7324542236016
0.0915 99.7407076249052
0.093 99.7654410780219
0.0945 99.7727771452916
0.096 99.7993397474061
0.0975 99.8112551947513
0.099 99.8176775804651
0.1005 99.8259226645443
0.102 99.8479187477535
0.1035 99.8524999753316
0.105 99.8570959897534
0.1065 99.8616725992672
0.108 99.8690068428425
0.1095 99.8754227571137
0.111 99.879091289588
0.1125 99.8809140323746
0.114 99.8836642314319
0.1155 99.8909911400997
0.117 99.8937404237247
0.1185 99.9001600276351
0.12 99.912088378808
0.1215 99.912088378808
0.123 99.9139184603006
0.1245 99.9194068793889
0.126 99.9221570970436
0.1275 99.9230707562487
0.129 99.9249026816605
0.1305 99.9304003372148
0.132 99.9358942936708
0.1335 99.9377243714551
0.135 99.9432090989968
0.1365 99.9450456537833
0.138 99.9569573647778
0.1395 99.9569573647778
0.141 99.9633603882818
0.1425 99.9715934951637
0.144 99.9761682328199
0.1455 99.9844087210448
0.147 99.9899128181845
0.1485 100
0.15 100
};
\addplot [ultra thick, dashed, darkorange25512714]
table {%
0 63.2872061937912
0.0015 63.3257937691775
0.003 63.5029985820476
0.0045 63.8562793475192
0.006 64.0958773355048
0.0075 64.7714969227338
0.009 65.4335776123682
0.0105 66.2871089856193
0.012 67.0868314975317
0.0135 67.9892635729781
0.015 68.8987999157729
0.0165 69.5674408369531
0.018 70.3880018268584
0.0195 71.3866325442277
0.021 72.4122506703419
0.0225 73.3941570418858
0.024 74.0796982791265
0.0255 74.7184548593408
0.027 75.237531860262
0.0285 75.7481397018424
0.03 76.3031991308835
0.0315 76.919296037866
0.033 77.5256083348013
0.0345 78.1919496488265
0.036 78.7640684466356
0.0375 79.5431602141617
0.039 80.1836661898661
0.0405 80.7242855030505
0.042 81.414376829219
0.0435 81.9033023502117
0.045 82.3261372349325
0.0465 82.6685313499164
0.048 83.051857741486
0.0495 83.5075672761412
0.051 83.9029598353477
0.0525 84.3945439790794
0.054 84.9888725552633
0.0555 85.4344891998671
0.057 85.7684480314269
0.0585 86.1592022787808
0.06 86.496368577331
0.0615 86.8326808438198
0.063 87.2298518290145
0.0645 87.6173582115383
0.066 87.987116199531
0.0675 88.2707211953121
0.069 88.6071331770038
0.0705 88.8606238649692
0.072 89.0835010044758
0.0735 89.2777466156127
0.075 89.4796785712188
0.0765 89.6335042748515
0.078 89.7994238142735
0.0795 90.0424849526938
0.081 90.175163217196
0.0825 90.3501915210532
0.084 90.5666072069471
0.0855 90.7868414237197
0.087 91.0677846607635
0.0885 91.2538411447847
0.09 91.4198614422892
0.0915 91.5586831563131
0.093 91.9073067363329
0.0945 92.1407853169098
0.096 92.4153076761621
0.0975 92.6781445040729
0.099 92.8967610724501
0.1005 93.0823564694153
0.102 93.3323113726001
0.1035 93.5873237509984
0.105 93.8123242964938
0.1065 93.9258950664783
0.108 94.0567754682418
0.1095 94.2716490198321
0.111 94.5220404820941
0.1125 94.5948345226282
0.114 94.770697682393
0.1155 94.8577862526481
0.117 95.0422429973284
0.1185 95.1672324331746
0.12 95.3636353745629
0.1215 95.5720722082164
0.123 95.7469947459563
0.1245 96.0561514092007
0.126 96.3315416515112
0.1275 96.6433986115411
0.129 96.8172310475166
0.1305 97.0834559942452
0.132 97.252853150831
0.1335 97.4270176326627
0.135 97.6962965714717
0.1365 98.0069385419717
0.138 98.1708406811048
0.1395 98.3993046110812
0.141 98.7685846674754
0.1425 99.0265916401134
0.144 99.286195611681
0.1455 99.4983256173369
0.147 99.6782215084182
0.1485 100
0.15 100
};
\end{axis}

\end{tikzpicture}

%% file: SupplimentFigures/pckNoise.tex
\begin{tikzpicture}

\definecolor{darkorange25512714}{RGB}{255,127,14}
\definecolor{darkslategray38}{RGB}{38,38,38}
\definecolor{forestgreen4416044}{RGB}{44,160,44}
\definecolor{lightgray204}{RGB}{204,204,204}
\definecolor{steelblue31119180}{RGB}{31,119,180}

\Large
\begin{axis}[
axis line style={lightgray204},
legend cell align={left},
legend style={
  fill opacity=0.8,
  draw opacity=1,
  text opacity=1,
  at={(0.53,0.03)},
  anchor=south west,
  draw=lightgray204
},
tick align=outside,
tick pos=left,
xtick = {0,0.05,0.10,0.15},
ytick = {40,60,80,100},
x grid style={lightgray204},
xlabel=\textcolor{darkslategray38}{Geodesic Error},
xmajorgrids,
xmin=0, xmax=0.15,
xtick style={color=darkslategray38},
xticklabel style={yshift= 5pt},
y grid style={lightgray204},
ylabel=\textcolor{darkslategray38}{\% Correspondences},
ymajorgrids,
ymin=40, ymax=100,
ytick style={color=darkslategray38}
]
\addplot [ultra thick, steelblue31119180]
table {%
0 93.4019477644976
0.0015 93.4021691013723
0.003 93.4074811863656
0.0045 93.4229747675963
0.006 93.4645861000443
0.0075 93.6702080566622
0.009 93.9803010181496
0.0105 94.4156706507304
0.012 94.7726870296591
0.0135 95.1188579017264
0.015 95.4712262062859
0.0165 95.8612217795485
0.018 96.199203187251
0.0195 96.4645861000443
0.021 96.8025675077468
0.0225 97.1790615316512
0.024 97.4922532093847
0.0255 97.7833111996458
0.027 98.0336432049579
0.0285 98.2386011509517
0.03 98.4420097388225
0.0315 98.6250553342187
0.033 98.7990261177512
0.0345 98.9389110225763
0.036 99.0894200973882
0.0375 99.1963258078796
0.039 99.2682602921646
0.0405 99.3530323151836
0.042 99.4156706507303
0.0435 99.4873837981407
0.045 99.5559982293049
0.0465 99.6177512173528
0.048 99.6757414785303
0.0495 99.7153607791058
0.051 99.7563081009296
0.0525 99.7886232846392
0.054 99.8065515714918
0.0555 99.8247011952191
0.057 99.839973439575
0.0585 99.847941567065
0.06 99.8601150951748
0.0615 99.8634351482957
0.063 99.8691899070385
0.0645 99.8747233289066
0.066 99.881584772023
0.0675 99.8851261620185
0.069 99.8860115095174
0.0705 99.8906595838866
0.072 99.893536963258
0.0735 99.8968570163789
0.075 99.9026117751217
0.0765 99.9052678176184
0.078 99.9061531651173
0.0795 99.9065958388667
0.081 99.9096945551128
0.0825 99.9163346613546
0.084 99.9196547144754
0.0855 99.9225320938468
0.087 99.9260734838424
0.0885 99.9267374944666
0.09 99.9280655157149
0.0915 99.9285081894643
0.093 99.9285081894643
0.0945 99.9293935369632
0.096 99.9318282425852
0.0975 99.9322709163346
0.099 99.9324922532093
0.1005 99.9329349269588
0.102 99.935148295706
0.1035 99.9393536963258
0.105 99.9393536963258
0.1065 99.9424524125719
0.108 99.9424524125719
0.1095 99.9431164231961
0.111 99.9446657813191
0.1125 99.945551128818
0.114 99.9459938025675
0.1155 99.9464364763169
0.117 99.9475431606906
0.1185 99.9510845506861
0.12 99.951969898185
0.1215 99.9528552456839
0.123 99.9532979194334
0.1245 99.954404603807
0.126 99.9552899513059
0.1275 99.95595396193
0.129 99.9561752988048
0.1305 99.9572819831784
0.132 99.9581673306773
0.1335 99.9586100044267
0.135 99.9586100044267
0.1365 99.9586100044267
0.138 99.9586100044267
0.1395 99.9588313413014
0.141 99.9590526781762
0.1425 99.9610447100487
0.144 99.9612660469234
0.1455 99.9612660469234
0.147 99.9612660469234
0.1485 99.9614873837981
0.15 99.9614873837981
};
\addlegendentry{$\sigma^2=0$}
\addplot [ultra thick, steelblue31119180, dashed, forget plot]
table {%
0 91.6682988047811
0.0015 91.8573559096949
0.003 92.1534072598499
0.0045 92.2699092518816
0.006 92.4444147853034
0.0075 92.6561049136786
0.009 92.9077742363879
0.0105 93.1210084108013
0.012 93.3243311199647
0.0135 93.5868308986276
0.015 93.820584329349
0.0165 93.9781297034086
0.018 94.1964696768483
0.0195 94.4627162461269
0.021 94.6034532979197
0.0225 94.7307100486944
0.024 94.866814962373
0.0255 95.0051606905713
0.027 95.1204501992035
0.0285 95.2372828685262
0.03 95.3412611775124
0.0315 95.4511212926075
0.033 95.5294431164233
0.0345 95.6038857901726
0.036 95.6777923860115
0.0375 95.7374249667994
0.039 95.7899269588312
0.0405 95.8569738822487
0.042 95.9315214696767
0.0435 96.0127498893314
0.045 96.1053359893757
0.0465 96.1861797255421
0.048 96.2939172200086
0.0495 96.4090708277998
0.051 96.5369482071712
0.0525 96.667708277999
0.054 96.7856184152279
0.0555 96.8869433377599
0.057 96.9861580345284
0.0585 97.0896980965028
0.06 97.2038640991587
0.0615 97.3278587870737
0.063 97.4450053120849
0.0645 97.5617963700753
0.066 97.6993984063747
0.0675 97.8068924302792
0.069 97.9534856131036
0.0705 98.0742872952638
0.072 98.1945254537411
0.0735 98.304020362993
0.075 98.389926516158
0.0765 98.4715276671097
0.078 98.5557777777782
0.0795 98.6202328463926
0.081 98.6725170429397
0.0825 98.7215692784422
0.084 98.7778260292168
0.0855 98.8353948649849
0.087 98.8738255865432
0.0885 98.9146892430284
0.09 98.9640730411692
0.0915 99.0067091633472
0.093 99.0378357680395
0.0945 99.0886843736173
0.096 99.1400743691906
0.0975 99.1785024347063
0.099 99.2314125719351
0.1005 99.2930987162467
0.102 99.3449207613994
0.1035 99.395184594954
0.105 99.457755201417
0.1065 99.503370075255
0.108 99.5598401947769
0.1095 99.5897419212044
0.111 99.6271522797702
0.1125 99.6498583444005
0.114 99.6764285081898
0.1155 99.6943700752548
0.117 99.7095489154496
0.1185 99.7231677733513
0.12 99.737783089863
0.1215 99.7508472775567
0.123 99.7730004426739
0.1245 99.786519698982
0.126 99.7996024789731
0.1275 99.8113652058434
0.129 99.8235599822932
0.1305 99.8379654714476
0.132 99.8480455953963
0.1335 99.8579132359452
0.135 99.8658946436477
0.1365 99.8769853917663
0.138 99.8854218680833
0.1395 99.8899667994688
0.141 99.8975090748119
0.1425 99.907606020363
0.144 99.9174860557769
0.1455 99.920591854803
0.147 99.924586542718
0.1485 99.9319065958389
0.15 99.9319065958389
};
\addplot [ultra thick, darkorange25512714]
table {%
0 86.7906153165118
0.0015 86.792828685259
0.003 86.8105356352369
0.0045 86.8503762726871
0.006 87.0030987162462
0.0075 87.2155821159806
0.009 87.8087649402391
0.0105 88.3975210270031
0.012 89.1522797698097
0.0135 90.0442673749447
0.015 90.9915891987605
0.0165 91.7441345728198
0.018 92.647189021691
0.0195 93.3753873395308
0.021 94.0150509074812
0.0225 94.6724214254094
0.024 95.2988047808765
0.0255 95.9096945551128
0.027 96.3789287295263
0.0285 96.896857016379
0.03 97.220008853475
0.0315 97.5586542718017
0.033 97.8353253652058
0.0345 98.140770252324
0.036 98.377600708278
0.0375 98.6498450641877
0.039 98.7715803452855
0.0405 98.9420097388224
0.042 99.0526781761841
0.0435 99.1389995573262
0.045 99.2054006197432
0.0465 99.2540947321823
0.048 99.3293492695883
0.0495 99.4090305444887
0.051 99.4577246569278
0.0525 99.5174856131031
0.054 99.5617529880478
0.0555 99.5816733067728
0.057 99.6060203629924
0.0585 99.6126604692341
0.06 99.6170872067286
0.0615 99.6281540504647
0.063 99.630367419212
0.0645 99.6525011066843
0.066 99.6635679504205
0.0675 99.6923417441345
0.069 99.7100486941123
0.0705 99.7122620628596
0.072 99.7343957503319
0.0735 99.7432492253209
0.075 99.7675962815405
0.0765 99.7742363877822
0.078 99.79194333776
0.0795 99.7941567065073
0.081 99.7963700752545
0.0825 99.7985834440017
0.084 99.800796812749
0.0855 99.8052235502434
0.087 99.8096502877379
0.0885 99.8162903939796
0.09 99.8185037627268
0.0915 99.8185037627268
0.093 99.8229305002213
0.0945 99.8251438689685
0.096 99.8317839752102
0.0975 99.8317839752102
0.099 99.8317839752102
0.1005 99.8339973439575
0.102 99.8339973439575
0.1035 99.8339973439575
0.105 99.8339973439575
0.1065 99.8362107127047
0.108 99.8362107127047
0.1095 99.8362107127047
0.111 99.8362107127047
0.1125 99.8362107127047
0.114 99.8362107127047
0.1155 99.8561310314298
0.117 99.8561310314298
0.1185 99.8561310314298
0.12 99.8561310314298
0.1215 99.8561310314298
0.123 99.8561310314298
0.1245 99.8561310314298
0.126 99.8561310314298
0.1275 99.8561310314298
0.129 99.8561310314298
0.1305 99.8561310314298
0.132 99.8561310314298
0.1335 99.8561310314298
0.135 99.858344400177
0.1365 99.858344400177
0.138 99.858344400177
0.1395 99.858344400177
0.141 99.858344400177
0.1425 99.858344400177
0.144 99.858344400177
0.1455 99.858344400177
0.147 99.858344400177
0.1485 99.858344400177
0.15 99.858344400177
};
\addlegendentry{$\sigma^2=0.01$}
\addplot [ultra thick, darkorange25512714, dashed, forget plot]
table {%
0 54.4931385568836
0.0015 54.4931385568836
0.003 54.497565294378
0.0045 54.5329791943338
0.006 54.6480743691899
0.0075 55.0774679061532
0.009 55.3917662682603
0.0105 55.7149181053563
0.012 56.555998229305
0.0135 57.7600708277999
0.015 58.3532536520584
0.0165 59.0261177512173
0.018 59.6901283753873
0.0195 60.5533421868083
0.021 61.5493581230633
0.0225 62.4745462594068
0.024 63.4484285081895
0.0255 64.3868968570164
0.027 65.3917662682603
0.0285 66.1752988047809
0.03 66.8570163789287
0.0315 67.6139884904825
0.033 68.5568835768039
0.0345 69.4333776007083
0.036 70.0619743249225
0.0375 70.9915891987605
0.039 71.651173085436
0.0405 72.1779548472775
0.042 72.5542275343072
0.0435 72.9216467463479
0.045 73.3421868083223
0.0465 73.7273129703409
0.048 74.1301460823373
0.0495 74.5374059318282
0.051 74.9579459938026
0.0525 75.5157149181054
0.054 75.9362549800797
0.0555 76.3612217795485
0.057 76.7242142540947
0.0585 77.1668880035414
0.06 77.5033200531209
0.0615 77.950420540062
0.063 78.4949092518814
0.0645 79.0748118636565
0.066 79.5661797255423
0.0675 80.0929614873838
0.069 80.6463036741922
0.0705 81.2040725984949
0.072 81.6069057104914
0.0735 82.2000885347499
0.075 82.5984949092519
0.0765 83.0854360336432
0.078 83.4262948207171
0.0795 83.7892872952634
0.081 84.1389995573262
0.0825 84.5683930942895
0.084 84.9977866312528
0.0855 85.5157149181054
0.087 86.0956175298805
0.0885 86.5692784417884
0.09 86.9765382912794
0.0915 87.3749446657814
0.093 87.7290836653387
0.0945 88.016821602479
0.096 88.3665338645419
0.0975 88.6100044267375
0.099 88.9907038512616
0.1005 89.2518813634351
0.102 89.5617529880478
0.1035 89.867197875166
0.105 90.0442673749446
0.1065 90.2877379371403
0.108 90.5090748118636
0.1095 90.6861443116423
0.111 90.863213811421
0.1125 91.0137228862328
0.114 91.2129260734838
0.1155 91.32802124834
0.117 91.5095174856131
0.1185 91.792828685259
0.12 92.2797698096503
0.1215 92.4833997343958
0.123 92.647189021691
0.1245 92.8729526339089
0.126 93.054448871182
0.1275 93.2979194333776
0.129 93.6741921204073
0.1305 93.842408145197
0.132 93.975210270031
0.1335 94.1389995573263
0.135 94.3249225320938
0.1365 94.4621513944223
0.138 94.6613545816733
0.1395 94.89597166888
0.141 95.1350154935812
0.1425 95.2368304559539
0.144 95.3386454183267
0.1455 95.4006197432492
0.147 95.6086764054891
0.1485 95.7149181053563
0.15 95.7149181053563
};
\addplot [ultra thick, forestgreen4416044]
table {%
0 69.5484727755644
0.0015 69.5573262505533
0.003 69.6436476316954
0.0045 69.7189021691014
0.006 69.8716246126604
0.0075 70.1040283311199
0.009 70.5356352368305
0.0105 71.5449313855688
0.012 72.5453740593183
0.0135 73.680832226649
0.015 74.8738379814077
0.0165 75.8853474988933
0.018 77.4280655157149
0.0195 78.7915006640106
0.021 80.1482957060646
0.0225 81.5250110668437
0.024 82.6228419654714
0.0255 83.8733953076582
0.027 84.9291722000885
0.0285 85.7414785303231
0.03 86.5382912793271
0.0315 87.3306772908366
0.033 87.8818061088977
0.0345 88.444001770695
0.036 88.995130588756
0.0375 89.5307658255865
0.039 90.0774679061531
0.0405 90.4758742806551
0.042 90.8676405489154
0.0435 91.1000442673749
0.045 91.2306330234616
0.0465 91.5936254980079
0.048 91.7950420540062
0.0495 92.0385126162018
0.051 92.2244355909694
0.0525 92.34395750332
0.054 92.476759628154
0.0555 92.5984949092518
0.057 92.7888446215139
0.0585 92.8950863213811
0.06 92.9836210712704
0.0615 93.1186365648516
0.063 93.1651173085435
0.0645 93.2293050022133
0.066 93.2979194333775
0.0675 93.3665338645417
0.069 93.4373616644532
0.0705 93.4993359893758
0.072 93.5502434705621
0.0735 93.5812306330234
0.075 93.6033643204957
0.0765 93.6277113767153
0.078 93.6564851704293
0.0795 93.6830455953961
0.081 93.7405931828242
0.0825 93.767153607791
0.084 93.8401947764497
0.0855 93.8822487826471
0.087 93.8933156263833
0.0885 93.9552899513058
0.09 93.9641434262947
0.0915 94.0128375387339
0.093 94.0371845949534
0.0945 94.0593182824258
0.096 94.0925188136343
0.0975 94.1257193448428
0.099 94.1567065073041
0.1005 94.1899070385126
0.102 94.2408145196989
0.1035 94.260734838424
0.105 94.3160690571049
0.1065 94.3979637007526
0.108 94.4200973882249
0.1095 94.4289508632138
0.111 94.4510845506861
0.1125 94.4555112881806
0.114 94.4665781319168
0.1155 94.4931385568836
0.117 94.5042054006197
0.1185 94.5196989818504
0.12 94.5484727755644
0.1215 94.5573262505533
0.123 94.5661797255423
0.1245 94.5706064630368
0.126 94.5750332005312
0.1275 94.5816733067729
0.129 94.5861000442674
0.1305 94.6569278441788
0.132 94.7100486941125
0.1335 94.7144754316069
0.135 94.7277556440903
0.1365 94.7410358565737
0.138 94.7565294378044
0.1395 94.7675962815405
0.141 94.7808764940239
0.1425 94.8074369189907
0.144 94.8140770252324
0.1455 94.8317839752103
0.147 94.8627711376716
0.1485 94.867197875166
0.15 94.867197875166
};
\addlegendentry{$\sigma^2=0.02$}
\addplot [ultra thick, forestgreen4416044, dashed, forget plot]
table {%
0 41.9256308100929
0.0015 41.9256308100929
0.003 41.9300575475874
0.0045 41.9300575475874
0.006 42.2443559096946
0.0075 43.0146082337317
0.009 43.6653386454183
0.0105 44.2363877822045
0.012 45.4537405931828
0.0135 47.1226206285967
0.015 48.1407702523241
0.0165 49.4289508632138
0.018 50.6551571491811
0.0195 52.2177954847278
0.021 53.3377600708278
0.0225 54.7808764940239
0.024 56.4099158919876
0.0255 57.7689243027889
0.027 59.1456396635679
0.0285 60.4205400619743
0.03 61.6998671978751
0.0315 63.2093846834882
0.033 64.4444444444445
0.0345 65.5821159805224
0.036 66.6932270916335
0.0375 67.7910579902612
0.039 68.5922974767596
0.0405 69.4953519256308
0.042 70.1859229747676
0.0435 70.9650287737937
0.045 71.7441345728198
0.0465 72.6073483842408
0.048 73.2890659583887
0.0495 73.8733953076583
0.051 74.6746347941567
0.0525 75.369632580788
0.054 76.1708720672864
0.0555 76.7020805666224
0.057 77.2244355909694
0.0585 77.6272687029659
0.06 77.9327135900841
0.0615 78.4683488269145
0.063 79.2164674634794
0.0645 79.7698096502878
0.066 80.2833111996459
0.0675 80.7702523240372
0.069 81.3590084108013
0.0705 81.8459495351926
0.072 82.2089420097389
0.0735 82.8906595838867
0.075 83.3466135458168
0.0765 83.784860557769
0.078 84.2363877822045
0.0795 84.7321823815848
0.081 85.1969898185038
0.0825 85.7547587428066
0.084 86.1752988047809
0.0855 86.6888003541391
0.087 86.9499778663126
0.0885 87.2687029659142
0.09 87.5298804780877
0.0915 87.9946879150067
0.093 88.3621071270474
0.0945 88.6542718016822
0.096 88.8756086764055
0.0975 89.1279327135901
0.099 89.4953519256308
0.1005 89.6635679504205
0.102 90.0044267374944
0.1035 90.2346170872067
0.105 90.3762726870297
0.1065 90.4957945993802
0.108 90.6507304116866
0.1095 90.8455068614431
0.111 91.0402833111996
0.1125 91.2262062859672
0.114 91.3988490482514
0.1155 91.6910137228863
0.117 91.9300575475875
0.1185 92.1646746347942
0.12 92.4391323594511
0.1215 92.7357237715804
0.123 92.9659141212927
0.1245 93.1518370960602
0.126 93.3156263833555
0.1275 93.4971226206286
0.129 93.7981407702524
0.1305 93.9884904825144
0.132 94.0947321823816
0.1335 94.2363877822045
0.135 94.373616644533
0.1365 94.5949535192563
0.138 94.6923417441346
0.1395 94.8339973439575
0.141 95.0243470562196
0.1425 95.0907481186366
0.144 95.1969898185038
0.1455 95.2501106684373
0.147 95.4493138556883
0.1485 95.5201416555998
0.15 95.5201416555998
};
\end{axis}

\end{tikzpicture}